\documentclass[acmtog]{acmart}
\acmSubmissionID{348}

\usepackage{booktabs} 
\usepackage{amsmath}
\usepackage{bbm}

\usepackage[ruled]{algorithm2e} 

\SetAlFnt{\small}
\SetAlCapFnt{\small}
\SetAlCapNameFnt{\small}
\SetAlCapHSkip{0pt}

\acmJournal{TOG}

\usepackage{bm}
\usepackage{multirow}

\begin{document}
\title{NeRO: \underline{Ne}ural Geometry and BRDF Reconstruction of \underline{R}eflective \underline{O}bjects from Multiview Images}
\newcommand{\ly}[1]{{\color{blue}{Yuan: #1}}}
\newcommand{\WP}[1]{{\color{green}{[\textbf{WP:} #1]}}}
\newcommand{\LXX}[1]{{\color{orange}{[\textbf{LXX:} #1]}}}
\newcommand{\JP}[1]{{\color{pink}{[\textbf{JP:} #1]}}}
\newcommand{\rv}[1]{{\color{red}{#1}}}
\author{Yuan Liu}
\affiliation{
  \institution{The University of Hong Kong}
  \city{Hong Kong}
  \country{China}
}

\author{Peng Wang}
\affiliation{
  \institution{The University of Hong Kong}
  \city{Hong Kong}
  \country{China}
}

\author{Cheng Lin}
\affiliation{
  \institution{Tencent Games}
  \city{Shen Zhen}
  \country{China}
}

\author{Xiaoxiao Long}
\affiliation{
  \institution{The University of Hong Kong}
  \city{Hong Kong}
  \country{China}
}
\author{Jiepeng Wang}
\affiliation{
  \institution{The University of Hong Kong}
  \city{Hong Kong}
  \country{China}
}
\author{Lingjie Liu}
\affiliation{
  \institution{Max Planck Institute for Informatics}
  \country{Germany}
}
\affiliation{
  \institution{University of Pennsylvania}
  \country{USA}
}

\author{Taku Komura}
\affiliation{
  \institution{The University of Hong Kong}
  \city{Hong Kong}
  \country{China}
}

\author{Wenping Wang}
\affiliation{%
  \institution{Texas A\&M University}
  \city{Texas}
  \country{U.S.A}
}

\begin{abstract}
We present a neural rendering-based method called NeRO for reconstructing the geometry and the BRDF of reflective objects from multiview images captured in an unknown environment. Multiview reconstruction of reflective objects is extremely challenging because specular reflections are view-dependent and thus violate the multiview consistency, which is the cornerstone for most multiview reconstruction methods. Recent neural rendering techniques can model the interaction between environment lights and the object surfaces to fit the view-dependent reflections, thus making it possible to reconstruct reflective objects from multiview images. However, accurately modeling environment lights in the neural rendering is intractable, especially when the geometry is unknown. Most existing neural rendering methods, which can model environment lights, only consider direct lights and rely on object masks to reconstruct objects with weak specular reflections. Therefore, these methods fail to reconstruct reflective objects, especially when the object mask is not available and the object is illuminated by indirect lights. We propose a two-step approach to tackle this problem. First, by applying the split-sum approximation and the integrated directional encoding to approximate the shading effects of both direct and indirect lights, we are able to accurately reconstruct the geometry of reflective objects without any object masks. Then, with the object geometry fixed, we use more accurate sampling to recover the environment lights and the BRDF of the object. Extensive experiments demonstrate that our method is capable of accurately reconstructing the geometry and the BRDF of reflective objects from only posed RGB images without knowing the environment lights and the object masks. Codes and datasets are available at \href{https://github.com/liuyuan-pal/NeRO}{https://github.com/liuyuan-pal/NeRO}.

\end{abstract}




\begin{teaserfigure}
  \centering
  \includegraphics[width=\textwidth]{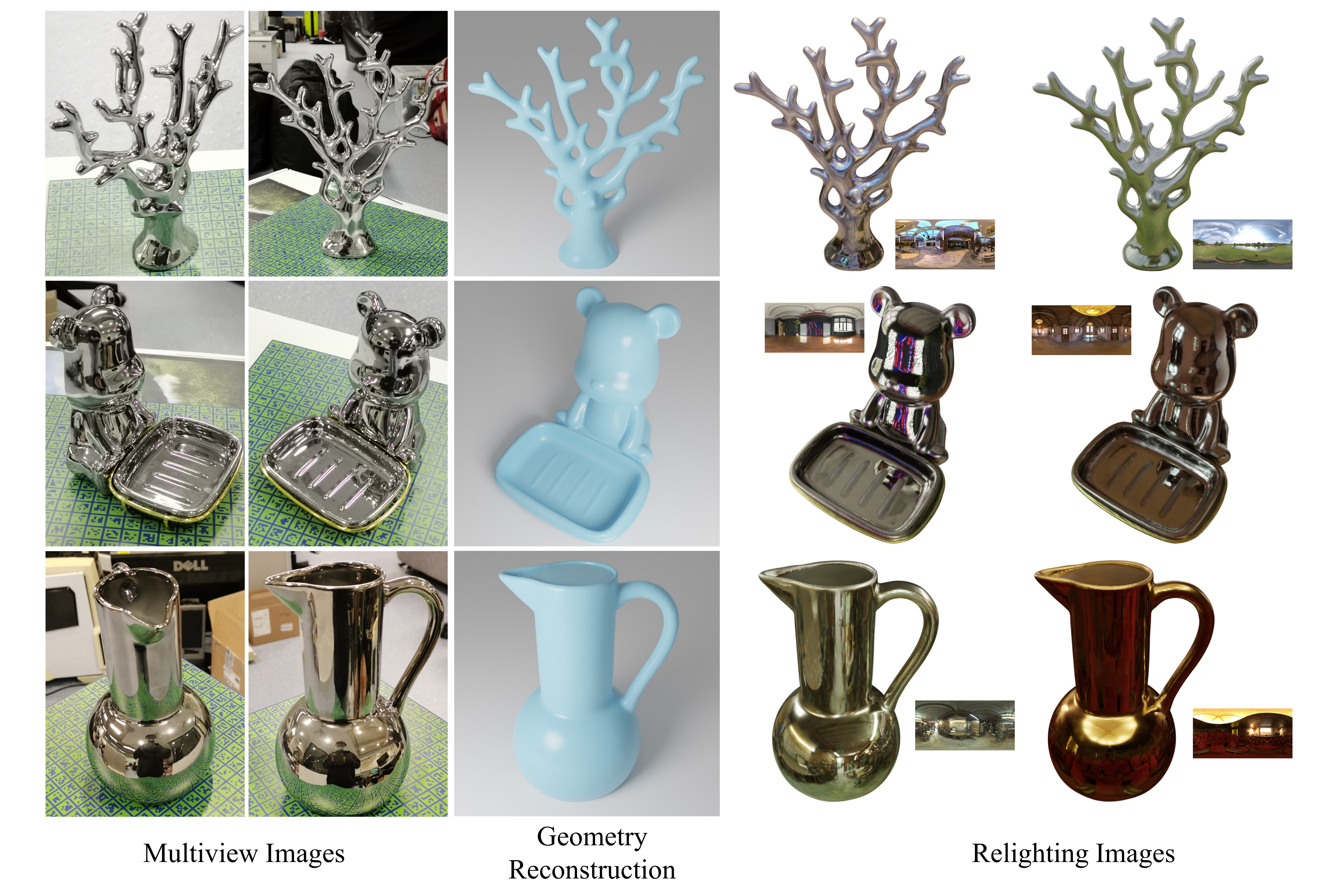}
  \caption{\textbf{NeRO.} We present NeRO for reconstructing the geometry and the BRDF of reflective objects with strong reflective appearances. NeRO only requires multiview input images of the reflective object under an unknown illumination condition. The output of NeRO is a triangular mesh with material parameters, which can easily be used in rendering software for relighting and other applications.}
  \label{fig:teaser}
\end{teaserfigure}

\maketitle


\section{Introduction}
\begin{figure*}
    \centering
    \setlength\tabcolsep{1pt}
    \begin{tabular}{cccccc}
        \includegraphics[width=0.16\textwidth]{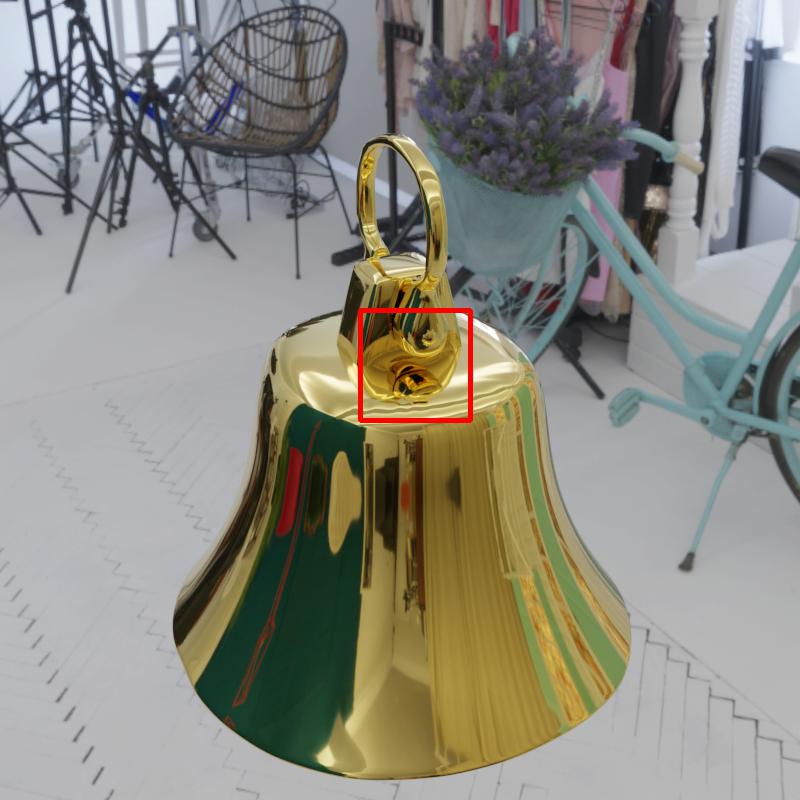} &
        \includegraphics[width=0.16\textwidth]{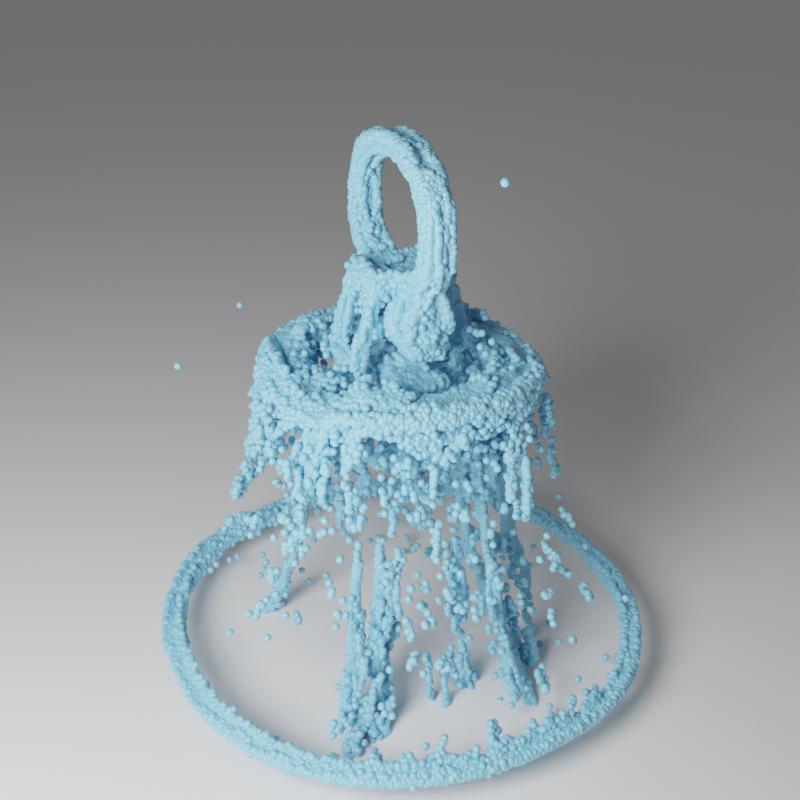} &
        \includegraphics[width=0.16\textwidth]{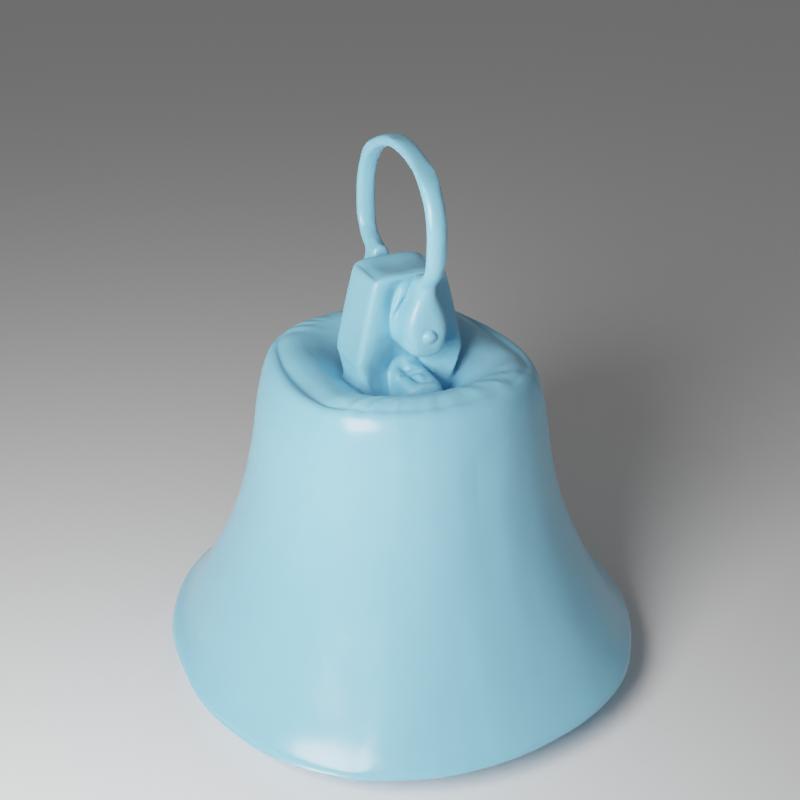} &
        \includegraphics[width=0.16\textwidth]{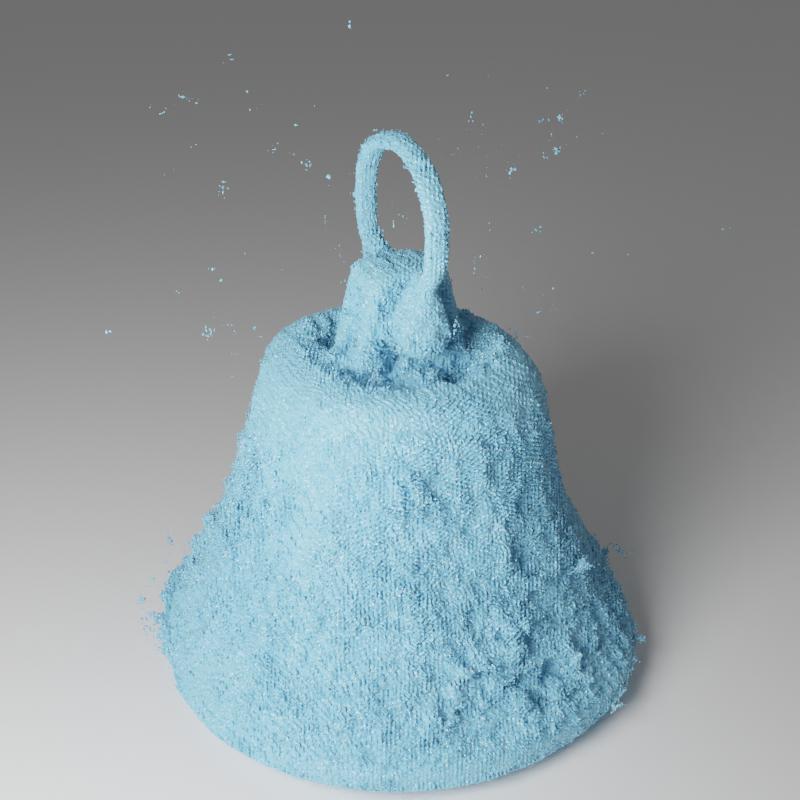} &
        \includegraphics[width=0.16\textwidth]{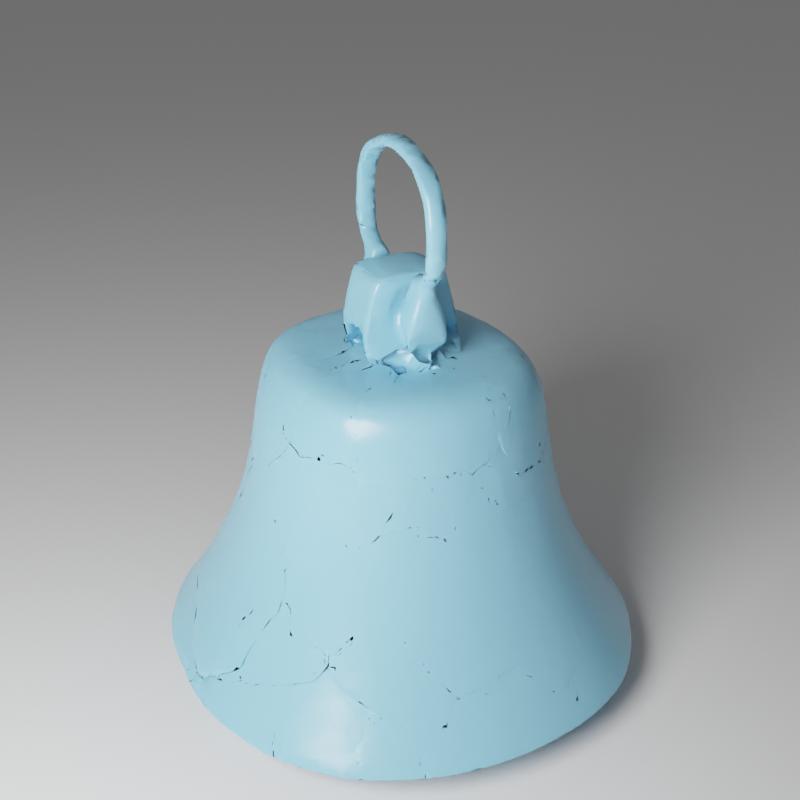} &
        \includegraphics[width=0.16\textwidth]{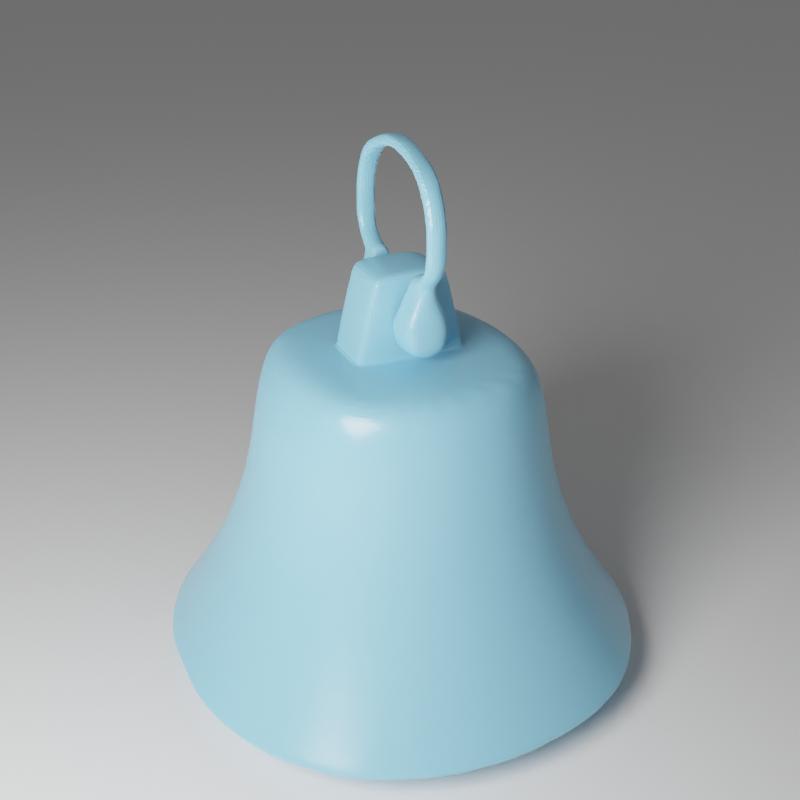} \\
        \includegraphics[width=0.16\textwidth]{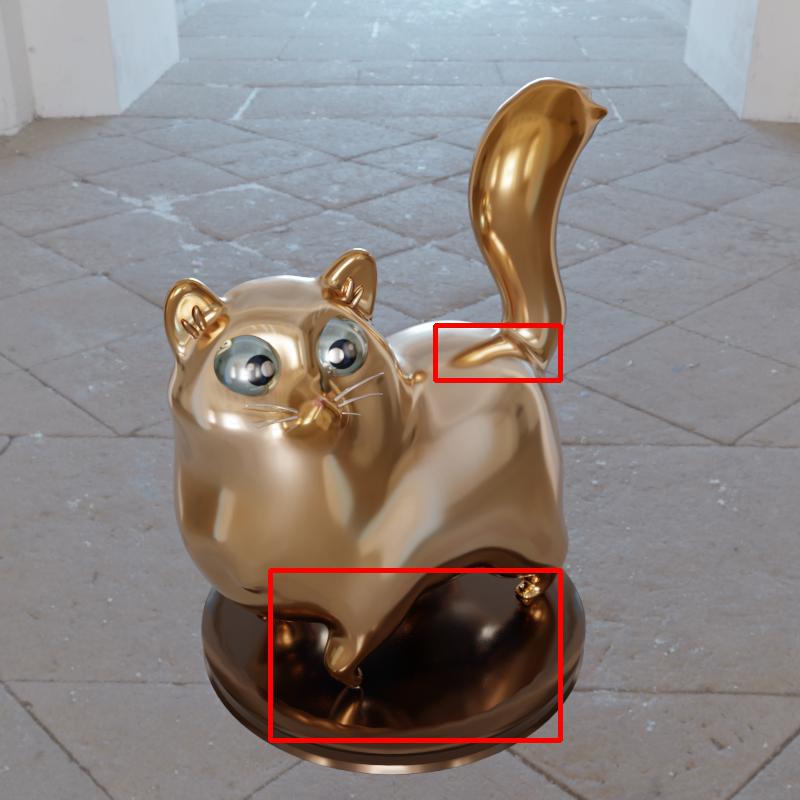} &
        \includegraphics[width=0.16\textwidth]{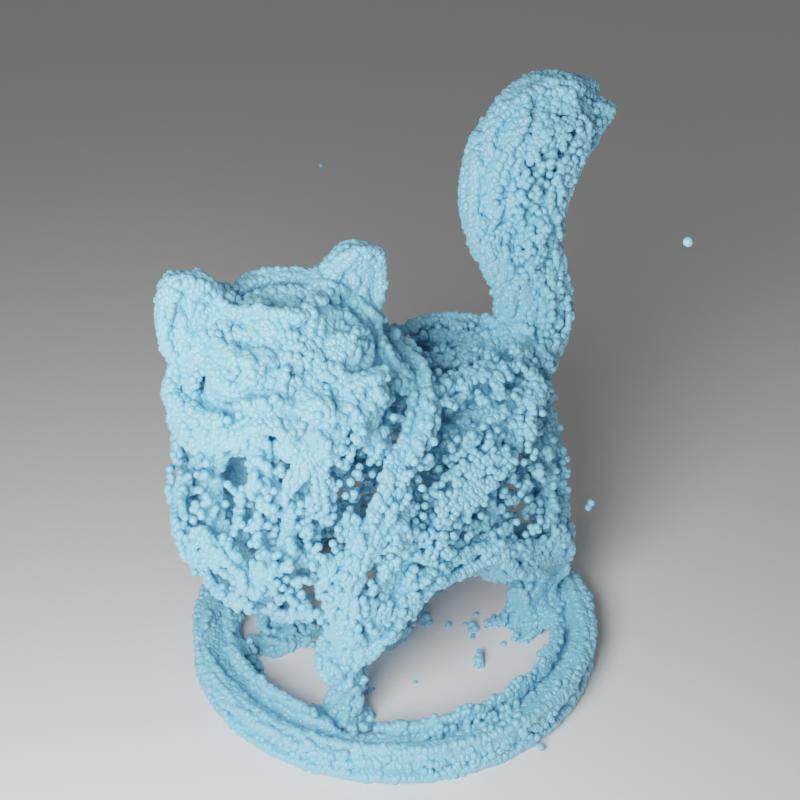} &
        \includegraphics[width=0.16\textwidth]{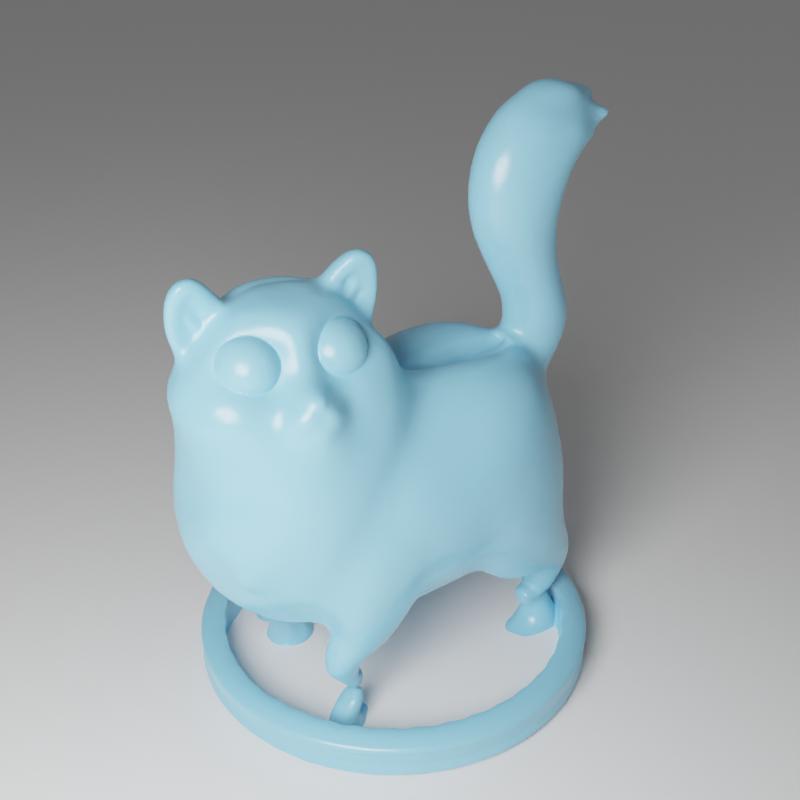} &
        \includegraphics[width=0.16\textwidth]{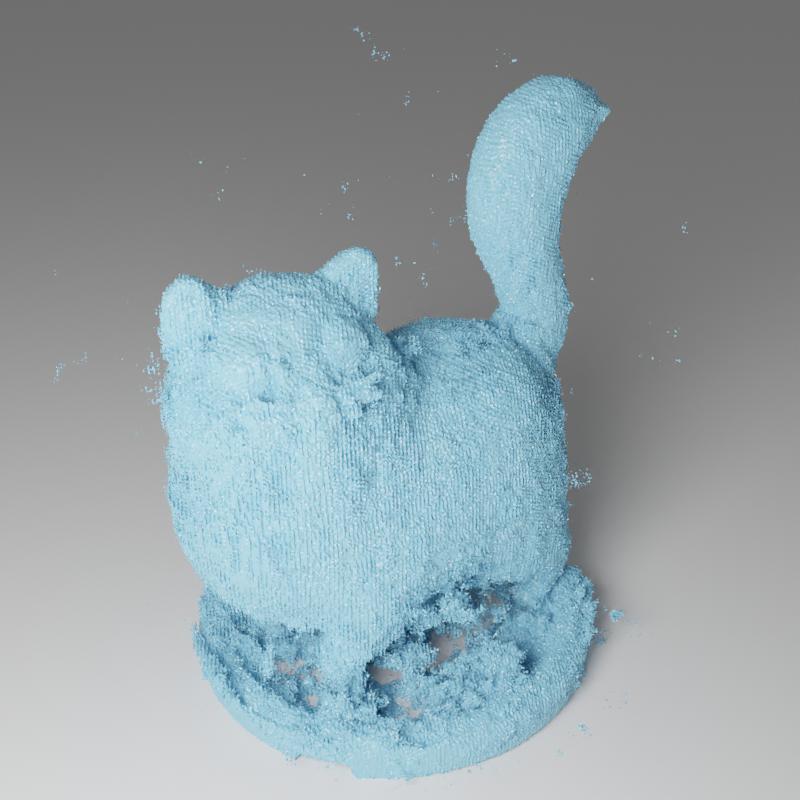} &
        \includegraphics[width=0.16\textwidth]{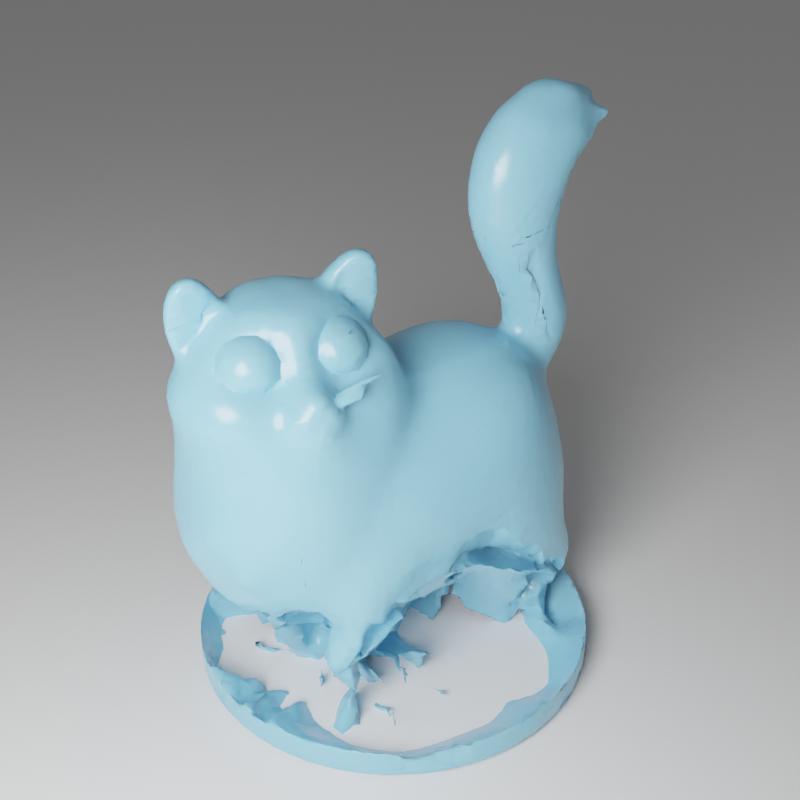} &
        \includegraphics[width=0.16\textwidth]{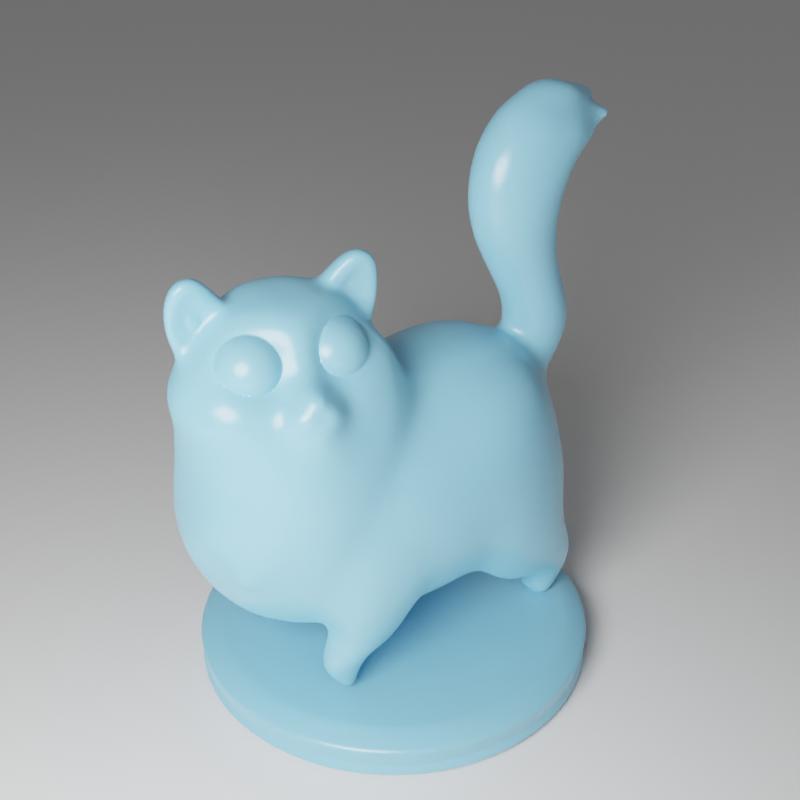} \\
        \includegraphics[width=0.16\textwidth]{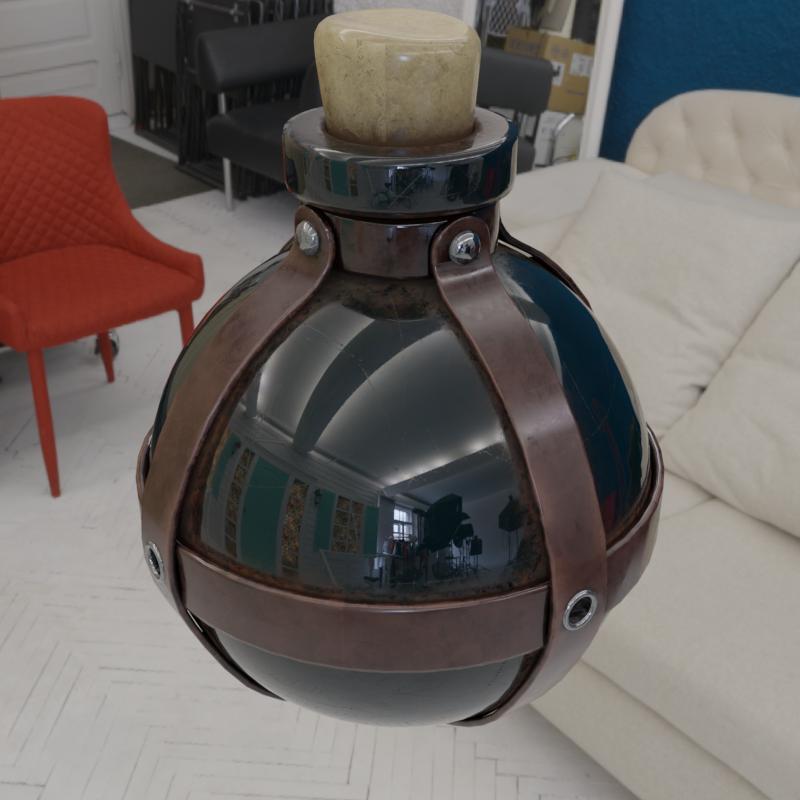} &
        \includegraphics[width=0.16\textwidth]{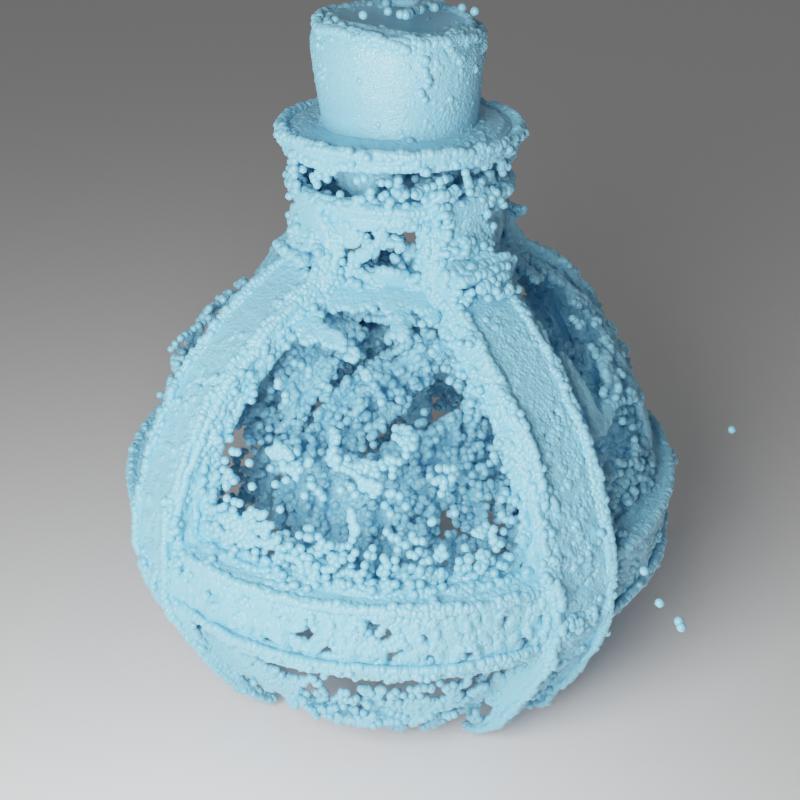} &
        \includegraphics[width=0.16\textwidth]{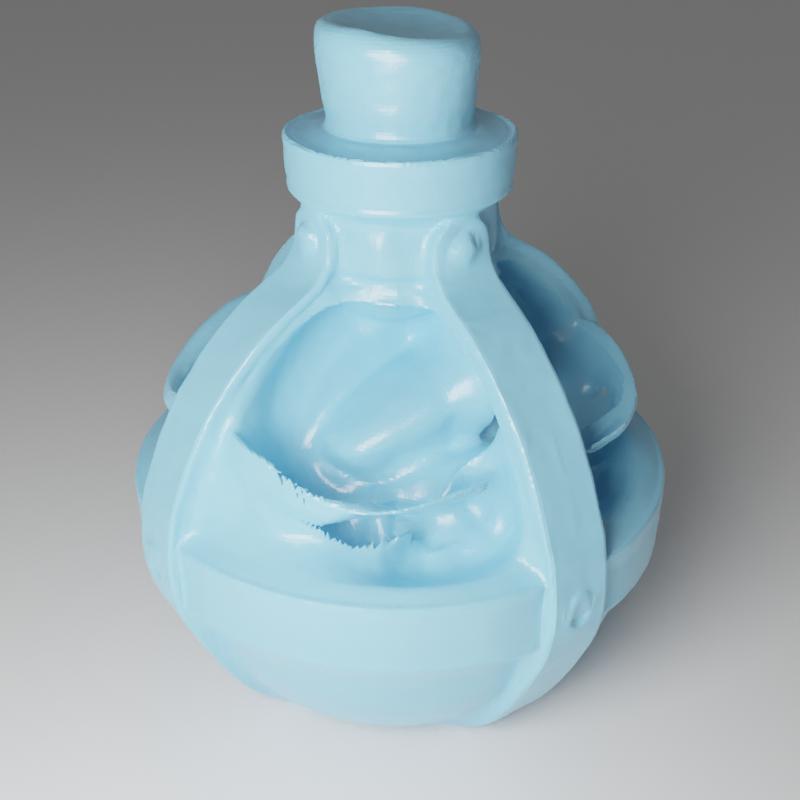} &
        \includegraphics[width=0.16\textwidth]{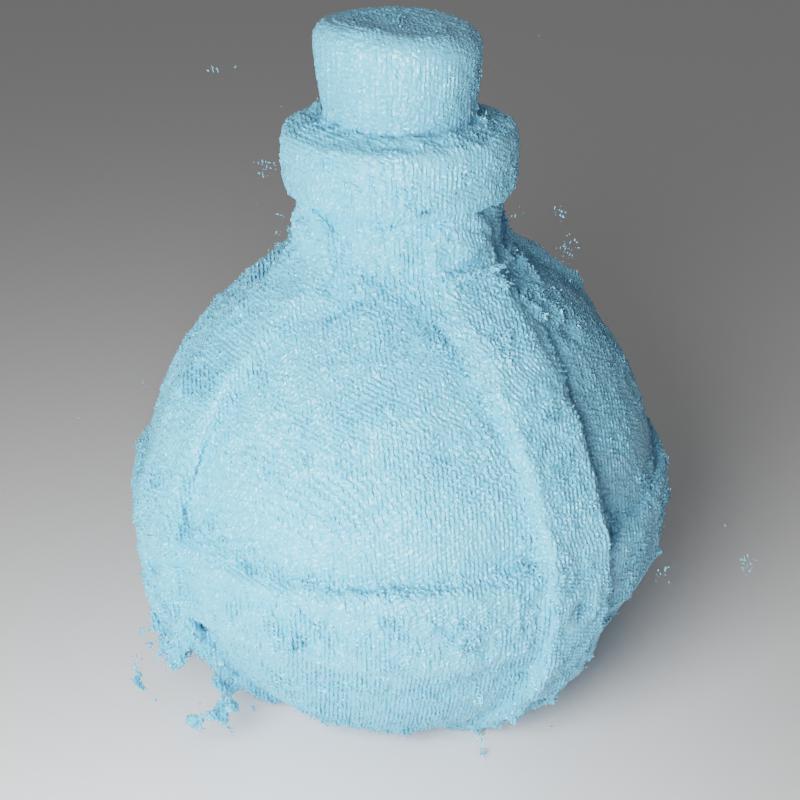} &
        \includegraphics[width=0.16\textwidth]{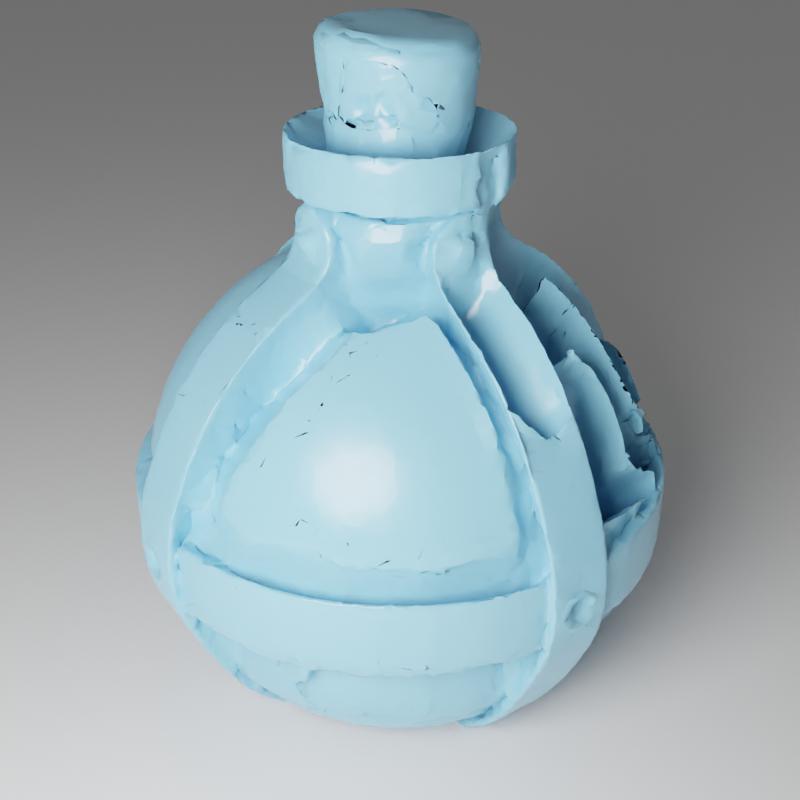} &
        \includegraphics[width=0.16\textwidth]{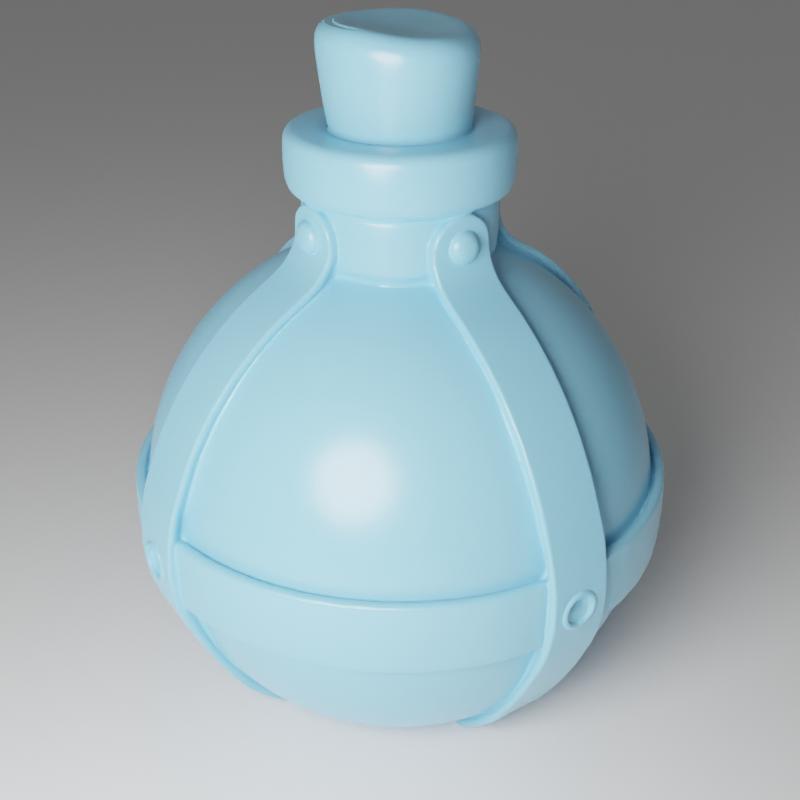} \\
        (a) Image & (b) COLMAP & (c) NeuS & (d) Ref-NeRF & (e) NDRMC$^*$ & (f) Ours
    \end{tabular}
    \caption{We apply different multiview reconstruction methods to reconstructing reflective objects. (a) The input images. Reconstruction results of (b) COLMAP~\cite{schoenberger2016mvs}, (c) NeuS~\cite{wang2021neus}, (d) Ref-NeRF~\cite{verbin2022ref}, (e) NDRMC~\cite{hasselgren2022shape} and (f) our method. In images of column (a), we use red bounding boxes to indicate regions illuminated by indirect lights. $^*$NDRMC uses ground-truth object masks for training while all the other methods are trained without object masks.}
    \label{fig:bell}
\end{figure*}

Multiview 3D reconstruction, a fundamental task in computer graphics and vision ~\cite{hartley2003multiple}, has witnessed tremendous progress in recent years ~\cite{schoenberger2016mvs,wang2021patchmatchnet,yao2018mvsnet,yariv2020multiview,yariv2021volume,wang2021neus,oechsle2021unisurf}. Despite the compelling results achieved, the reconstruction of reflective objects, which are frequently seen in the real-world environment, remains a challenging and outstanding problem. Reflective objects usually have glossy surfaces on which some or all of the lights that strike the object are reflected. The reflection leads to inconsistent colors when observing the objects from different views. However, most multiview reconstruction methods rely heavily on view consistency for stereo matching. This constitutes a significant barrier to the reconstruction quality of existing techniques. Fig.~\ref{fig:bell} (b) shows the reconstructions of widely-used COLMAP~\cite{schoenberger2016mvs} on reflective objects.



As an emerging trend for multi-view reconstruction, modeling surfaces based on neural rendering exhibits a powerful ability for tackling complex objects ~\cite{yariv2020multiview,yariv2021volume,wang2021neus,oechsle2021unisurf}. In these so-called neural reconstruction methods, the underlying surface geometry is represented as an implicit function, e.g., a signed distance function (SDF) encoded by a multi-layer perception (MLP). 
To reconstruct the geometry, these methods optimize the neural implicit function by modeling the view-dependent colors and minimizing the difference between the rendered and the input images. However, neural reconstruction methods still struggle to reconstruct reflective objects. 
Examples are provided in Fig.~\ref{fig:bell} (c). 
The reason is that the color function used in these methods only correlates the color with the view direction and surface geometry, rather than explicitly considering the underlying shading mechanism for reflections. Consequently, fitting the specular color variations in different view directions on the surface leads to erroneous geometry, even with higher frequency in positional encoding, or deeper and wider MLP networks.


To address the challenging surface reflections, we propose to explicitly incorporate the formulation of the rendering equation~\cite{kajiya1986rendering} into the neural reconstruction framework. 
The rendering equation enables us to consider the interaction between the surface Bidirectional Reflectance Distribution Function (BRDF)~\cite{nicodemus1965directional} and the environment lights. 
Since the appearances of reflective objects are strongly affected by the environment lights, the view-dependent specular reflection can be well-explained by the rendering equation.
With the explicit rendering function, the representation ability of the existing neural reconstruction framework is substantially enhanced to capture the high-frequency specular color variations, which significantly benefits the geometry reconstruction of reflective objects. 


Explicitly incorporating the rendering equation in a neural reconstruction framework is not trivial. 
Accurately evaluating the rendering equation on a surface point requires computing the integral of environment lights, which is intractable with unknown surface locations and unknown environment lights. 
In order to tractably evaluate the rendering equation, existing material estimation methods~\cite{munkberg2022extracting,zhang2021physg,boss2021nerd,boss2021neural,zhang2022modeling,verbin2022ref,zhang2021nerfactor,hasselgren2022shape} strongly rely on object masks to obtain a correct surface reconstruction and are mainly designed for material estimation of objects without strong specular reflections, which perform much worse on reflective objects as shown in Fig.~\ref{fig:bell} (d,e). 
Moreover, most of these methods further simplify the rendering process to only consider the lights from distant regions (\textit{direct lights})~\cite{munkberg2022extracting,zhang2021physg,boss2021nerd,boss2021neural,verbin2022ref}, which thus struggle to reconstruct surfaces illuminated by reflected lights from the object itself or nearby regions (\textit{indirect lights}). 
Although there are methods~\cite{zhang2022modeling,zhang2021nerfactor,hasselgren2022shape}  considering indirect lights in the rendering, they either require a reconstructed radiance field with known geometry~\cite{zhang2022modeling,zhang2021nerfactor} or only use very few ray samples to compute the lights~\cite{hasselgren2022shape}, which results in unstable convergence on reflective objects or additional dependence on object masks. 
Thus, considering both direct and indirect lights to correctly reconstruct the unknown surfaces of reflective objects is still challenging.
By incorporating the rendering equation in a neural reconstruction framework, we propose a method called NeRO for reconstructing both the geometry and the BRDF of reflective objects from only posed RGB images. 
The key component of NeRO is a novel light representation. 
In this light representation, we use two individual MLPs to encode the radiance of direct and indirect lights respectively and compute an occlusion probability to determine whether direct or indirect lights should be used in the rendering. 
Such a light representation efficiently accommodates both direct lights and indirect lights for accurate surface reconstruction of reflective objects.
Based on the proposed light representation, NeRO adopts a two-stage strategy for a tractable evaluation of the rendering equation in the neural reconstruction.
The first stage of NeRO employs a split-sum approximation and the integrated directional encoding~\cite{verbin2022ref} to evaluate the rendering equation, which produces accurate geometry reconstruction with compromised environment lights and surface BRDF estimation. Then, with the reconstructed geometry fixed, the second stage of NeRO improves the estimated BRDF by more accurately evaluating the rendering equation with Monte Carlo sampling. 
With the light representation and the two-stage design, the proposed method essentially extends the representational power of neural rendering methods on reflective objects, making it achieve the full potential of learning geometric surfaces.

To evaluate the performance of NeRO, we introduce a synthetic dataset and a real dataset, both of which contain reflective objects illuminated by complex environment lights. On both datasets, NeRO successfully reconstructs both the geometry and the surface BRDF of reflective objects, on which the baseline MVS methods and neural reconstruction methods fail. The output of our method is a triangular mesh with the estimated BRDF parameters, which can be easily used in downstream applications such as relighting.

\section{Related works}

\subsection{Multiview 3D reconstruction}

Multiview 3D reconstruction or Multiview Stereo (MVS) has been studied for decades~\cite{campbell2008using,strecha2006combined,furukawa2009accurate}. Traditional multiview reconstruction methods mainly rely on the multiview consistency of 3D points to build correspondences and estimate the depth values on different views~\cite{schoenberger2016mvs,bleyer2011patchmatch,hosni2012fast,richardt2010real,gallup2007real,barron2016fast,campbell2008using,strecha2006combined,furukawa2009accurate}. With the advances of deep learning techniques, many recent works~\cite{yao2018mvsnet,wang2021patchmatchnet,yan2020dense,yang2020cost,cheng2020deep} try to introduce neural networks to estimate correspondences for the MVS task, which demonstrates impressive reconstruction quality on widely-used benchmarks~\cite{jensen2014large,scharstein2002taxonomy,geiger2013vision}. In this paper, we aim to reconstruct reflective objects with strong specular reflections. The strong specular reflections violate the multiview consistency so these correspondence-based methods do not perform well on reflective objects.

\textbf{Neural surface reconstruction}. Neural rendering and neural representations~\cite{mildenhall2020nerf,tewari2020state,tewari2022advances,park2019deepsdf,sitzmann2019scene} have attracted much attention due to their strong representation ability and impressive improvements on the novel-view-synthesis task. DVR~\cite{niemeyer2020differentiable} first introduces the neural rendering and neural surface representation in the multiview reconstruction. IDR~\cite{yariv2020multiview} improves the reconstruction quality with the differentiable sphere tracing and the Eikonal regularization~\cite{gropp2020implicit}. UNISURF~\cite{oechsle2021unisurf}, VolSDF~\cite{yariv2021volume} and NeuS~\cite{wang2021neus} introduce the differentiable volume rendering in the multiview surface reconstruction with improved robustness and quality. 
Subsequent works improve the volume-rendering-based multiview reconstruction framework in various aspects, such as introducing Manhattan or normal priors~\cite{guo2022neural,wang2022neuris}, utilizing symmetry~\cite{zhang2021ners,insafutdinov2022snes}, extracting image features~\cite{darmon2022improving,long2022sparseneus}, improving fidelity~\cite{fu2022geo,wang2022hf} and efficiency~\cite{wu2022voxurf,li2022vox,wang2022neus2,sun2022neural,zhao2022human}. 
Similar to these works, we also follow the volume rendering framework for surface reconstruction but we focus on reconstructing reflective objects with strong specular reflections, an outstanding problem that has not been explored by existing neural reconstruction methods.

\subsection{Reflective object reconstruction}

Only a few works try to reconstruct reflective objects in the multiview stereo setting by using additional object masks~\cite{godard2015multi} or removing the reflections~\cite{wu2018specular}. Other than the uncontrolled multiview reconstruction, some works~\cite{roth2006specular,han2016mirror} resort to constrained settings with known specular flows~\cite{roth2006specular} or known environments~\cite{han2016mirror} for the reconstruction of ideal mirror-like objects. Some other works utilize additional ray information by encoding rays~\cite{tin20163d} or utilizing polarization images~\cite{rahmann2001reconstruction,kadambi2015polarized,dave2022pandora} to reconstruct the objects with specular reflections. \cite{whelan2018reconstructing} reconstructs mirror planes in a scene by utilizing reflected images of the scanner. These methods are limited to a relatively strict setting with specially-designed capturing devices. In contrast, we aim to directly reconstruct the reflective objects from posed multiview images, which can be easily captured using a cellphone camera. 

Some image-based rendering methods~\cite{rodriguez2020image,sinha2012image} are specially designed for glossy or reflective objects for the NVS task. NeRFRen~\cite{guo2022nerfren} reconstructs a neural density field of a scene with the existence of mirror-like planes. Neural Point Catacaustics~\cite{kopanas2022neural} applies a warp field to improve the rendering quality on reflective objects. Ref-NeRF~\cite{verbin2022ref} proposes integrated direction encoding (IDE) to improve the NVS quality on reflective materials. Our method incorporates the IDE in reconstructing reflective objects with a neural SDF for the surface reconstruction. A concurrent work ORCA~\cite{tiwary2022orca} extends to reconstruct the radiance field of a scene from the reflections on a glossy object, which also reconstructs the object in the pipeline. Since the target of ORCA is mainly to reconstruct the radiance field of the scene, it relies on object masks for the reconstruction of the reflective objects. In comparison, our method does not require object masks and our main target is to reconstruct the geometry and BRDF of the object. 

\subsection{BRDF estimation}
Estimating the surface BRDF from images is mainly based on the inverse rendering techniques~\cite{barron2014shape,nimier2019mitsuba}. Some methods~\cite{li2020inverse,gao2019deep,guo2020materialgan,li2018learning,wimbauer2022rendering,ye2022intrinsicnerf} rely on an object prior or a scene prior to directly estimating BRDF and lighting. Differentiable renderers~\cite{nimier2019mitsuba,liu2019soft,chen2019learning,chen2021dib,kato2018neural} allow direct optimization of the BRDF from image losses. To enable more accurate BRDF estimation, most methods~\cite{bi2020neural,bi2020deep,zhang2022iron,schmitt2020joint,nam2018practical,yang2022s3,yang2022ps,li2022neural,kuang2022neroic,li2022self,cheng2021multi} require multiple images of the object to be illuminated by different collocated flashlights. In this paper, we estimate the BRDF in a static scene with moving cameras, which is also the setting adopted by \cite{zhang2021physg,boss2021nerd,boss2021neural,munkberg2022extracting,hasselgren2022shape,zhang2021nerfactor,zhang2022modeling,deng2022dip,boss2022samurai}. Among these works, PhySG~\cite{zhang2021physg}, NeRD~\cite{boss2021nerd}, Neural-PIL~\cite{boss2021neural} and NDR~\cite{munkberg2022extracting} consider the interaction between the direct environment lights and the surfaces for the BRDF estimation. Subsequent works MII~\cite{zhang2022modeling}, NDRMC~\cite{hasselgren2022shape}, DIP~\cite{deng2022dip} and NeILF~\cite{yao2022neilf} add indirect lights, which improves the quality of estimated BRDF. These methods mainly aim to reconstruct the BRDF of common objects without too many specular reflections, which produces low-quality BRDF on reflective objects. Some other methods~\cite{duchene2015multi,shih2013data,yu2019inverserendernet,yu2020self,gao2020deferred,zheng2021neural,chen2022relighting4d,nestmeyer2020learning,philip2019multi,philip2021free,liu2021relighting,rudnev2022nerf,zhao2022factorized,lyu2022neural} are mainly targeted to the relighting task but not designed for reconstructing the surface geometry or the BRDF. NeILF~\cite{yao2022neilf} is the most similar work to the Stage II of our method, both of which fix the geometry to optimize BRDF with MC sampling. However, NeILF does not use importance sampling on the specular lobe and simply predicts lights from a position and a direction without considering the occlusions. In comparison, our method explicitly distinguishes direct and indirect lights and uses importance sampling on diffuse and specular lobes for a better BRDF estimation on reflective objects.





\section{Method}
\subsection{Overview}

Given a set of RGB images with known camera poses as input, our target is to reconstruct the surface and BRDF of the reflective object in the images. Note that our method does not require knowing the object masks or environment lights. 
The pipeline of NeRO consists of two stages. In Stage I (Sec.~\ref{sec:stage1}), we reconstruct the geometry of the reflective object by optimizing a neural SDF with the volume rendering, in which approximate direct and indirect lights are estimated to model the view-dependent specular colors. In Stage II (Sec.~\ref{sec:stage2}), we fix the object geometry and finetune the direct and indirect lights to compute an accurate BRDF of the reflective object. In the following, we begin with a brief review of NeuS~\cite{wang2021neus} and a micro-facet BRDF model~\cite{torrance1967theory,cook1982reflectance}.

\subsection{Preliminaries}
\label{sec:pre}

\textbf{NeuS~\cite{wang2021neus} for surface reconstruction}. We follow NeuS to represent the object surface by an SDF encoded by an MLP network $g_{\rm sdf}(\mathbf{x})$. The surface is the zero-level set with $\{\mathbf{x}\in \mathbb{R}^3|g_{\rm sdf}(\mathbf{x})=0\}$. Then, volume rendering~\cite{mildenhall2020nerf} is applied to render images from the neural SDF. Given a camera ray $\mathbf{o}+t\mathbf{v}$ emitting from the camera center $\mathbf{o}$ to the space along the direction $\mathbf{v}$, we sample $n$ points on the ray $\{\mathbf{p}_j=\mathbf{o}+t_j \mathbf{v}|t_j>0, t_{j-1}<t_j \}$. Then, the rendered color for this camera ray is computed by
\begin{equation}
    \mathbf{\hat{c}} = \sum_n w_j \mathbf{c}_j,
    \label{eq:vr}
\end{equation}
where $w_j$ is the weight for the $j$-th point, which is derived from the SDF value via the opaque density proposed in \cite{wang2021neus}. $\mathbf{c}_j$ is the color for this point, which is decoded from an MLP network by $\mathbf{c}_j = g_{\rm color}(\mathbf{p}_j, \mathbf{v})$ in NeuS. Then, by minimizing the difference between the rendered color $\hat{\mathbf{c}}$ and the input ground-truth color $\mathbf{c}$, the parameters of two MLP networks $g_{\rm sdf}$ and $g_{\rm color}$ are learned. The reconstructed surface is extracted from the zero-level set of $g_{\rm sdf}$.  
To enable the color function to correctly represent the specular colors on the reflective surfaces, NeRO replaces the color function of NeuS with the shading function using a Micro-facet BRDF.

\begin{figure}
    \centering
    \includegraphics[width=0.8\linewidth]{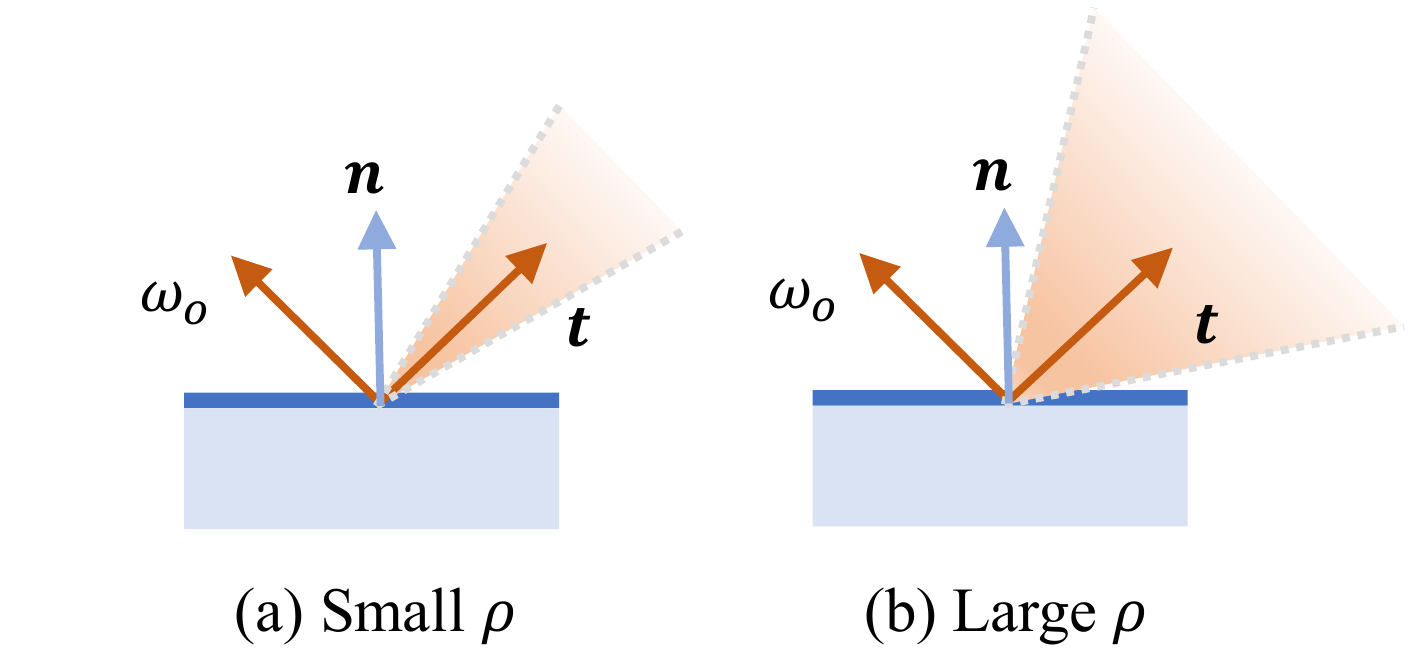}
    \caption{\textbf{Specular lobe} in Eq.~\ref{eq:app_spe} is determined by the roughness $\rho$ and the reflective direction $\mathbf{t}$. (a) A smooth surface with small $\rho$ has a smaller specular lobe while (b) a rougher surface with large $\rho$ has a larger specular lobe. }
    \label{fig:sp_lobe}
\end{figure}

\textbf{Micro-facet BRDF~\cite{cook1982reflectance}}. On the point $\mathbf{p}_j$, we compute its colors $\mathbf{c}_j$ by the rendering equation (the subscript $j$ is omitted in the following discussion for simplicity)
\begin{equation}
    \mathbf{c}(\omega_o) = \int_\Omega L(\omega_i) f(\omega_i, \omega_o) (\omega_i \cdot \mathbf{n}) d \omega_i,
    \label{eq:render}
\end{equation}
where $\omega_o=-\mathbf{v}$ is the outgoing view direction, $\mathbf{c}(\omega_o)$ is the output color $\mathbf{c}_j$ for this point $\mathbf{p}_j$ in the view direction $\omega_o$, $\mathbf{n}$ is the surface normal, $\omega_i$ is the input light direction on the upper half sphere $\Omega$, $f(\omega_i,\omega_o) \in [0,1]^3$ is the BRDF function, $L(\omega_i)\in [0,+\infty)^3$ is the radiance of incoming lights. In NeRO, the normal $\mathbf{n}$ is computed from the gradient of the SDF.
The BRDF function consists of a diffuse and a specular part
\begin{equation}
    f(\omega_i,\omega_o) = \underbrace{(1-m) \frac{\mathbf{a}}{\pi}}_{\rm diffuse} + \underbrace{\frac{D F G}{4 (\omega_i \cdot \mathbf{n}) (\omega_o \cdot \mathbf{n})}}_{\rm specular},
    \label{eq:brdf}
\end{equation}
where $m\in [0,1]$ is the metalness of the point, $1-m$ is the weight for the diffuse part, $\mathbf{a}\in[0,1]^3$ is the albedo color of the point, 
$D$ is the normal distribution function, $F$ is the Fresnel term and $G$ is the geometry term. $D$, $F$ and $G$ are all determined by the metalness $m$, the roughness $\rho\in[0,1]$ and the albedo $\mathbf{a}$. The expressions of $D$, $F$, and $G$ can be found in Sec.~\ref{sec:app_brdf} of the Appendix.
In summary, the BRDF of the point is specified by the metalness $m$, the roughness $\rho$, and the albedo $\mathbf{a}$, all of which are predicted by a material MLP $g_{\rm material}$ in NeRO, i.e., $[m,\rho,\mathbf{a}] = g_{\rm material}(\mathbf{p})$. 

Combining Eq.~\ref{eq:render} and Eq.~\ref{eq:brdf}, we have
\begin{equation}
    \mathbf{c}(\omega_o) = \mathbf{c}_{\rm diffuse} + \mathbf{c}_{\rm specular},
    \label{eq:combine}
\end{equation}
\begin{equation}
    \mathbf{c}_{\rm diffuse} = \int_\Omega (1-m) \frac{\mathbf{a}}{\pi} L(\omega_i) (\omega_i \cdot \mathbf{n}) d \omega_i,
    \label{eq:diffuse}
\end{equation}
\begin{equation}
    \mathbf{c}_{\rm specular} = \int_\Omega \frac{D F G}{4 (\omega_i \cdot \mathbf{n}) (\omega_o \cdot \mathbf{n})} L(\omega_i) (\omega_i \cdot \mathbf{n}) d \omega_i.
    \label{eq:specular}
\end{equation}
As explained before, accurately evaluating the integrals of Eq.~\ref{eq:diffuse} and Eq.~\ref{eq:specular} for every sample point in the volume rendering is intractable. 
Therefore, we propose a two-step framework to approximately compute these two integrals. In the first stage, our priority is to faithfully reconstruct the geometric surface.

\subsection{Stage I: Geometry reconstruction}
\label{sec:stage1}

In order to reconstruct surfaces of reflective objects, we adopt the same neural SDF representation and the volume rendering algorithm (Eq.~\ref{eq:vr}) as NeuS~\cite{wang2021neus} but with a different color function. In NeRO, we predict a metalness $m$, a roughness $\rho$ and an albedo $\mathbf{a}$ to compute a color $\mathbf{c}_j$ (i.e., $\mathbf{c}(\omega_o)$) using the micro-facet BRDF (Eq.~\ref{eq:combine}-\ref{eq:specular}). 
To make the computation tractable in the volume rendering of NeuS, we adopt the split-sum approximation~\cite{karis2013real}, which separates the integral of the product of lights and BRDFs into two individual integrals.

\begin{figure}
    \centering
    \includegraphics[width=0.7\linewidth]{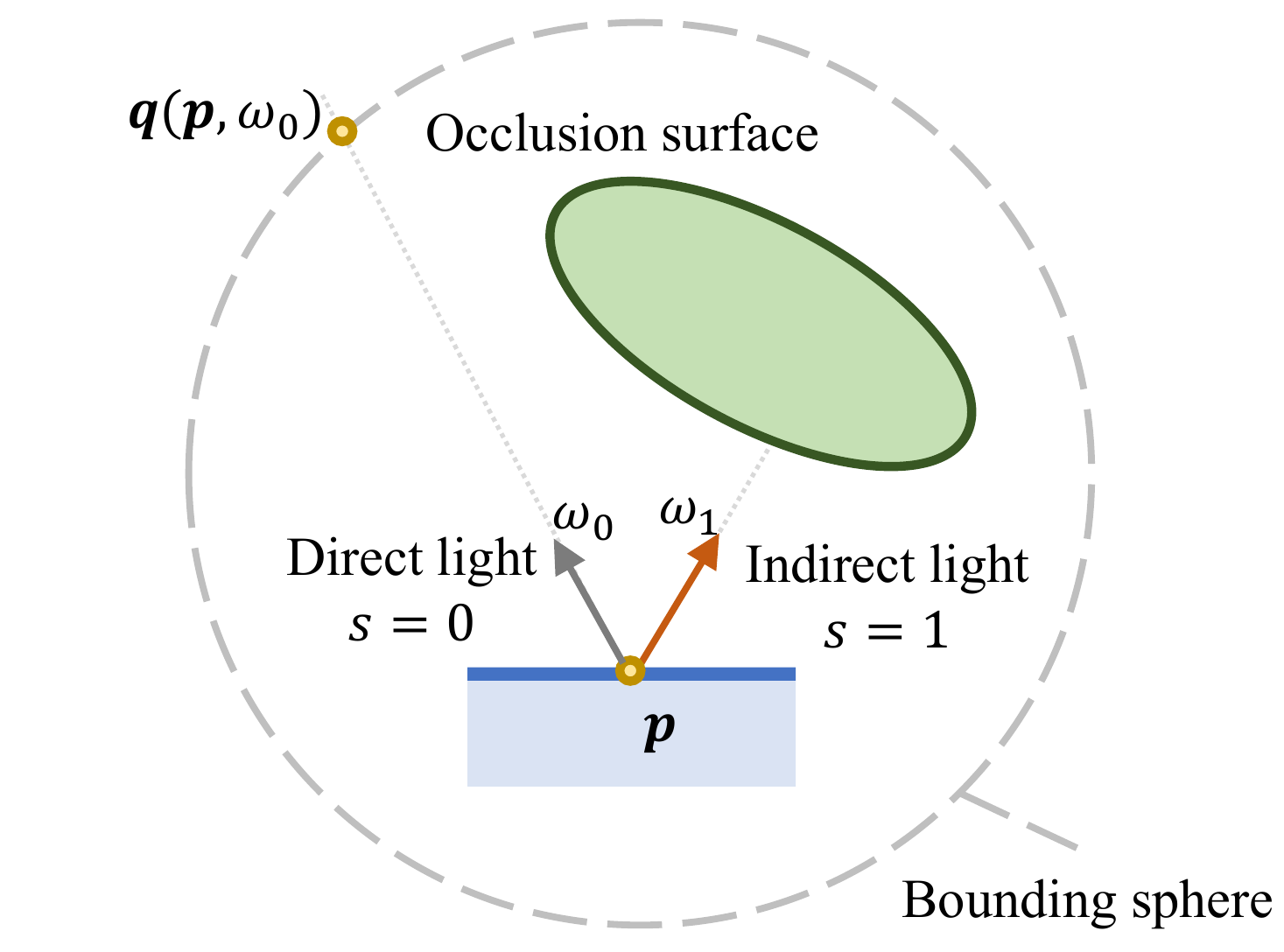}
    \caption{\textbf{Direct and indirect lights for a point $\mathbf{p}$.} The direct light in the direction $\omega_0$ is not occluded while the indirect light in the direction $\omega_1$ is occluded by surfaces inside the unit sphere. $s$ is the occlusion probability. $\mathbf{q}(\mathbf{p},\omega)$ is the intersection point on the bounding sphere of the ray emitting from $\mathbf{p}$ along the direction $\omega$.}
    \label{fig:light_rep}
\end{figure}

\textbf{Split-sum approximation}. For the specular color in Eq.~\ref{eq:specular}, we follow \cite{karis2013real} to approximate it with
\begin{equation}
    \mathbf{c}_{\rm specular} \approx \underbrace{\int_{\Omega} L(\omega_i) D(\rho, \mathbf{t}) d\omega_i}_{L_{\rm specular}} \cdot \underbrace{\int_\Omega \frac{D F G}{4(\omega_o \cdot \mathbf{n})} d\omega_i}_{M_{\rm specular}},
    \label{eq:app_spe}
\end{equation}
where $L_{\rm specular}$ is the integral of lights on the normal distribution function $D(\rho, \mathbf{t})\in [0,1]$ (also called the \textit{specular lobe}) as shown in Fig.~\ref{fig:sp_lobe}, $\mathbf{t}$ is the reflective direction, $M_{\rm specular}$ denotes the integral of BRDF. Note that a rougher surface has a larger specular lobe while a smoother surface has a smaller lobe. 
The integral of BRDF can be directly computed by $M_{\rm specular}=((1-m)*0.04+m*\mathbf{a})*F_1+F_2$, where $F_1$ and $F_2$ are two pre-computed scalars depending on the roughness $\rho$, the view direction $\omega_o$ and the normal $\mathbf{n}$ as introduced in more details in Sec.~\ref{sec:app_brdf} in Appendix. 
The diffuse color in Eq.~\ref{eq:diffuse} can be written as
\begin{equation}
    \mathbf{c}_{\rm diffuse} = \mathbf{a}(1-m) \underbrace{\int_\Omega L(\omega_i) \frac{\omega_i \cdot \mathbf{n}}{\pi}d\omega_i}_{L_{\rm diffuse}},
    \label{eq:app_diff}
\end{equation}
where we use $L_{\rm diffuse}$ to denote the diffuse light integral.

Split-sum has already been proven to be a good approximation and is widely used in real-time rendering. 
With predicted material parameters albedo $\mathbf{a}$, roughness $\rho$ and metalness $m$ from the material MLP, the only two unknowns in Eq.~\ref{eq:app_diff} and Eq.~\ref{eq:app_spe} are light integrals $L_{\rm diffuse}$ and $L_{\rm specular}$. 
However, to compute the light integrals, we do not prefilter environment lights as previous methods~\cite{boss2021neural,munkberg2022extracting} but use the integrated directional encoding~\cite{verbin2022ref}.
In the following, we first introduce our light representation $L(\omega_i)$.

\textbf{Light representation}. 
In NeRO, we define a \textit{bounding sphere} around the object to build the neural SDF. 
Since we only aim to reconstruct the surfaces inside the bounding sphere, we call all lights coming from the outside of the bounding sphere as \textit{direct lights} while we name lights reflected by the surfaces inside the bounding sphere as \textit{indirect lights}, which is shown in Fig.~\ref{fig:light_rep}.
Then, we represent the light $L(\omega_i)$ by 
\begin{equation}
    L(\omega_i) = [1-s(\omega_i)]g_{\rm direct} (SH(\omega_i)) + s(\omega_i) g_{\rm indirect} (SH(\omega_i), \mathbf{p}),
    \label{eq:light}
\end{equation}
where $g_{\rm direct}$ and $g_{\rm indirect}$ are two MLPs for the direct lights and the indirect lights respectively, $s(\omega_i)\in [0,1]$ is the occlusion probability that the ray from the point $\mathbf{p}$ to the direction $\omega_i$ is occluded by the surfaces inside the bounding sphere, $SH$ is the directional encoding using spherical harmonics (SH) as basis functions. Note that $s(\omega_i)=g_{\rm occ}(SH(\omega_i), \mathbf{p})$ is also predicted by an MLP $g_{\rm occ}$. 

\textbf{Motivation of the light representation design}. The direct light $g_{\rm direct}(\omega_i)$ only depends on the direction $\omega_i$ so that all points are illuminated by the same direct environment light. This provides a strong global prior to explaining the view-dependent colors of the reflective object. The indirect light $g_{\rm indirect}(\omega_i,\mathbf{p})$ additionally takes the point position $\mathbf{p}$ as input because indirect lights vary in the space. Additional discussion about the light representation is provided in Sec.~\ref{sec:app_light_rep} of the Appendix.


\begin{figure}
    \centering
    \includegraphics[width=\linewidth]{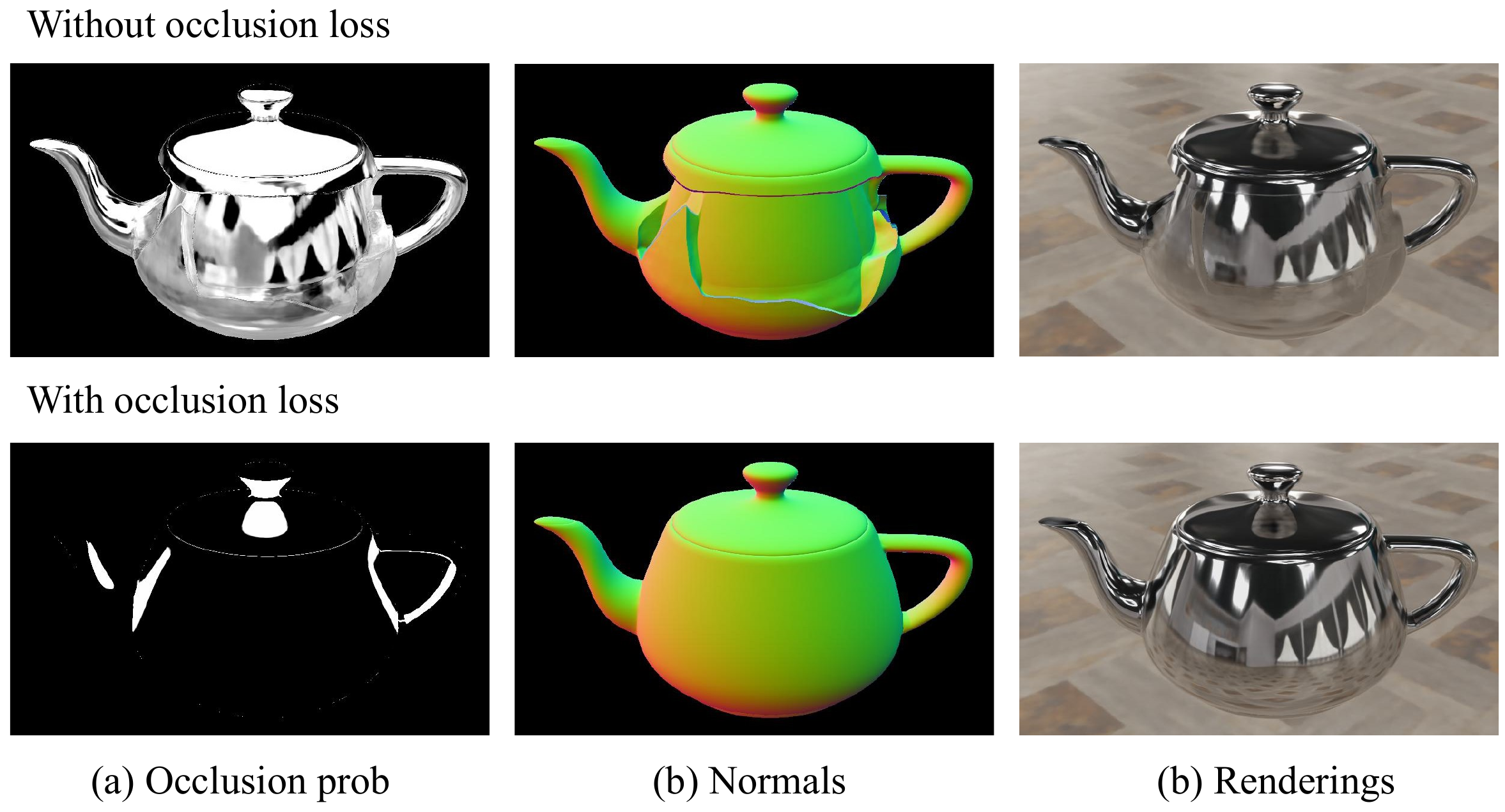}
    \caption{\textbf{Effects of occlusion loss}. (Top) Without $\ell_{\rm occ}$, the occlusion probability predicted by $g_{\rm occ}$ will be completely inconsistent with the reconstructed geometry and causes incorrect reconstruction. (Bottom) With $\ell_{\rm occ}$, the predicted occlusion probability is accurate and the reconstruction is correct.}
    \label{fig:occ_loss}
\end{figure}
\textbf{Light integral approximation}. We use the integrated directional encoding to approximate the light integrals $L_{\rm specular}$ and $L_{\rm diffuse}$. The specular light integral is
\begin{equation}
\begin{split}
    L_{\rm specular} \approx & [1-s(\mathbf{t})] \int_\Omega g_{\rm direct}(SH(\omega_i))D(\rho,\mathbf{t})d\omega_i + \\
                       & s(\mathbf{t}) \int_\Omega g_{\rm indirect} (SH(\omega_i), \mathbf{p}) D(\rho,\mathbf{t}) d\omega_i \\
               \approx & [1-s(\mathbf{t})] g_{\rm direct}( \int_\Omega SH(\omega_i) D(\rho,\mathbf{t})d\omega_i) + \\  
                             & s(\mathbf{t}) g_{\rm indirect} (\int_\Omega SH(\omega_i) D(\rho,\mathbf{t})d\omega_i, \mathbf{p}). \\
\end{split}
\label{eq:ide}
\end{equation}
In the first approximation, we use the occlusion probability $s(\mathbf{t})$ of the reflective direction $\mathbf{t}$ to replace occlusion probabilities $s(\omega_i)$ of different rays. In the second approximation, we exchange the order of the MLP and the integral. Discussion about the rationale of two approximations is provided in Sec.~\ref{sec:app_light_int}. 
With Eq.~\ref{eq:ide}, we only need to evaluate the MLP networks $g_{\rm direct}$ and $g_{\rm indirect}$ once on the integrated directional encoding $\int_\Omega SH(\omega_i) D(\rho,\mathbf{t})d\omega_i$. By choosing the normal distribution function $D$ to be a von Mises–Fisher (vMF) distribution (Gaussian distribution on a sphere), Ref-NeRF~\cite{verbin2022ref} has shown that $\int_\Omega SH(\omega_i) D(\rho,\mathbf{t}) d\omega_i$ has an approximated closed-form solution. In this case, we use this closed-form solution here to approximate the integral of lights. 

Similarly, for $L_{\rm diffuse}$, the cosine lobe $\frac{\omega_i \cdot \mathbf{n}}{\pi}$ is also a probability distribution, which can be approximated by
\begin{equation}
    \frac{\omega_i \cdot \mathbf{n}}{\pi} \approx D(1.0, \mathbf{n}).
    \label{eq:app_diff_light}
\end{equation}
Thus, we can also compute the diffuse light integral as Eq.~\ref{eq:ide}.
With the integrals of lights, we are able to compute the diffuse colors Eq.~\ref{eq:diffuse} and specular colors Eq.~\ref{eq:specular} and composite them as the color for each sample point. Note that both the split-sum approximation and the light integral approximation are only used in Stage I to enable tractable computation and will be replaced by more accurate Monte Carlo sampling in Stage II.

\textbf{Occlusion loss}. In the light representation, we use an occlusion probability $s$ predicted by an MLP $g_{\rm occ}$ to determine whether direct lights or indirect lights will be used in rendering. However, as shown in Fig.~\ref{fig:occ_loss}, if we put no constraint on the occlusion probability $s$ and just let the network $g_{\rm occ}$ learn $s$ from rendering loss, the predicted occlusion probability will be completely inconsistent with the reconstructed geometry and cause unstable convergence. 
Thus, we use the neural SDF to constrain the predicted occlusion probability.
Given the ray emitting from a sample point $\mathbf{p}$ to its reflective direction $\mathbf{t}$, we compute its occlusion probability $s_{\rm march}$ by ray-marching in the neural SDF $g_{\rm SDF}$ and enforce the consistency between the computed probability $s_{\rm march}$ and the predicted probability $s$ with
\begin{equation}
    \ell_{occ} = \|s_{\rm march} - s\|_1,
    \label{eq:occ_loss}
\end{equation}
where $\ell_{occ}$ is the loss for this occlusion probability regularization. 

\begin{figure}
    \centering
    \includegraphics[width=\linewidth]{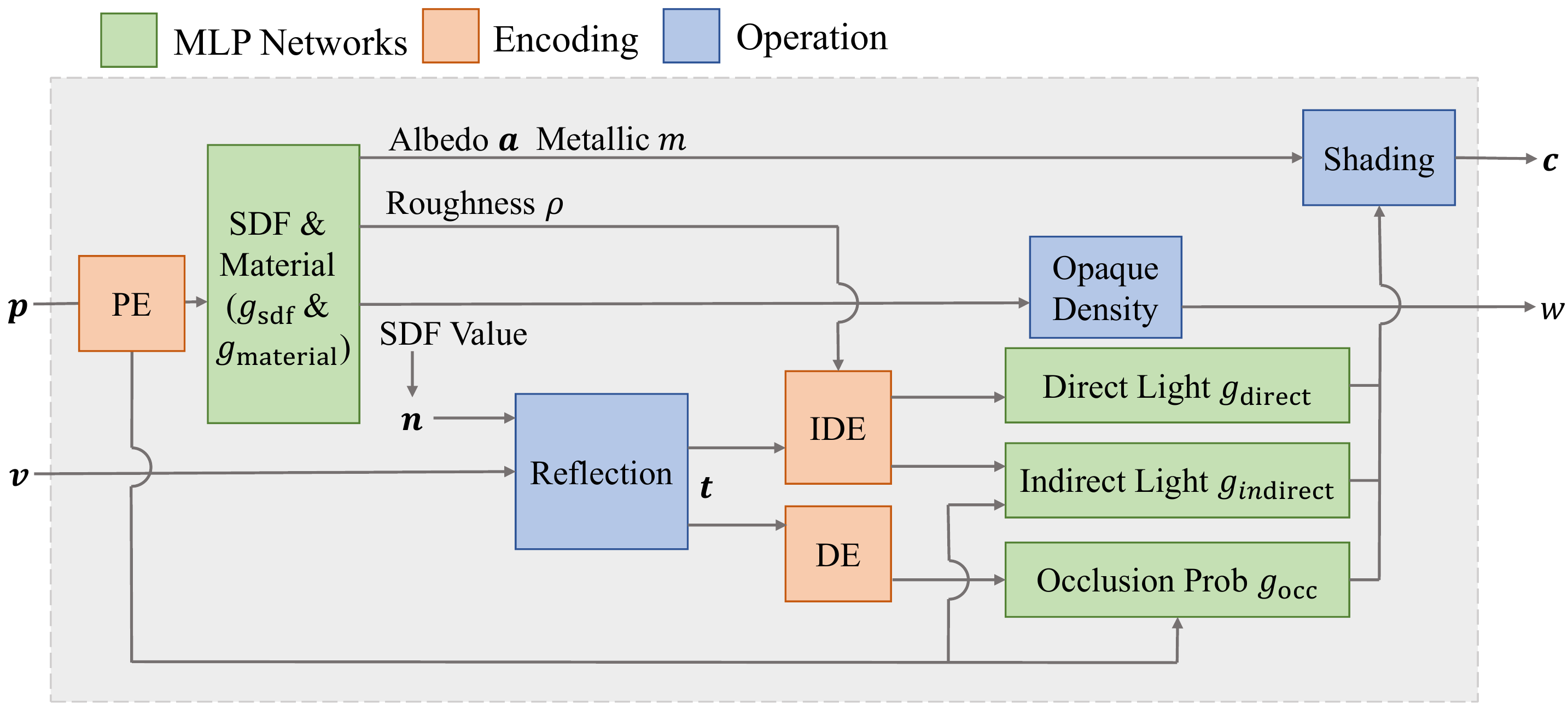}
    \caption{\textbf{Architecture of networks in Stage I.} ``PE''is positional encoding~\cite{mildenhall2020nerf} while ``IDE'' and ``DE'' are integrated direction encoding~\cite{verbin2022ref} and vanilla directional encoding, respectively.}
    \label{fig:arch1}
\end{figure}

\textbf{Training Losses}. Based on the volume rendering Eq.~\ref{eq:vr}, we compute the color for the camera ray and compute the Charbonier loss~\cite{barron2022mip,charbonnier1994two} between the rendered color and the input ground-truth color as the rendering loss $\ell_{\rm render}$. Meanwhile, we observe that the first few training steps for the SDF are not stable, which either extremely enlarges the surface or squashes the surface too small. A stabilization regularization loss $\ell_{\rm stable}$ is applied for the first 1k steps. (This is discussed in detail in Sec.~\ref{sec:app_stab}.) In summary, the final loss is
\begin{equation}
    \ell = \ell_{\rm render}+\lambda_{\rm eikonal}\ell_{\rm eikonal}+\lambda_{\rm occ}\ell_{occ}+\mathbbm{1}(\rm step<1000)\ell_{\rm stable},
\end{equation}
where we also adopt the Eikonal loss~\cite{gropp2020implicit} to regularize the norms of SDF gradients to be 1, $\mathbbm{1}$ is the indicator function, $\lambda_{eikonal}$ and $\lambda_{\rm occ}$ are two predefined scalars. The overview of the network architecture is shown in Fig.~\ref{fig:arch1}.

\begin{figure}
    \centering
    \includegraphics[width=\linewidth]{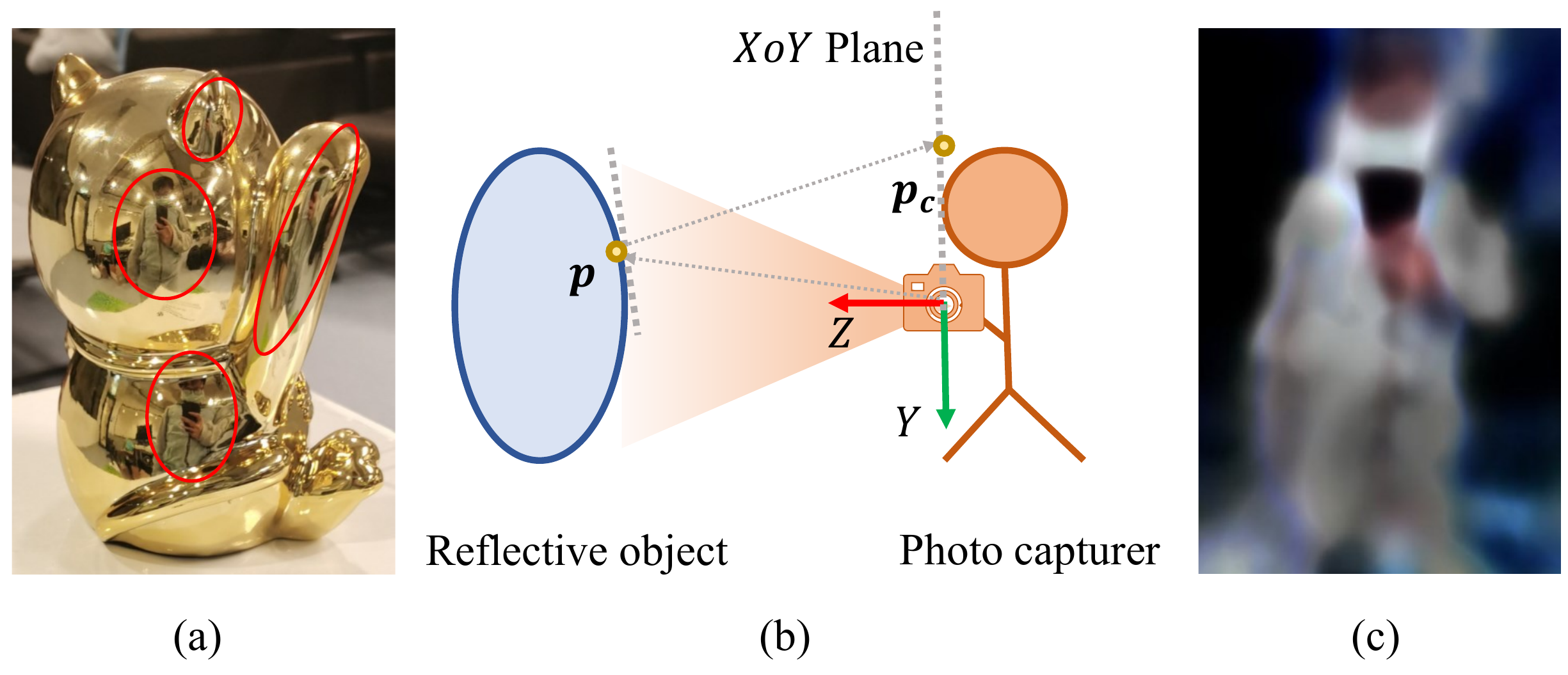}
    \caption{(a) A reflective object shows the reflections of the photo capturer. (b) We build a 2D NeRF on the $XoY$ plane to model the lights occluded by the photo capturer. (c) The estimated 2D radiance field of the capturer.}
    \label{fig:ref_human}
\end{figure}

\textbf{Reflection of the capturer}. We assume a static illumination environment in our model. However, in reality, there is always a person holding a camera to capture images around a reflective object. The moving person will be visible in the reflection of the object, thus violating the assumption of static illumination, as shown in the red circle of Fig.~\ref{fig:ref_human} (a). Since the photo capturer is relatively static to the camera, we build a 2D NeRF on the $XoY$ plane in the camera coordinate system as shown in Fig.~\ref{fig:ref_human} (b). In the computation of the direct light $g_{\rm direct}$ in Eq.~\ref{eq:light}, we additionally check whether the ray hits the $XoY$ plane of the camera coordinate system or not. If a hit point $\mathbf{p}_c$ exists, we will use an MLP $g_{\rm camera}$ to compute an alpha value $\alpha_{\rm camera}$ and a color $\mathbf{c}_{\rm camera}$ by 
\begin{equation}
    [\alpha_{\rm camera},\mathbf{c}_{\rm camera}]=g_{\rm camera}(\mathbf{p}_c),
    \label{eq:cap}
\end{equation}
where $\alpha_{\rm camera}$ indicates whether the ray is occluded by the capturer or not while the $\mathbf{c}_{\rm camera}$ represents the color of the capturer on this point.
Then, the direct light is $(1-\alpha_{\rm camera})g_{\rm direct}(\omega_i) + \alpha_{\rm camera} \mathbf{c}_{\rm camera}$.

\subsection{Stage II: BRDF estimation}
\label{sec:stage2}

So far, after Stage I, we have faithfully reconstructed the geometry of the reflective object but obtained only a rough BRDF estimation, which needs to be further refined. In Stage II, we aim to accurately evaluate the rendering equation so as to precisely estimate the surface BRDF i.e. metalness $m$, albedo $\mathbf{a}$, and $\rho$. With the fixed geometry from Stage I, we only need to evaluate the rendering equation on surface points. Thus, it is now feasible to apply Monte Carlo sampling to compute the diffuse colors in Eq.~\ref{eq:diffuse} and the specular colors in Eq.~\ref{eq:specular}. In the MC sampling, we conduct importance sampling on both the diffuse lobe and the specular lobe as follows.

\textbf{Importance sampling}. In Monte Carlo sampling, the diffuse color $\mathbf{c}_{\rm diffuse}$ is computed by sampling $N_{d}$ rays with a cosine-weighted hemisphere probability
\begin{equation}
    \mathbf{c}_{\rm diffuse} = \frac{1}{N_{d}}\sum_i^{N_{d}} (1-m) \mathbf{a} L(\omega_i),
    \label{eq:diff_mc}
\end{equation}
where $i$ is $i$-th sample ray and $\omega_i$ is the direction of this sample ray. 
For the specular color $\mathbf{c}_{\rm specular}$, we apply the GGX distribution as normal distribution $D$.
Then, the specular color $\mathbf{c}_{\rm specular}$ is computed by sampling $N_{s}$ rays with DDX distribution~\cite{cook1982reflectance}
\begin{equation}
    \mathbf{c}_{\rm specular} = \frac{1}{N_s} \sum_{i}^{N_{s}} \frac{FG(\omega_o \cdot \mathbf{h})}{(\mathbf{n}\cdot \mathbf{h})(\mathbf{n}\cdot\mathbf{\omega_o})} L(\omega_i),
    \label{eq:spe_mc}
\end{equation}
where $\mathbf{h}$ is the half-way vector between $\omega_i$ and $\omega_o$. To evaluate Eq.~\ref{eq:diff_mc} and Eq.~\ref{eq:spe_mc}, we still use the same material MLP $[m,\rho,\mathbf{a}]=g_{\rm material}$ as Stage I to compute the metalness $m$, roughness $\rho$ and albedo $\mathbf{a}$. The light representation $L(\omega_i)$ in Stage II is also Eq.~\ref{eq:light} with the reflected lights of the capturer in Eq.~\ref{eq:cap}, which is the same as Stage I. Since the geometry is fixed, we directly compute the occlusion probability $s$ by tracing rays in the given geometry instead of predicting it from the MLP $g_{\rm occ}$. Meanwhile, for real data, we add the intersection point on the bounding sphere $\mathbf{q}_{\mathbf{p}, \omega}$ of the ray emitting from $\mathbf{p}$ along the direction $\omega$, as shown in Fig.~\ref{fig:light_rep}, as an additional input to the direct light MLP $g_{\rm direct}$. More details are discussed in Sec.~\ref{sec:app_light_rep}


\textbf{Regularization terms}. 
We follow previous works~\cite{hasselgren2022shape,munkberg2022extracting} to impose two regularization terms on the final loss. The first is a smoothness regularization $\ell_{\rm smooth}$
\begin{equation}
    \ell_{\rm smooth} = \| g_{\rm material} (\mathbf{p}) - g_{\rm material}(\mathbf{p}+\epsilon)\|_2,
\end{equation}
where $\epsilon=5e-3$. The $\ell_{\rm smooth}$ makes the predicted materials (roughness, metallic, and albedo) more smooth in the space. Furthermore, a light regularization $\ell_{\rm light}$ is imposed to make the diffuse lights $L_{\rm diffuse}=\frac{1}{N_d}\sum_i L(\omega_i)$ to be neutral white lighting. We compute the mean of RGB values of the diffuse light and minimize the difference between the RGB values and their mean,
\begin{equation}
    \ell_{\rm light} = \sum_c^3 ([L_{\rm diffuse}]_c - \frac{1}{3}\sum_c^3 [L_{\rm diffuse}]_c),
\end{equation}
where $c$ is the index of RGB channel of $L_{\rm diffuse}$. Along with the rendering loss, the final loss for Stage II is
\begin{equation}
    \ell = \ell_{\rm render} + \lambda_{\rm smooth} \ell_{\rm smooth} + \lambda_{\rm light} \ell_{\rm light},
\end{equation}
where $\lambda_{\rm smooth}$ and $\lambda_{\rm light}$ are two predefined scalars.

\section{Experiments}

\subsection{Experiment protocol}

\begin{figure}
    \centering
    \setlength\tabcolsep{1pt}
    \begin{tabular}{cccc}
    \includegraphics[width=0.11\textwidth]{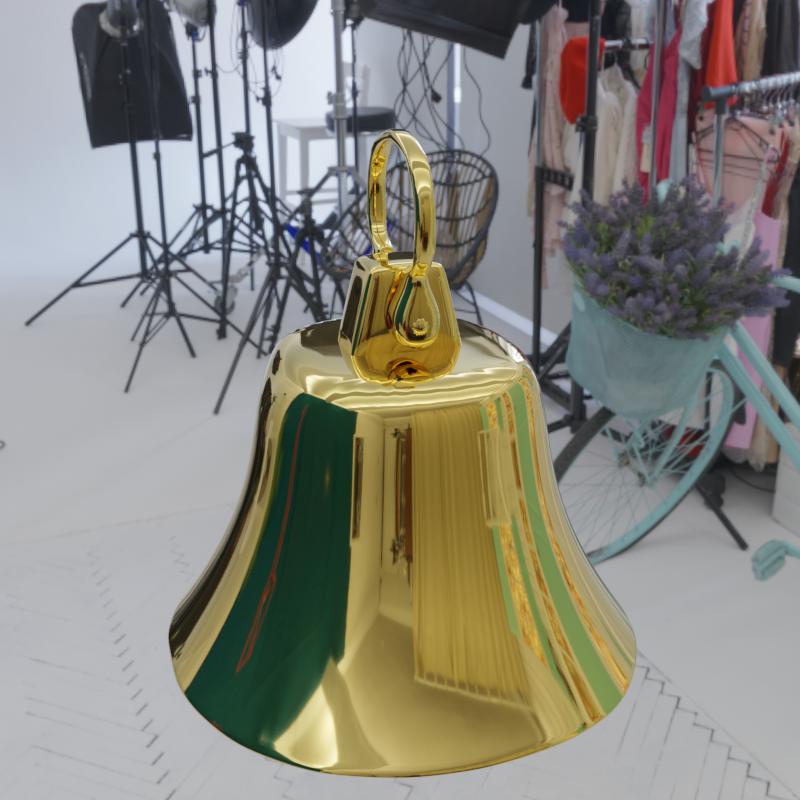} & 
    \includegraphics[width=0.11\textwidth]{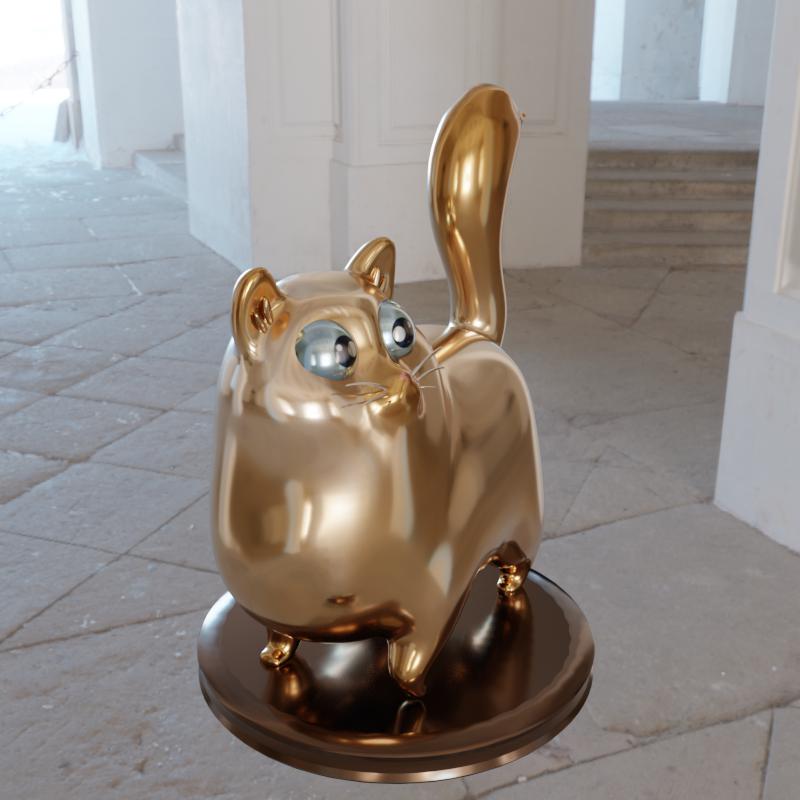} & 
    \includegraphics[width=0.11\textwidth]{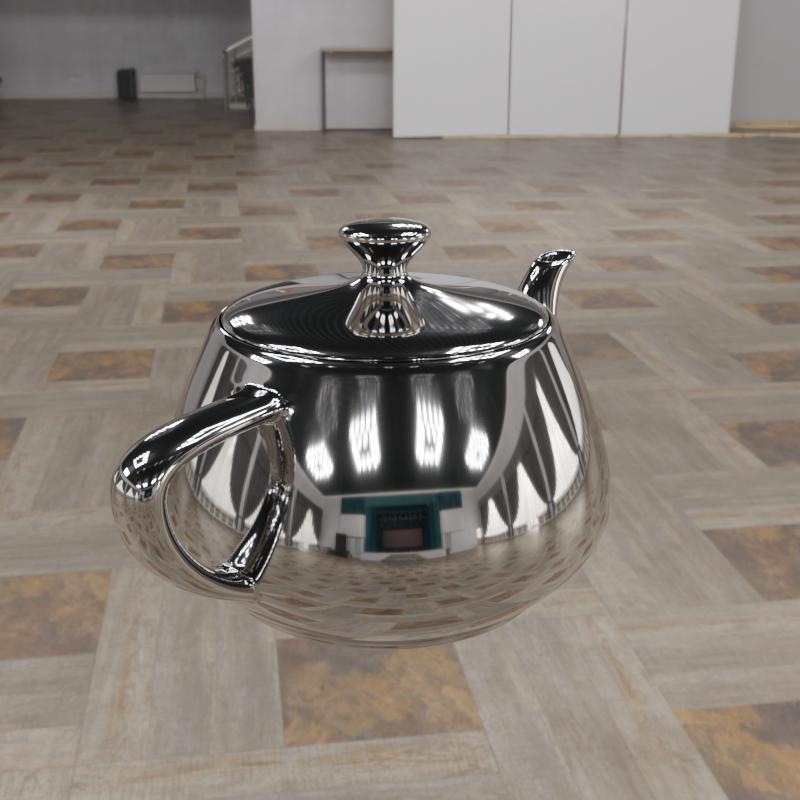} & 
    \includegraphics[width=0.11\textwidth]{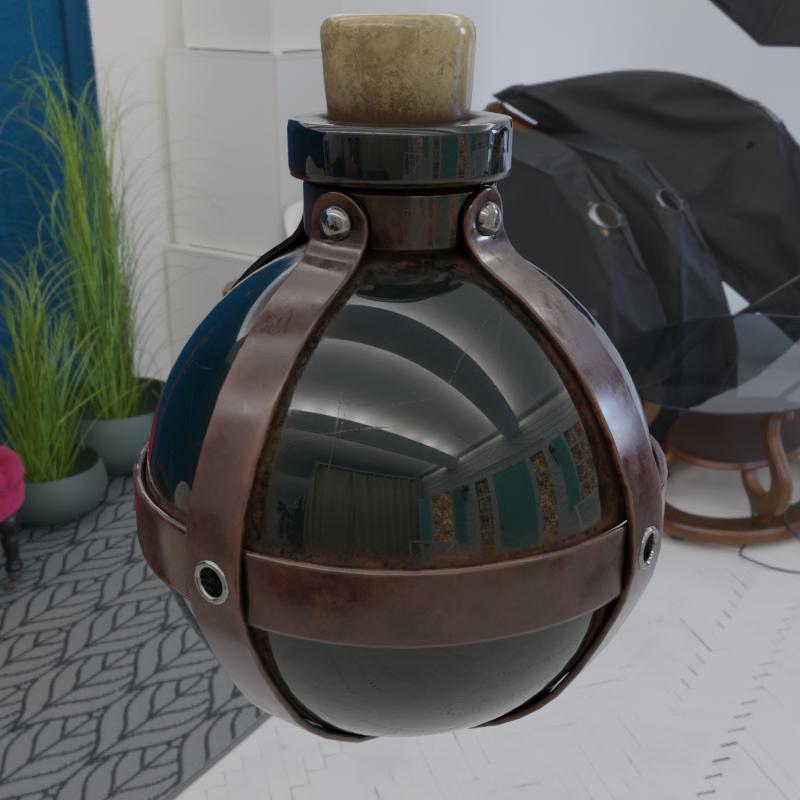} \\
    Bell & Cat & Teapot & Potion \\
    \includegraphics[width=0.11\textwidth]{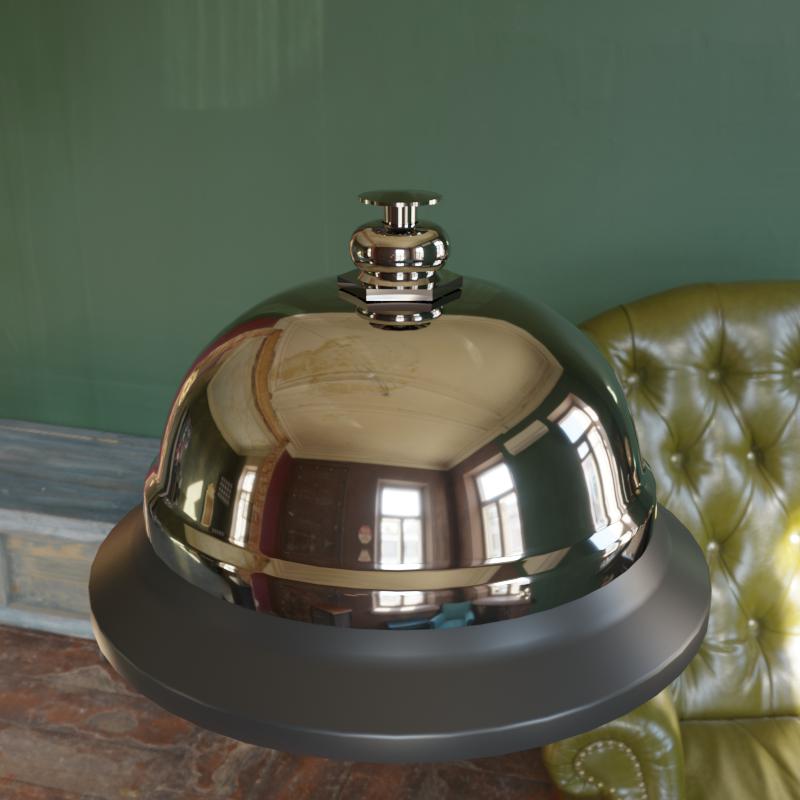} & 
    \includegraphics[width=0.11\textwidth]{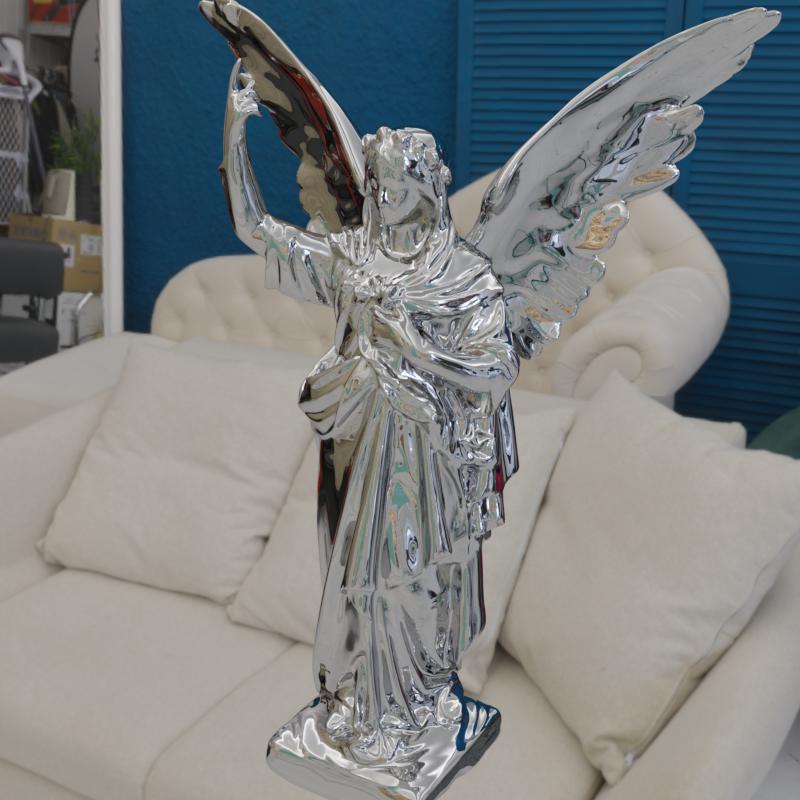} & 
    \includegraphics[width=0.11\textwidth]{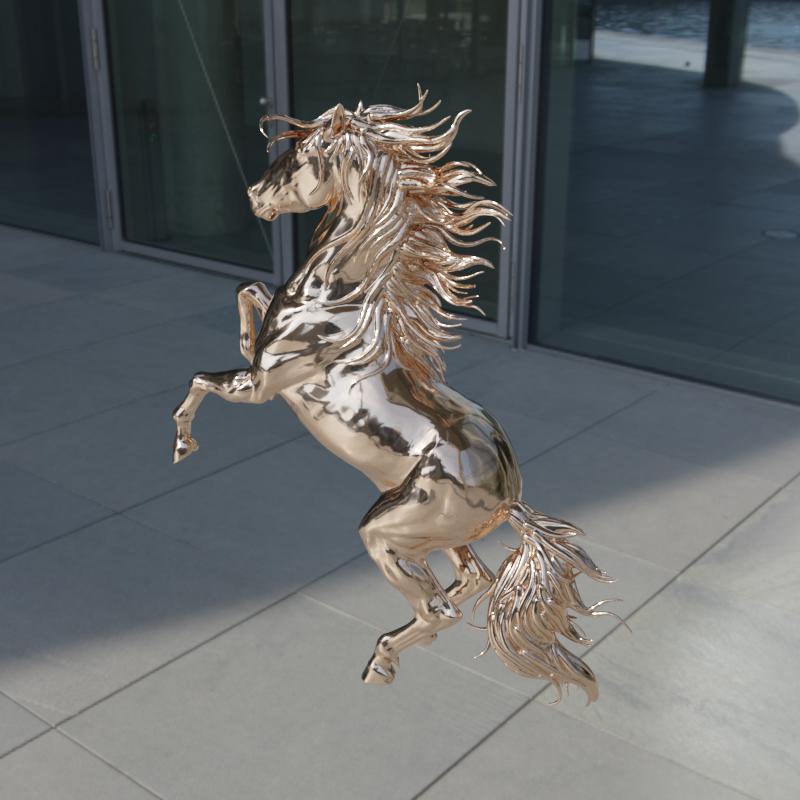} & 
    \includegraphics[width=0.11\textwidth]{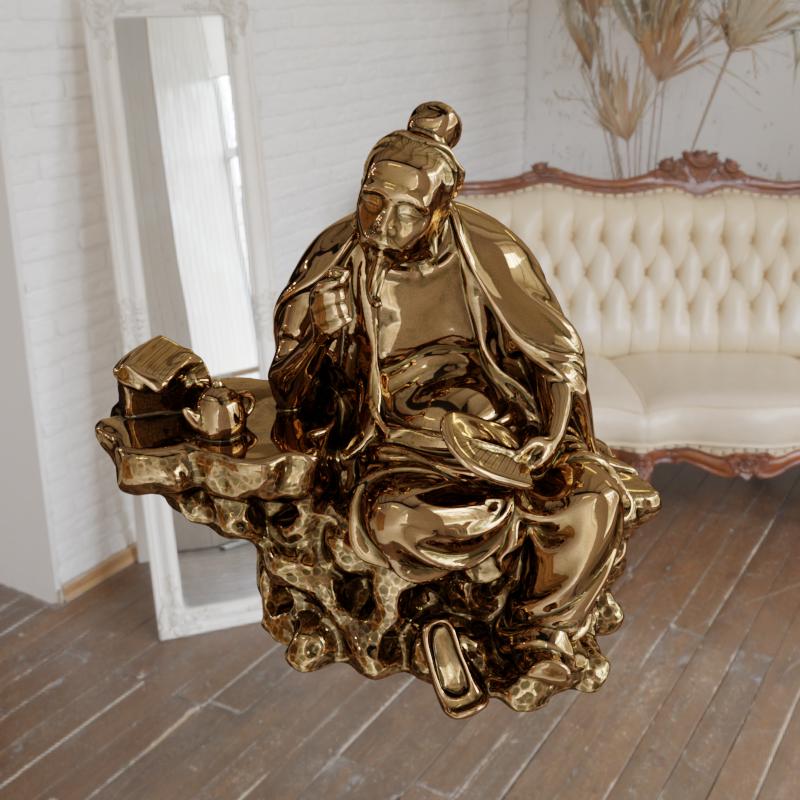} \\
    TBell & Angel & Horse & Luyu \\
    \end{tabular}
    \caption{\textbf{Objects from the Glossy-Blender dataset}. ``TBell'' means ``Table Bell'' and ``Luyu'' is a Chinese tea master worshiped as the Sage of Tea.}
    \label{fig:syn_example}
\end{figure}

\begin{figure}
    \centering
    \includegraphics[width=\linewidth]{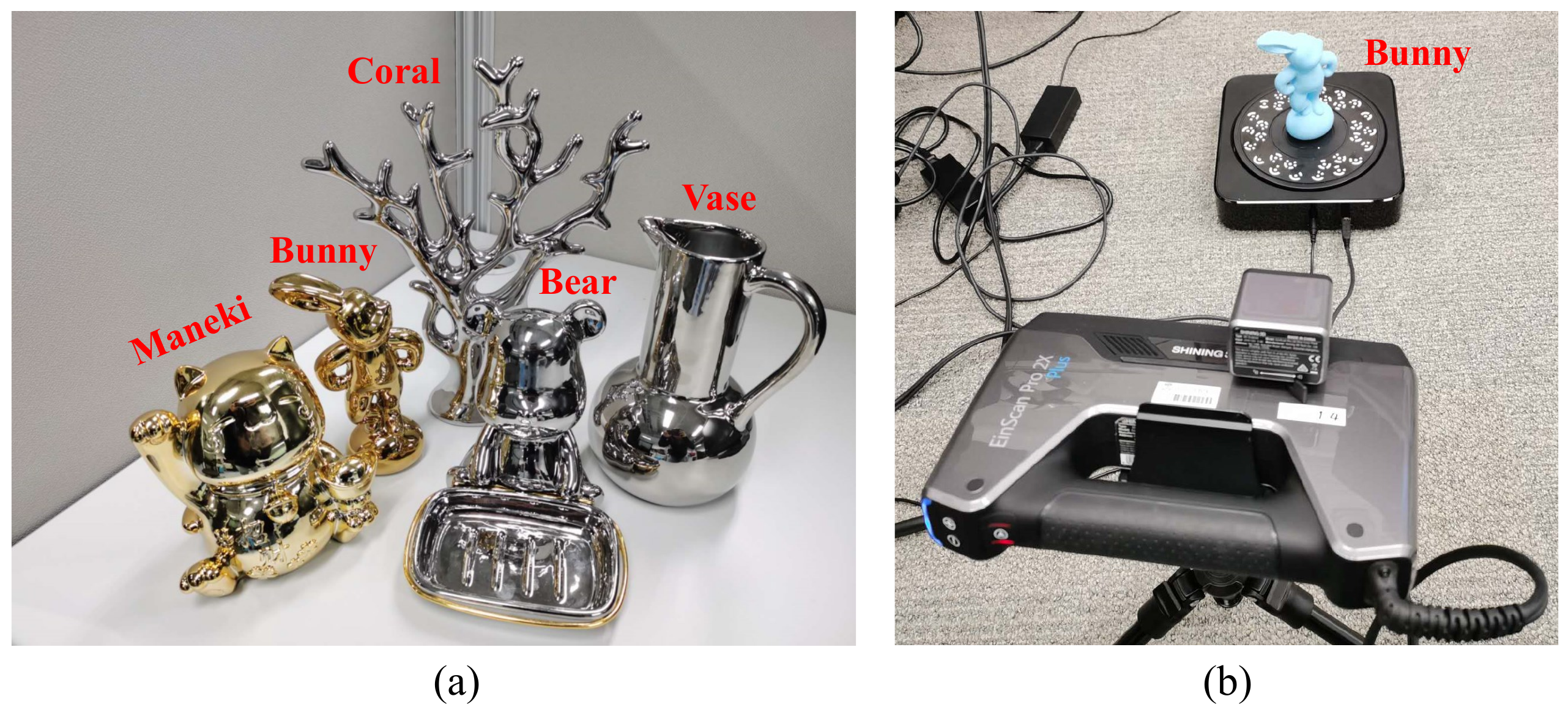}
    \caption{\textbf{The Glossy-Real dataset}. (a) Objects and their names. (b) We paint non-reflective substances on objects and scan objects with an EinScan Pro 2X scanner.
    \vspace{-5pt}}
    \label{fig:real_example}
\end{figure}

\textbf{Datasets}. To evaluate the performance of NeRO, we propose a synthetic dataset called the Glossy-Blender dataset and a real dataset called the Glossy-Real dataset. The Glossy-Blender dataset consists of 8 objects with low roughness and strong reflective appearances. The copyright information of these objects are included in Sec.~\ref{sec:app_copy}. For each object, we uniformly rendered 128 images of resolution $800\times 800$ around the object with the Cycles renderer in Blender. In rendering, we randomly select indoor HDR images from PolyHeaven~\cite{polyheaven} as sources of environment lights. Among 8 objects, Bell, Cat, Teapot, Potion, and TBell have large smooth surfaces with strong specular effects while Angel, Luyu, and Horse contain more complex geometry with small reflective fragments, as shown in Fig.~\ref{fig:syn_example}. The Glossy-Real dataset contains 5 objects (Coral, Maneki, Bear, Bunny and Vase), as shown in Fig.~\ref{fig:real_example} (a). We capture about 100-130 images around each object and use COLMAP~\cite{schoenberger2016sfm} to track the camera poses for all images. To enable robust camera tracking for COLMAP, we place the object on a calibration board with strong textures. All images are captured by a cellphone camera with a resolution of $1024\times 768$. To get the ground-truth surfaces, we use the structure light-based RGBD sensor EinScan Pro 2X to scan these objects. Since the scanner is unable to reconstruct reflective objects, we manually paint non-reflective substances on these objects before scanning, as shown in Fig.~\ref{fig:real_example} (b). Sec.~\ref{sec:stats} includes the detailed dataset statistics.

\begin{figure}
    \centering
    \includegraphics[width=\linewidth]{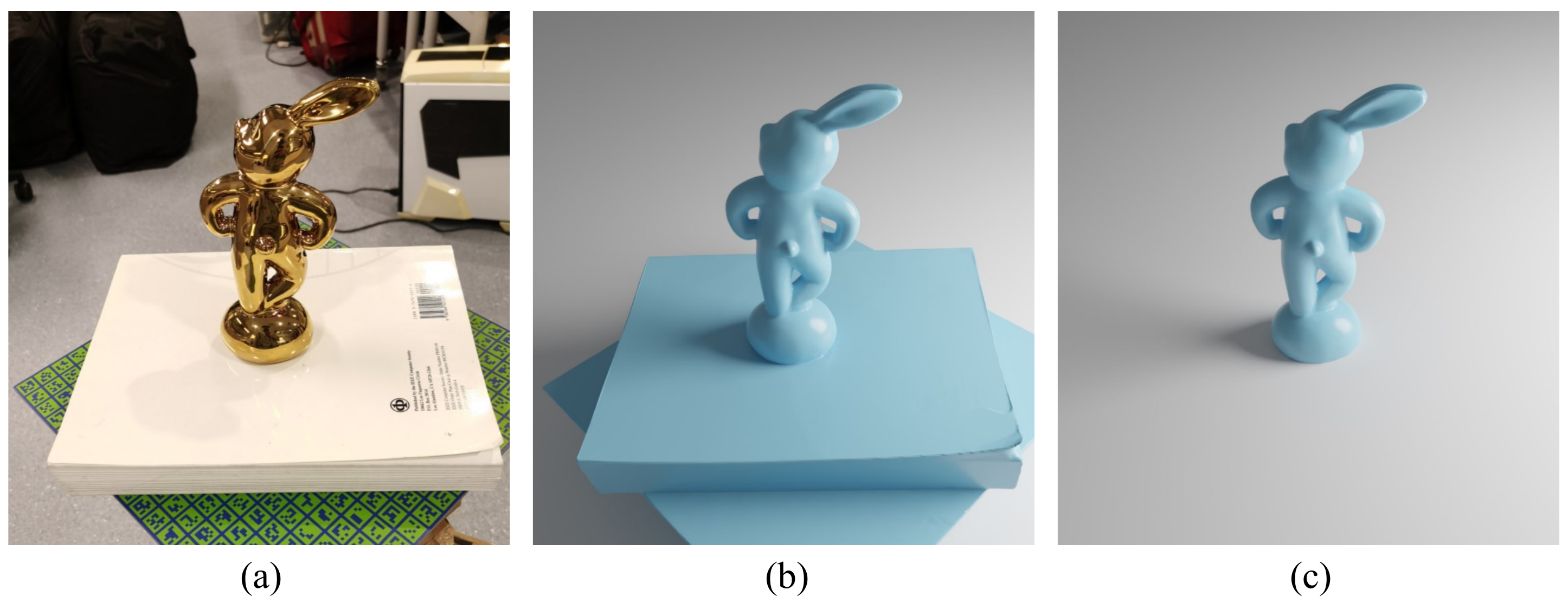}
    \caption{(a) An input image. (b) The completed reconstruction of our method. (c) The cropped mesh for the evaluation.}
    \label{fig:mesh_crop}
\end{figure}

\noindent{\textbf{Geometry evaluation}}. For the evaluation of geometry, we adopt the Chamfer distance (CD) between the reconstructed surfaces and the ground-truth surfaces as the metric. Note that we only evaluate the reconstructed surfaces that are visible in images. To achieve this, we first select 16 input cameras using the farthest point sampling on the camera locations. Then, we project the reconstructed surface or the ground-truth surface onto each sampled camera to get a depth map. Finally, we fuse all the depth maps into a complete point cloud. The Chamfer distance is computed between the point cloud of the reconstructed surface and the one of the ground-truth surface. On the real data, NeRO reconstructs all surfaces inside the bounding sphere rather than only reconstructing the surfaces of the reflective object, as shown in Fig.~\ref{fig:mesh_crop} (b). From the reconstructed mesh, we manually crop the mesh of the object for the evaluation, as shown in Fig.~\ref{fig:mesh_crop} (c).

\noindent{\textbf{BRDF evaluation}}. Directly evaluating the quality of BRDF is difficult because different baseline methods adopt different BRDF models. 
Instead, we evaluate the quality of relighted images to reveal the quality of the estimated BRDF. 
For each object in the Glossy-Blender dataset, we use three new HDR images as the environment lights to illuminate the object. For each environment light, we render 16 evenly-distributed relighted images around the object. Finally, we compute PSNR, SSIM~\cite{wang2004image}, and LPIPS~\cite{zhang2018unreasonable} between the relighted images of different methods and the images rendered by Blender. 
There is an indeterminable scale factor between the albedo and the lights. The indeterminable light-albedo scales may be different for different baselines. For a fair comparison, we normalize the relighted images to match the average colors of ground-truth images before computing the metrics. On all the real data, we do not adopt such normalization and provide a visual comparison.

\noindent{\textbf{Baselines for geometry estimation}}. For geometry reconstruction, we compare our method with COLMAP~\cite{schoenberger2016mvs}, Ref-NeRF~\cite{verbin2022ref}, NeuS~\cite{wang2021neus}, NDR~\cite{munkberg2022extracting}, and NDRMC~\cite{hasselgren2022shape}. COLMAP~\cite{schoenberger2016mvs} adopts the traditional MVS algorithm patch match stereo~\cite{bleyer2011patchmatch} for the reconstruction. Since COLMAP fails to reconstruct complete meshes on reflective objects, we only provide a qualitative comparison with COLMAP. Ref-NeRF represents the surface with a density field and uses the reflective direction and IDE in its color function. For Ref-NeRF, we use grid search to find the best density threshold in $[0,15]$ to extract a level set as the output surface. NDR adopts the differentiable marching tetrahedral (DMTet)~\cite{shen2021deep} as the surface representation and uses prefiltered split-sum to approximate direct illuminations. NDRMC replaces the prefiltered split-sum of NDR with a differentiable MC sampling. Note that both NDR and NDRMC use ground-truth object masks for training. NeuS uses volume rendering on SDF to extract the surfaces. For all baseline methods, we adopt their official implementations.

\begin{figure*}
    \centering
    \setlength\tabcolsep{1pt}
    \renewcommand{\arraystretch}{0.5} 
    \begin{tabular}{cccccc}
        \includegraphics[width=0.16\textwidth]{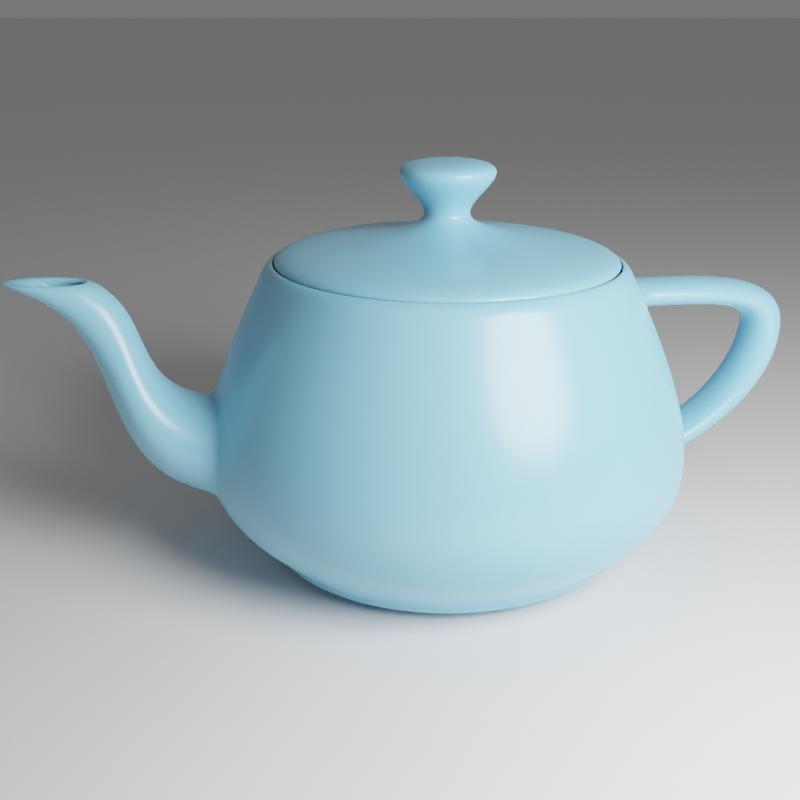} &
        \includegraphics[width=0.16\textwidth]{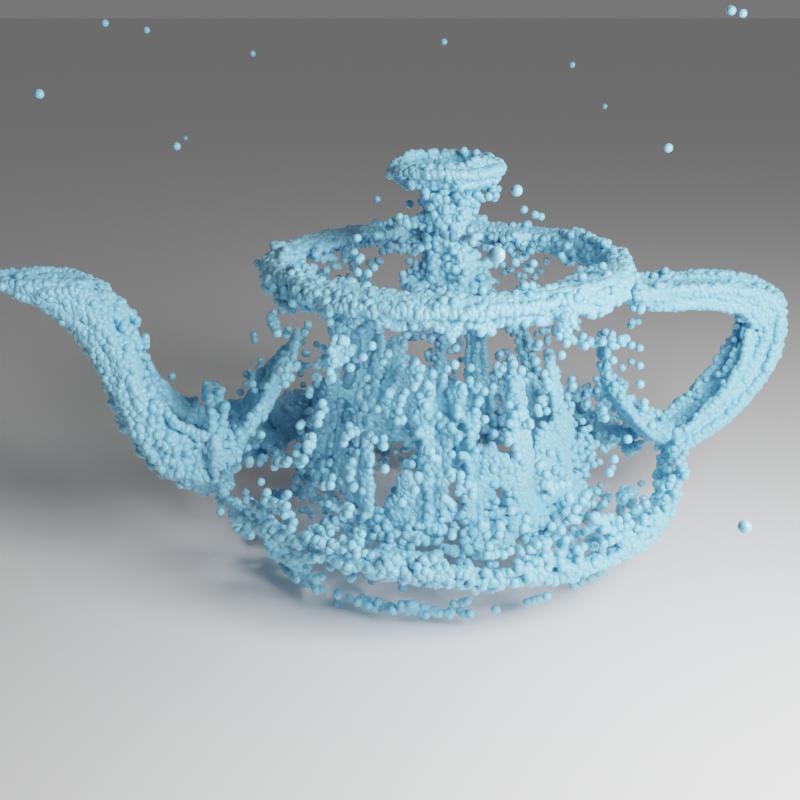} &
        \includegraphics[width=0.16\textwidth]{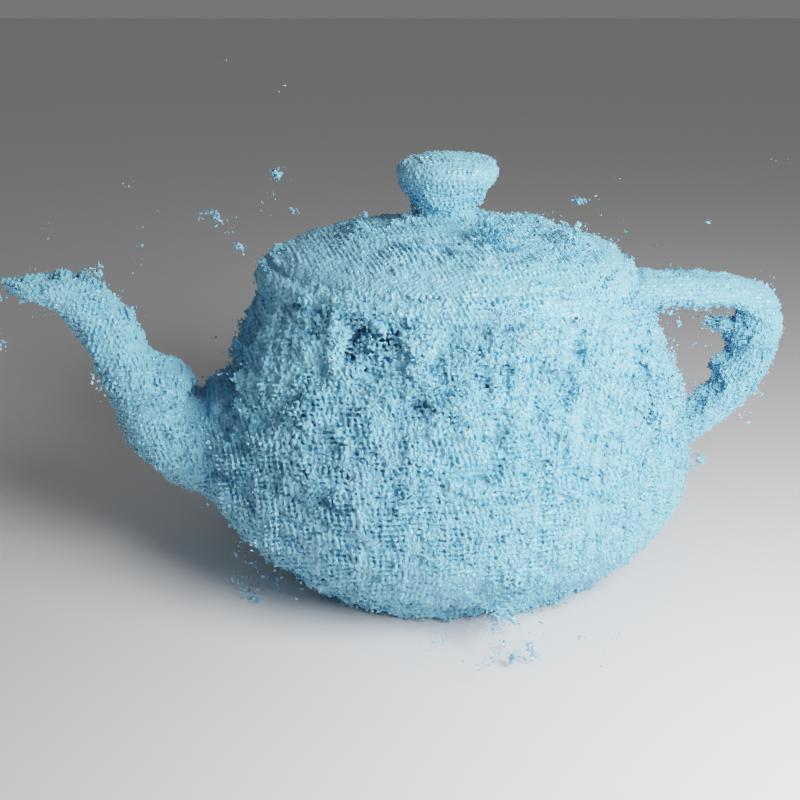} &
        \includegraphics[width=0.16\textwidth]{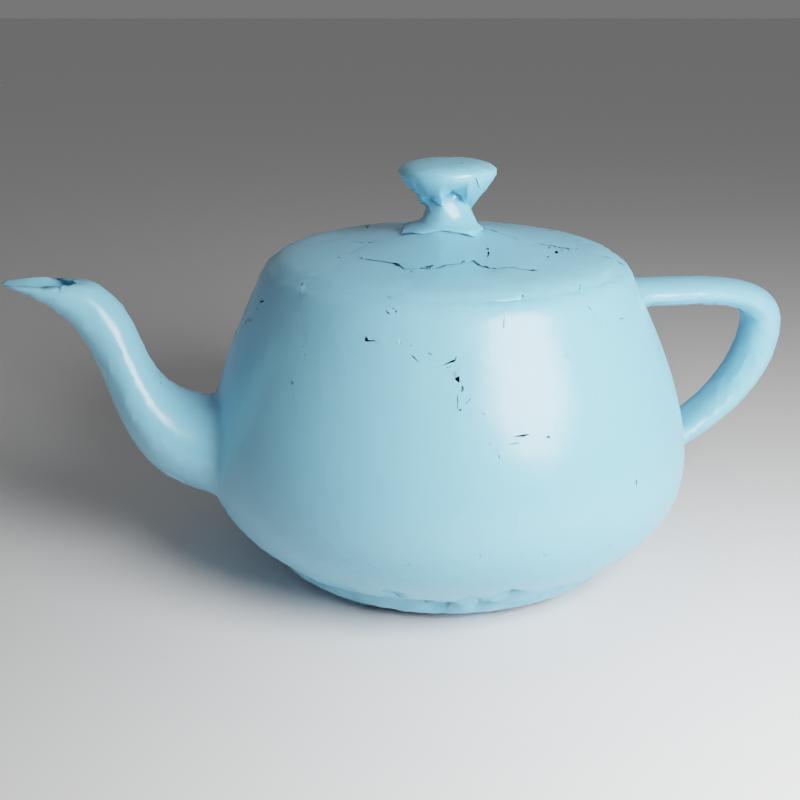} &
        \includegraphics[width=0.16\textwidth]{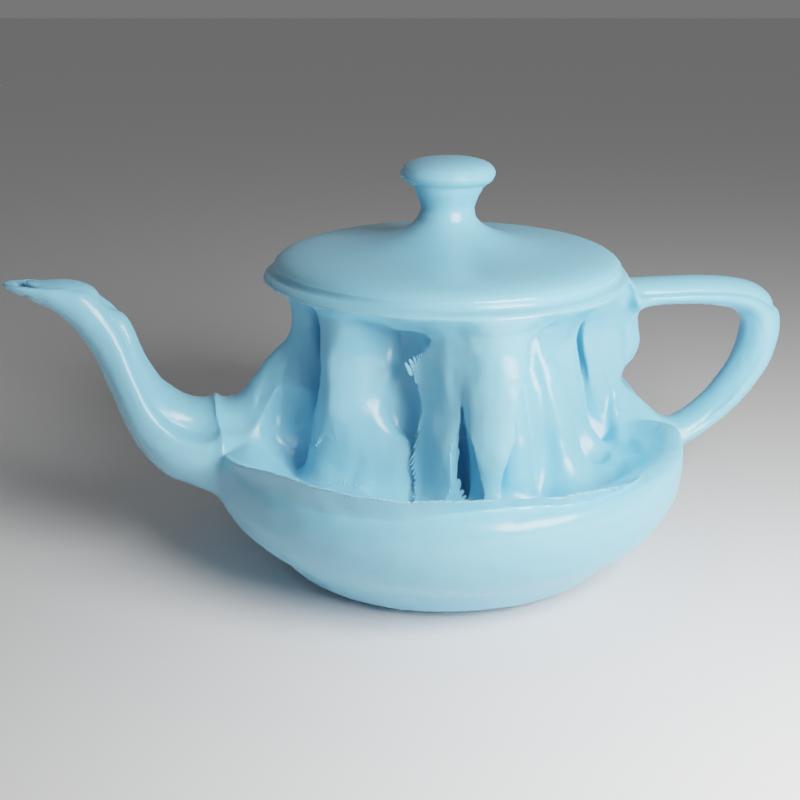} &
        \includegraphics[width=0.16\textwidth]{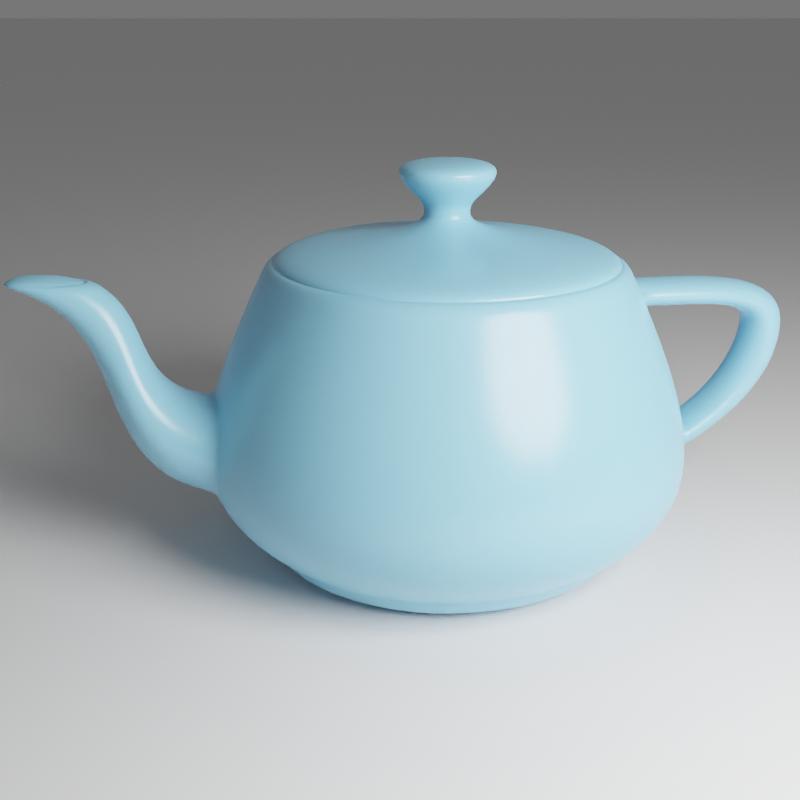} \\
        \includegraphics[width=0.16\textwidth]{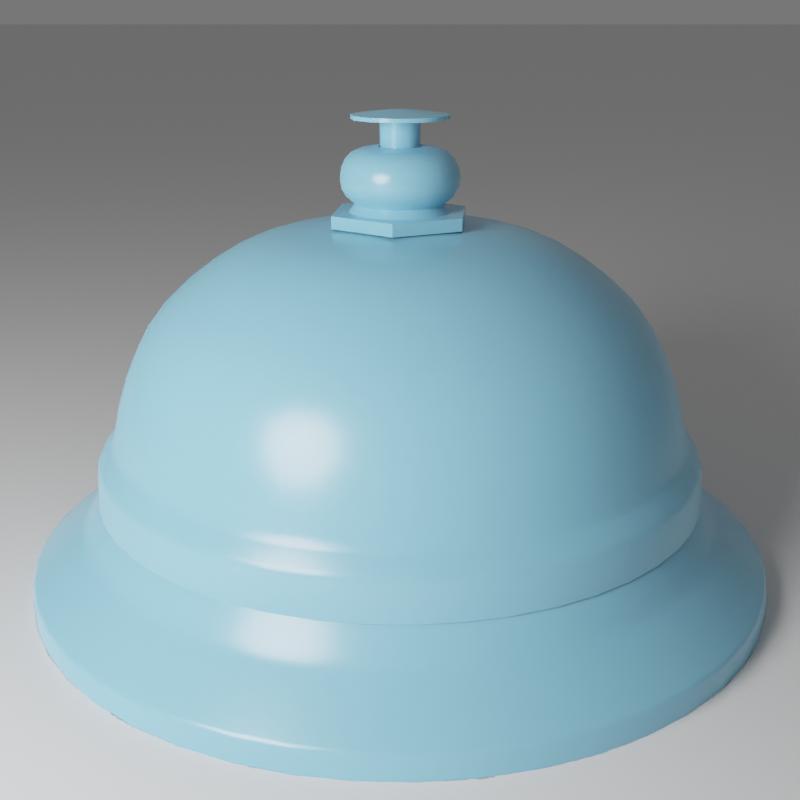} &
        \includegraphics[width=0.16\textwidth]{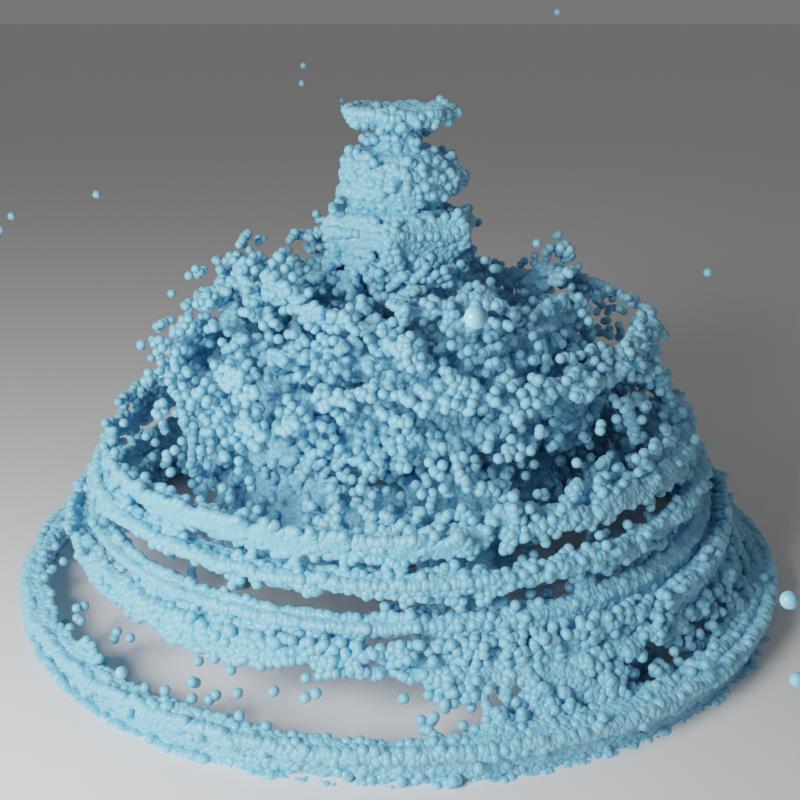} &
        \includegraphics[width=0.16\textwidth]{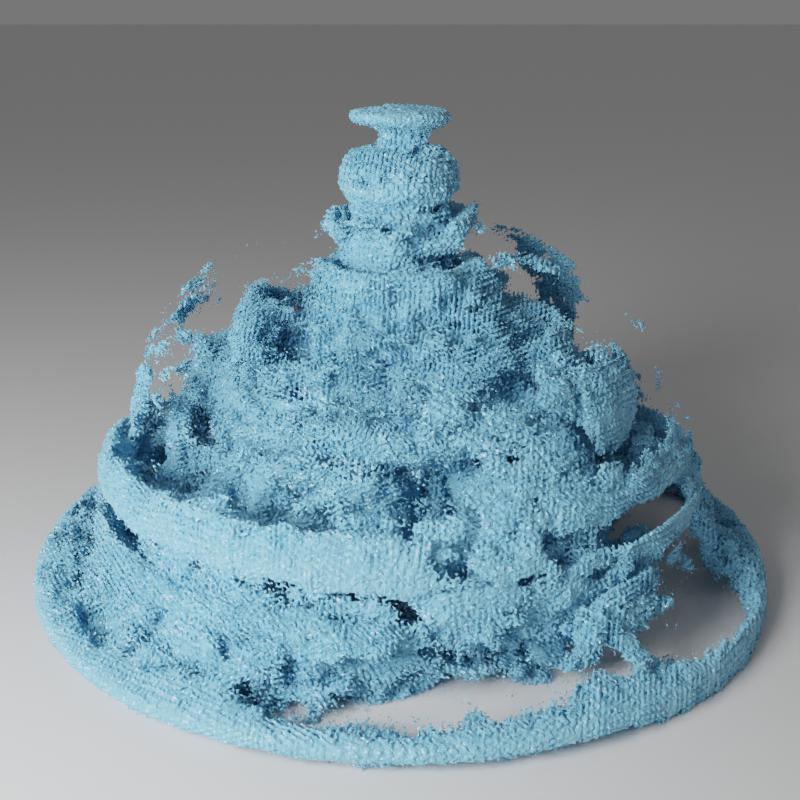} &
        \includegraphics[width=0.16\textwidth]{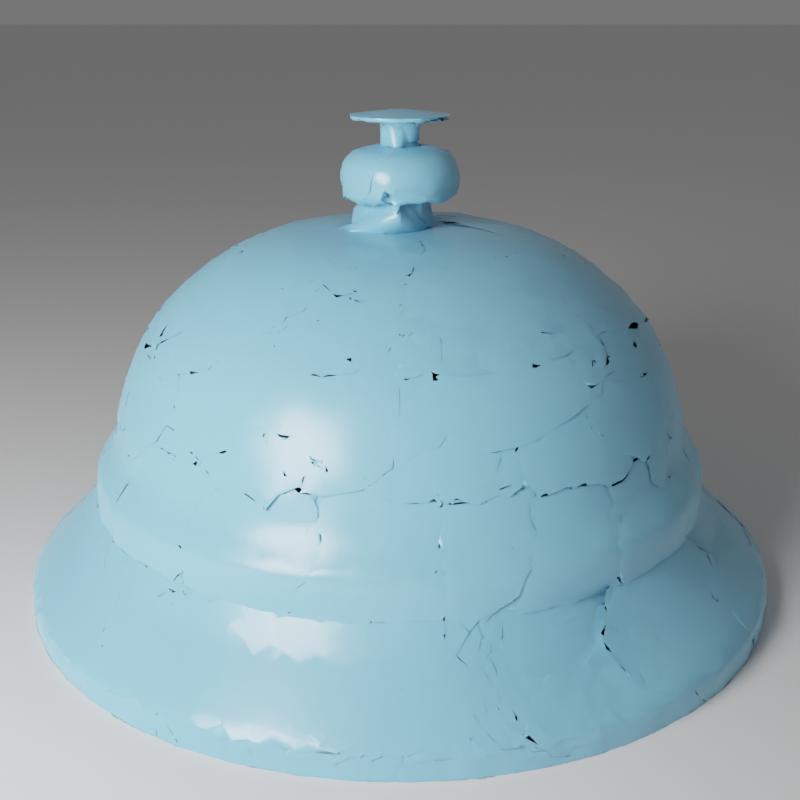} &
        \includegraphics[width=0.16\textwidth]{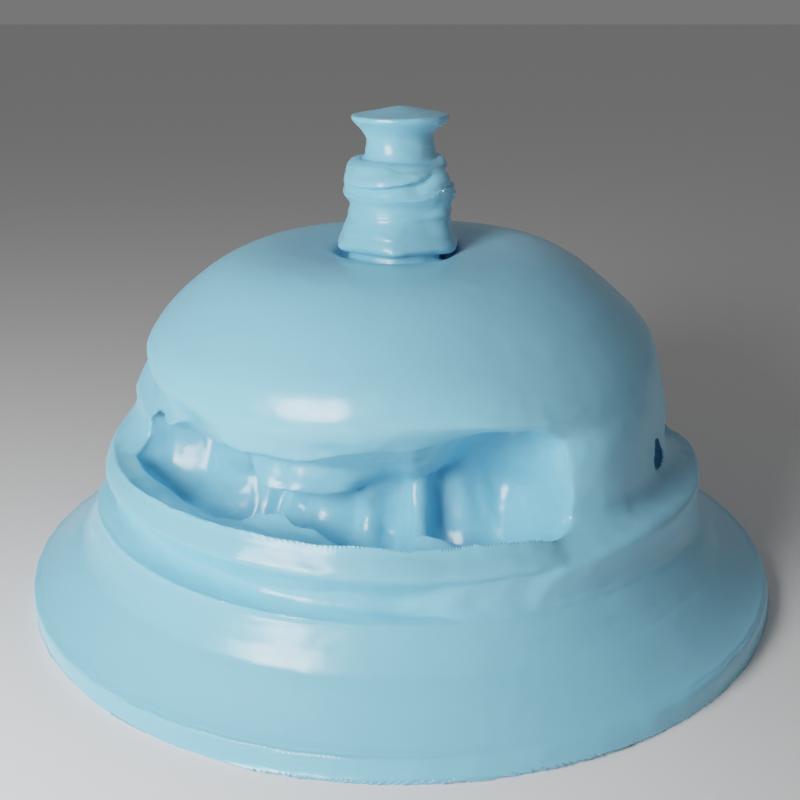} &
        \includegraphics[width=0.16\textwidth]{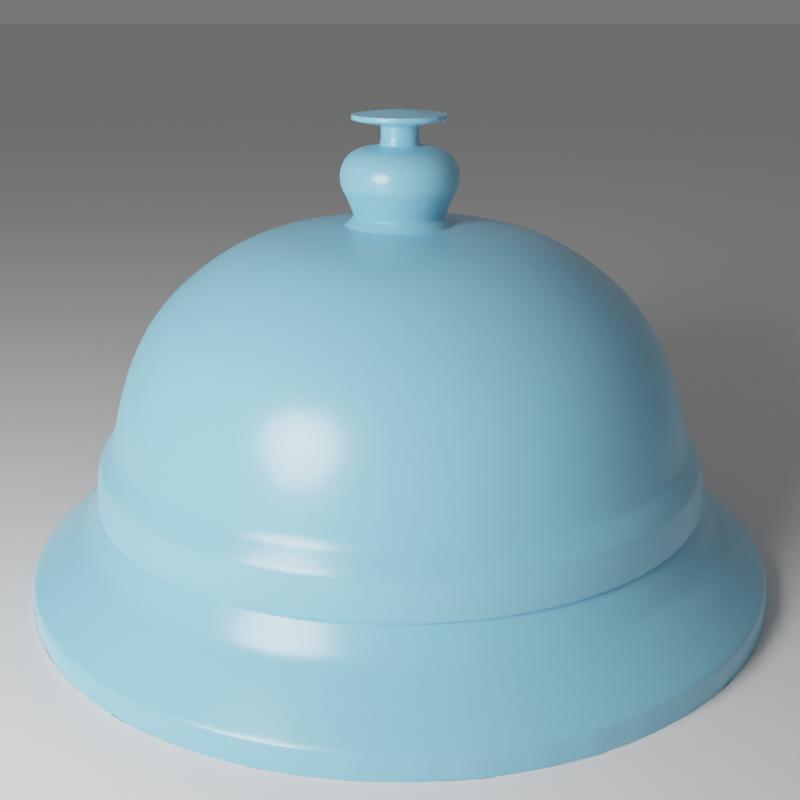} \\
        \includegraphics[width=0.16\textwidth]{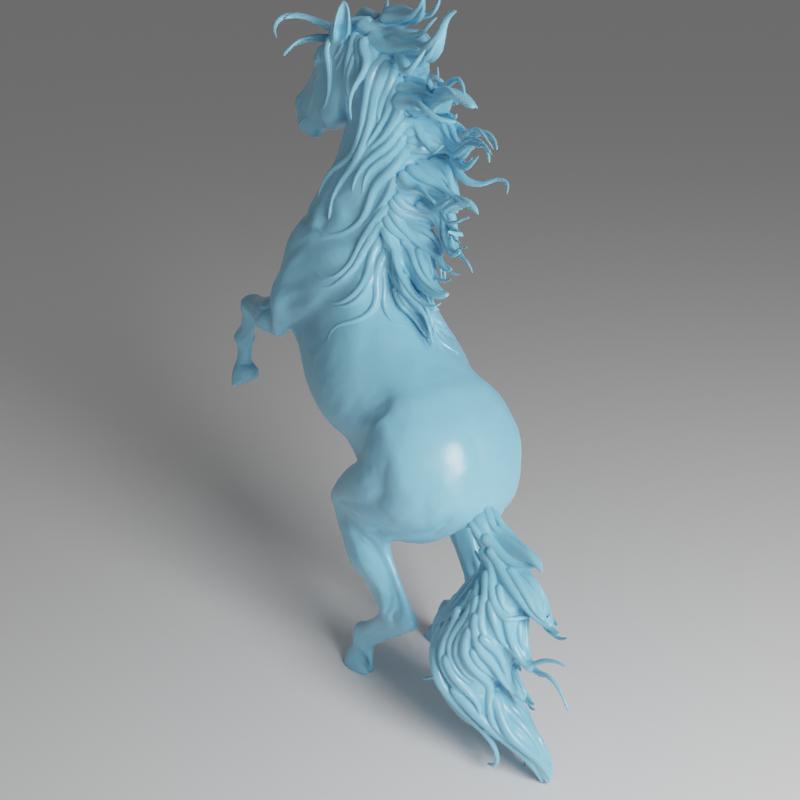} &
        \includegraphics[width=0.16\textwidth]{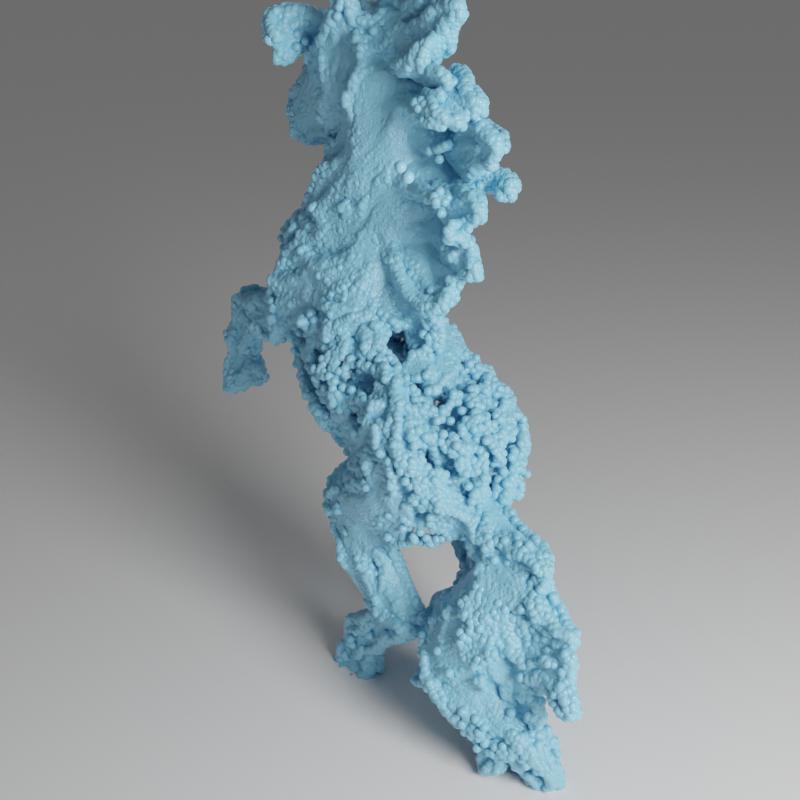} &
        \includegraphics[width=0.16\textwidth]{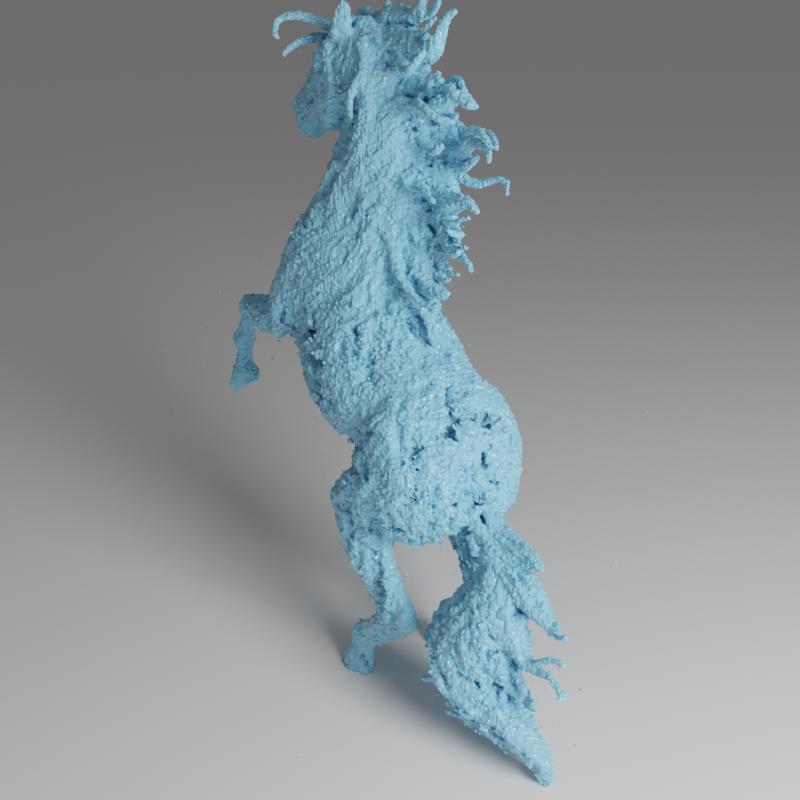} &
        \includegraphics[width=0.16\textwidth]{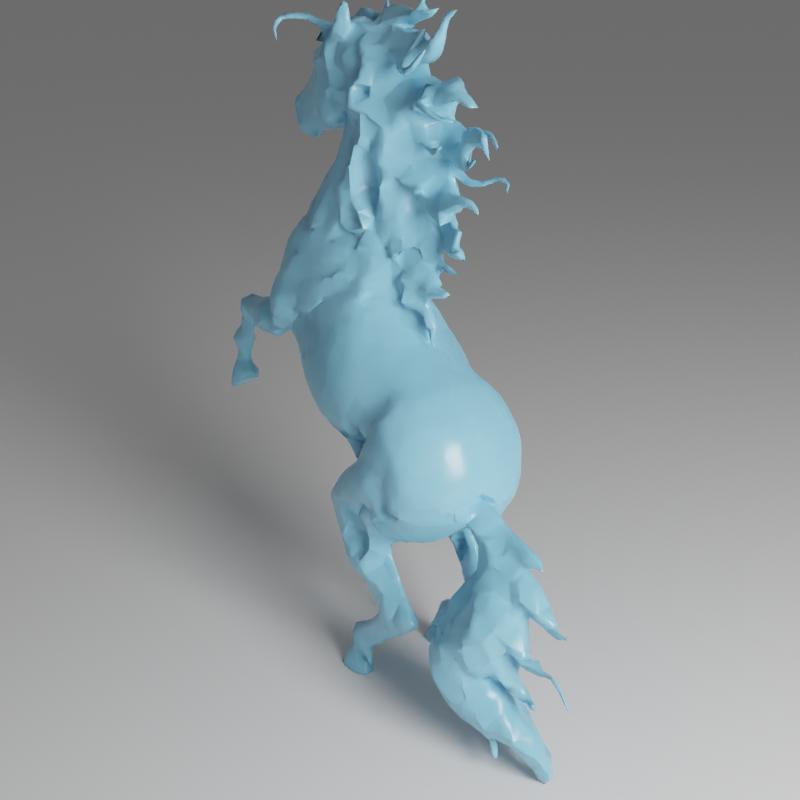} &
        \includegraphics[width=0.16\textwidth]{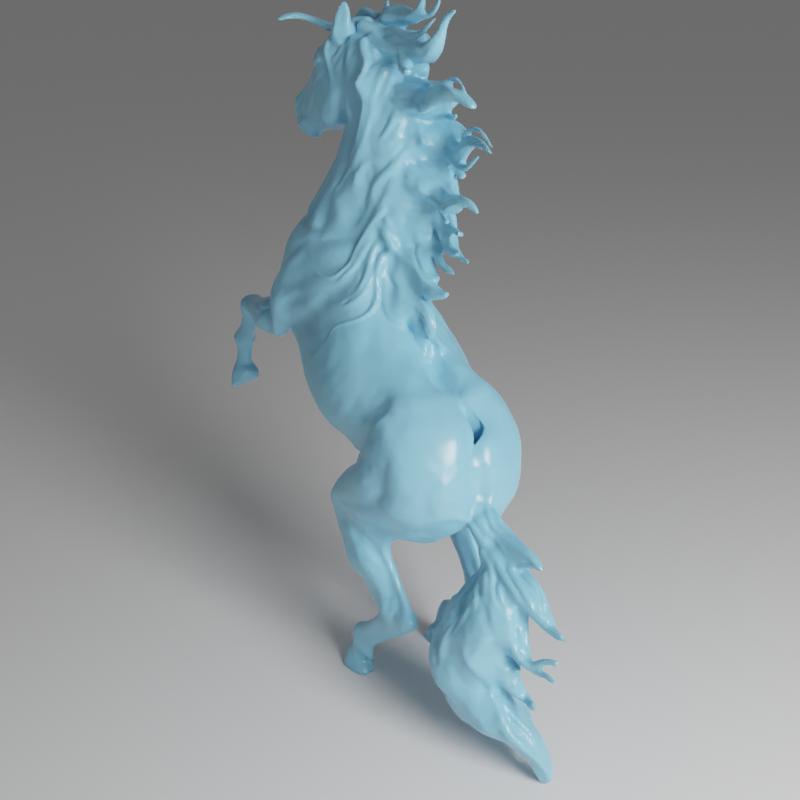} &
        \includegraphics[width=0.16\textwidth]{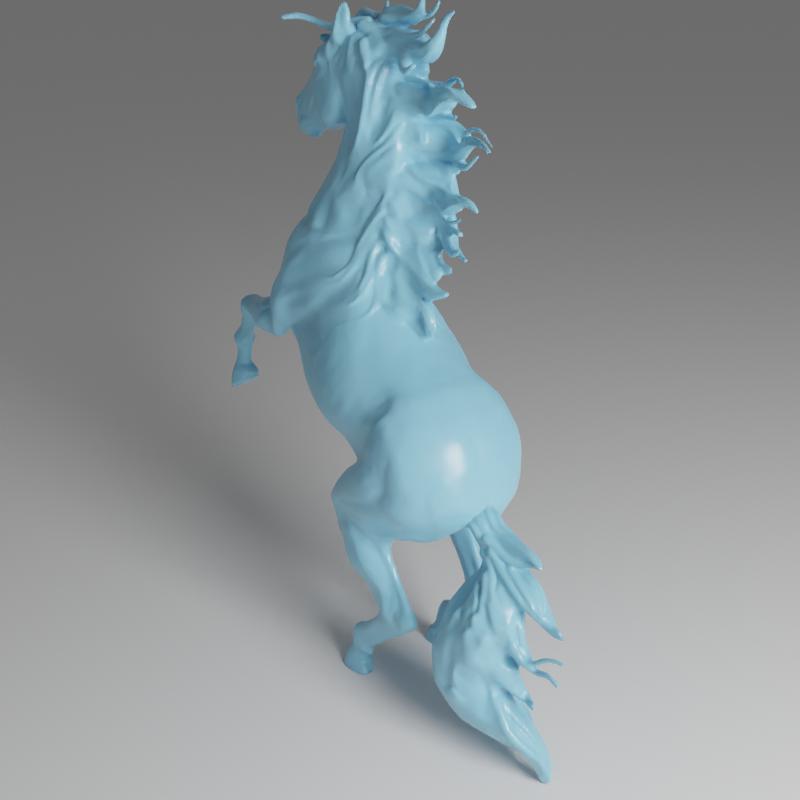} \\
        (a) Ground-truth & (b) COLMAP & (c) Ref-NeRF & (d) NDRMC$^*$ & (e) NeuS & (f) Ours
    \end{tabular}
    \caption{\textbf{Ground-truth and reconstructed surfaces of the Glossy-Blender dataset}. We compare our results with COLMAP~\cite{schoenberger2016mvs}, Ref-NeRF~\cite{verbin2022ref}, NDRMC~\cite{hasselgren2022shape}, and NeuS~\cite{wang2021neus}. $^*$NDRMC is trained with ground-truth object masks while the other methods do not use object masks.}
    \label{fig:syn_visual}
\end{figure*}

\noindent{\textbf{Baselines for BRDF estimation}}. We choose baseline methods NDR~\cite{munkberg2022extracting}, NDRMC~\cite{hasselgren2022shape}, MII~\cite{zhang2022modeling} and NeILF~\cite{yao2022neilf} for comparison. NDR only uses split-sum to consider direct lights while NDRMC, NeILF, and MII all consider indirect lights in the BRDF estimation. NDRMC uses the differentiable MC sampling with a denoiser. NeILF also adopts the MC sampling but with an MLP-based light model. MII uses the Spherical Gaussian to represent direct or indirect lights and the BRDF. In order to reconstruct reasonable geometry on reflective objects, MII, NDR and NDRMC all rely on the object masks in training. NeILF does not involve surface reconstruction in its pipeline and needs a reconstructed mesh as input. Thus, we use the mesh reconstructed by our method as the input to NeILF. 
MII can be regarded as a combined version of NeRFactor~\cite{zhang2021nerfactor} and PhySG~\cite{zhang2021physg}. NDR is similar to Neural-PIL~\cite{boss2021neural}, both of which are based on the prefiltered split-sum. In this case, we do not exhaustively include a comparison with NeRFactor, PhySG, and Neural-PIL.

\noindent{\textbf{Experimental details}}. We train NeRO on each object for 300k steps for the surface reconstruction and 100k steps for the BRDF estimation. On each training step, 512 camera rays are sampled for both stages. We adopt the Adam optimizer~\cite{kingma2014adam} ($\beta_1=0.9$, $\beta_2=0.999$) with a warm-up learning rate schedule, in which the learning rate first increases from 1e-5 to 5e-4 in 5k (or 1k) steps for the surface reconstruction (or the BRDF estimation) and then decreases to 1e-5 in the subsequent steps. To compute the MC sampling in Stage II, $N_d=512$ ray samples and $N_s=256$ ray samples are used on the diffuse lobe and the specular lobe respectively. The whole training process takes about 20 hours for the surface reconstruction and 5 hours for the BRDF estimation on a 2080Ti GPU.  Note that the most time-consuming part in training is that we sample a massive number of points in the volume rendering. Recent voxel-based representations could largely reduce sample points and thus could speed up the NeRO training, which we leave for future works. On the output mesh, we assign BRDF parameters to vertices and interpolate the parameters on each face. More details about the architecture can be found in Sec.~\ref{sec:app_imp}. 

\begin{table}[]
    \centering
    \resizebox{\linewidth}{!}{
    \begin{tabular}{cccccc}
    \toprule
          &NDR$^*$ & NDRMC$^*$ & NeuS &  Ref-NeRF & Ours \\
    \midrule
    Bell  & 0.0122 & \underline{0.0045} & 0.0146 & 0.0137  & \textbf{0.0032} \\
    Cat   & 0.0344 & 0.1299 & 0.0278 & \underline{0.0201}  & \textbf{0.0044} \\
    Teapot& 0.0530 & \underline{0.0052} & 0.0546 & 0.0143  & \textbf{0.0037} \\
    Potion& 0.0554 & 0.0415 & 0.0393 & \underline{0.0131}  & \textbf{0.0053} \\
    TBell & 0.0821 & \underline{0.0046} & 0.0348 & 0.0216  & \textbf{0.0035} \\
    Angel & 0.0056 & \textbf{0.0034} & \underline{0.0035} & 0.0291  & \textbf{0.0034} \\
    Horse & 0.0077 & \underline{0.0052} & 0.0053 & 0.0071   & \textbf{0.0049} \\
    Luyu  & 0.0131 & 0.0082 & \underline{0.0066} & 0.0141  & \textbf{0.0054} \\
    \midrule
    Avg.  & 0.0329 & 0.0253 & \underline{0.0233} & 0.0241 & \textbf{0.0042} \\
    \bottomrule
    \end{tabular}
    }
    \caption{\textbf{Reconstruction quality in Chamfer Distance (CD$\downarrow$) on the Glossy-Synthetic dataset}. We compare our method with NeuS~\cite{wang2021neus}, Ref-NeRF~\cite{verbin2022ref}, NDR~\cite{munkberg2022extracting} and NDRMC~\cite{hasselgren2022shape}. $^*$NDRMC and NDR use ground-truth object masks while the other methods do not use object masks. \textbf{Bold} means the best performance and \underline{underline} means the second best performance.}
    \label{tab:syn}
\end{table}

\begin{figure*}
    \centering
    \setlength\tabcolsep{1pt}
    \renewcommand{\arraystretch}{0.5} 
    \begin{tabular}{cccccc}
        \includegraphics[width=0.16\textwidth]{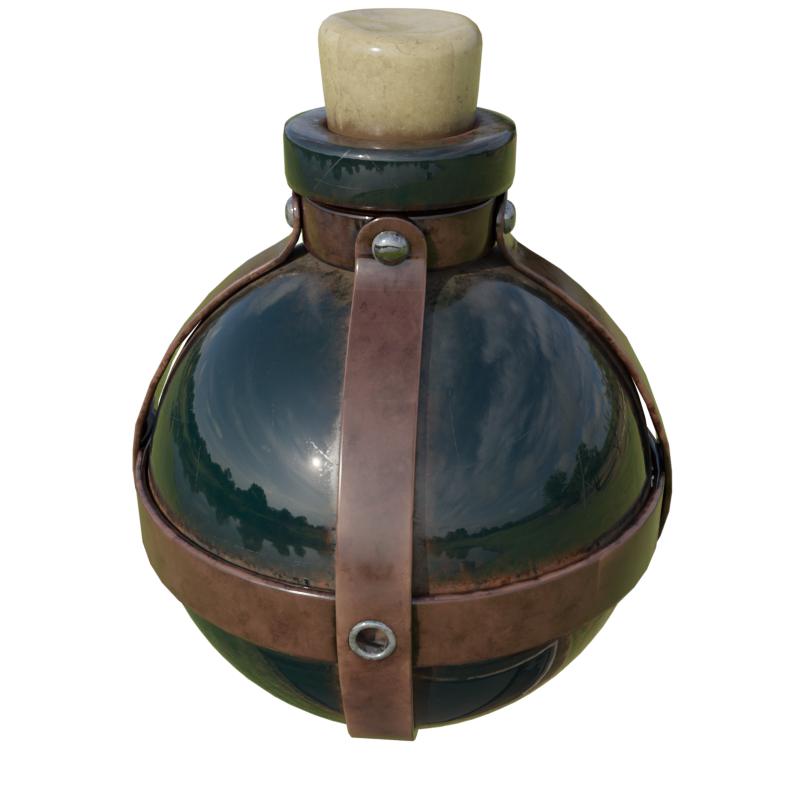} &
        \includegraphics[width=0.16\textwidth]{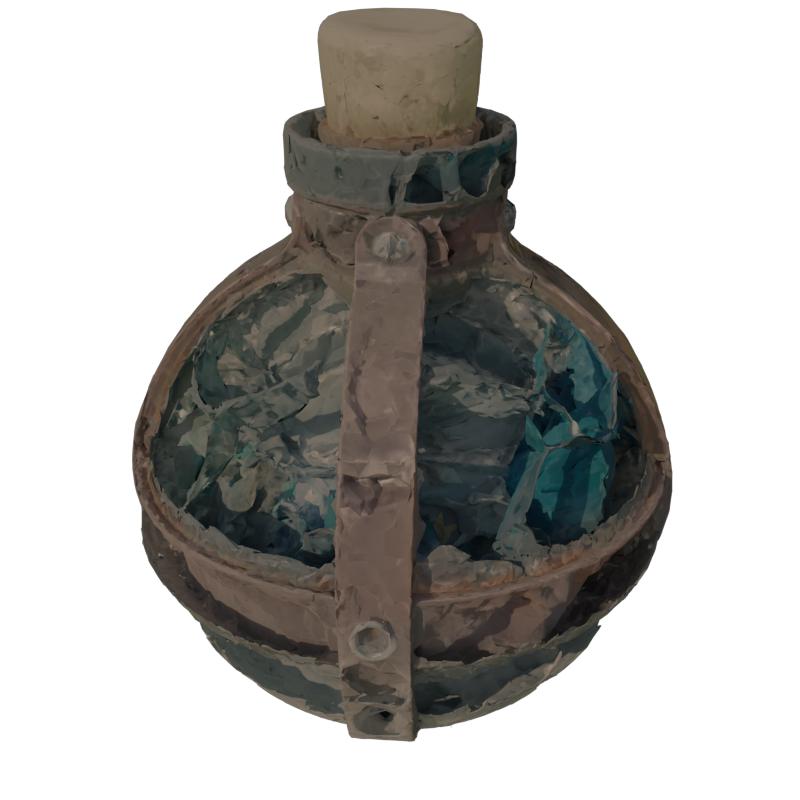} &
        \includegraphics[width=0.16\textwidth]{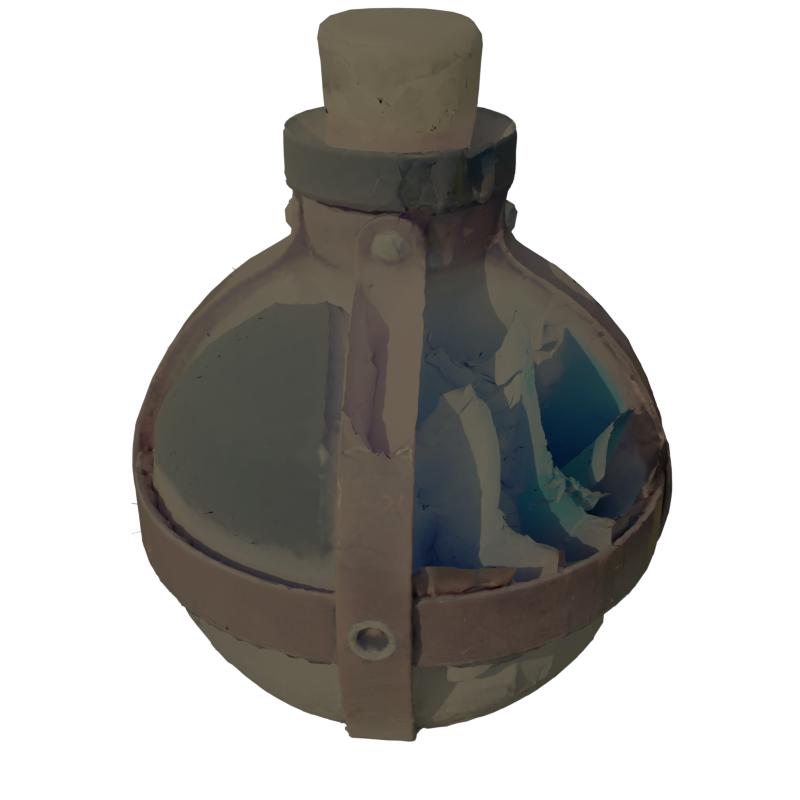} &
        \includegraphics[width=0.16\textwidth]{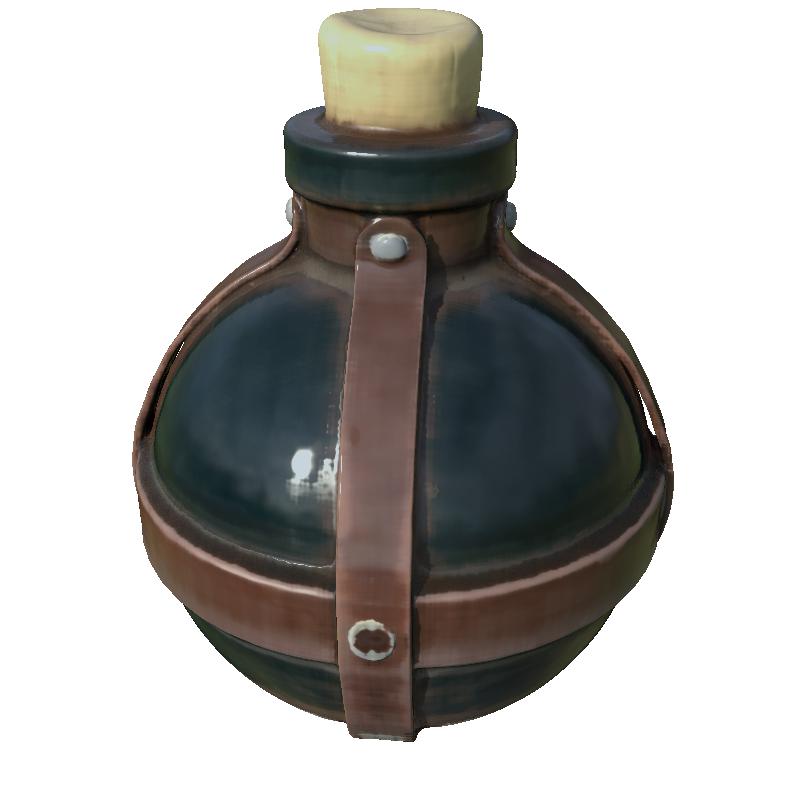} &
        \includegraphics[width=0.16\textwidth]{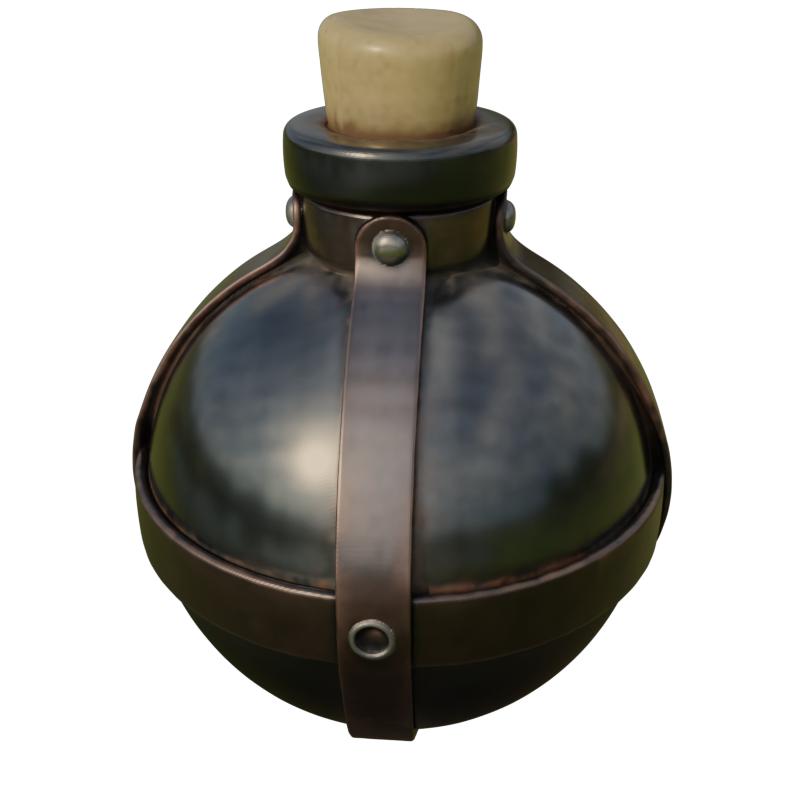} &
        \includegraphics[width=0.16\textwidth]{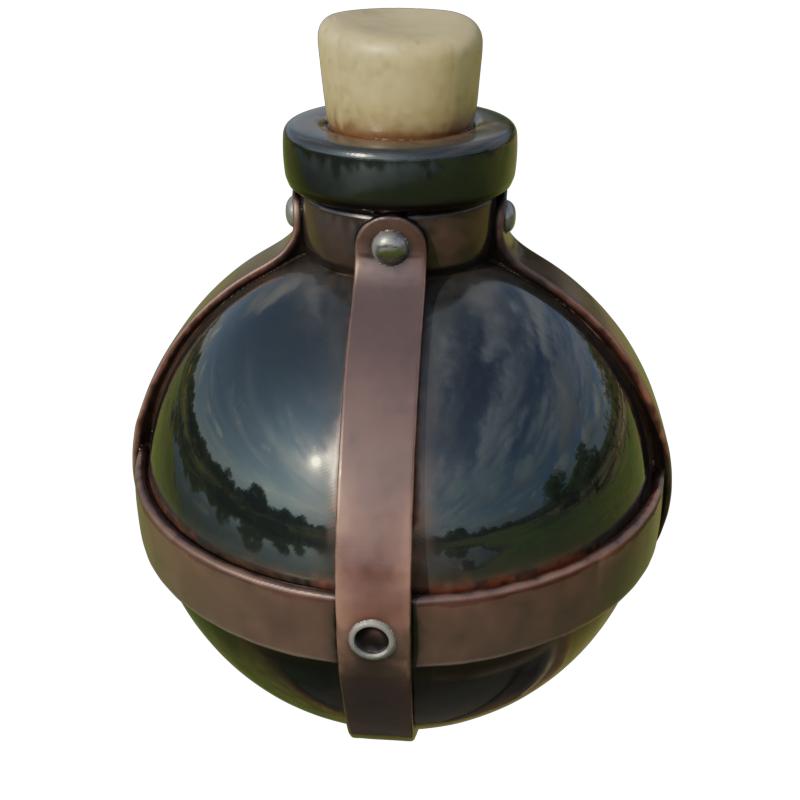} \\
        \includegraphics[width=0.16\textwidth]{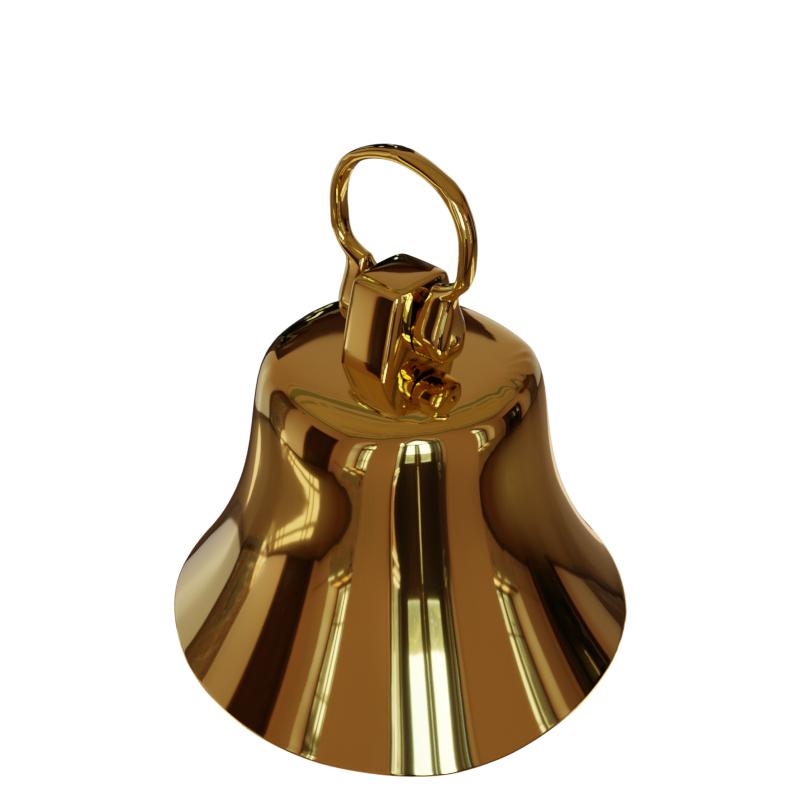} &
        \includegraphics[width=0.16\textwidth]{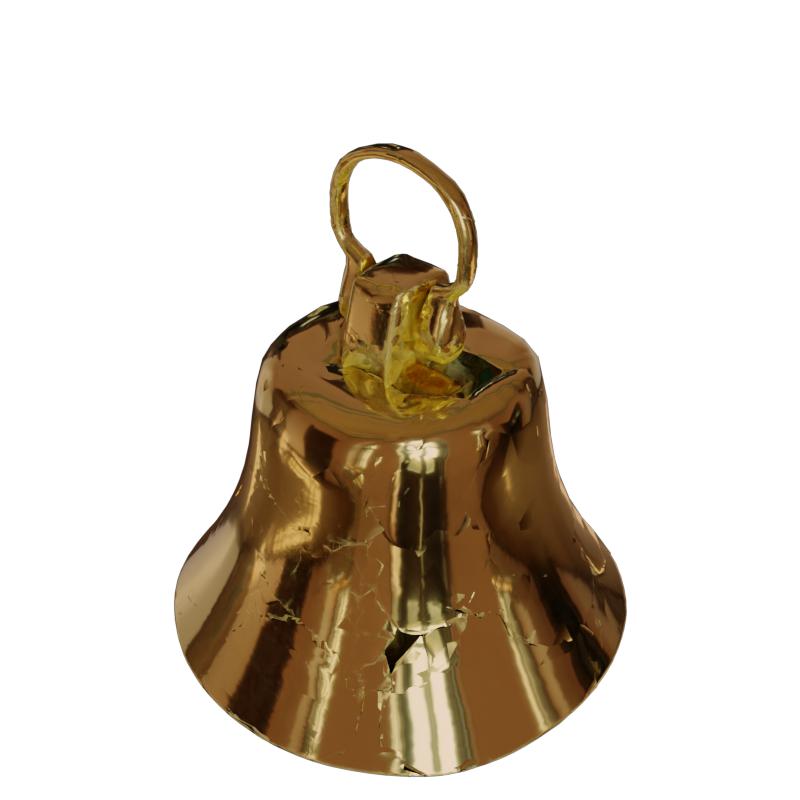} &
        \includegraphics[width=0.16\textwidth]{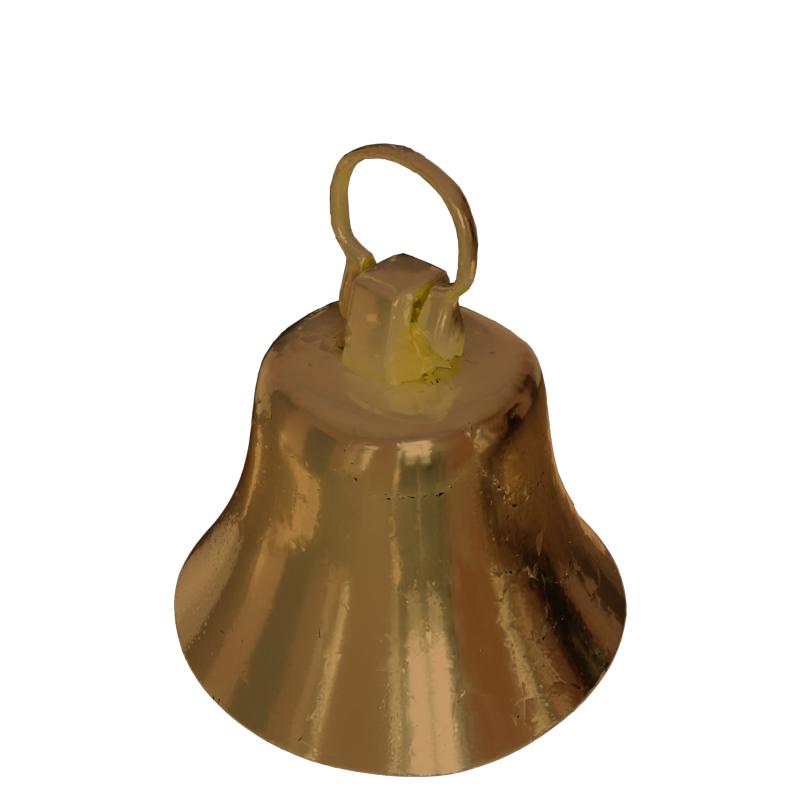} &
        \includegraphics[width=0.16\textwidth]{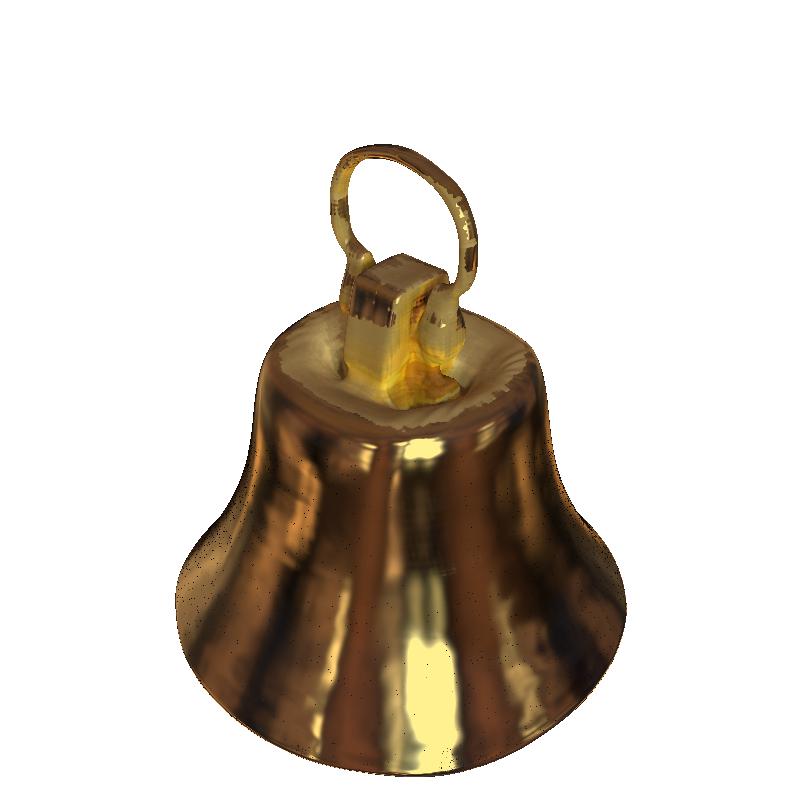} &
        \includegraphics[width=0.16\textwidth]{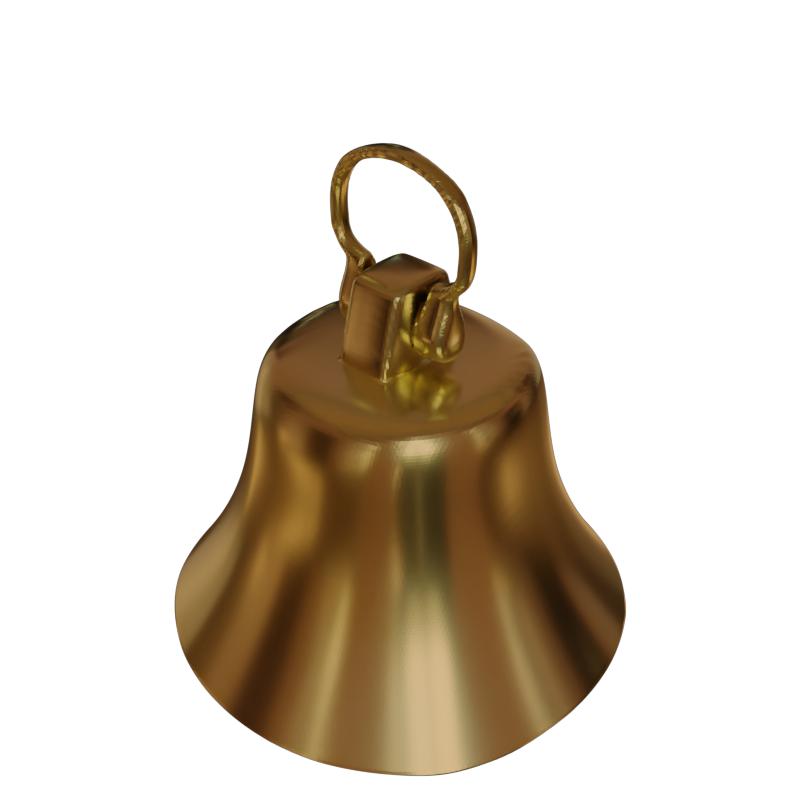} &
        \includegraphics[width=0.16\textwidth]{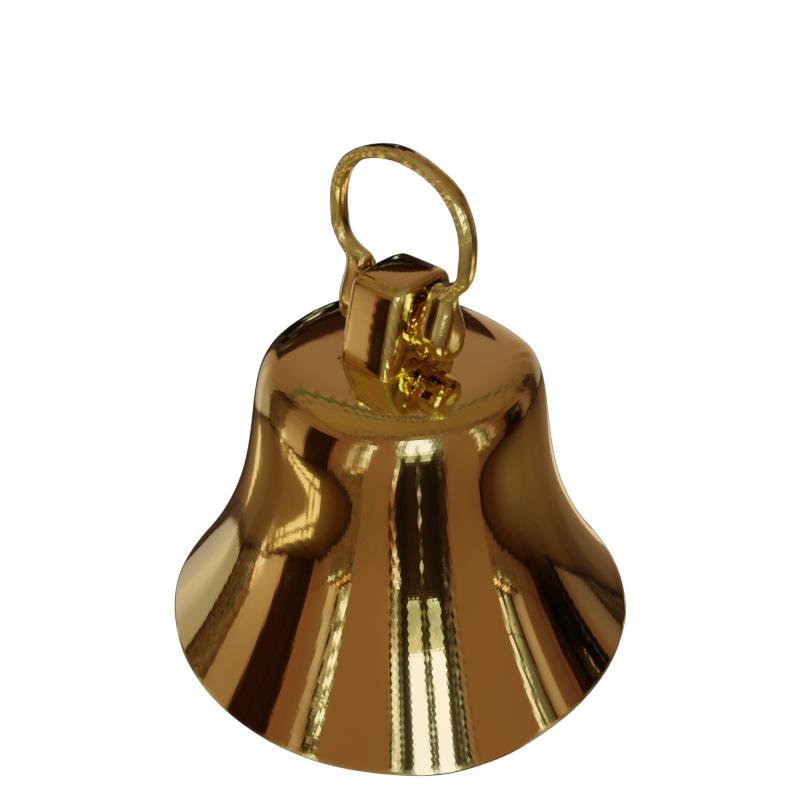} \\
        \includegraphics[width=0.16\textwidth]{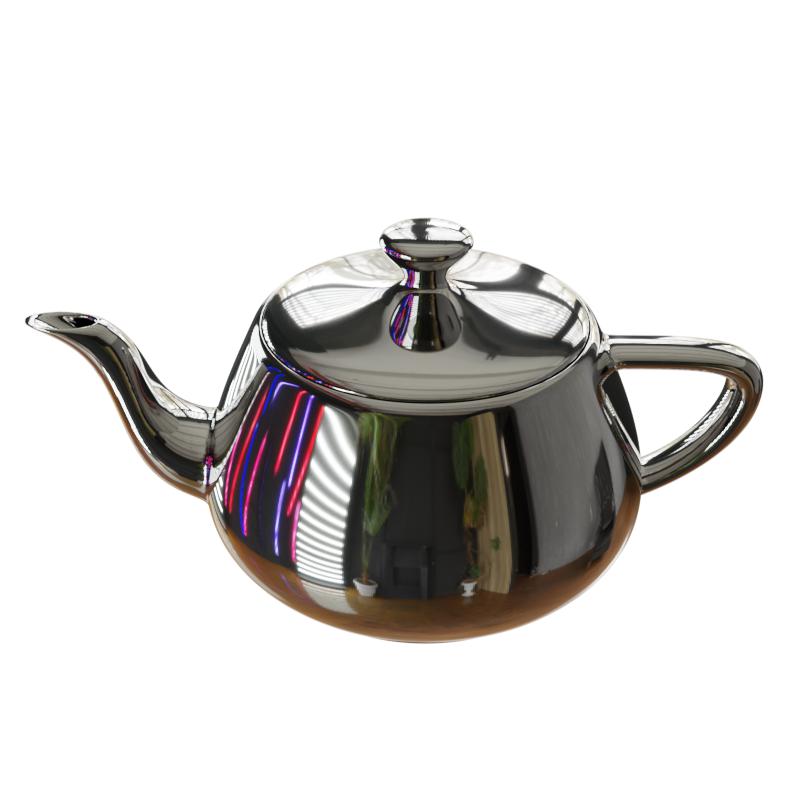} &
        \includegraphics[width=0.16\textwidth]{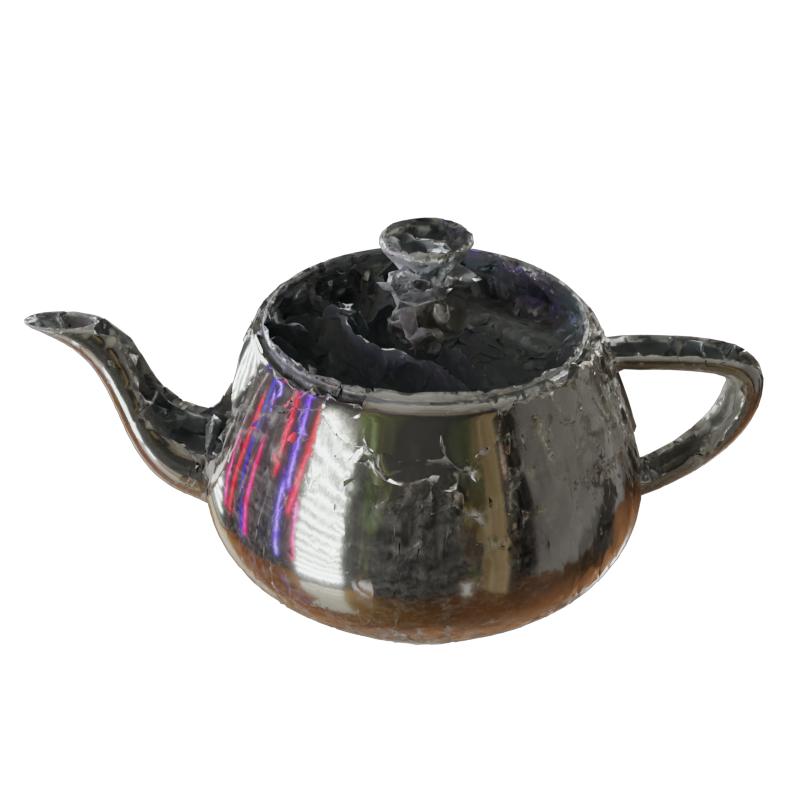} &
        \includegraphics[width=0.16\textwidth]{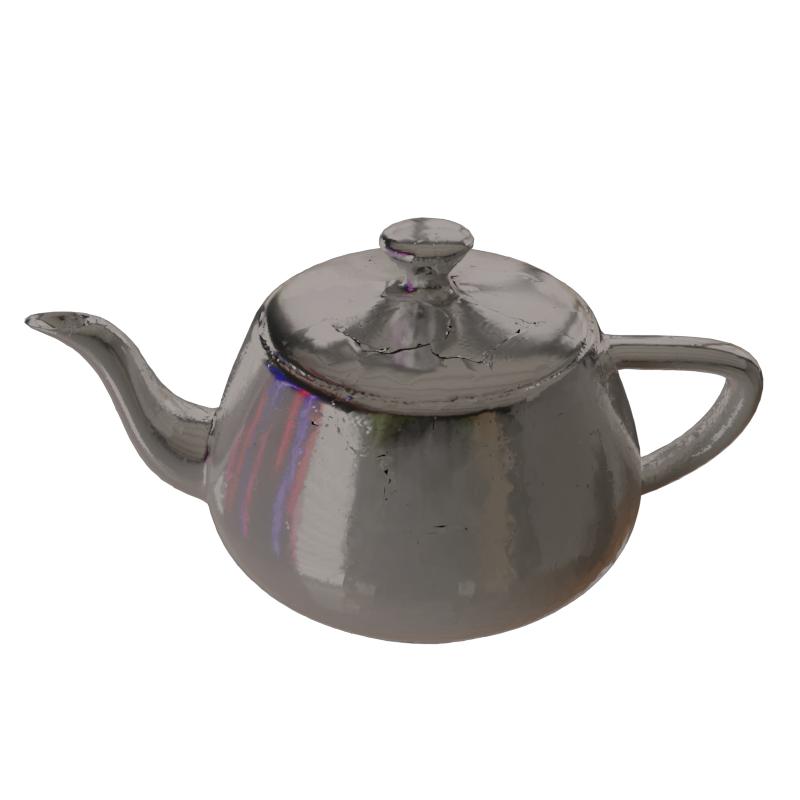} &
        \includegraphics[width=0.16\textwidth]{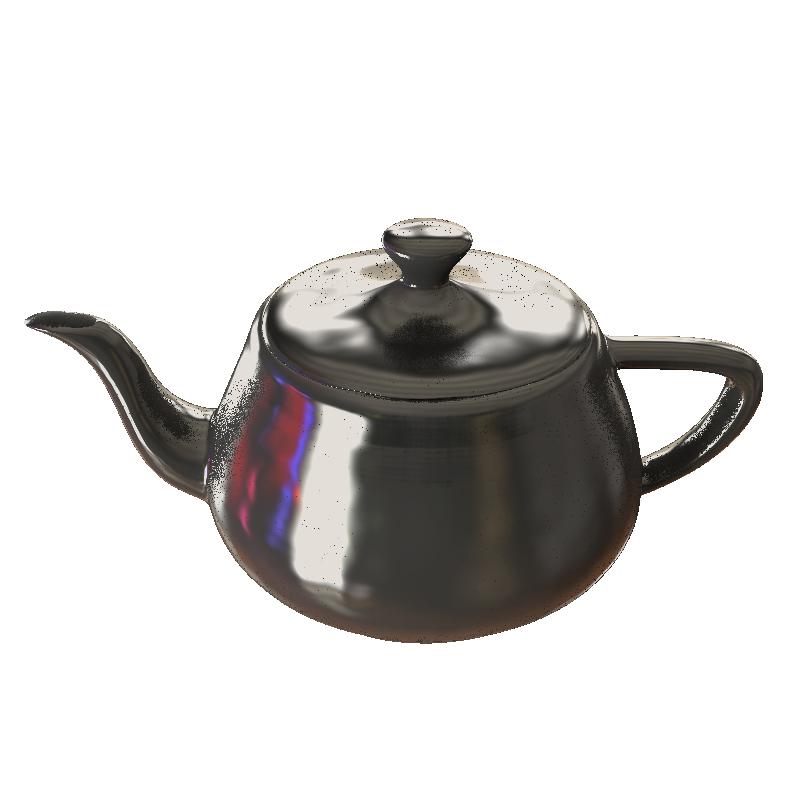} &
        \includegraphics[width=0.16\textwidth]{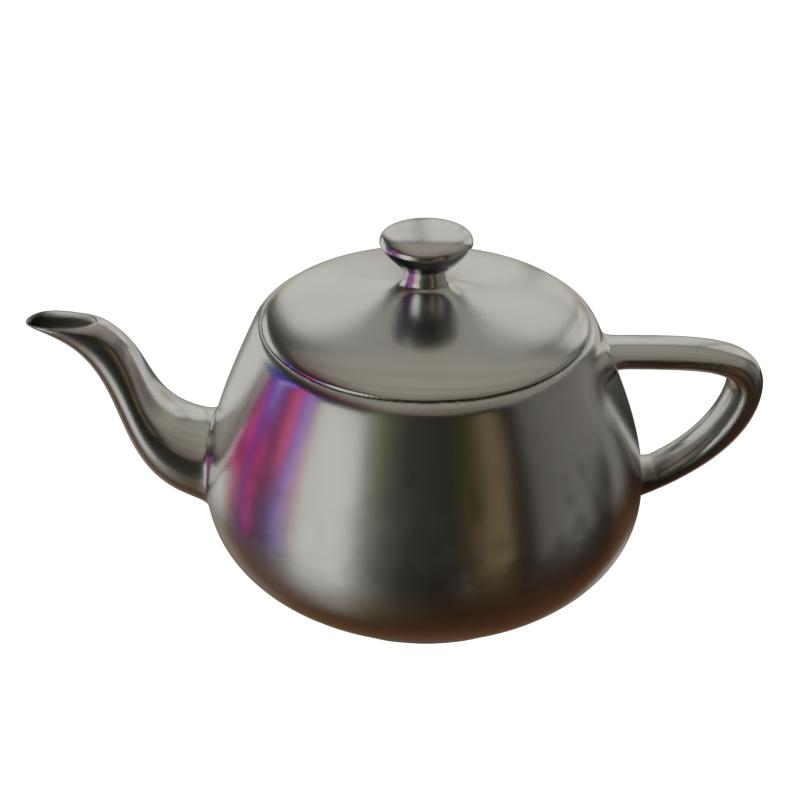} &
        \includegraphics[width=0.16\textwidth]{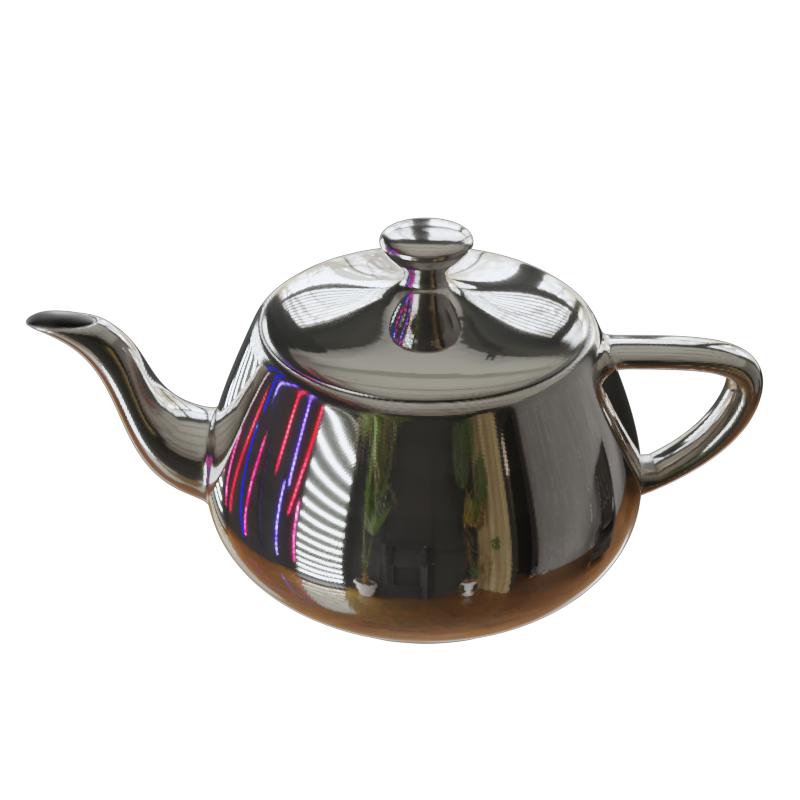} \\
        (a) Ground-truth & (b) NDR & (c) NDRMC & (d) MII & (e) NeILF & (f) Ours \\
    \end{tabular}
    \caption{\textbf{Relighting objects in the Glossy-Blender dataset}. We compare our method with NDR~\cite{munkberg2022extracting}, NDRMC~\cite{hasselgren2022shape}, MII~\cite{zhang2022modeling} and NeILF~\cite{yao2022neilf}. All the relighted images are normalized to match the average colors of the ground-truth images.}
    \label{fig:syn_relight}
\end{figure*}

\subsection{Results on Glossy-Blender}
\label{sec:results_syn}
\textbf{Geometry}. 
The quantitative results in CD are reported in Table~\ref{tab:syn}. Fig.~\ref{fig:bell} and Fig.~\ref{fig:syn_visual} show the visualization of reconstructed surfaces. We interpret the results in the following aspects: 
\begin{enumerate}
    \item On ``Luyu'', ``Horse'' and ``Angel'' which have complex geometry but only small reflective fragments, all methods achieve good reconstruction. The reason is that the surface reconstruction is mainly dominated by the silhouettes of these objects and the appearances of small reflective fragments are relatively less informative. However, on the other objects ``Bell'', ``Cat'', ``Teapot'', ``Potion'', and ``TBell'' which contain a large reflective surface, baseline methods suffer from view-dependent reflections and struggle to reconstruct correct surfaces. In comparison, our method can accurately reconstruct the surfaces.
    \item As a traditional MVS method, COLMAP fails to reconstruct the reflective objects due to the strong reflections, which either cannot reconstruct any points due to the lack of multiview consistency or only reconstructs some erroneous points inside the object surfaces.
    \item Ref-NeRF explicitly considers the direct environment lights by encoding colors in reflective directions. However, a density field is not a good representation for the surface reconstruction, which not only results in noisy surfaces but also causes errors in estimating surface normals when the reflections are strong, as shown by the ``TBell'' Row 2 in Fig.~\ref{fig:syn_visual}. 
    Meanwhile, for regions illuminated by indirect lights, Ref-NeRF also reconstructs incorrect surfaces, as shown by ``Bell'' Row 1 and ``Cat'' Row 2 in Fig.~\ref{fig:bell}.
    \item NDRMC produces good general shape reconstruction but the reconstructed surfaces are not smooth and contain holes or cracks. The CD of NDRMC in Table~\ref{tab:syn} is large because these small cracks or holes will produce points that are far away from the ground-truth surfaces. Moreover, it is still intractable to apply MC-based shading in volume rendering with multiple sample points on a ray. Thus, similar to IDR~\cite{yariv2020multiview}, NDRMC computes shading on one point per ray and strongly relies on object masks for shape reconstruction.
    \item NeuS incorrectly distorts the large reflective surfaces to fit the reflection colors, leading to erroneous reconstructions on the objects with large reflective surfaces.
    \item Our method can correctly reconstruct all these reflective objects by considering both direct and indirect lights to explain the reflective appearance. Especially on ``Teapot'', ``TBell'', ``Cat'' and ``Potion'' with large reflective surfaces, our method accurately reconstructs surfaces no matter whether the regions are illuminated by direct or indirect lights.
\end{enumerate}

\begin{table}[]
    \centering
    \begin{tabular}{cccccc}
        \toprule
                & NDR & NDRMC & MII & NeILF& Ours \\
         \midrule
         Bell   & 25.05 & 24.30 & 21.58 & \underline{25.40} & \textbf{31.05} \\
         Cat    & \underline{24.65} & 23.88 & 23.46 & 23.04 & \textbf{30.03} \\
         Teapot & 19.19 & 22.45 & 21.76 & \underline{24.49} & \textbf{30.95} \\
         Potion & 21.95 & 22.07 & \underline{25.62} & 24.86 & \textbf{31.17}  \\
         TBell  & 16.10 & 22.60 & 19.27 & \underline{22.97} & \textbf{27.48}  \\
         Angel  & 22.90 & 22.89 & 22.47 & \underline{24.56} & \textbf{25.49} \\
         Horse  & 25.56 & \underline{26.42} & 20.46 & 25.97 & \textbf{27.41}  \\
         Luyu   & 23.72 & 23.60 & 23.42 & \underline{24.62} & \textbf{26.61}  \\
         \midrule
         Avg.   & 22.39 & 23.53 & 22.28 & \underline{24.49} &  \textbf{28.77} \\
         \bottomrule
    \end{tabular}
    \caption{\textbf{Relighting quality in PSNR$\uparrow$ on the Glossy-Blender dataset}. We compare our method with NDR~\cite{munkberg2022extracting}, NDRMC~\cite{hasselgren2022shape}, MII~\cite{zhang2022modeling} and NeILF~\cite{yao2022neilf}. \textbf{Bold} means the best performance while \underline{underline} means the second best.}
    \label{tab:syn_relight}
\end{table}
\noindent{\textbf{BRDF}}. For the evaluation of BRDF estimation, we report the quality of relighted images of different methods. The quantitative results in PSNR are reported in Table~\ref{tab:syn_relight} and some qualitative results are shown in Fig.~\ref{fig:syn_relight}. The results of SSIM and LPIPS are included in Sec.~\ref{sec:ssim_lpips}. NDR~\cite{munkberg2022extracting} mainly suffers from incorrectly reconstructed surfaces, which prevents it from accurately estimating BRDF. NDRMC tends to produce rough materials with blurred reflections for these glossy objects. MII represents both lights and BRDF with Spherical Gaussian, which saves computations but is unable to represent high-frequency environment lights and smooth reflective BRDFs. NeILF also produces rough materials because it only uses a uniform sampling on the upper half sphere and has difficulty in converging to a smooth surface. 
In comparison with NDRMC using only 16 ray samples, we directly use MLP to decode indirect lights and thus are able to use 512 ray samples for the cosine lobe and 256 ray samples for the specular lobe, which makes our method accurately capture the materials.
In comparison with NeILF which only uses a Fibonacci ray sampling on the upper half-sphere, we adopt the importance sampling on both the cosine diffuse lobe and the specular lobe. This is essential for recovering the smooth material because the specular lobe is small but vital for the reflective appearance and Fibonacci ray sampling often fails to sample rays in the small specular lobe. 
Note that our method does not simply predict smooth materials but also correctly recovers the rough materials on the ``Potion'' row 1 of Fig.~\ref{fig:syn_relight}.

\subsection{Results on Glossy-Real}

\begin{figure*}
    \centering
    \setlength\tabcolsep{1pt}
    \renewcommand{\arraystretch}{0.5} 
    \begin{tabular}{cccccc}
        \includegraphics[width=0.16\textwidth]{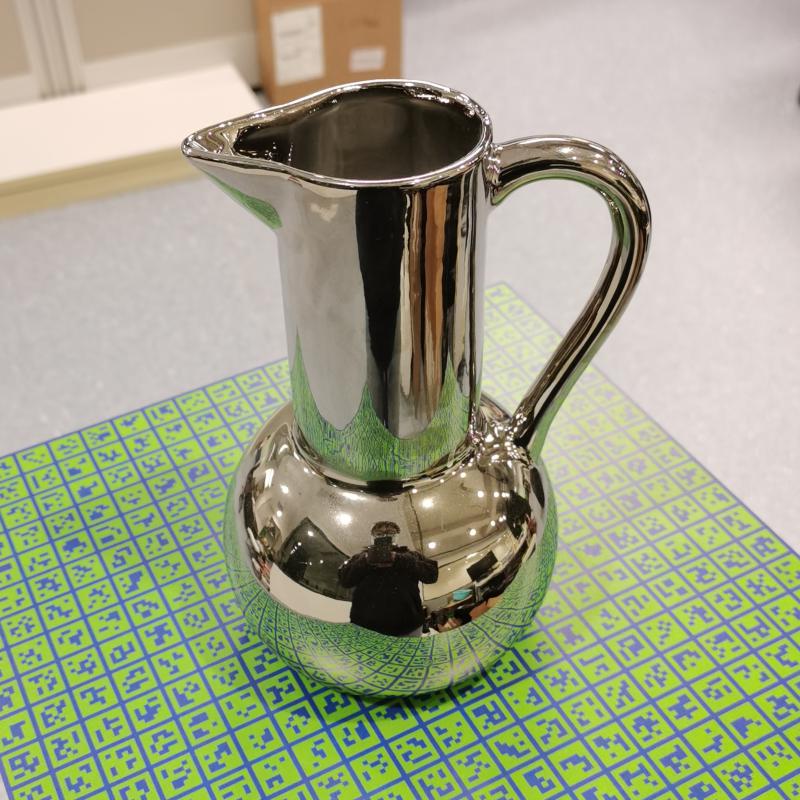} &
        \includegraphics[width=0.16\textwidth]{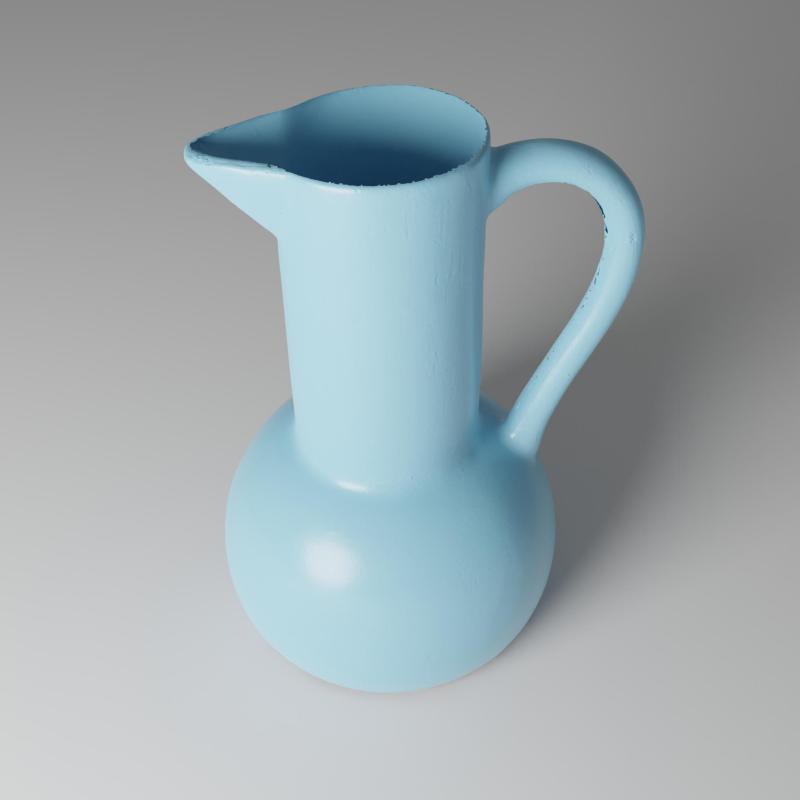} &
        \includegraphics[width=0.16\textwidth]{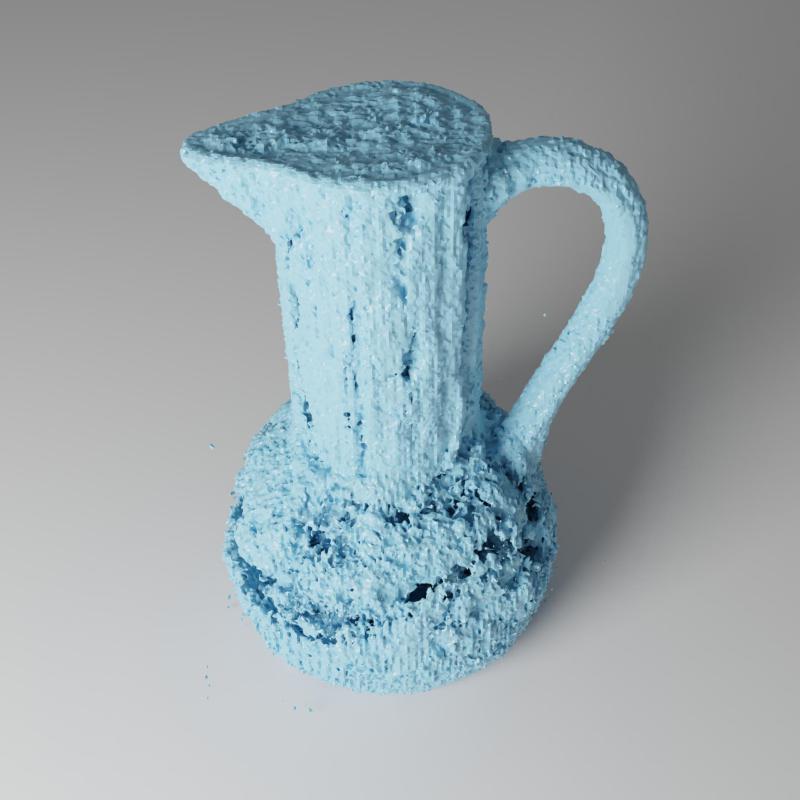} &
        \includegraphics[width=0.16\textwidth]{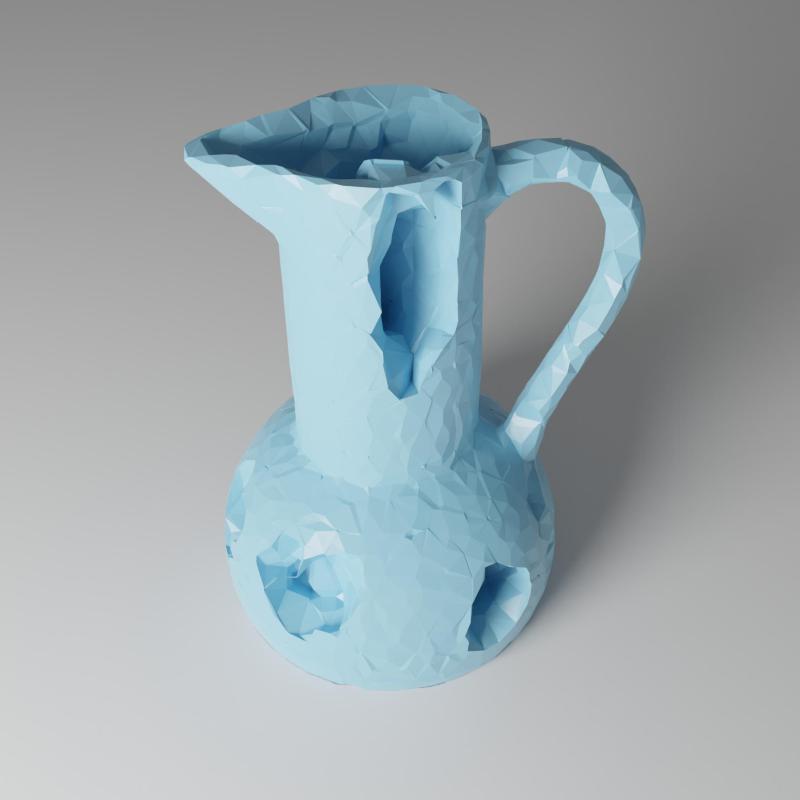} &
        \includegraphics[width=0.16\textwidth]{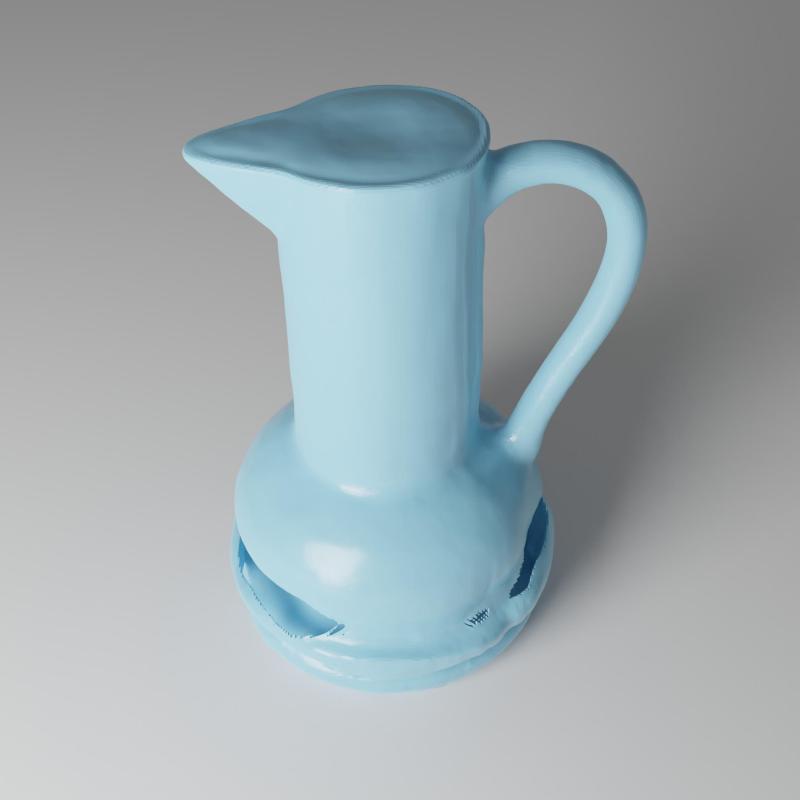} &
        \includegraphics[width=0.16\textwidth]{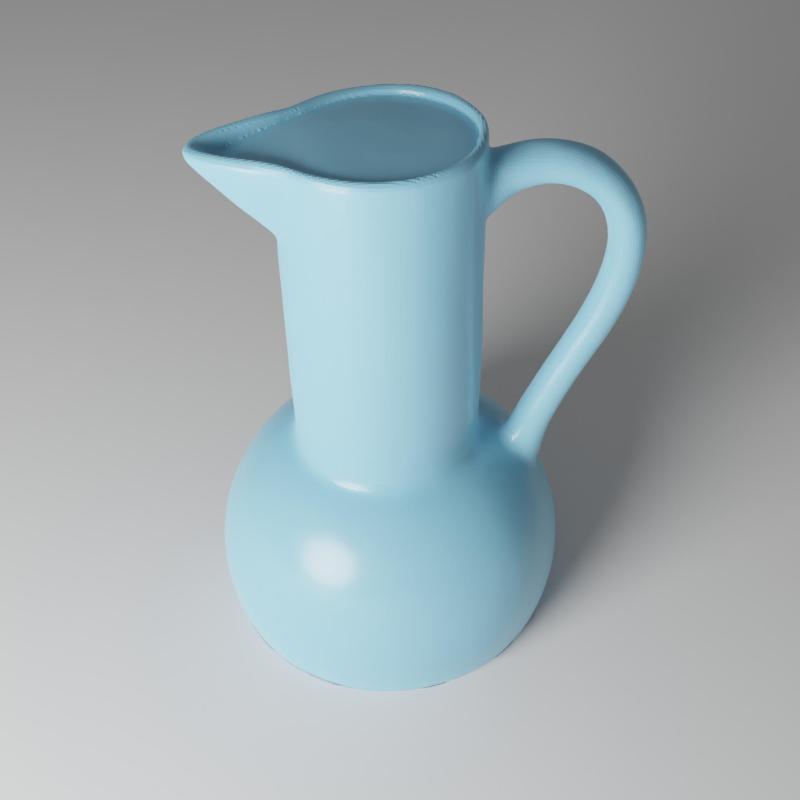} \\
        \includegraphics[width=0.16\textwidth]{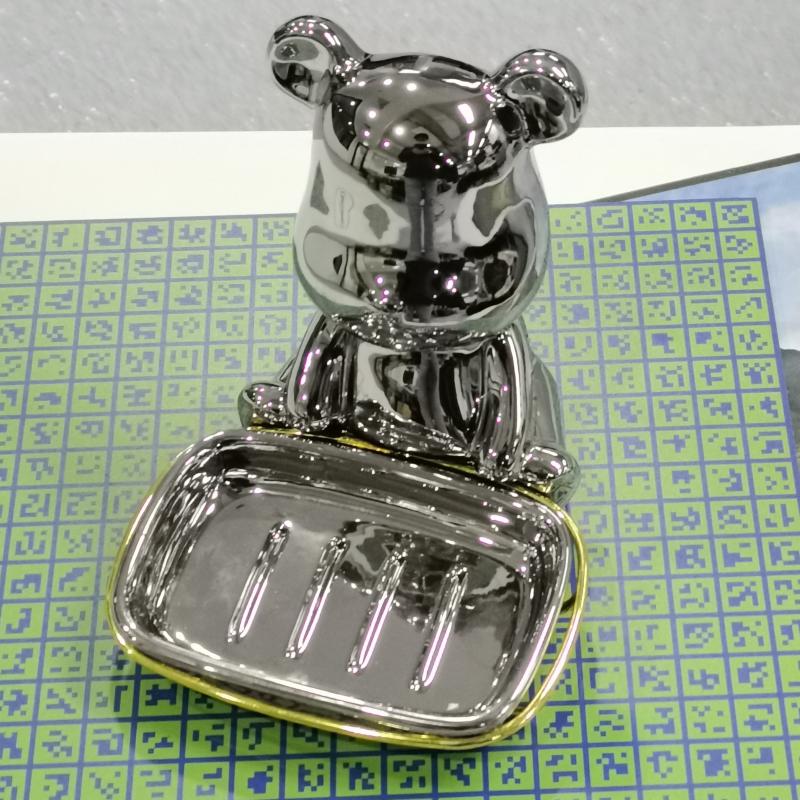} &
        \includegraphics[width=0.16\textwidth]{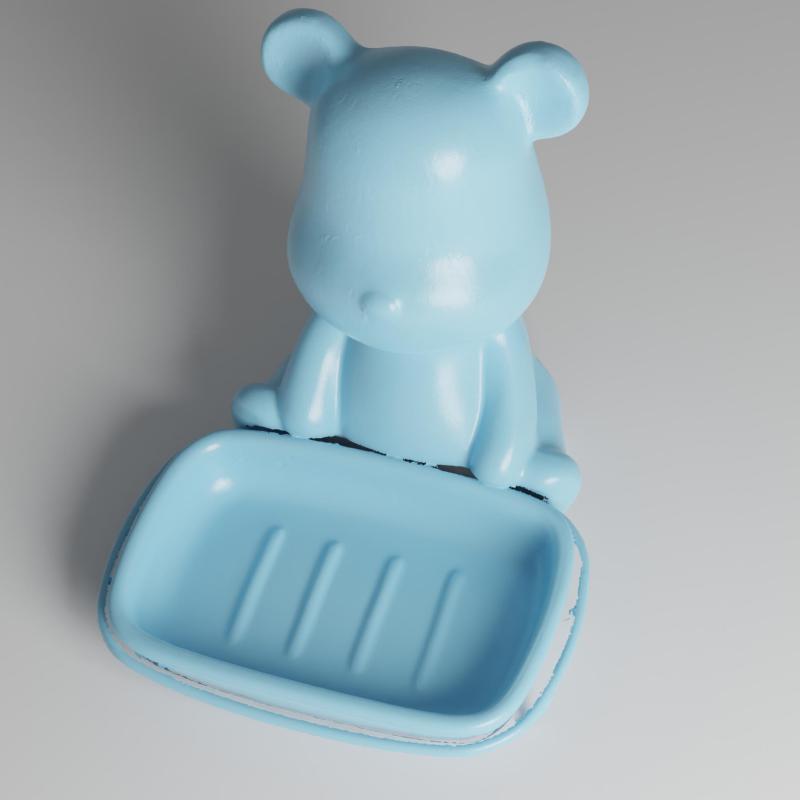} &
        \includegraphics[width=0.16\textwidth]{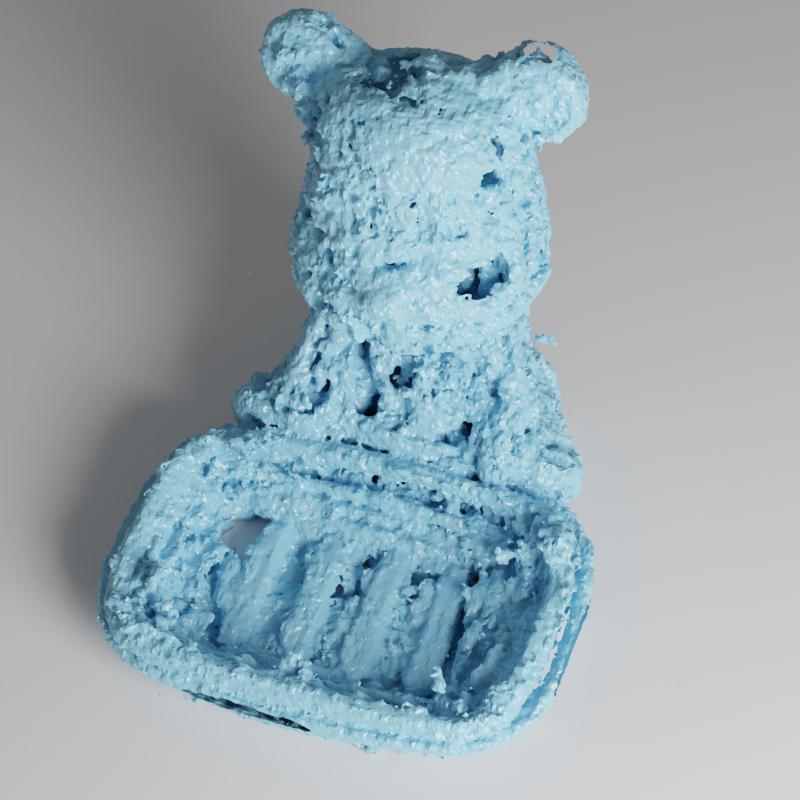} &
        \includegraphics[width=0.16\textwidth]{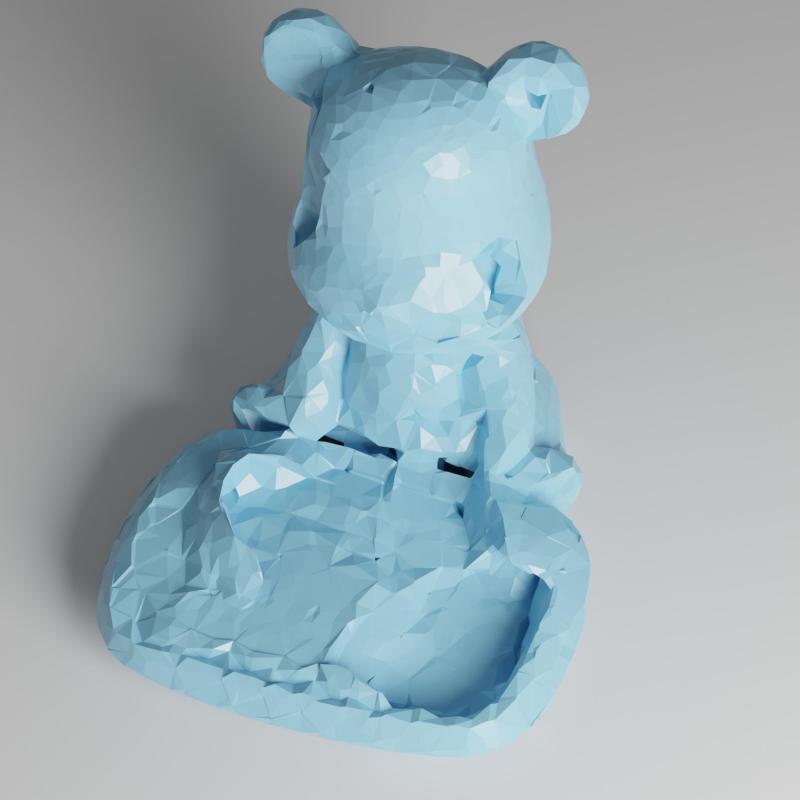} &
        \includegraphics[width=0.16\textwidth]{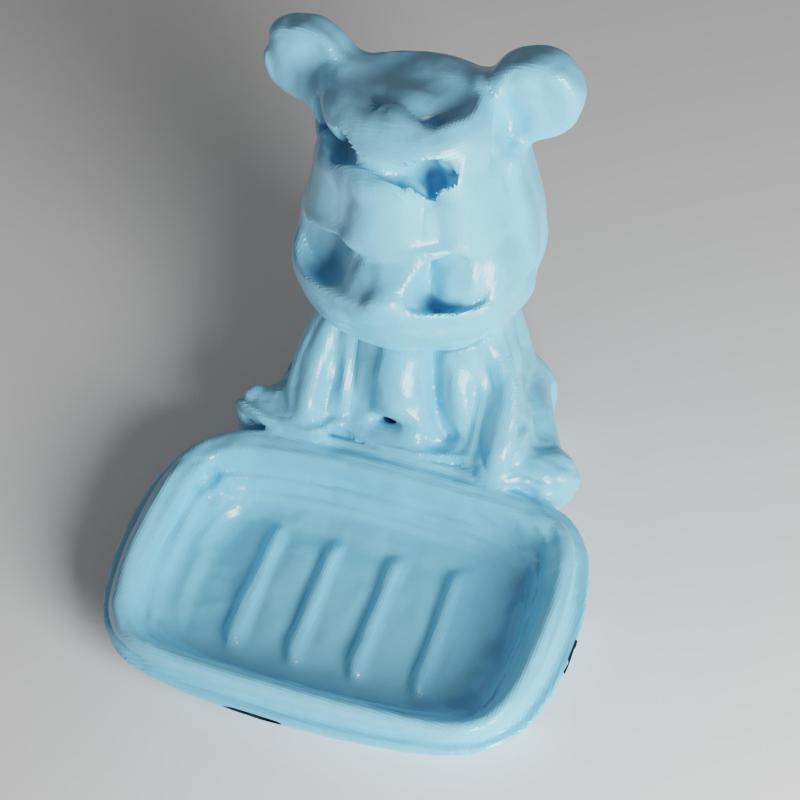} &
        \includegraphics[width=0.16\textwidth]{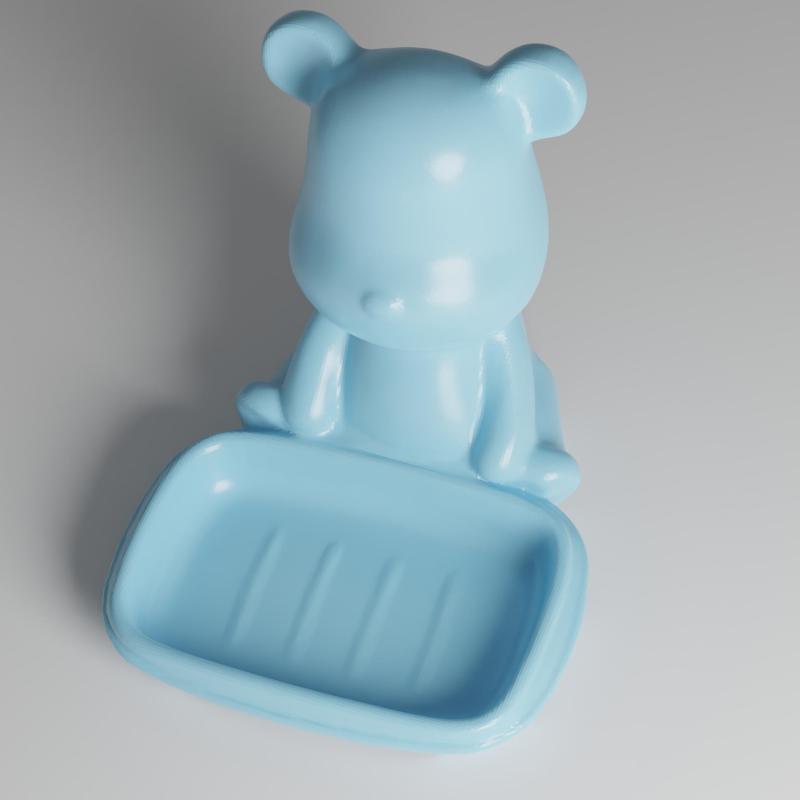} \\
        \includegraphics[width=0.16\textwidth]{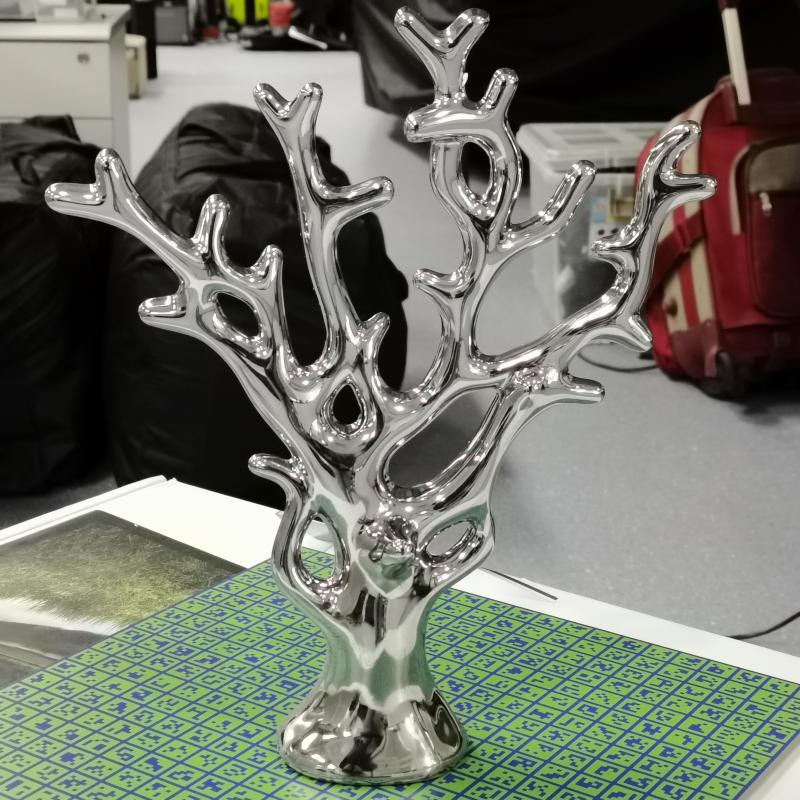} &
        \includegraphics[width=0.16\textwidth]{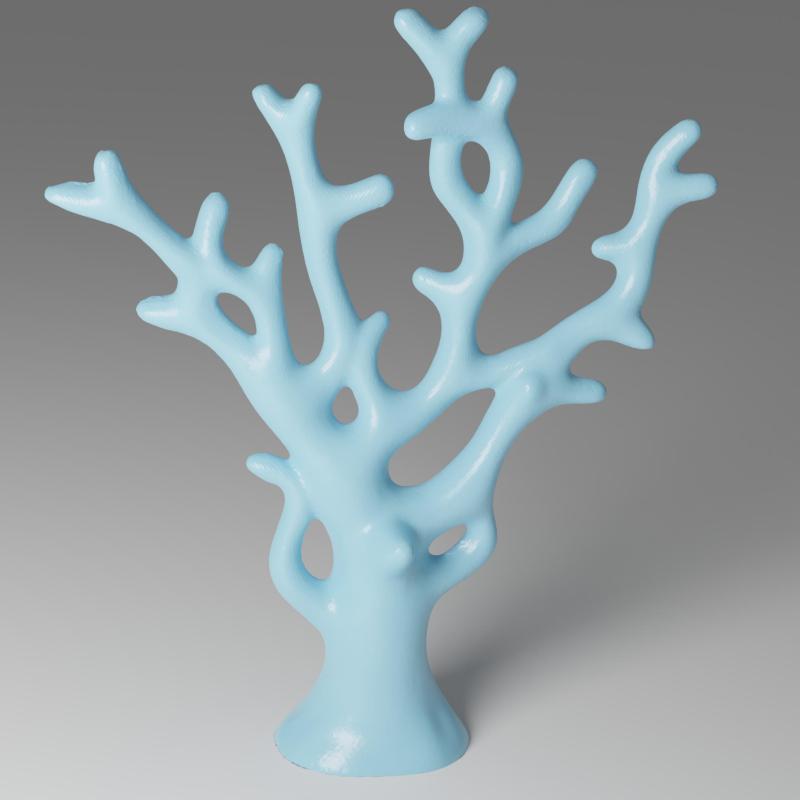} &
        \includegraphics[width=0.16\textwidth]{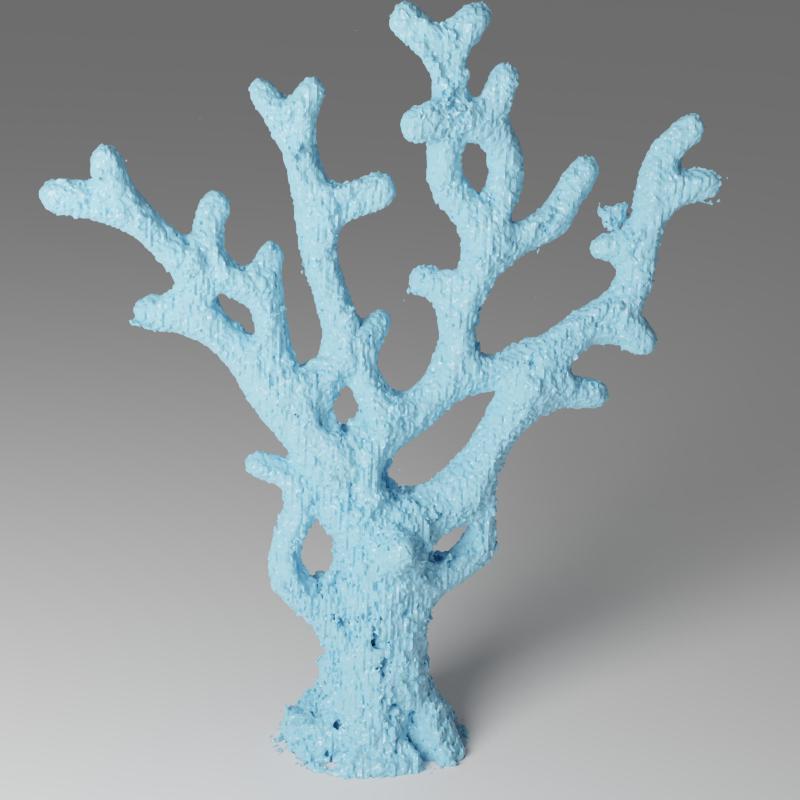} &
        \includegraphics[width=0.16\textwidth]{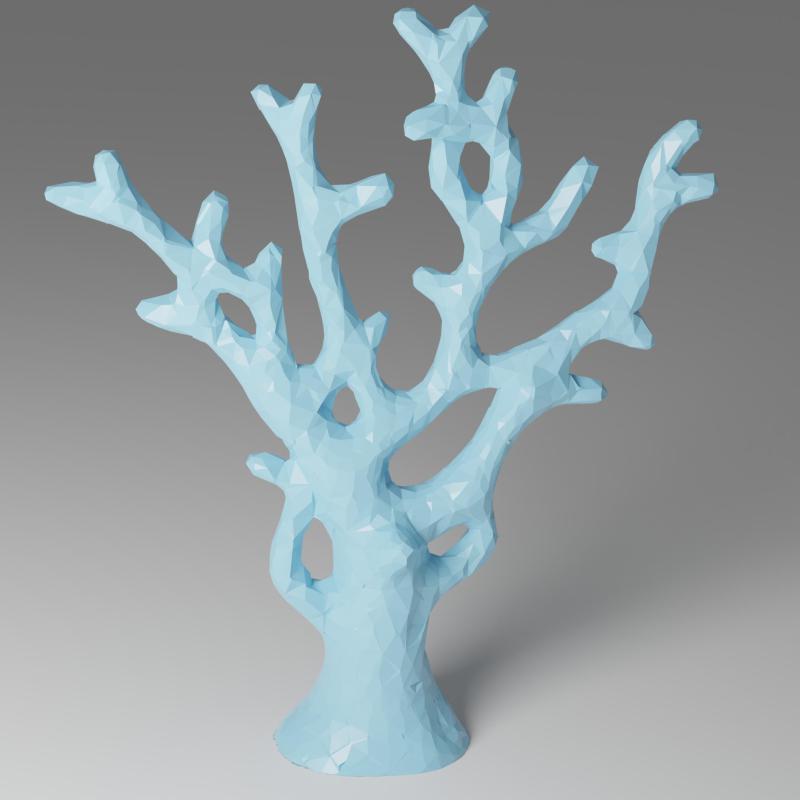} &
        \includegraphics[width=0.16\textwidth]{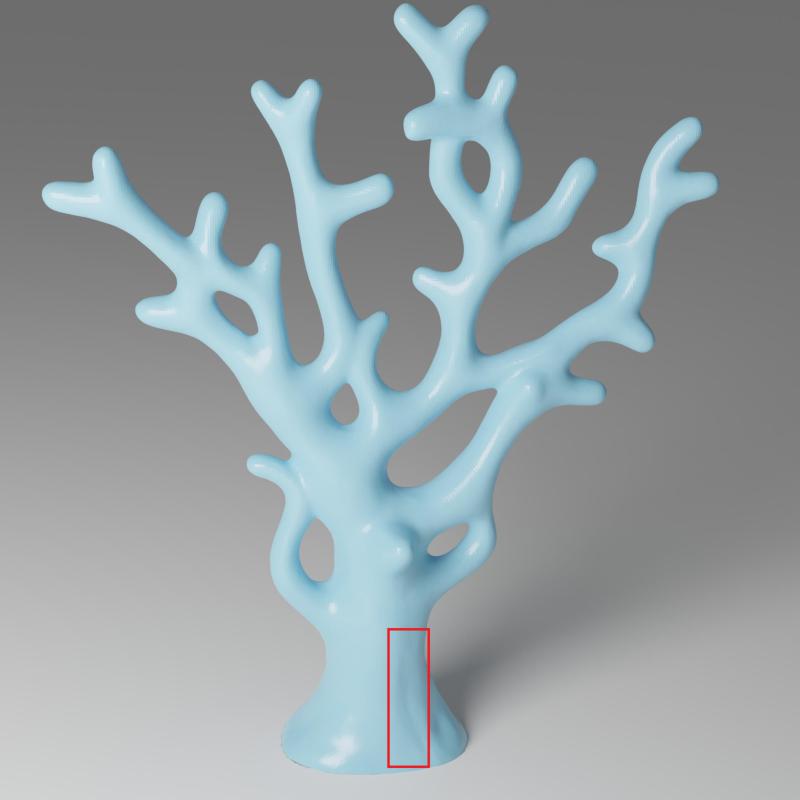} &
        \includegraphics[width=0.16\textwidth]{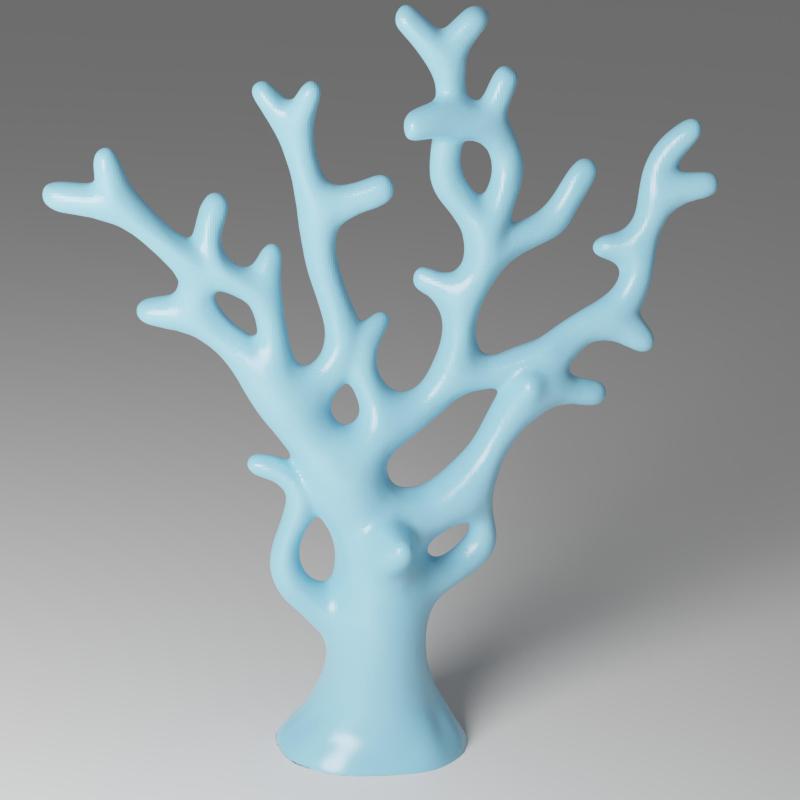} \\
        \includegraphics[width=0.16\textwidth]{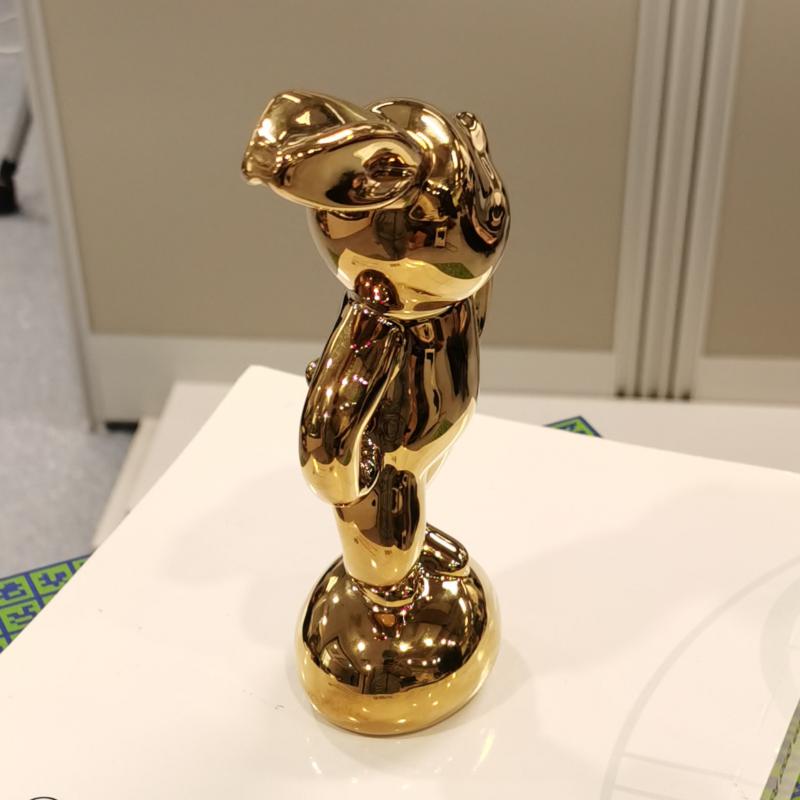} &
        \includegraphics[width=0.16\textwidth]{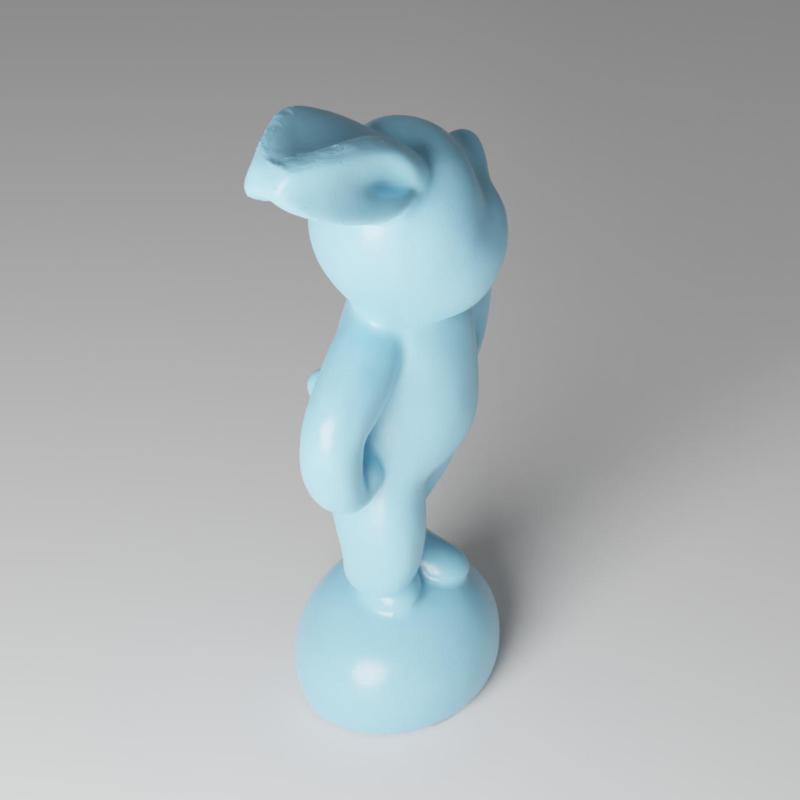} &
        \includegraphics[width=0.16\textwidth]{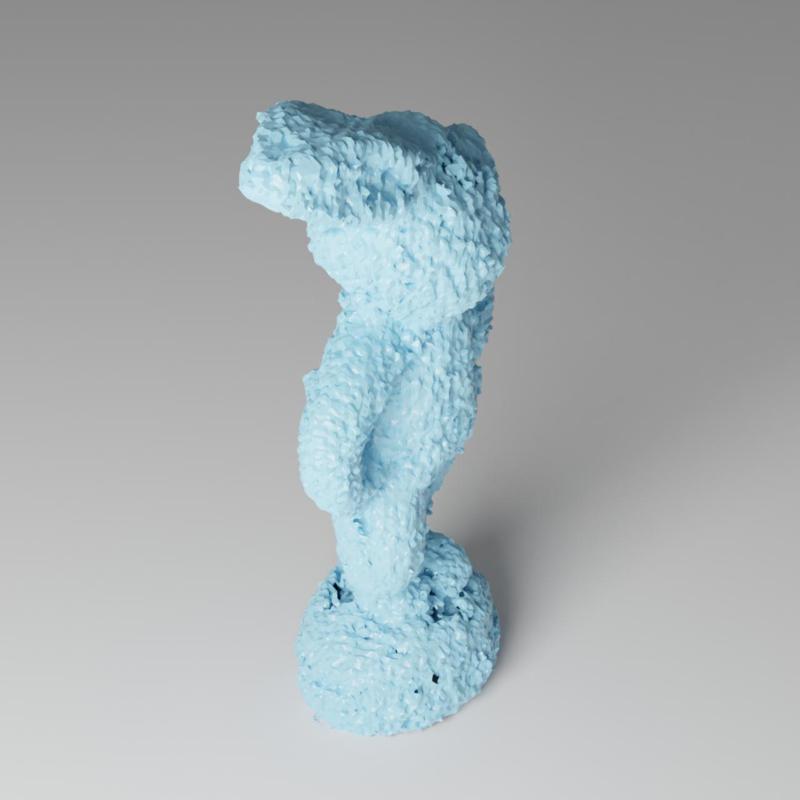} &
        \includegraphics[width=0.16\textwidth]{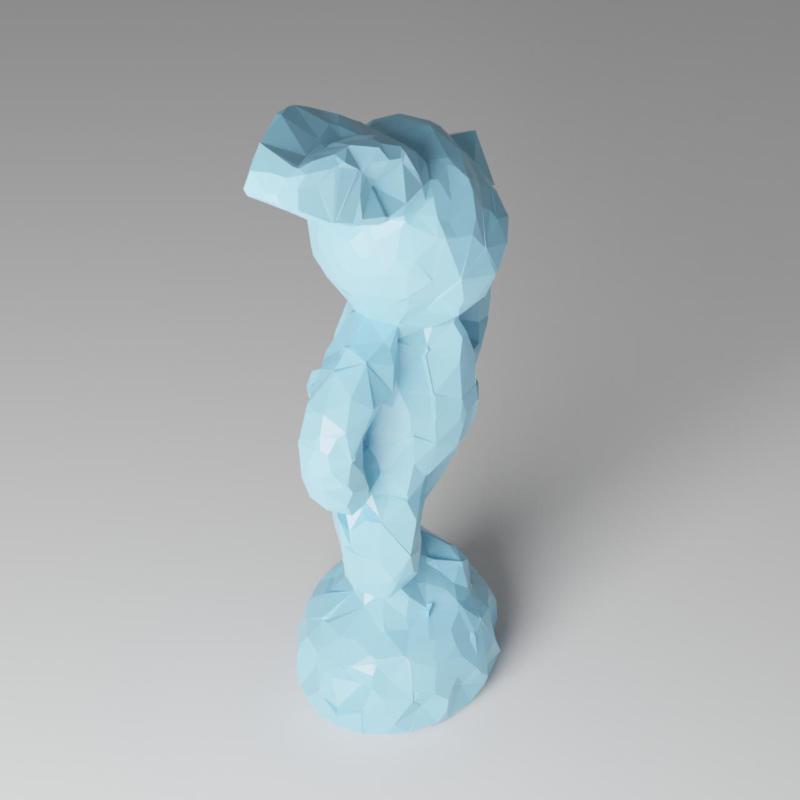} &
        \includegraphics[width=0.16\textwidth]{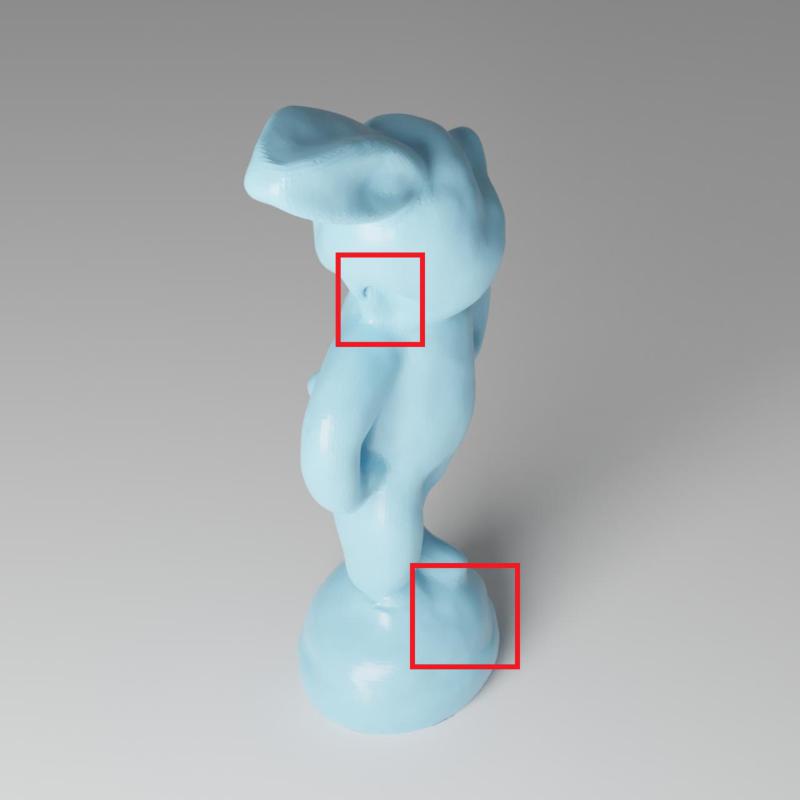} &
        \includegraphics[width=0.16\textwidth]{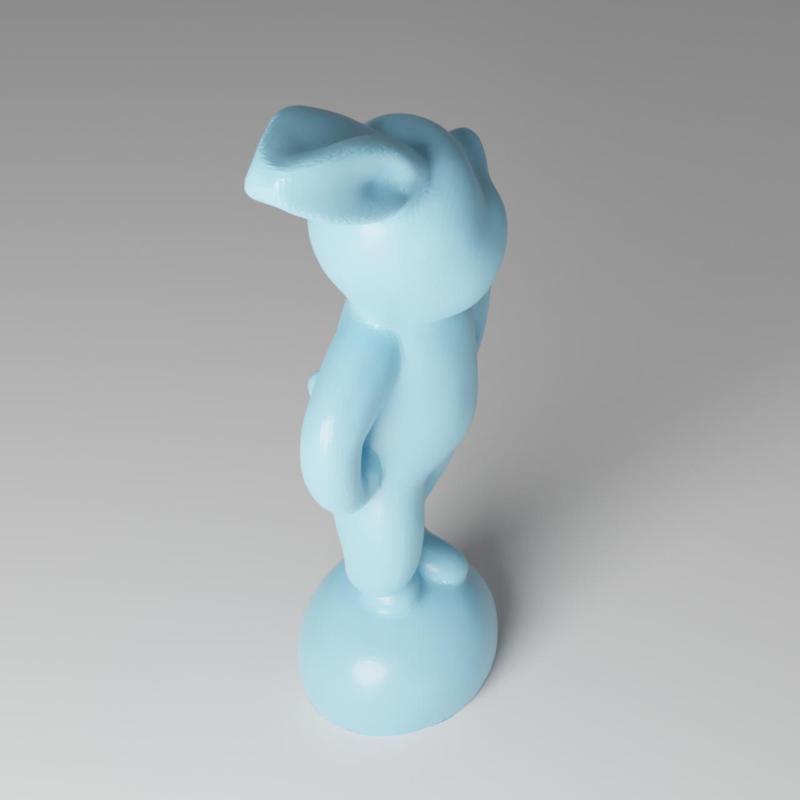} \\
        \includegraphics[width=0.16\textwidth]{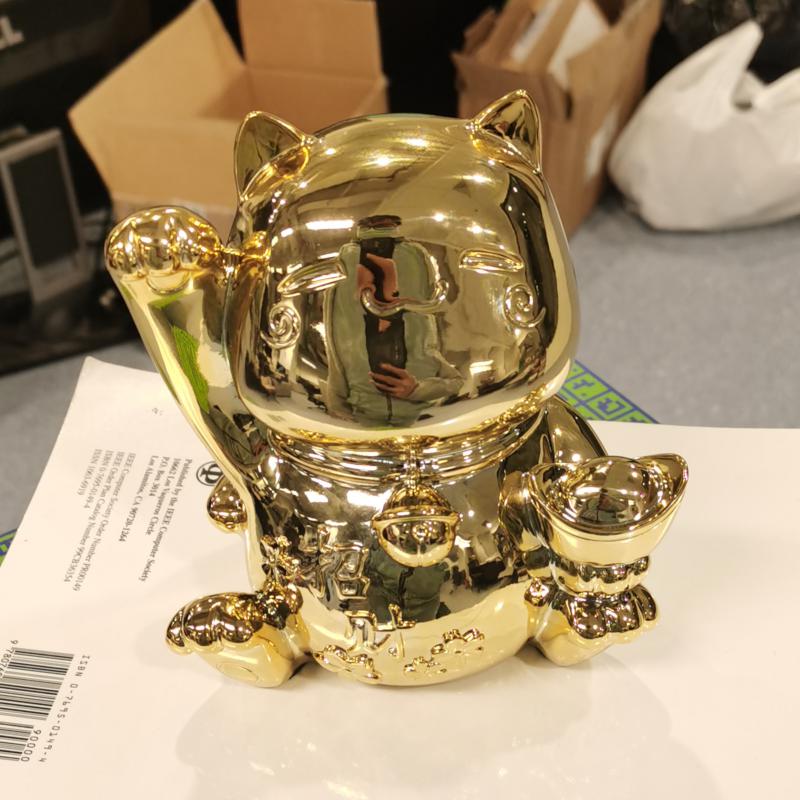} &
        \includegraphics[width=0.16\textwidth]{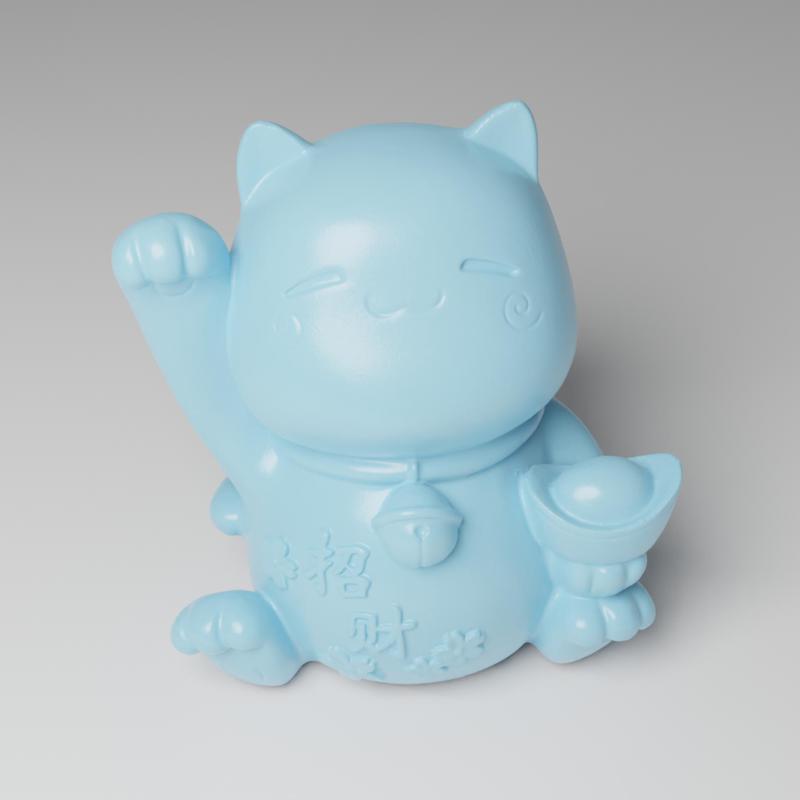} &
        \includegraphics[width=0.16\textwidth]{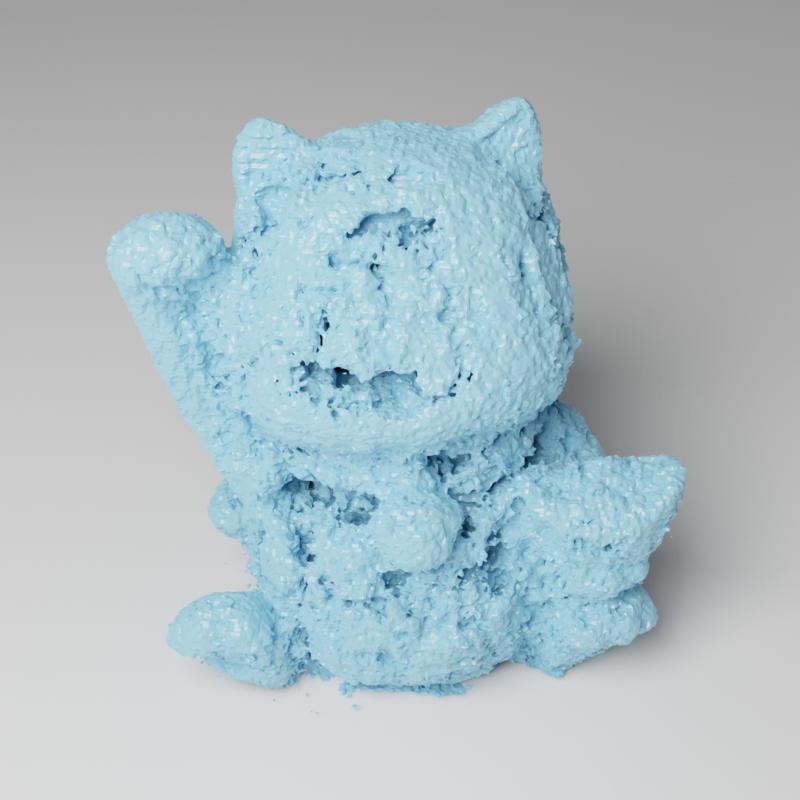} &
        \includegraphics[width=0.16\textwidth]{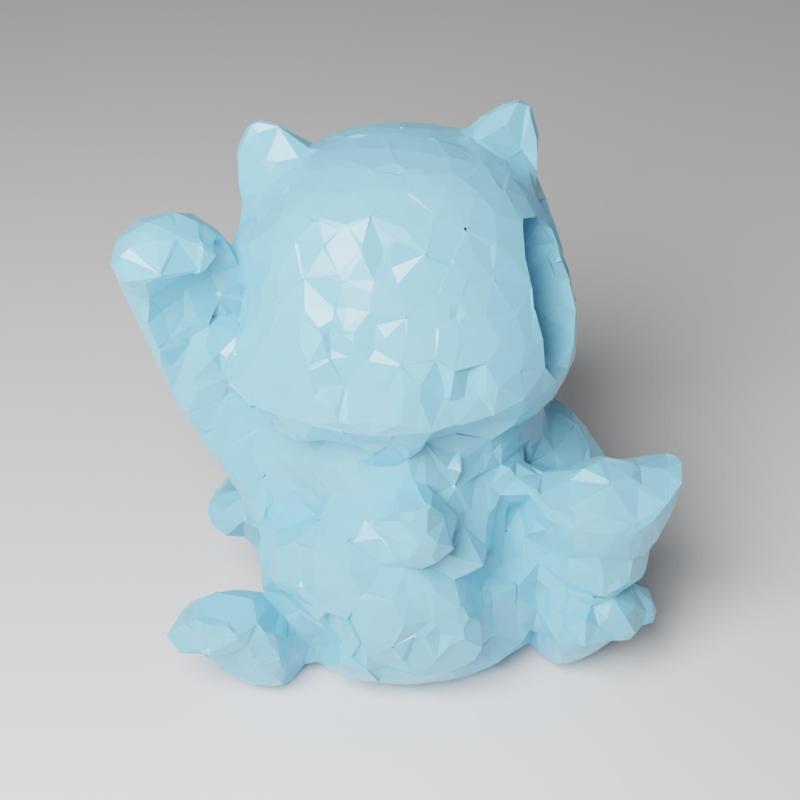} &
        \includegraphics[width=0.16\textwidth]{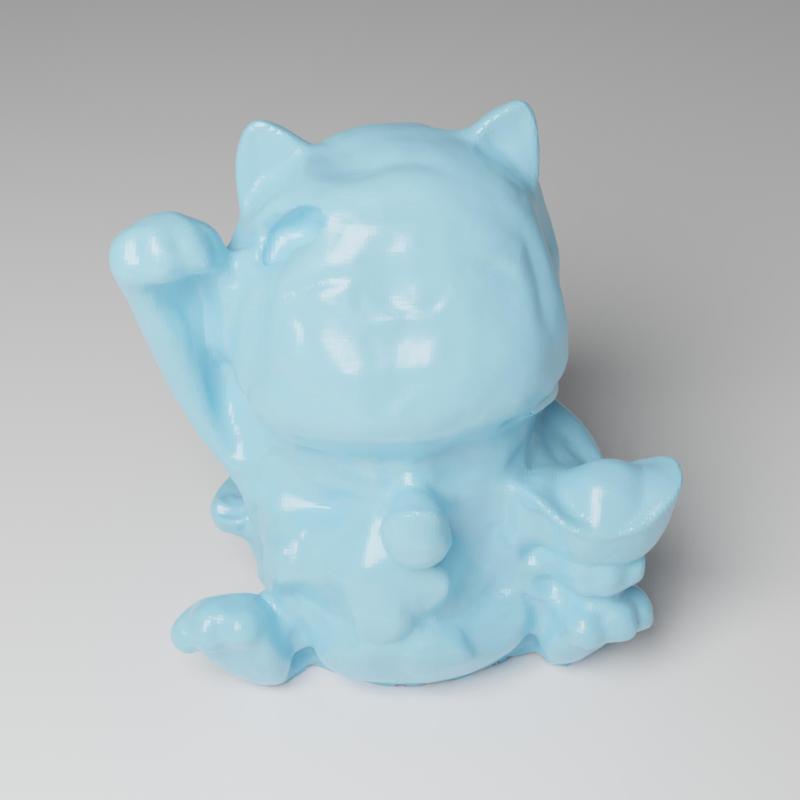} &
        \includegraphics[width=0.16\textwidth]{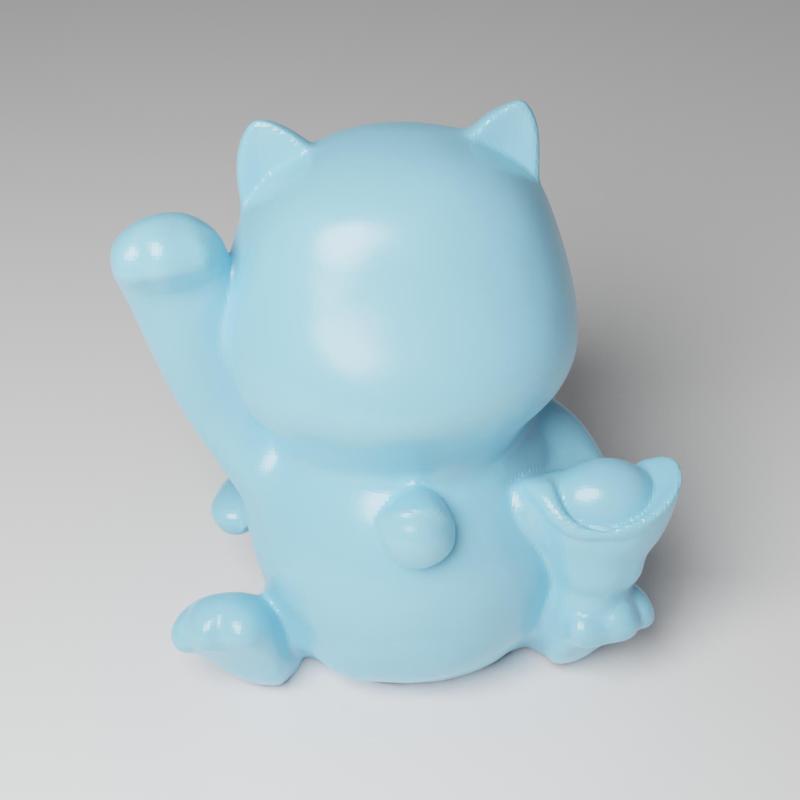} \\
        (a) Images & (b) Ground-truth & (c) Ref-NeRF & (d) NDRMC$^*$ & (e) NeuS & (f) Ours \\
    \end{tabular}
    \caption{\textbf{Images, ground-truth and reconstructed surfaces of the Glossy-Real dataset}. We compare our results with Ref-NeRF~\cite{verbin2022ref}, NDRMC~\cite{hasselgren2022shape}, and NeuS~\cite{wang2021neus}. $^*$NDRMC is trained with ground-truth object masks while the other methods do not use object masks.}
    \label{fig:real_visual}
\end{figure*}

\begin{table}[]
    \centering
    \resizebox{\linewidth}{!}{
    \begin{tabular}{cccccc}
    \toprule
          &NDR$^*$ & NDRMC$^*$ & NeuS &  Ref-NeRF & Ours \\
    \midrule
    Bunny  & 0.0047  &   0.0042  &  \underline{0.0022}    &   0.0054   &   \textbf{0.0012}   \\ 
    Coral  & 0.0025  &   0.0022  &  \underline{0.0016}    &   0.0052   &   \textbf{0.0014}   \\ 
    Maneki & 0.0148  &   0.0117  &  0.0091    &  \underline{0.0084}    &   \textbf{0.0024}   \\ 
    Bear   & 0.0104  &   0.0118  &  \underline{0.0074}    &   0.0109   &   \textbf{0.0033}   \\ 
    Vase   &  0.0201 &   \underline{0.0058}  &  0.0101    &   0.0091   &   \textbf{0.0011}   \\ 
    \midrule
    Avg.   &  0.0105 & 0.0071    &   \underline{0.0061}   &   0.0078   &   \textbf{0.0019}   \\ 
    \bottomrule
    \end{tabular}
    }
    \caption{\textbf{Reconstruction quality in Chamfer Distance (CD$\downarrow$) on the Glossy-Real dataset}. We compare our method with NeuS~\cite{wang2021neus}, Ref-NeRF~\cite{verbin2022ref}, NDR~\cite{munkberg2022extracting} and NDRMC~\cite{hasselgren2022shape}. $^*$NDR and NDRMC use ground-truth object masks while the other methods do not use object masks. \textbf{Bold} means the best performance and \underline{underline} means the second best performance.}
    \label{tab:real}
\end{table}

\textbf{Geometry}. The quantitative results in CD on the Glossy-Real dataset are reported in Table~\ref{tab:real} and qualitative results are shown in Fig.~\ref{fig:real_visual}. In general, the results on the Glossy-Real dataset are similar to the results on the Glossy-Blender dataset, where our method can accurately reconstruct all reflective surfaces while each baseline method fails to reconstruct several objects. On all the objects of the Glossy-Real dataset, surfaces reconstructed by Ref-NeRF are very noisy. On ``Vase'', ``Bear'' and ``Maneki'', NeuS incorrectly distorts the large reflective surfaces and on ``Coral'' and ``Bunny'', the reconstruction of NeuS is generally correct but still contains non-smooth artifacts on the regions highlighted by the red bounding boxes in Fig.~\ref{fig:real_visual}.
NDRMC can reconstruct the general shape with the help of object masks but also produces holes on the surfaces of ``Vase'', ``Bear'' and ``Maneki''. 
In comparison, our method can correctly reconstruct all objects. The main artifact of our method is that some details on the ``Maneki'' are missing.

\begin{figure*}
    \centering
    \setlength\tabcolsep{1pt}
    \renewcommand{\arraystretch}{0.5} 
    \begin{tabular}{cccccc}
    \multicolumn{6}{c}{\includegraphics[width=0.2\linewidth]{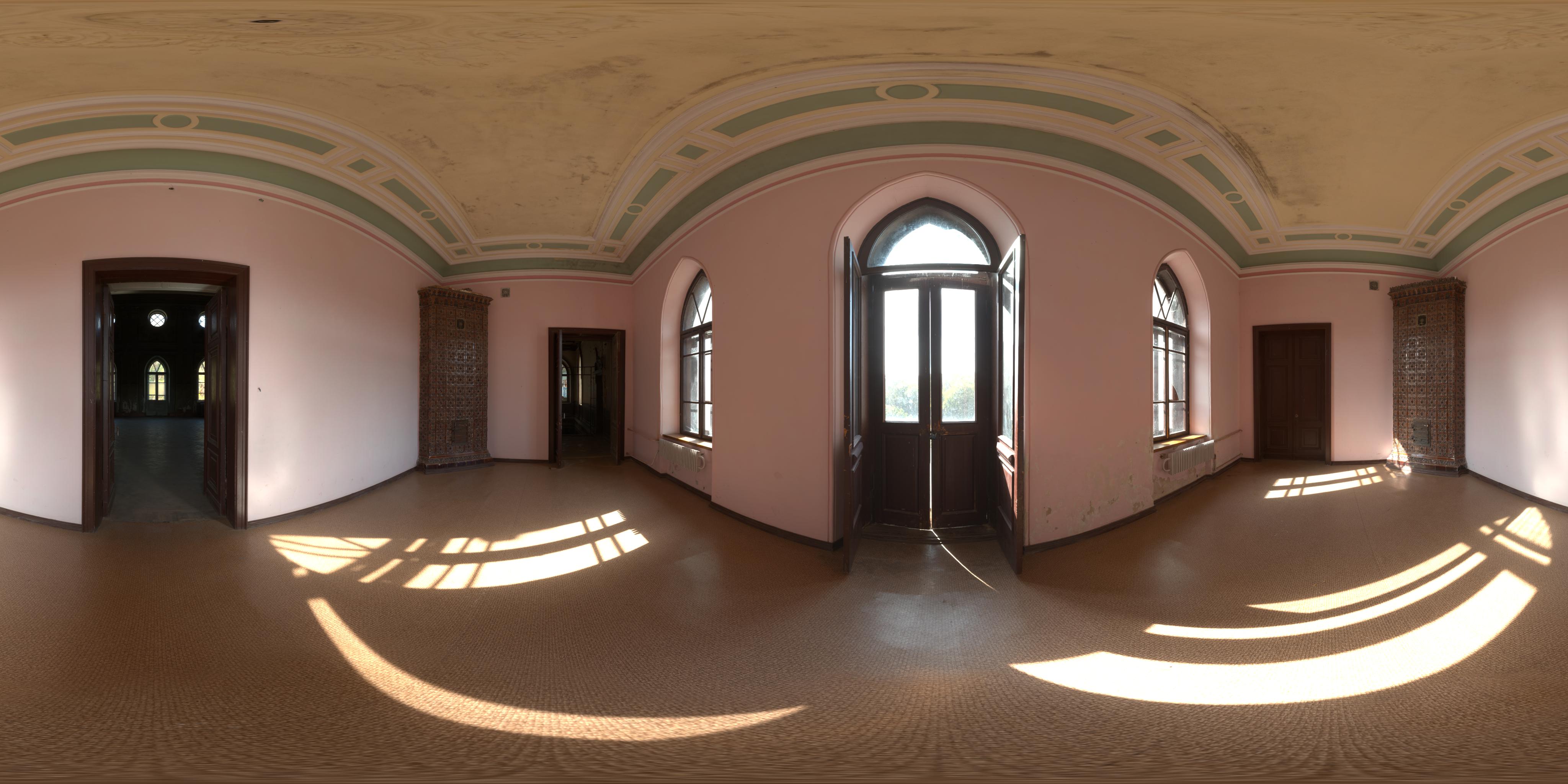}} \\
    \includegraphics[width=0.16\textwidth]{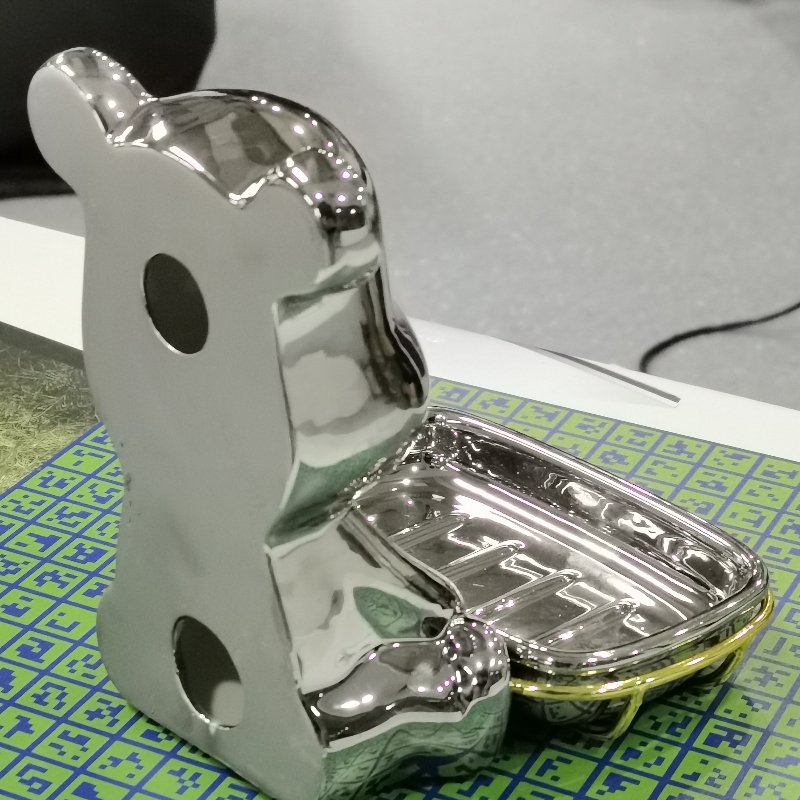} &
    \includegraphics[width=0.16\linewidth]{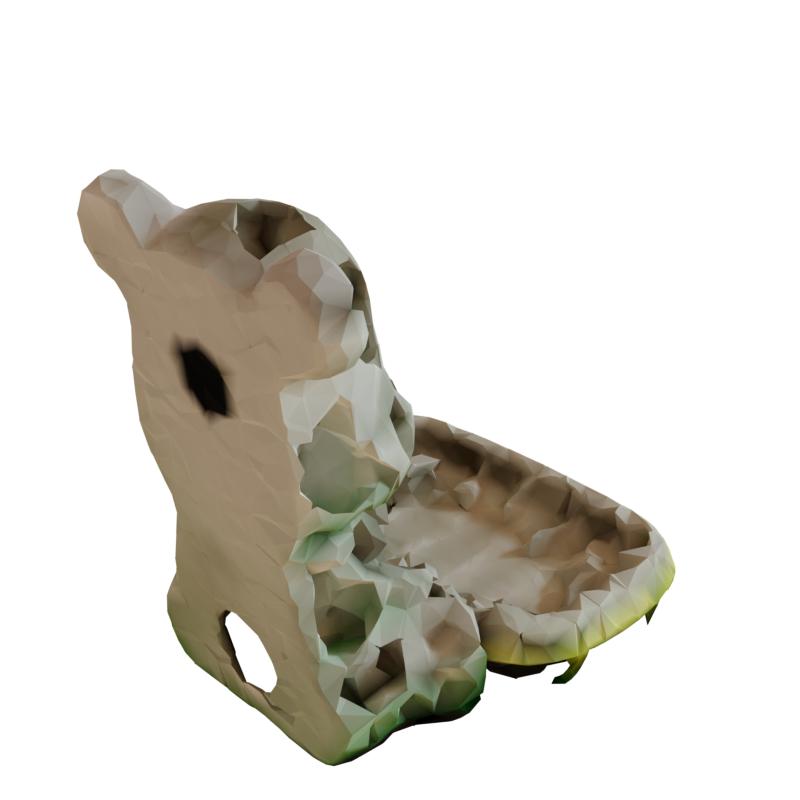} &
    \includegraphics[width=0.16\linewidth]{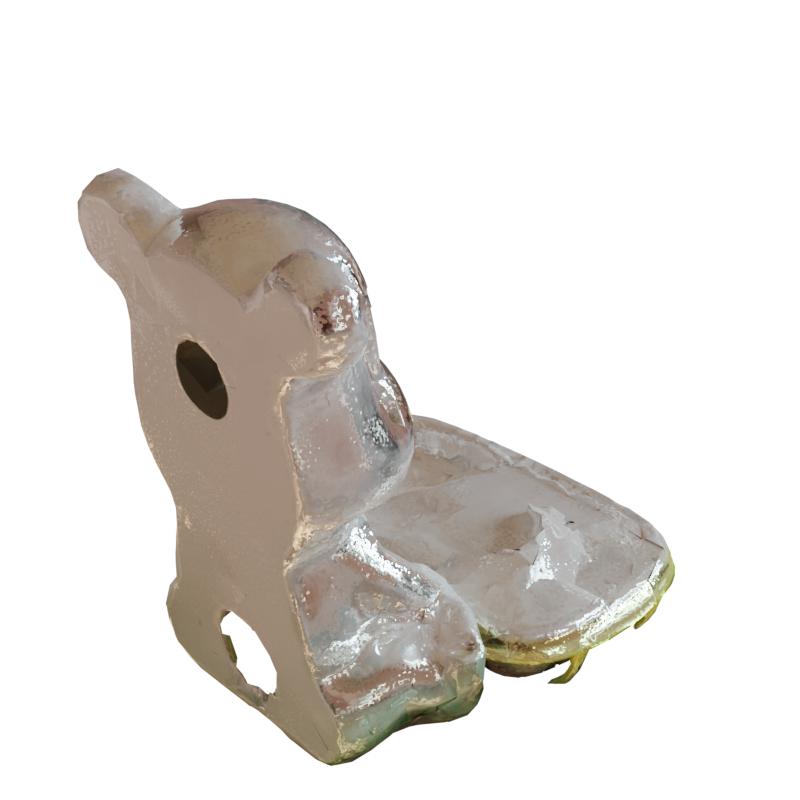} &
    \includegraphics[width=0.16\linewidth]{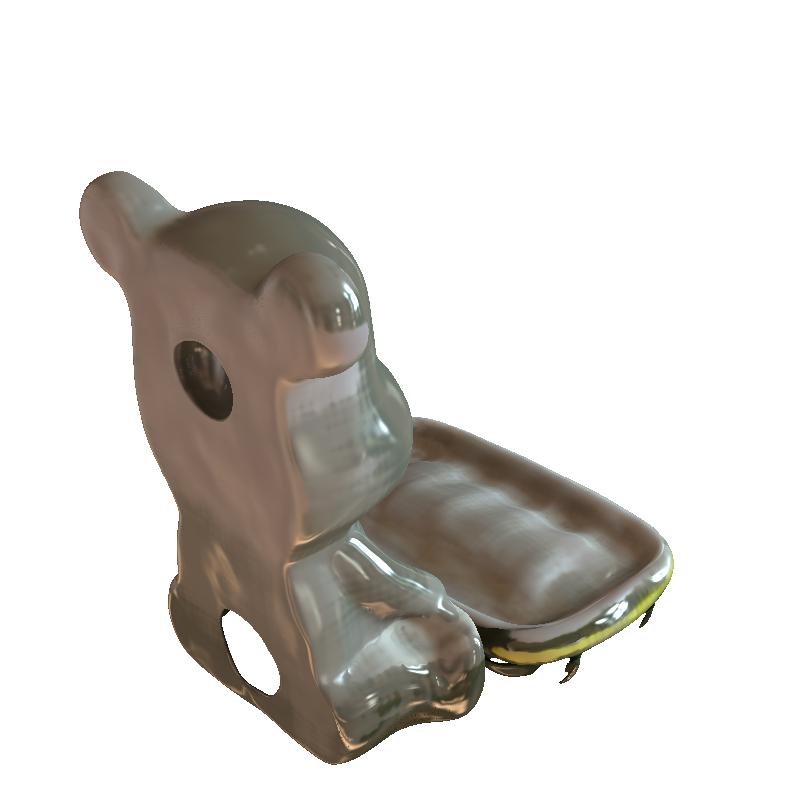} &
    \includegraphics[width=0.16\linewidth]{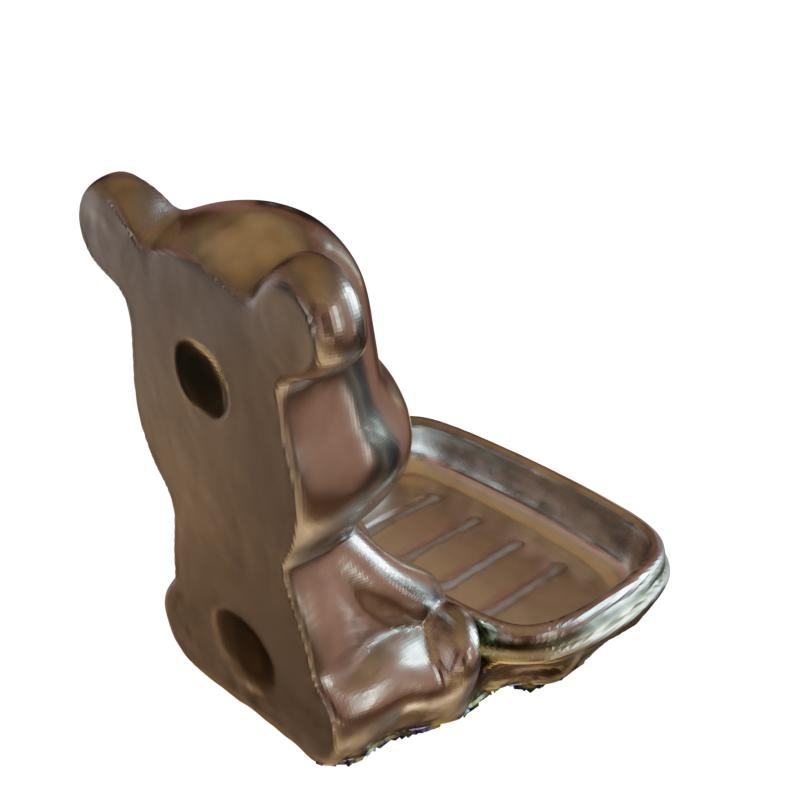} &
    \includegraphics[width=0.16\linewidth]{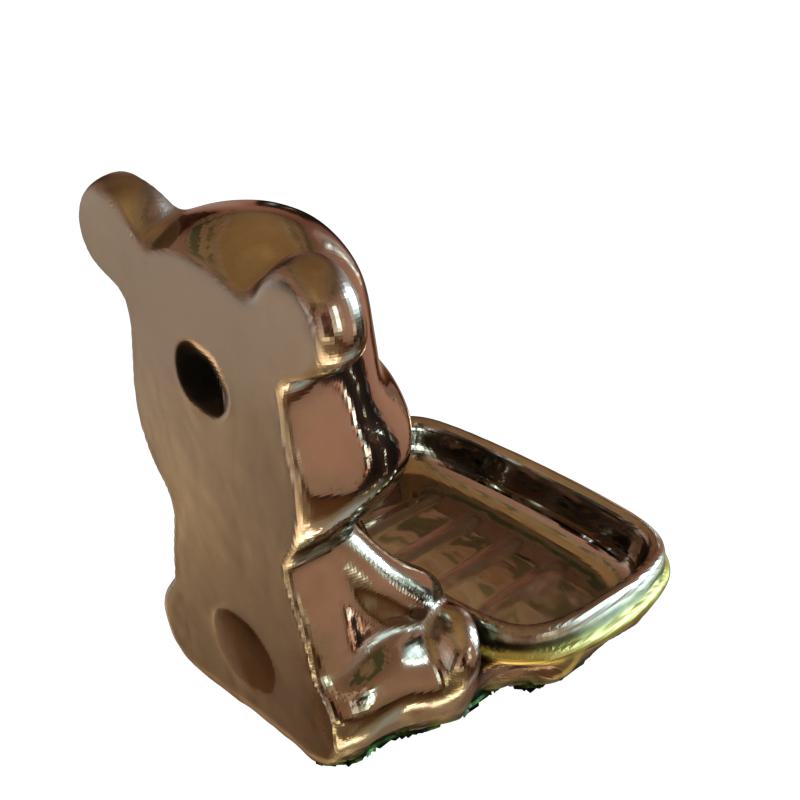} \\
    \multicolumn{6}{c}{\includegraphics[width=0.2\linewidth]{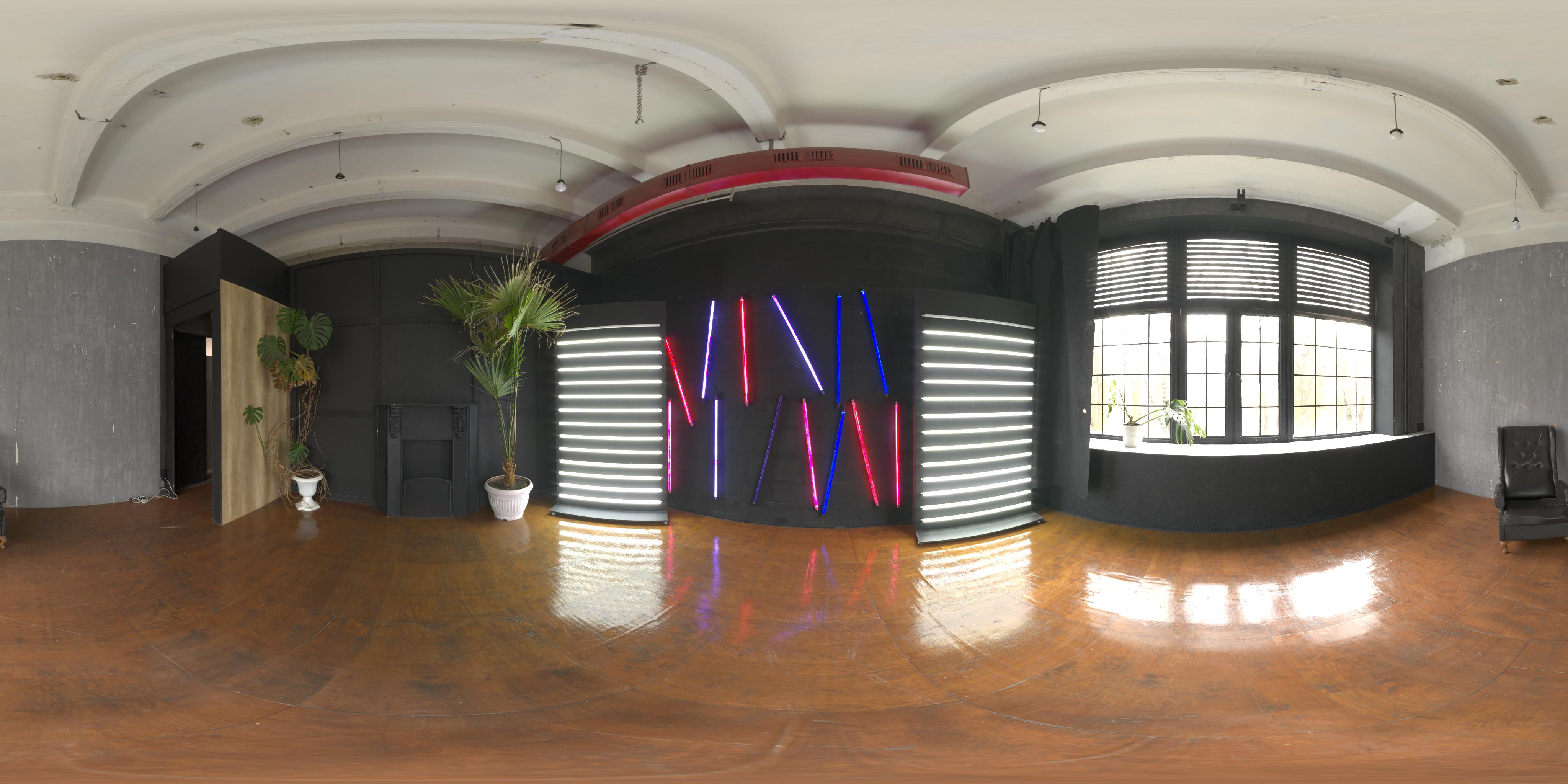}} \\
    \includegraphics[width=0.16\linewidth]{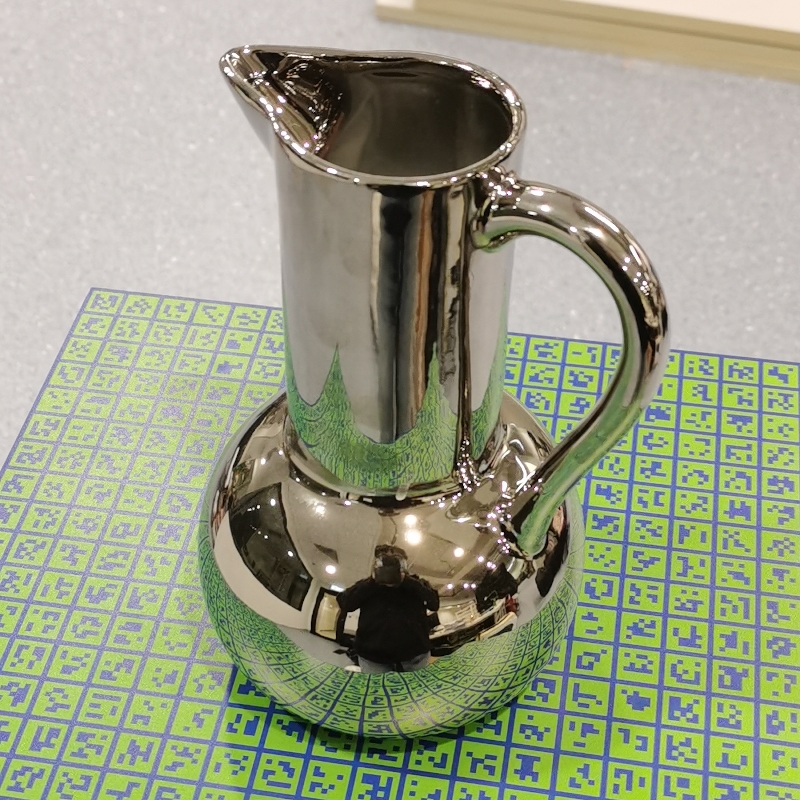} &
    \includegraphics[width=0.16\linewidth]{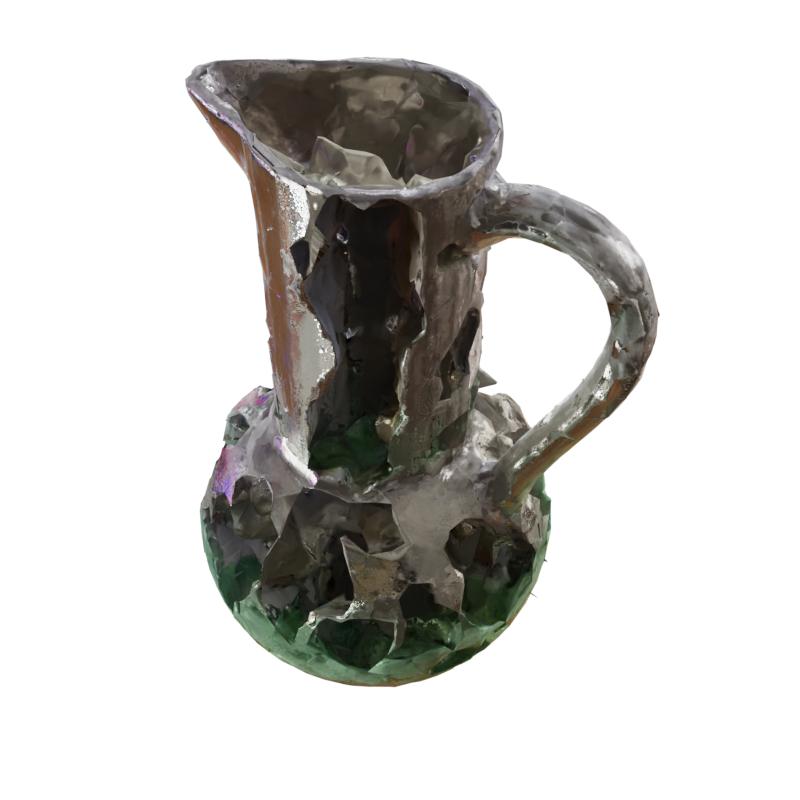} &
    \includegraphics[width=0.16\linewidth]{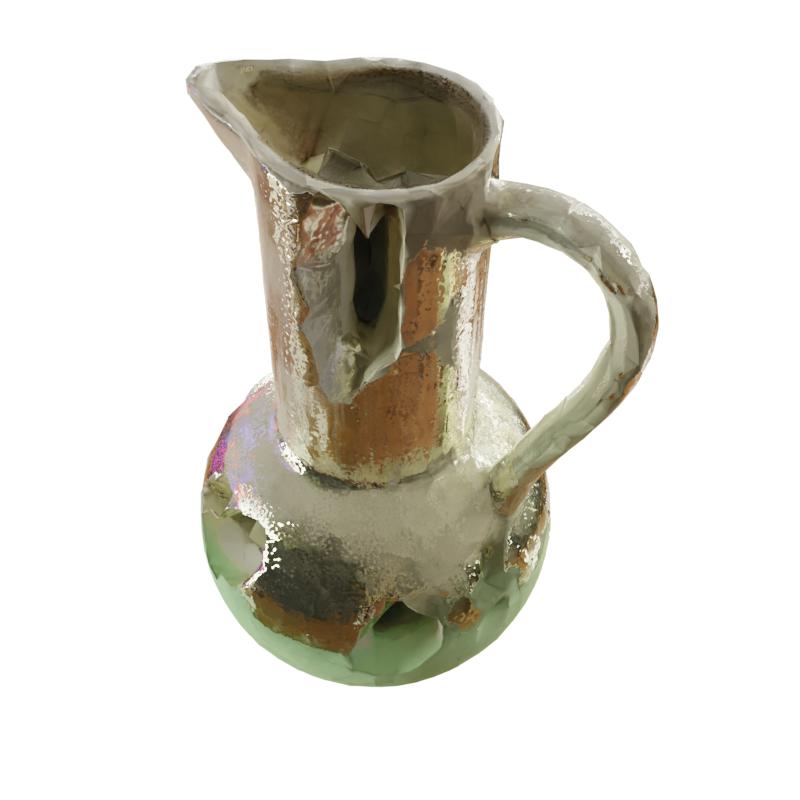} &
    \includegraphics[width=0.16\linewidth]{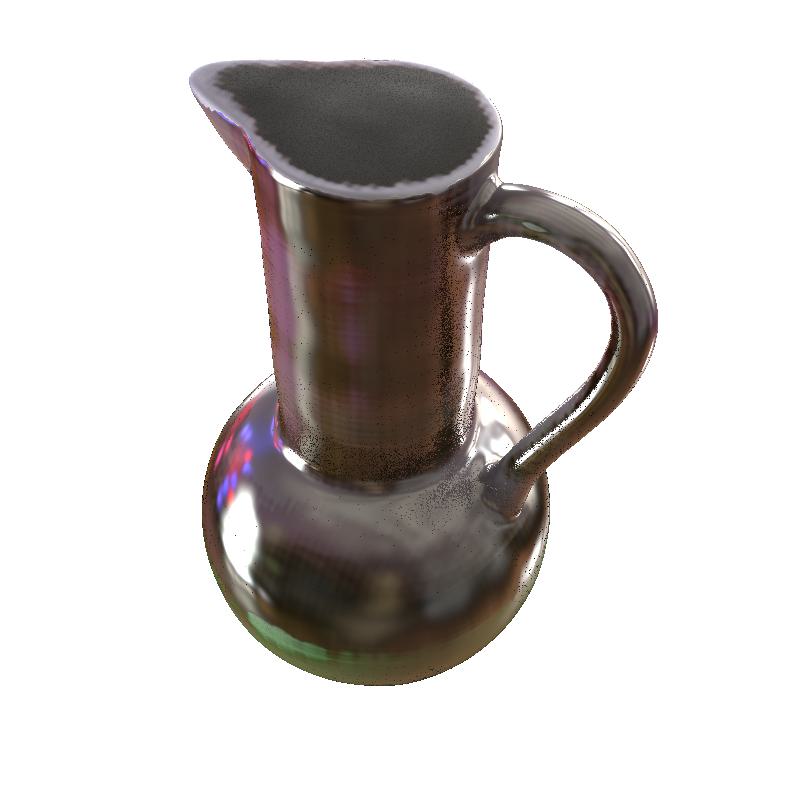} &
    \includegraphics[width=0.16\linewidth]{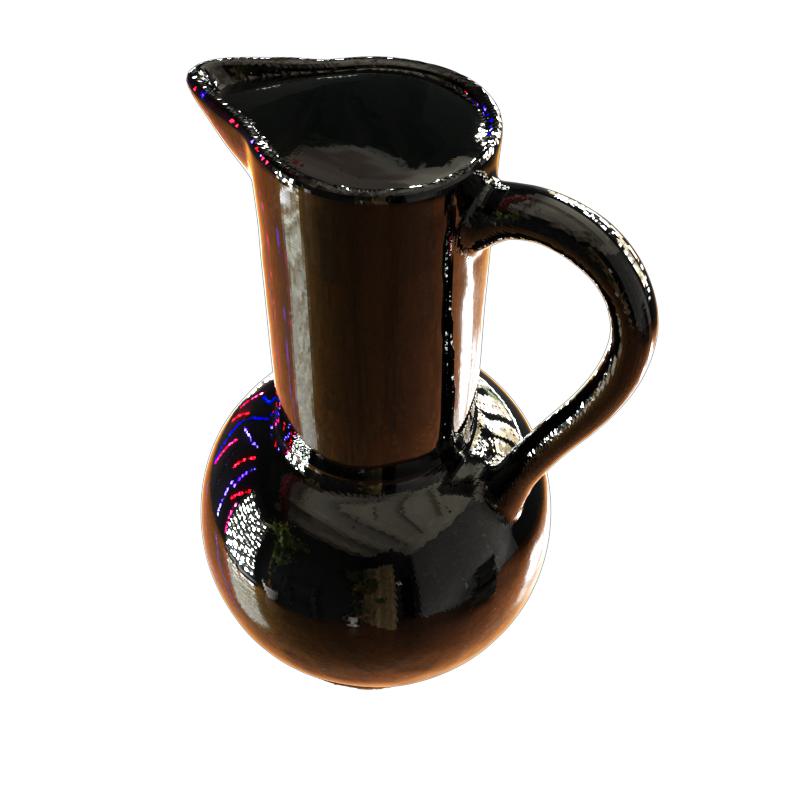} &
    \includegraphics[width=0.16\linewidth]{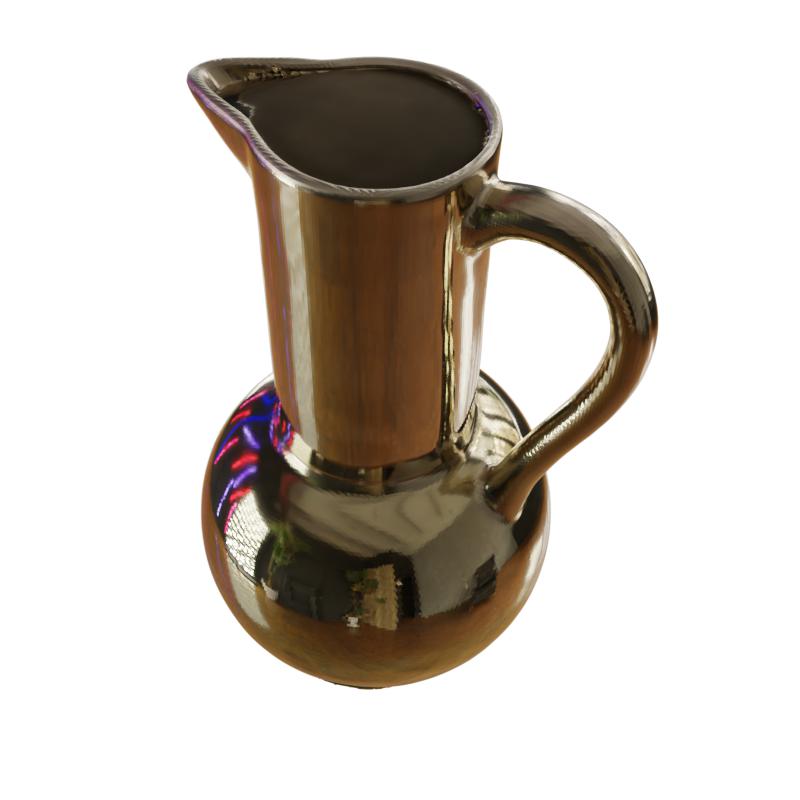} \\
    (a) Image & (b) NDR & (c) NDRMC & (d) MII & (e) NeILF & (f) Ours \\
    \end{tabular}
    \caption{\textbf{Relighting objects from the Glossy-Real dataset}. We provide a visual comparison with NDR~\cite{munkberg2022extracting}, NDRMC~\cite{hasselgren2022shape}, MII~\cite{zhang2022modeling}, and NeILF~\cite{yao2022neilf}. We provide the input image with the nearest viewpoint and the relighting HDR map as a reference.
    }
    \label{fig:real_relight}
\end{figure*}
\noindent{\textbf{BRDF}}. We provide qualitative relighting results in Fig.~\ref{fig:real_relight} to compare our method with MII~\cite{zhang2022modeling}, NeILF~\cite{yao2022neilf}, NDR~\cite{munkberg2022extracting} and NDRMC~\cite{hasselgren2022shape}. Similar to the relighting results on the synthetic dataset, our method produces more realistic relighting results than the two baselines. NDRMC tends to produce rough materials while MII estimates smooth but plastic-like materials for these smooth metallic objects. Our method accurately estimates the smooth metallic materials for all reflective surfaces and the rough materials for the back of ``Bear''.

\subsection{Ablation studies}
In this section, we conduct ablation studies on geometry reconstruction and BRDF estimation.

\begin{table}[]
    \centering
    \resizebox{\linewidth}{!}{
    \begin{tabular}{clcccc|c}
    \toprule
          ID & Description & Angel & Bell & Cat & Teapot  & Avg.\\
    \midrule
    0 & NeuS                               & 0.0035 & 0.0146 & 0.0278 & 0.0546 & 0.0251 \\
    1 & Only direct lights                 & 0.0033 & 0.0051 & 0.0217 & 0.0076 & 0.0094 \\
    2 & Only indirect lights               & 0.0037 & 0.0043 & 0.0210 & 0.0064 & 0.0089 \\
    3 & No occlusion loss $\ell_{\rm occ}$ & 0.0033 & 0.0036 & 0.0212 & 0.0287 & 0.0142 \\
    
    4 & No Eikonal loss $\ell_{\rm eikonal}$   & 0.0027 & 0.0030 & 0.0212 & 0.0184 & 0.0113 \\
    
    5 & Full model                         & 0.0034 & 0.0032 & 0.0044 & 0.0037 & 0.0037 \\
    \bottomrule
    \end{tabular}
    }
    \caption{\textbf{Ablation studies on the geometry reconstruction} using objects from the Glossy-Synthetic dataset. Chamfer Distance (CD$\downarrow$) is reported.}
    \label{tab:ab}
\end{table}
\begin{figure*}
    \centering
    \setlength\tabcolsep{1pt}
    \renewcommand{\arraystretch}{0.5} 
    \begin{tabular}{ccccccc}
        \includegraphics[width=0.14\textwidth]{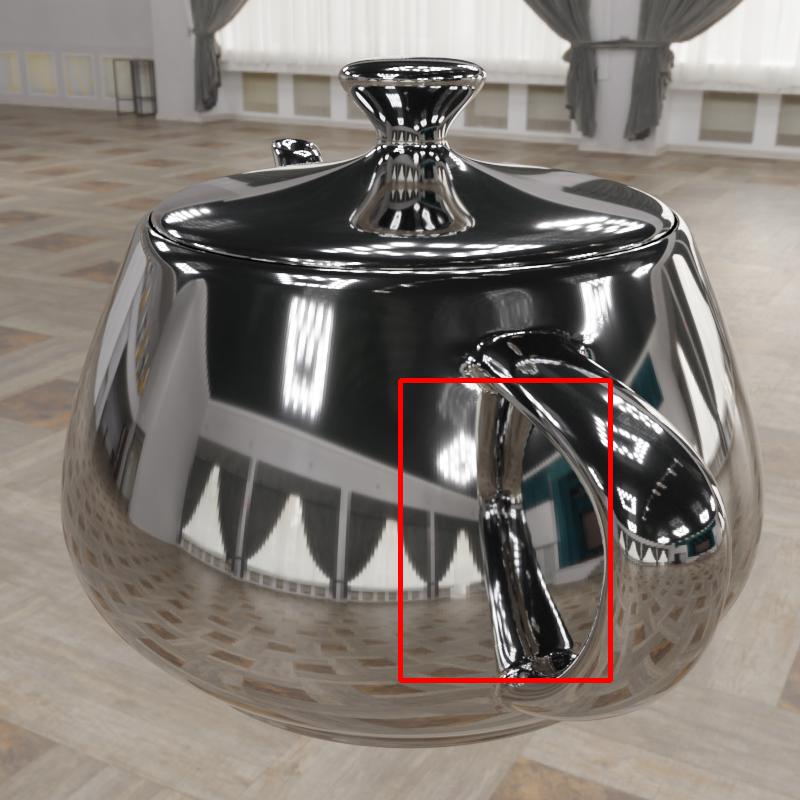} &
        \includegraphics[width=0.14\textwidth]{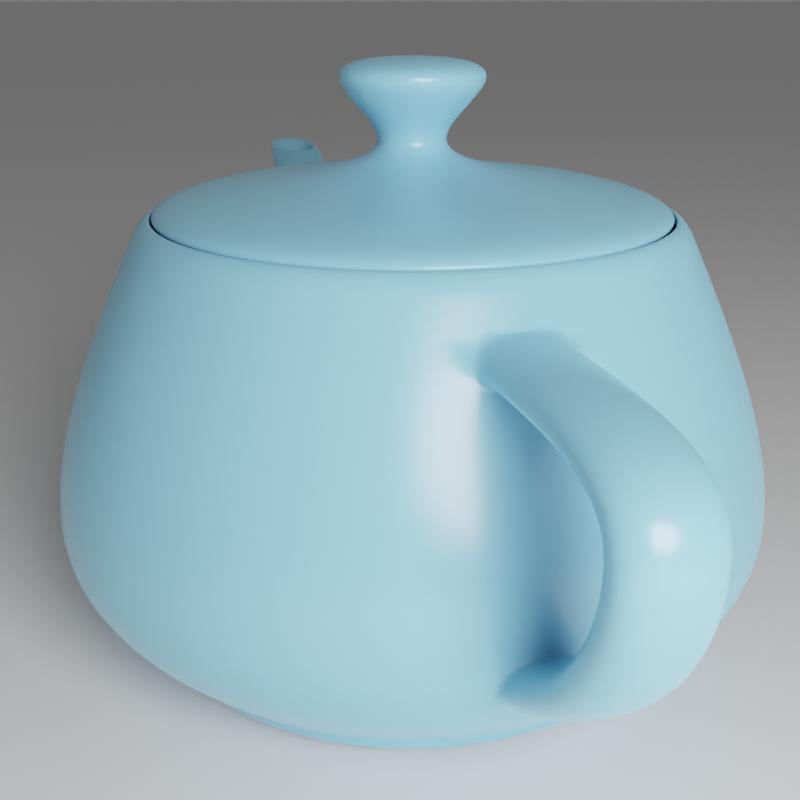} &
        \includegraphics[width=0.14\textwidth]{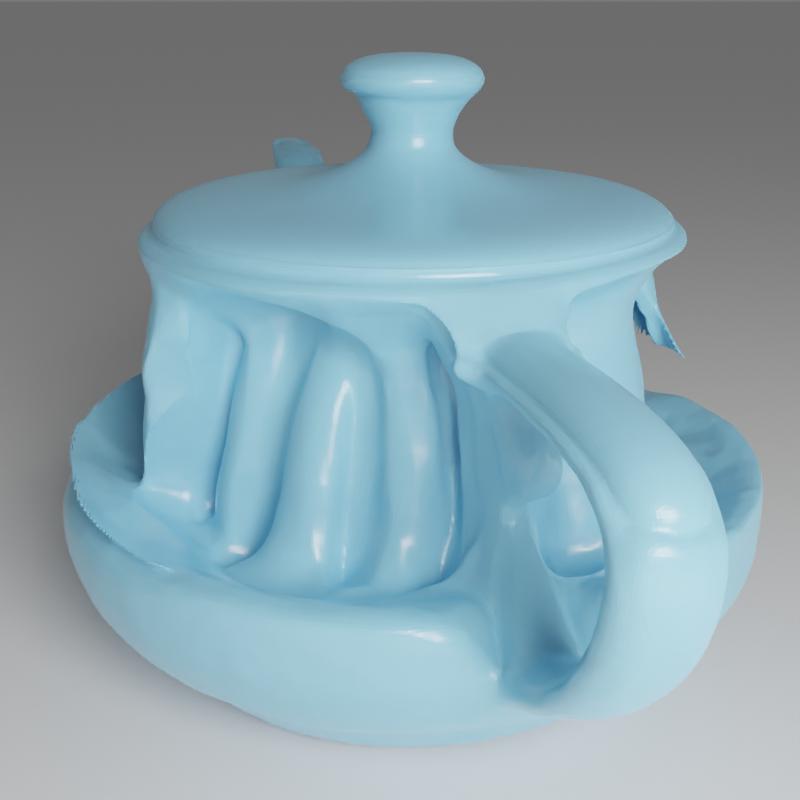} &
        \includegraphics[width=0.14\textwidth]{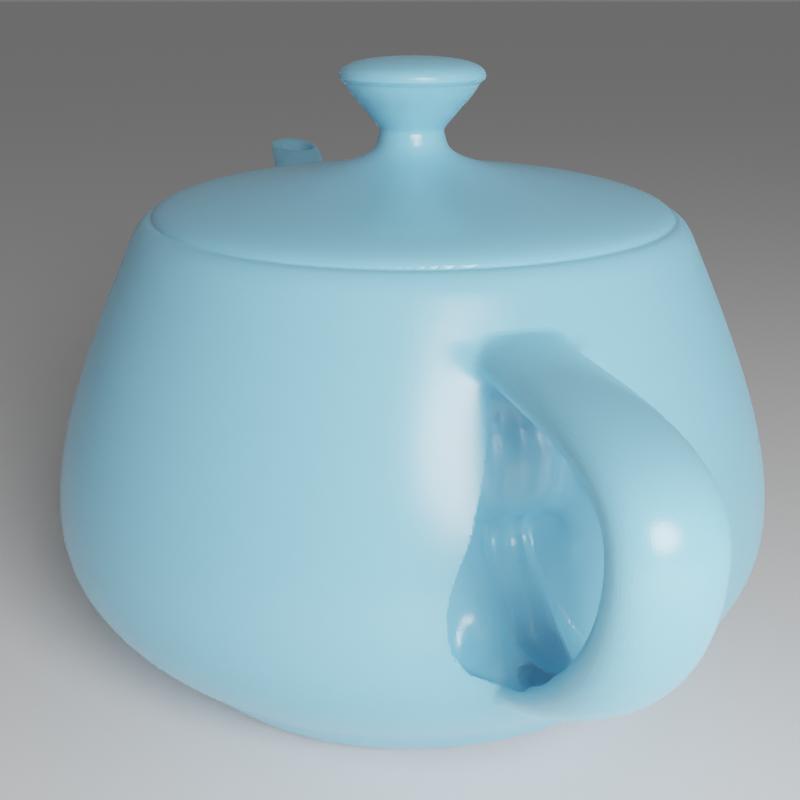} &
        \includegraphics[width=0.14\textwidth]{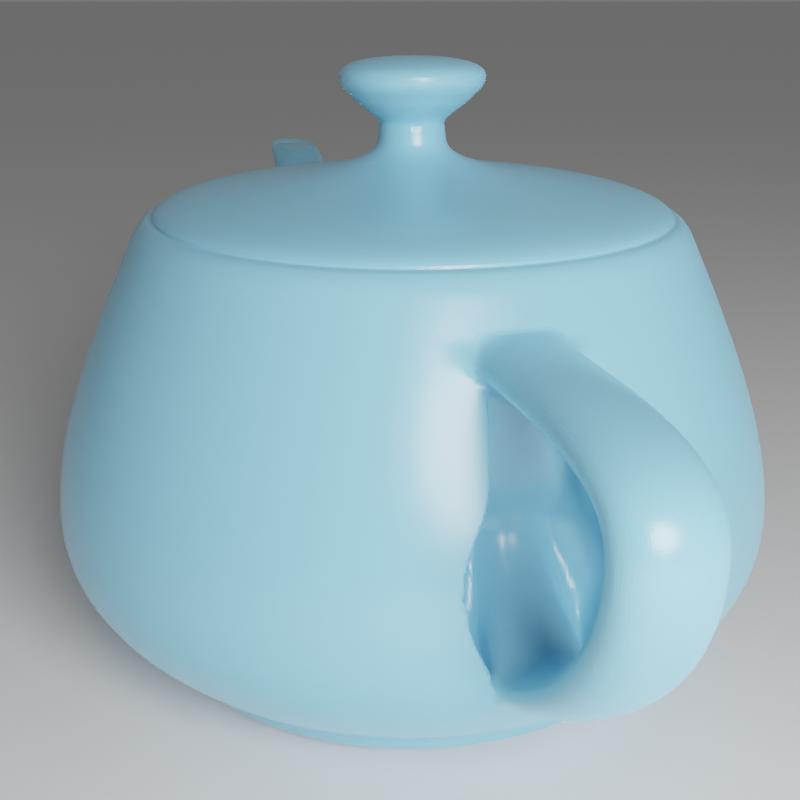} &
        \includegraphics[width=0.14\textwidth]{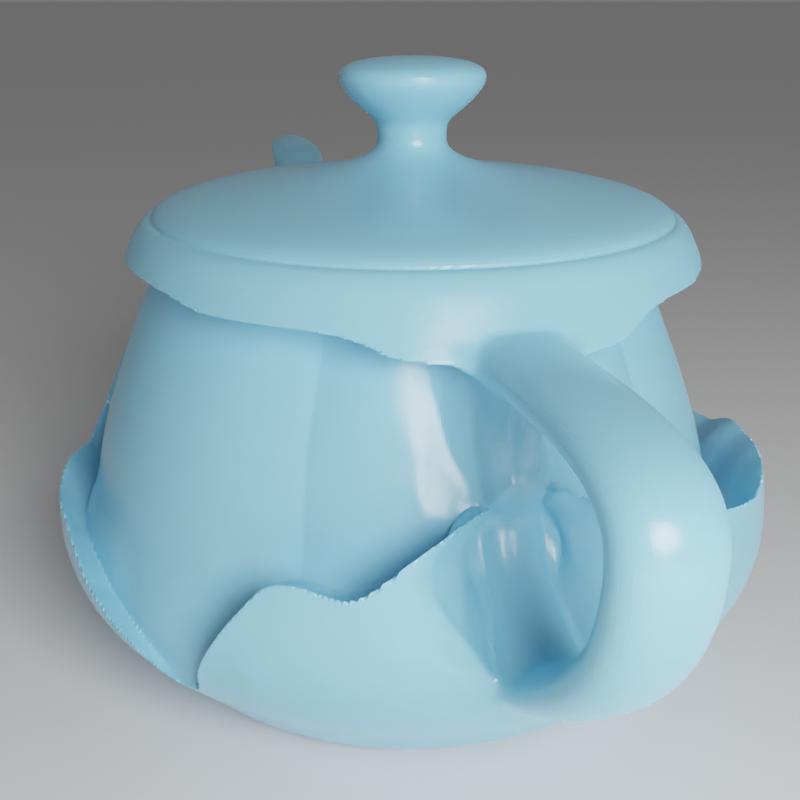} &
        \includegraphics[width=0.14\textwidth]{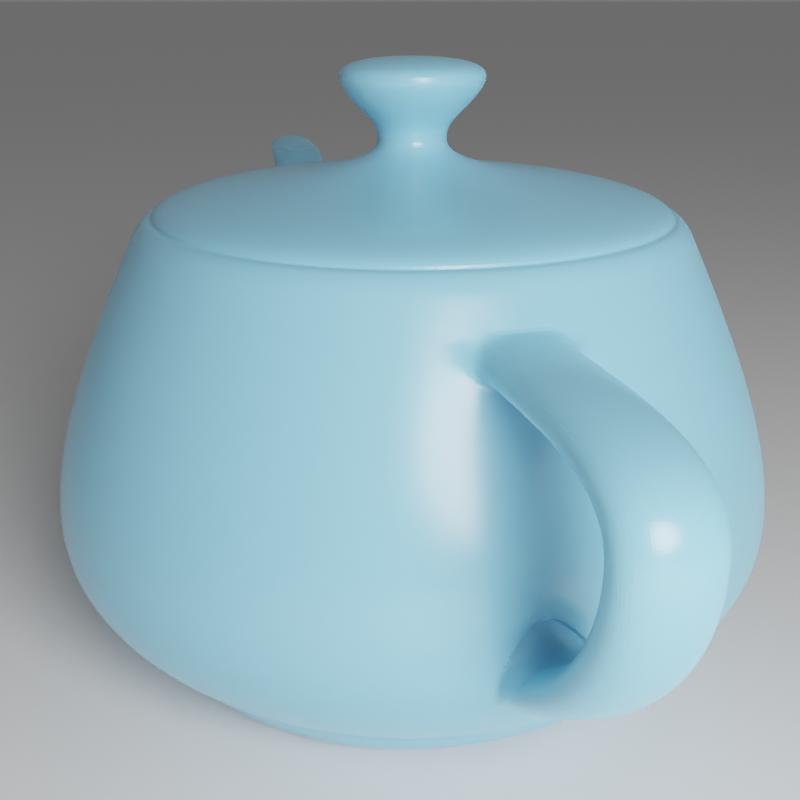} \\
        \includegraphics[width=0.14\textwidth]{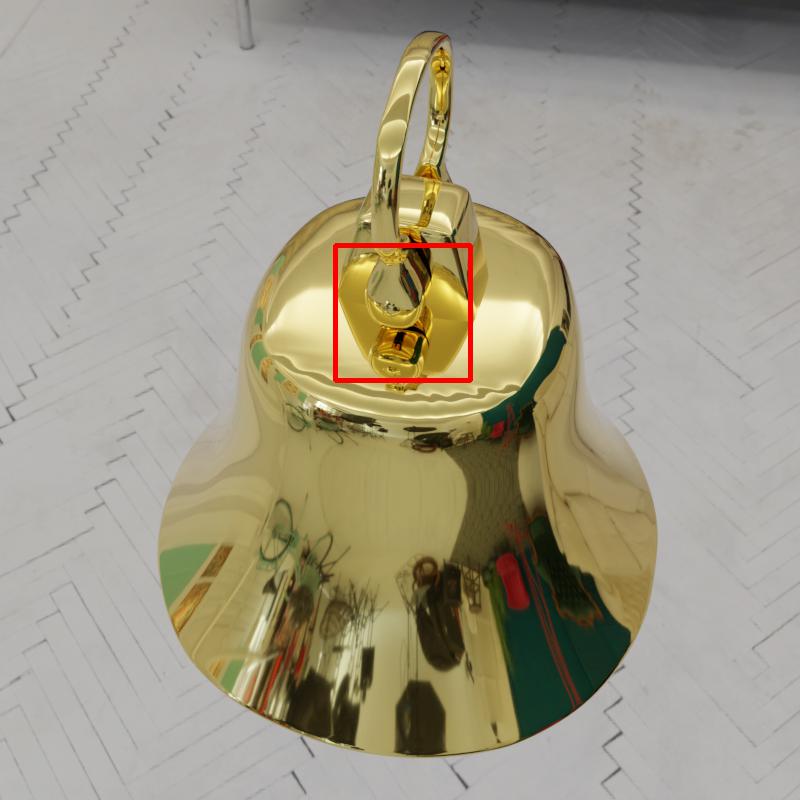} &
        \includegraphics[width=0.14\textwidth]{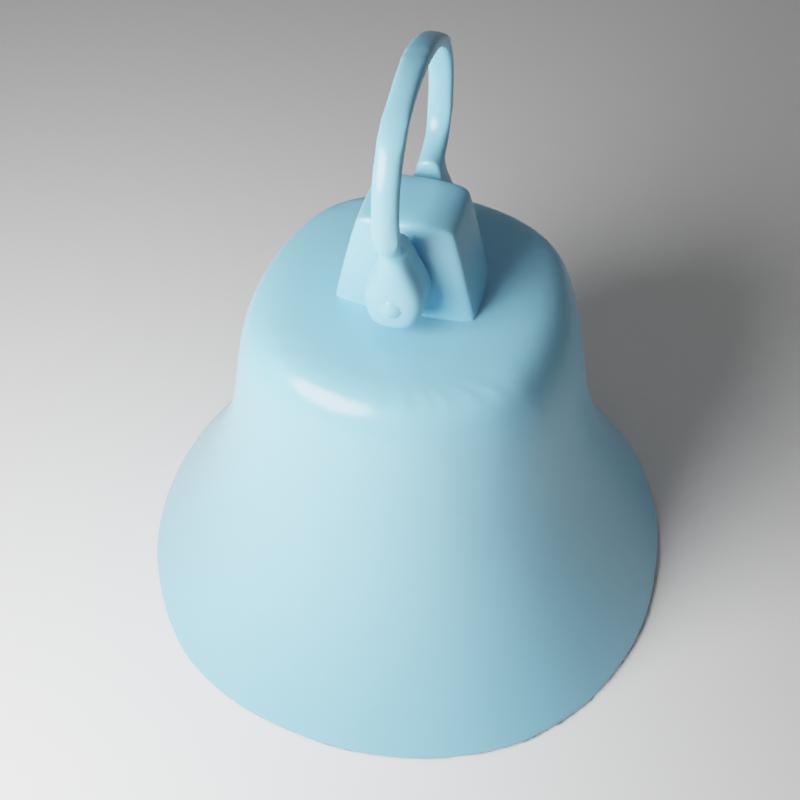} &
        \includegraphics[width=0.14\textwidth]{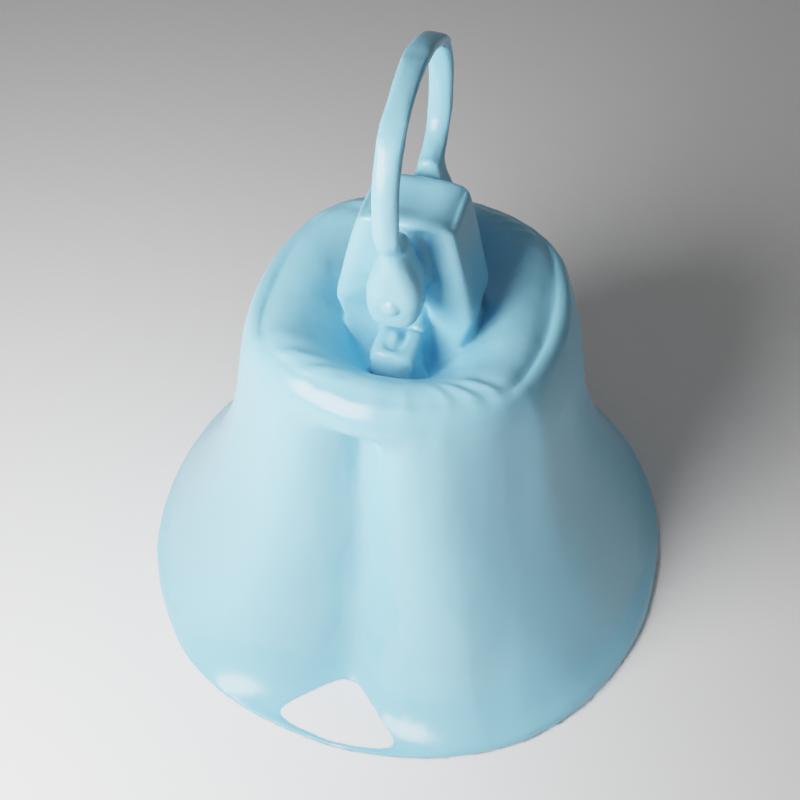} &
        \includegraphics[width=0.14\textwidth]{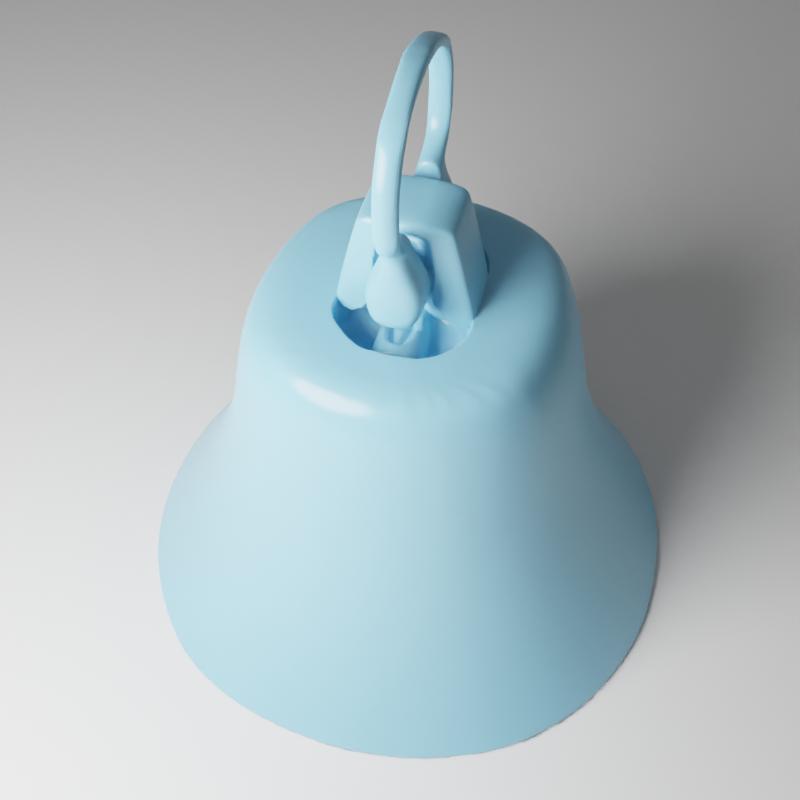} &
        \includegraphics[width=0.14\textwidth]{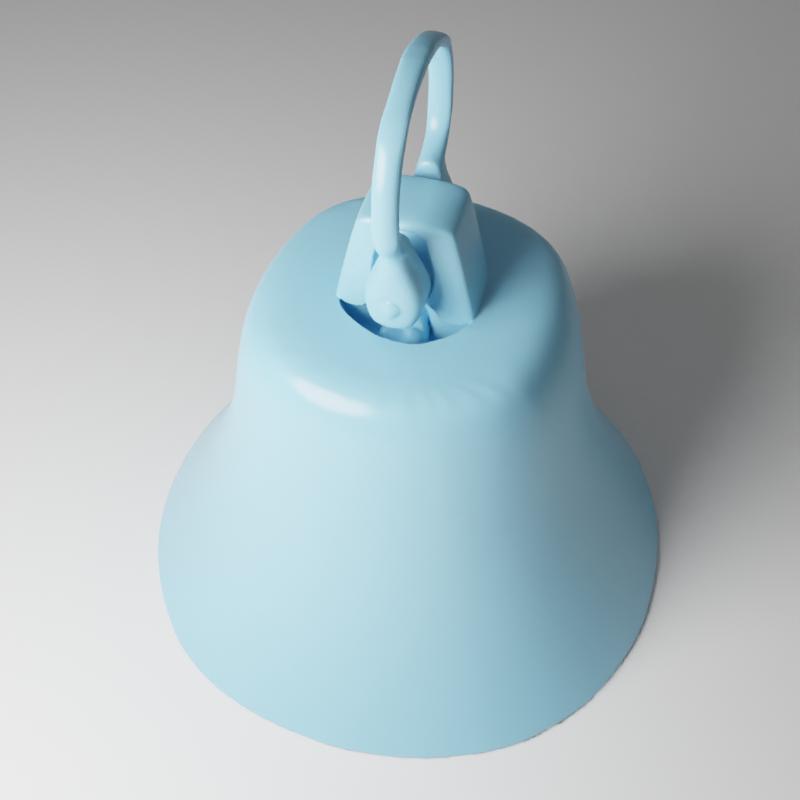} &
        \includegraphics[width=0.14\textwidth]{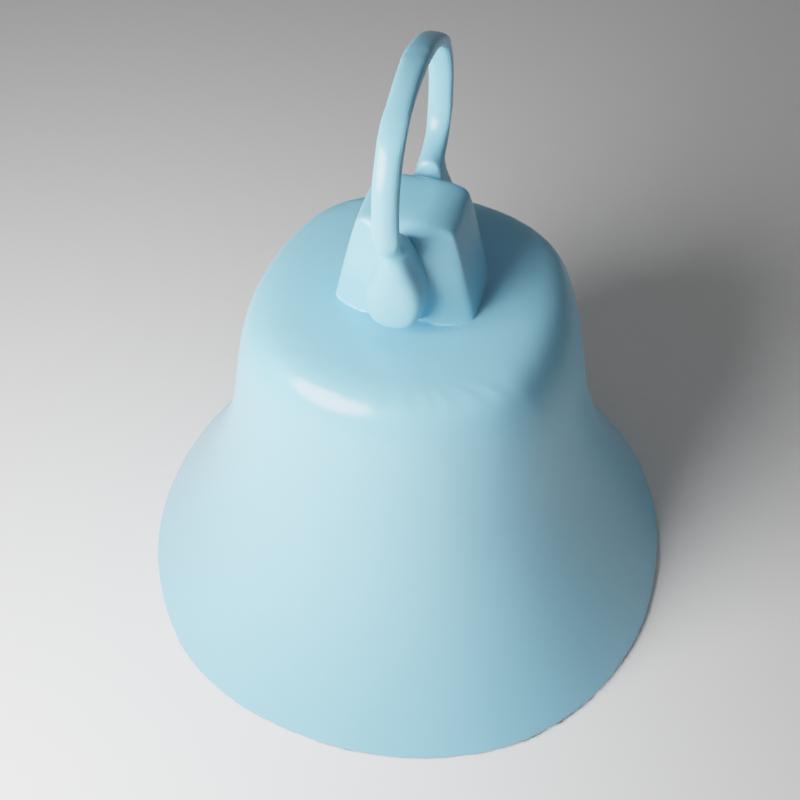} &
        \includegraphics[width=0.14\textwidth]{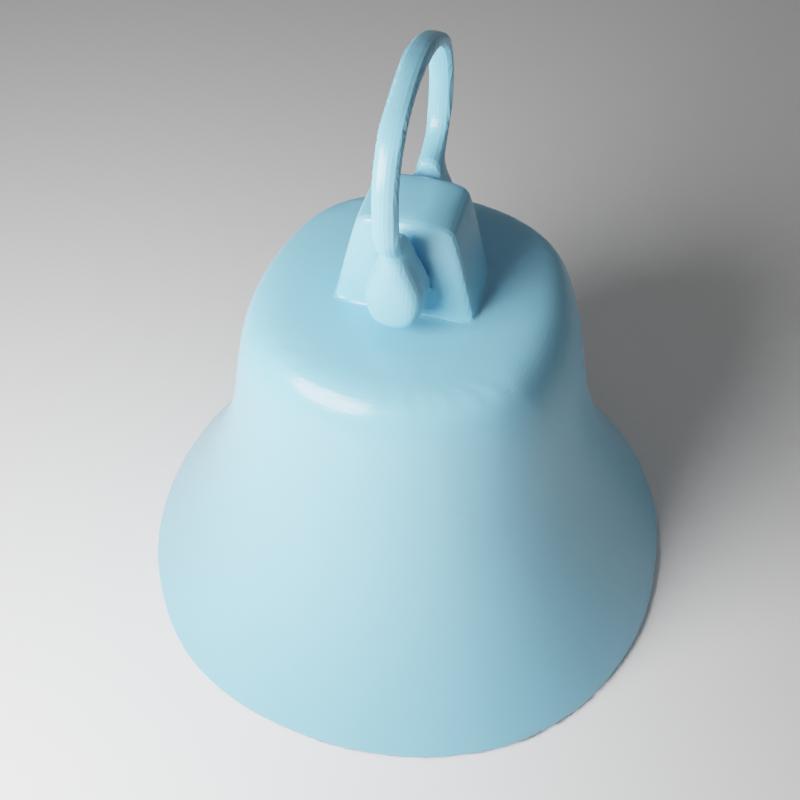} \\
        \includegraphics[width=0.14\textwidth]{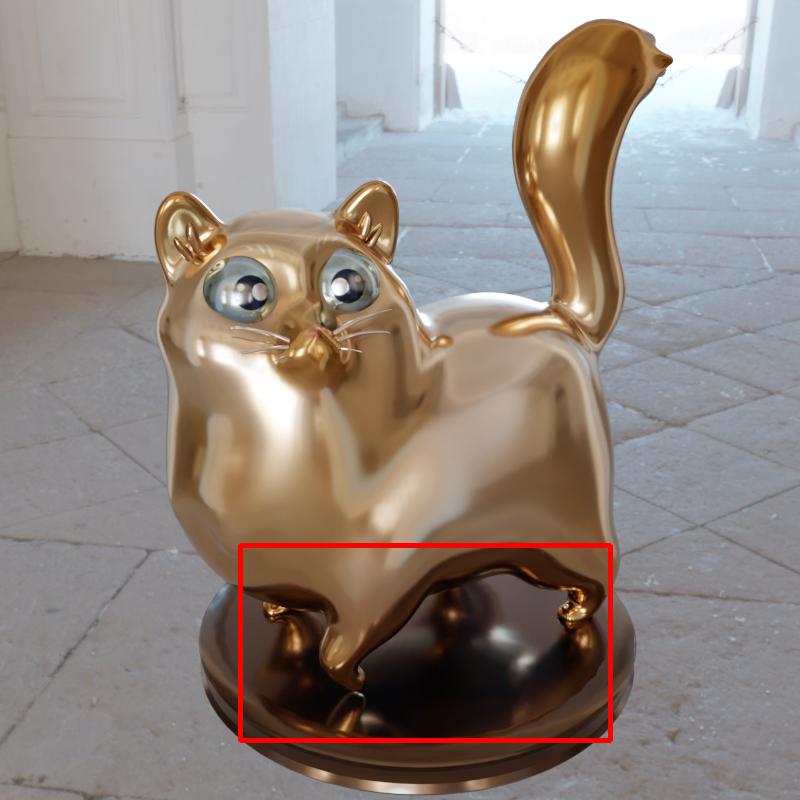} &
        \includegraphics[width=0.14\textwidth]{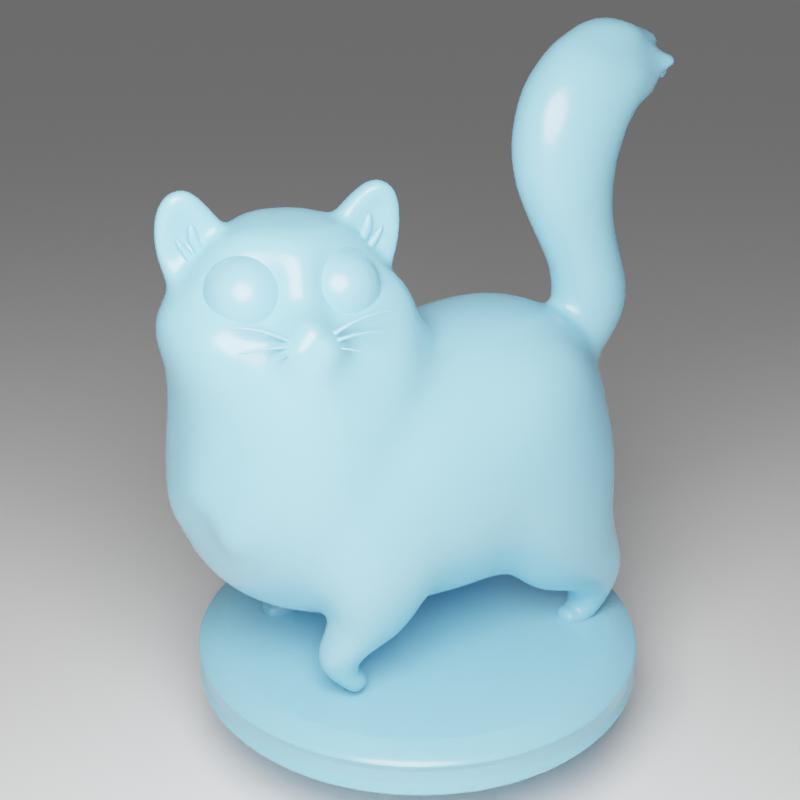} &
        \includegraphics[width=0.14\textwidth]{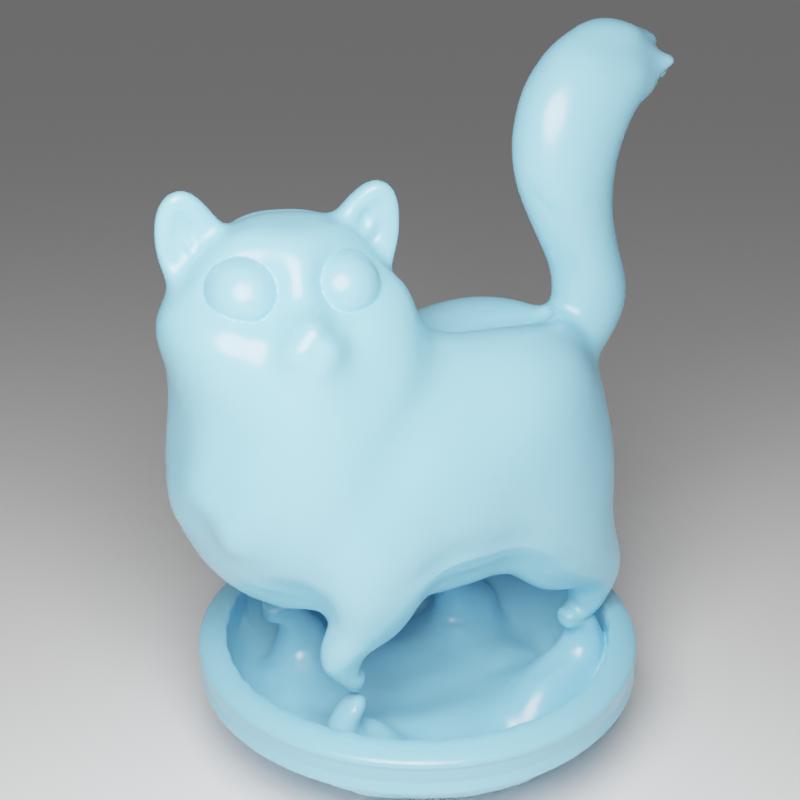} &
        \includegraphics[width=0.14\textwidth]{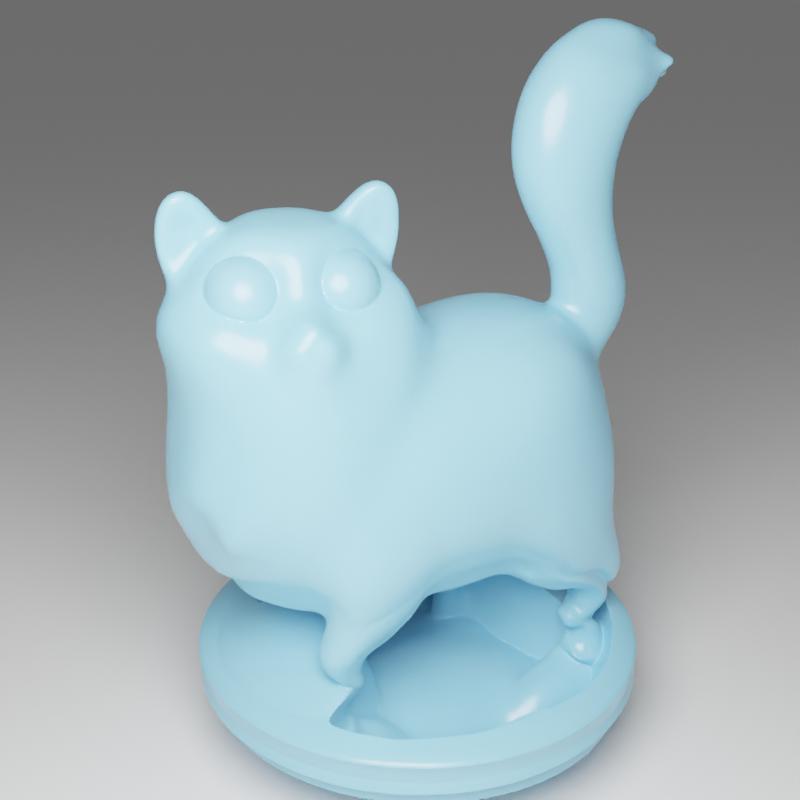} &
        \includegraphics[width=0.14\textwidth]{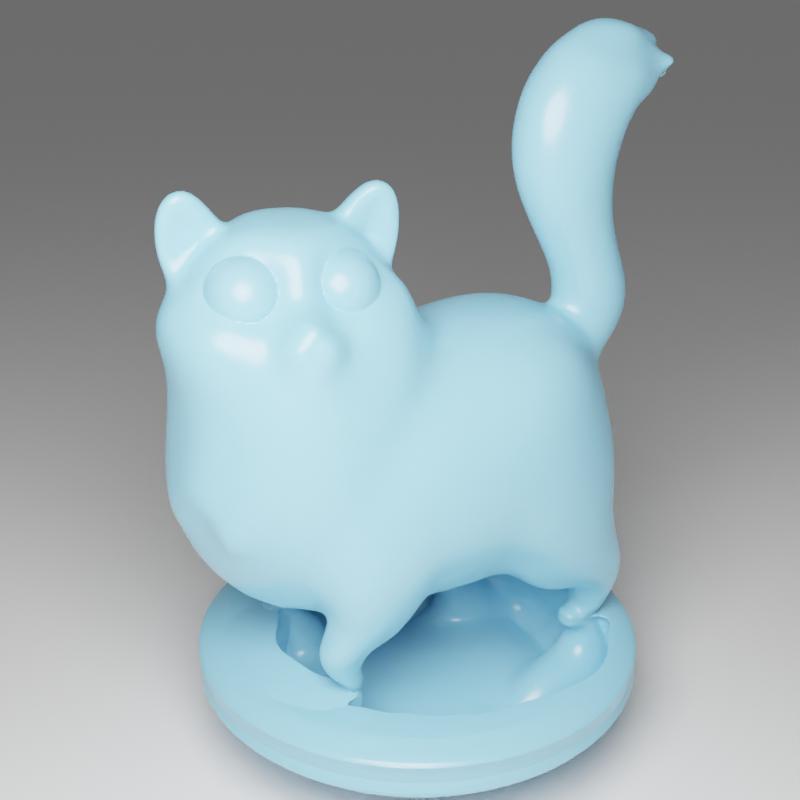} &
        \includegraphics[width=0.14\textwidth]{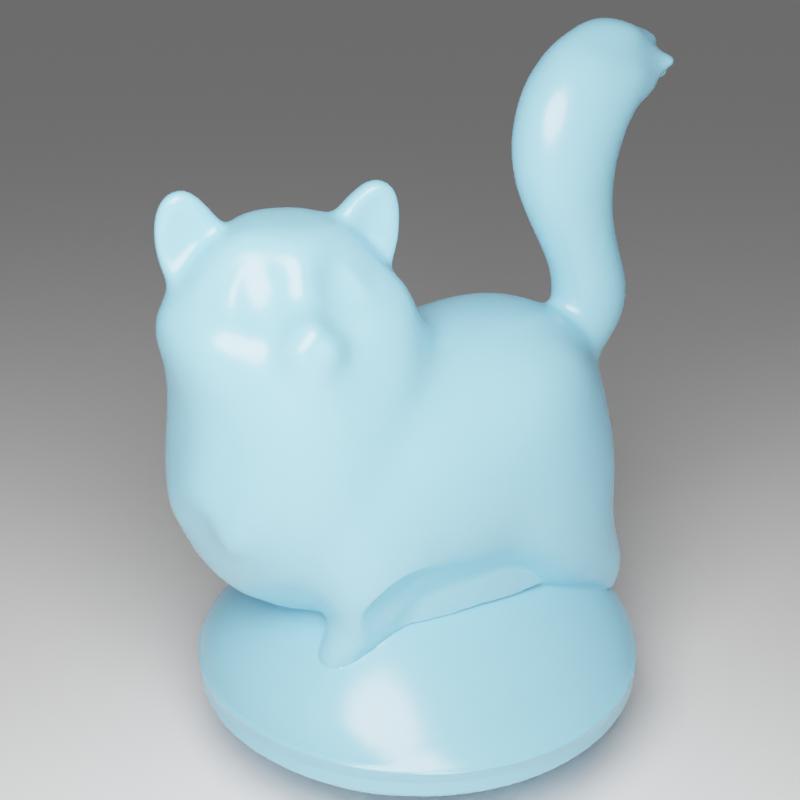} &
        \includegraphics[width=0.14\textwidth]{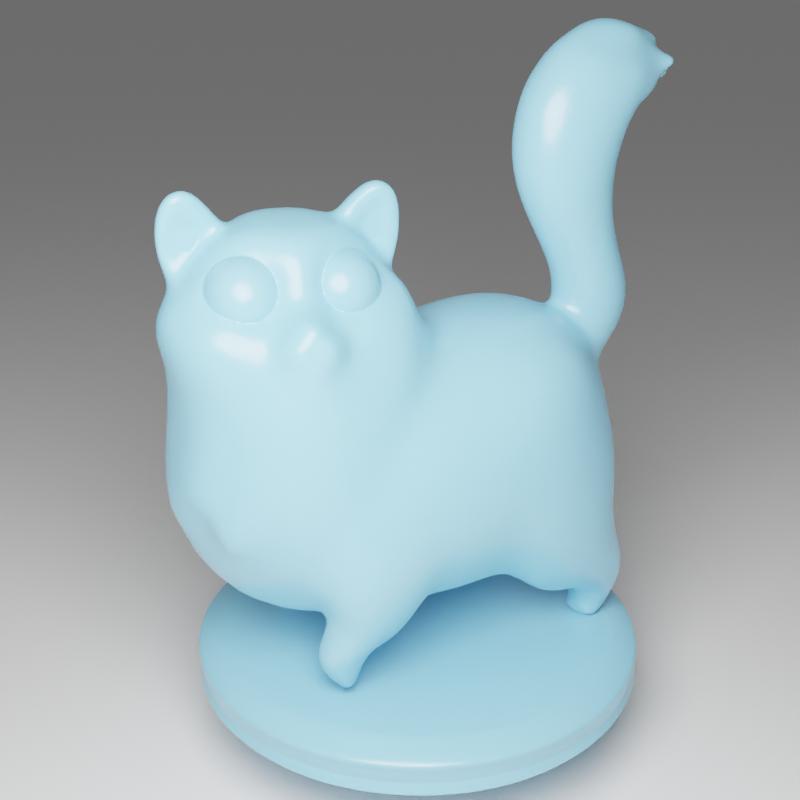} \\
        Image & Ground-truth & NeuS & Only direct & Only indirect & No $\ell_{\rm occ}$ loss & Full model \\
    \end{tabular}
    \caption{\textbf{Ablation studies on the surface reconstruction}. We use red bounding boxes to highlight regions illuminated by indirect lights in the images. Results of NeuS~\cite{wang2021neus} are provided for comparison. ``Only direct'' means the model using only the direct lights (Model 1 in Table~\ref{tab:ab}). ``Only indirect'' means the model using only indirect lights (Model 2 in Table~\ref{tab:ab}). ``No $\ell_{\rm occ}$ loss'' means the model uses both direct and indirect lights but not use $\ell_{\rm occ}$ on the occlusion probability (Model 3 in Table~\ref{tab:ab}). ``Full model'' contains all components (Model 4 in Table~\ref{tab:ab}).}
    \label{fig:ab}
\end{figure*}

\begin{figure}[]
    \centering
    \setlength\tabcolsep{1pt}
    \begin{tabular}{cccc}
         \multicolumn{2}{c}{With $\ell_{\rm eikonal}$} & \multicolumn{2}{c}{Without $\ell_{\rm eikonal}$}  \\
         \includegraphics[width=0.24\linewidth]{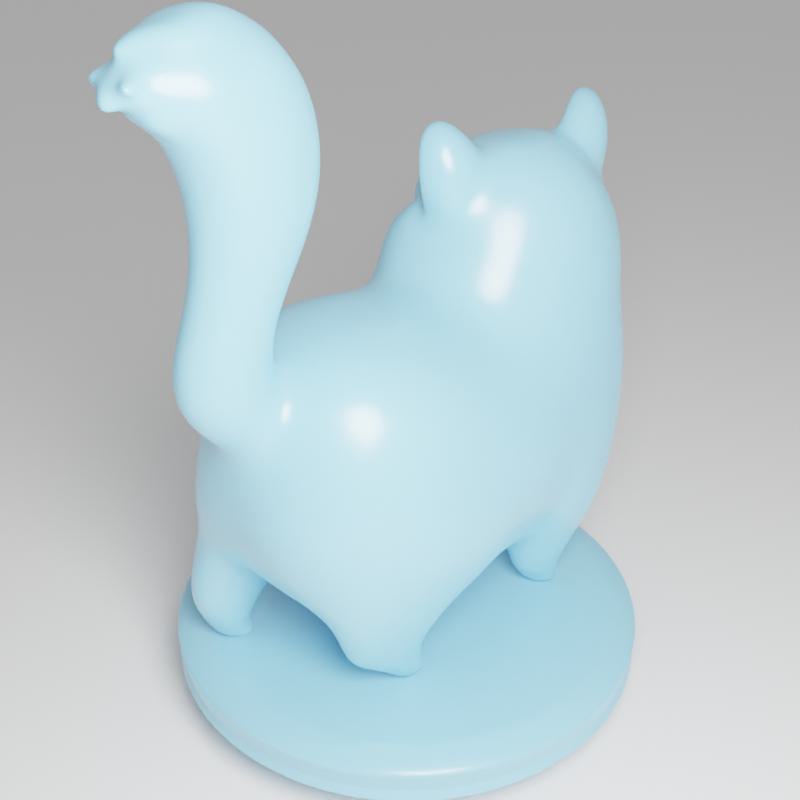} &
         \includegraphics[width=0.24\linewidth]{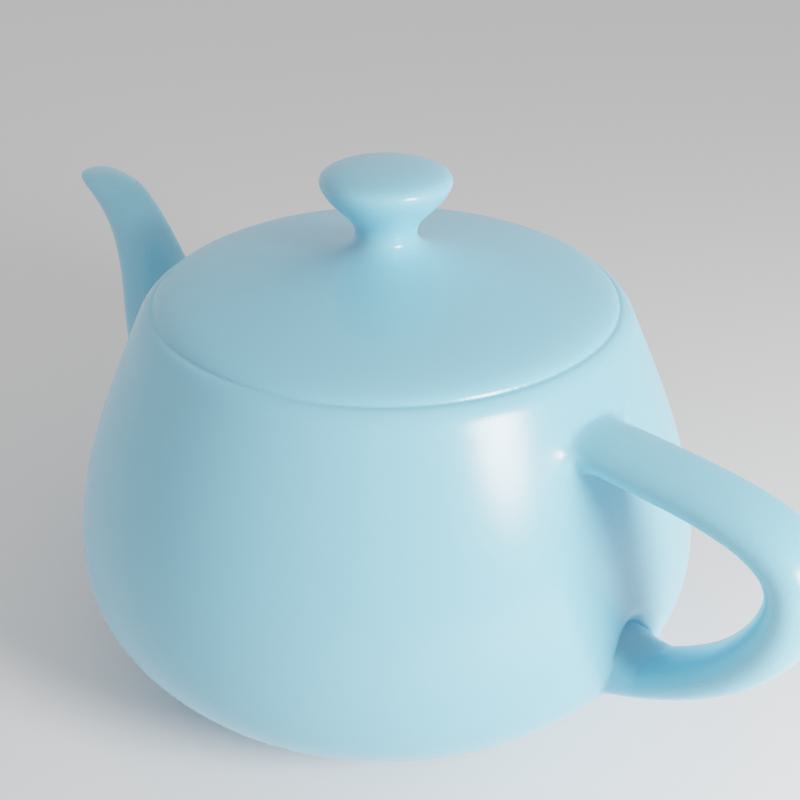} &
         \includegraphics[width=0.24\linewidth]{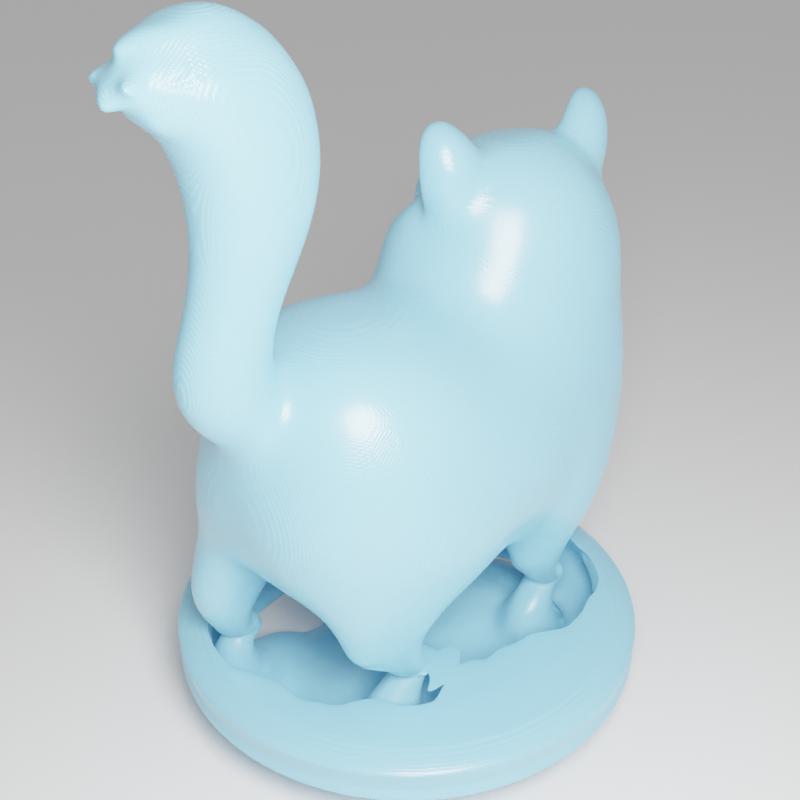} &
         \includegraphics[width=0.24\linewidth]{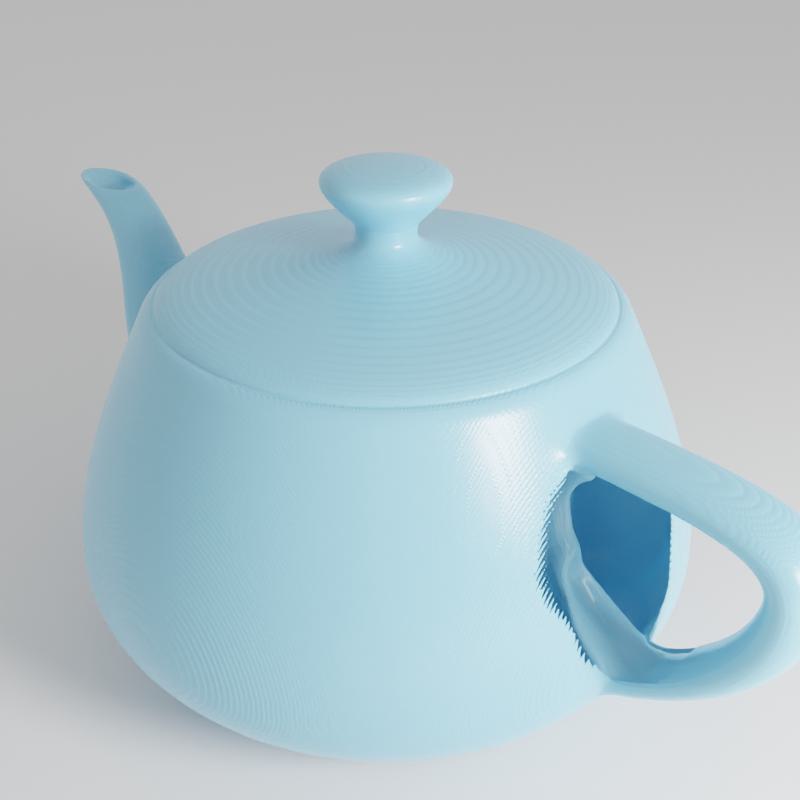} \\
    \end{tabular}
    \caption{\textbf{Ablation studies on the Eikonal loss} $\ell_{\rm eikonal}$.}
    \label{fig:ab_eik}
\end{figure}
\begin{figure*}
    \centering
    \setlength\tabcolsep{1pt}
    \renewcommand{\arraystretch}{0.5} 
    \begin{tabular}{cccccc}
        \includegraphics[width=0.16\textwidth]{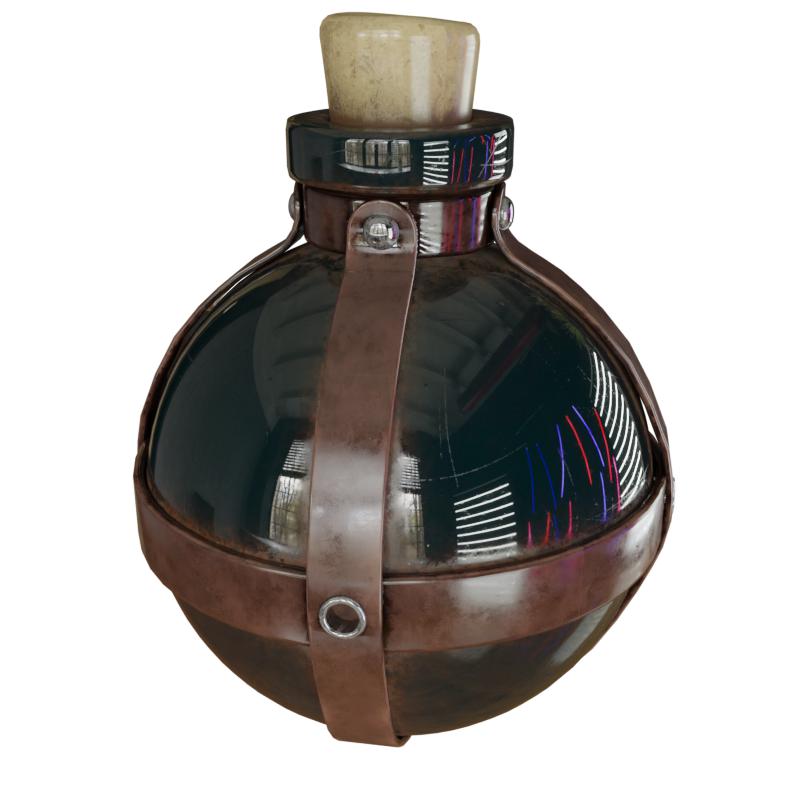} & 
        \includegraphics[width=0.16\textwidth]{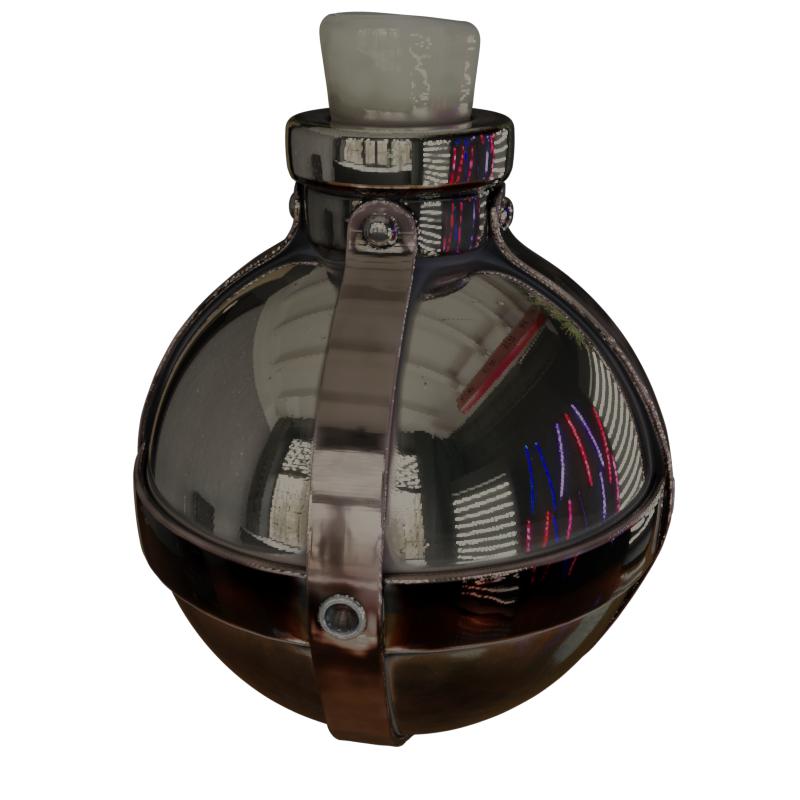} &
        \includegraphics[width=0.16\textwidth]{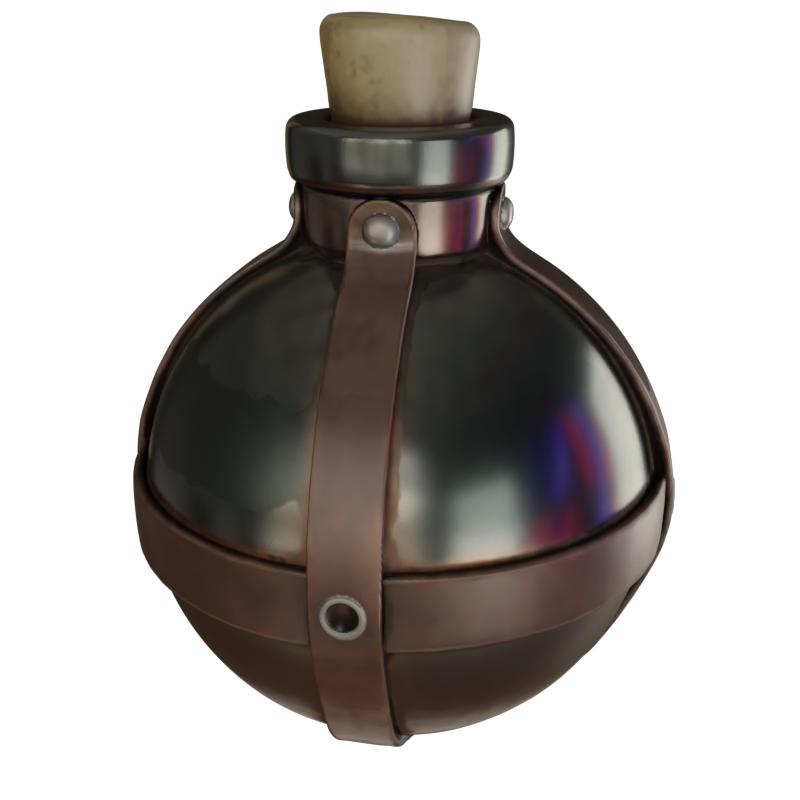} &
        \includegraphics[width=0.16\textwidth]{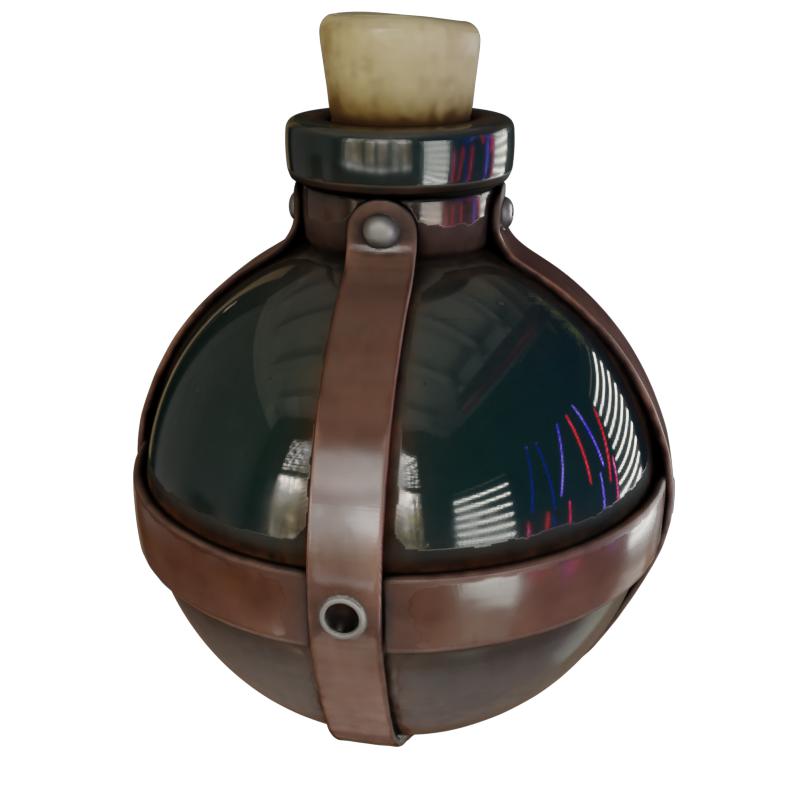} & 
        \includegraphics[width=0.16\textwidth]{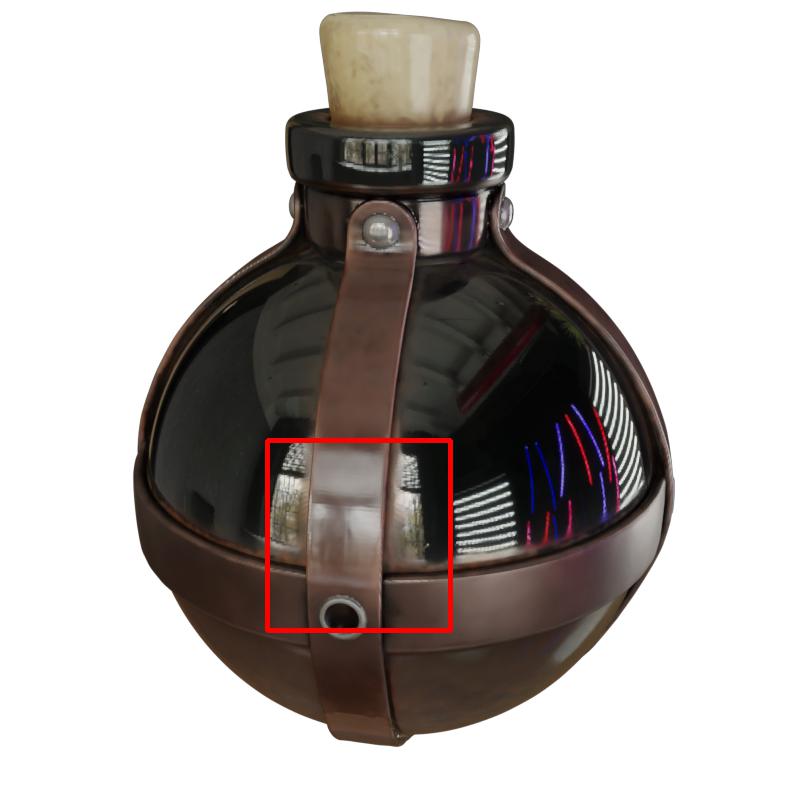} & 
        \includegraphics[width=0.16\textwidth]{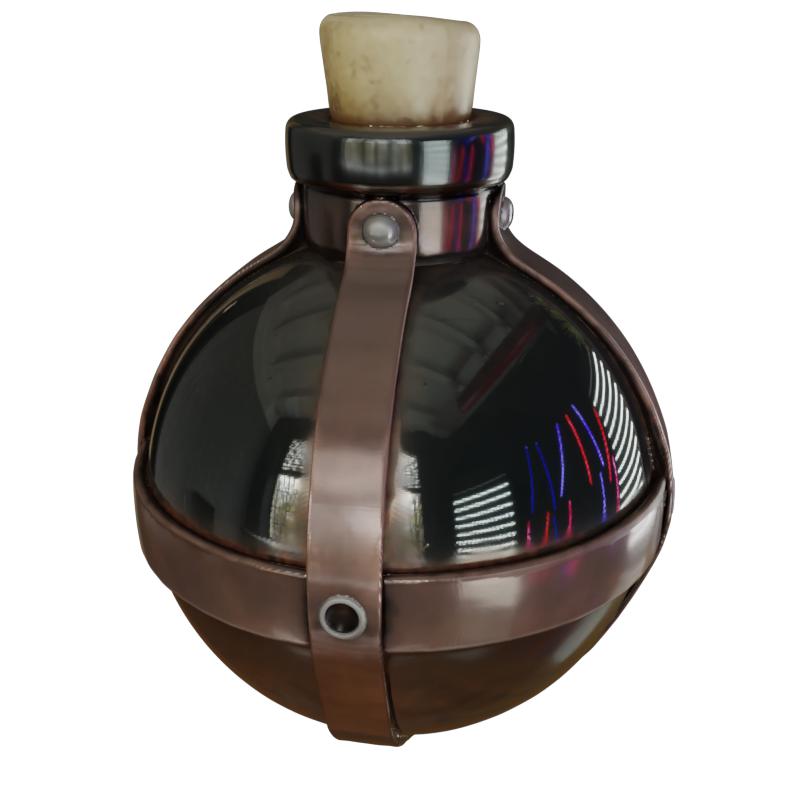} \\
        \includegraphics[width=0.16\textwidth]{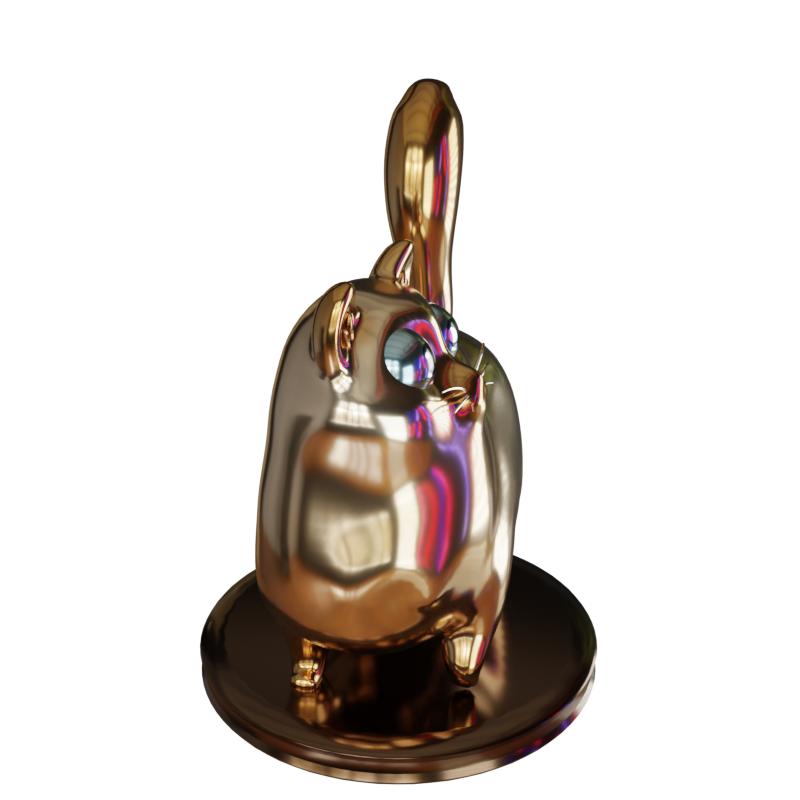} & 
        \includegraphics[width=0.16\textwidth]{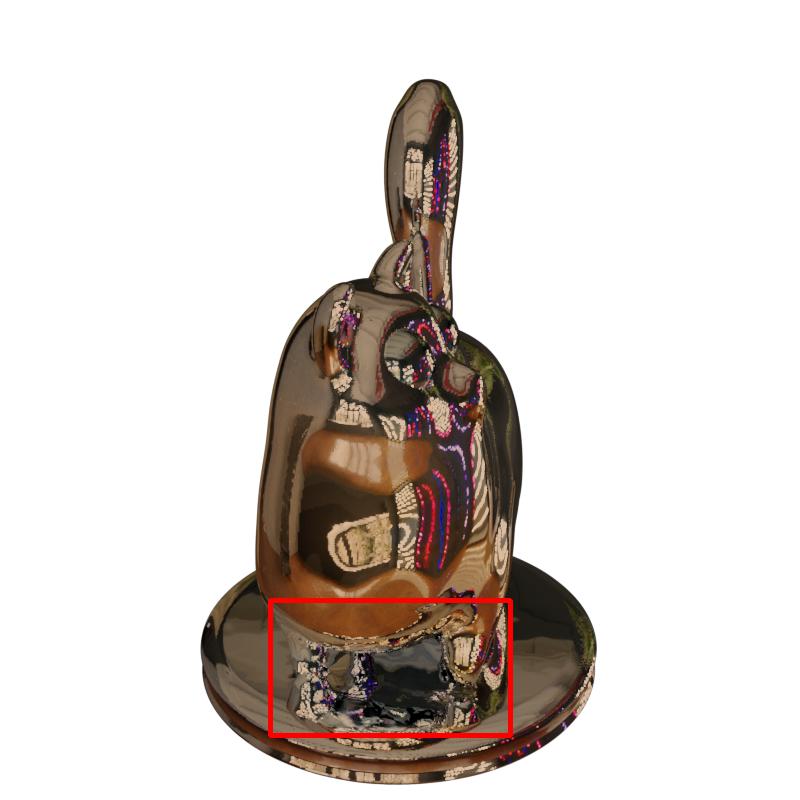} &
        \includegraphics[width=0.16\textwidth]{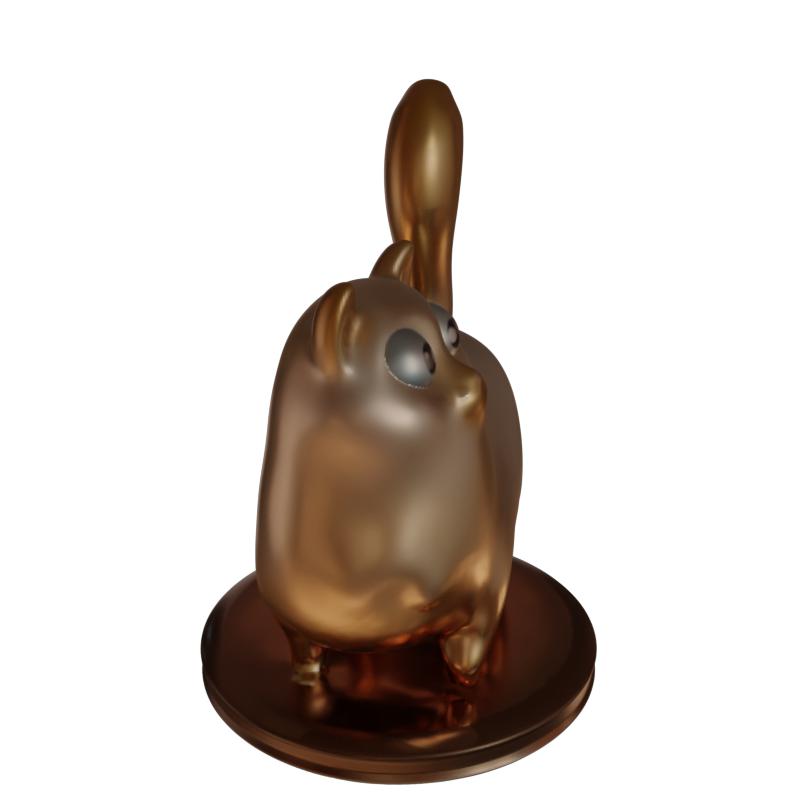} &
        \includegraphics[width=0.16\textwidth]{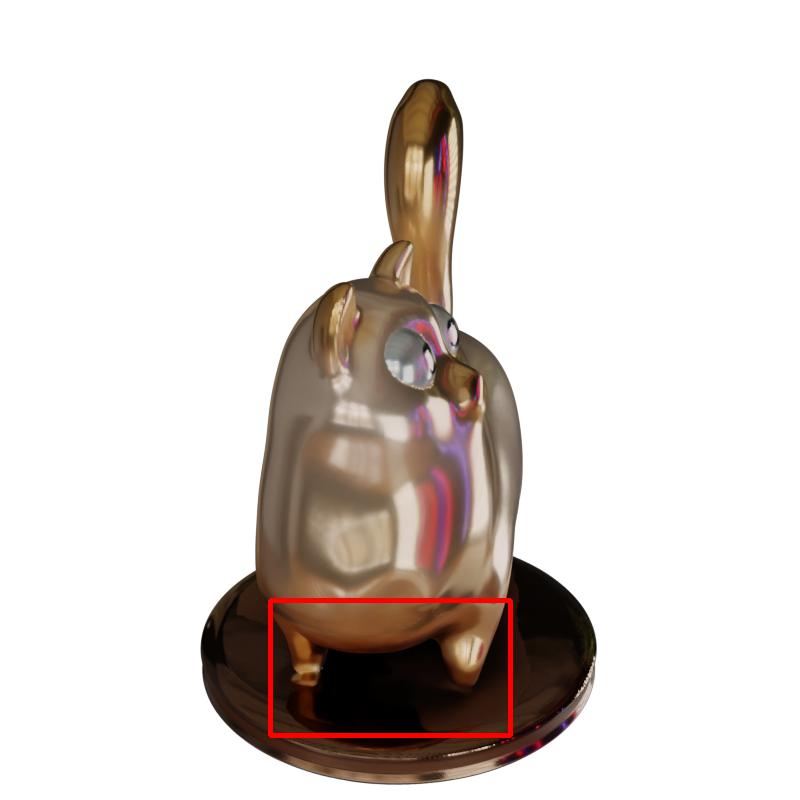} & 
        \includegraphics[width=0.16\textwidth]{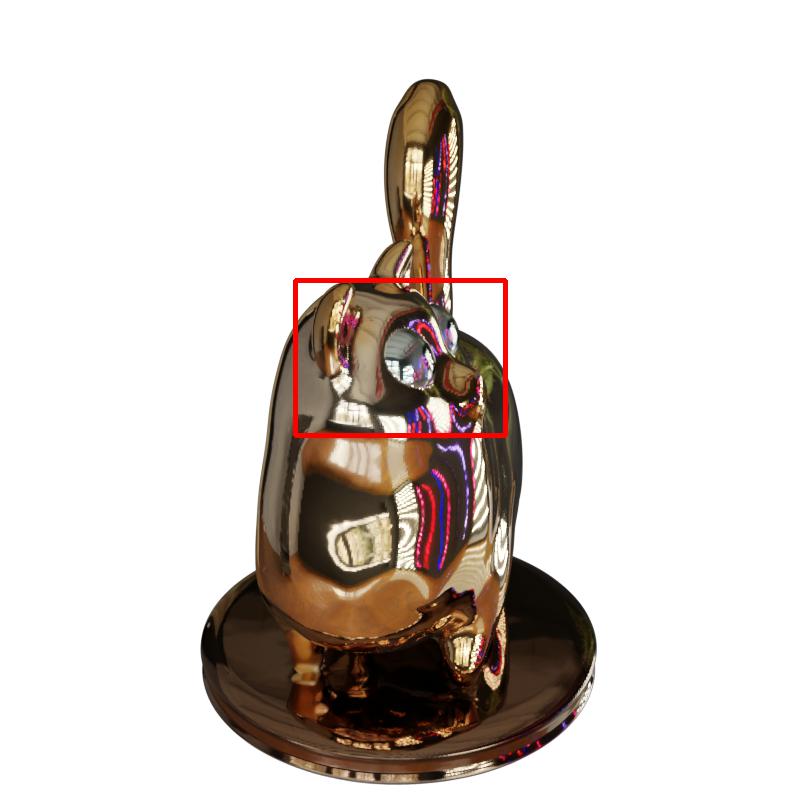} & 
        \includegraphics[width=0.16\textwidth]{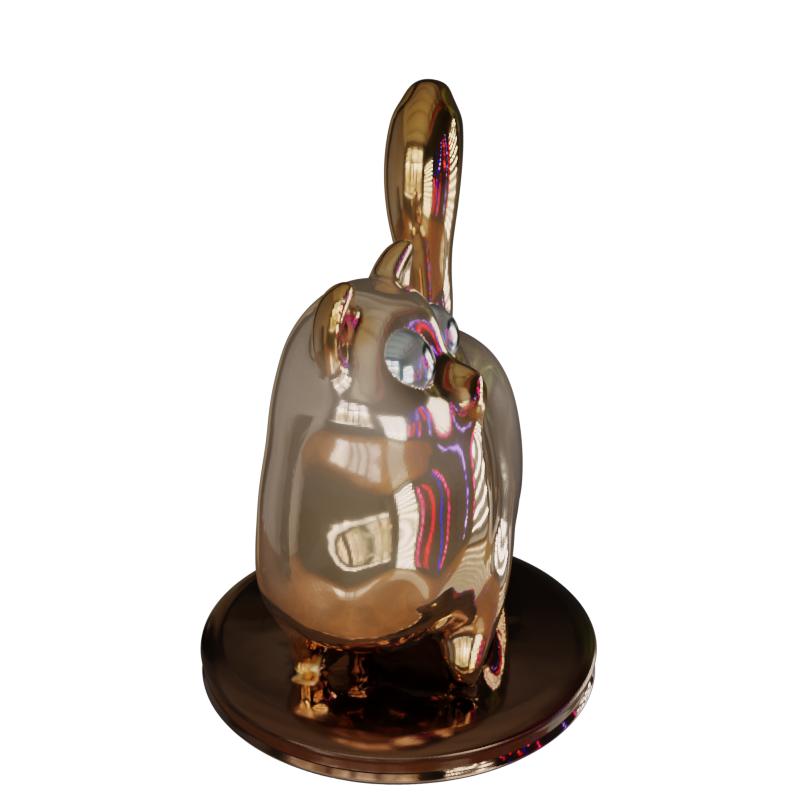} \\
        Ground-truth & Stage I & No sp. lobe & Only direct & Only indirect & Full model \\
    \end{tabular}
    \caption{\textbf{Ablation studies on the BRDF estimation}. ``Stage 1'' means the estimated BRDF in Stage I. ``No sp. lobe'' means not using the importance sampling on the specular lobe. ``Only direct'' means only using direct lights in the light representation while ``Only indirect'' means only using indirect lights. Note that all relighted images are normalized to match the average colors of the ground-truth images.}
    \label{fig:ab_brdf}
\end{figure*}

\textbf{Ablation studies on the geometry reconstruction}. The key component of our method for surface reconstruction is to consider the environmental lights for the reconstruction. Thus, we provide ablation studies on different environment light representations as shown in Table~\ref{tab:ab} and Fig.~\ref{fig:ab}. 
\begin{enumerate}
    \item In Model 1, we simply only consider the direct lights with $L(\omega_i)=g_{\rm direct}(\omega_i)$ in shading. In comparison with the original color function of NeuS, Model 1 is already able to reconstruct the surfaces illuminated by direct lights and avoids the undesirable surface distortion of NeuS. However, Model 1 is unable to correctly reconstruct the regions illuminated by indirect lights as shown in red boxes of Fig.~\ref{fig:ab}.
    \item Model 2 is designed only to consider the indirect lights with $L(\omega_i)=g_{\rm indirect}(\mathbf{p}, \omega_i)$, where $g_{\rm indirect}$ is supposed to predict location-dependent lights and is not limited to the direct environment lights. However, as shown in Fig.~\ref{fig:ab}, Model 2 still struggles to reconstruct the surfaces illuminated by the indirect lights. 
    The reason is that when the surfaces are mostly illuminated by the direct environment lights, Model 2 tends to neglect the input position $\mathbf{p}$ and only predicts the position-independent lights on all surface points. 
    \item Model 3 applies the combination of direct and indirect lights as Eq.~\ref{eq:light} but does not use the occlusion loss in Eq.~\ref{eq:occ_loss} to constrain the predicted occlusion probability. In this case, Model 3 learns the occlusion probability from only the rendering loss. As shown in Fig.~\ref{fig:ab}, the convergence of Model 3 is not stable due to the lack of clear supervision on the occlusion probability, which succeeds in reconstructing all surfaces of ``Bell'' but fails on ``Teapot'' and ``Cat''. 
    \item Model 4 does not use the Eikonal loss~\cite{gropp2020implicit}. Without Eikonal loss, the reconstructed surfaces are slightly better on ``Angel'' and ``Bell''. However, on ``Cat'' and ``Teapot'', Model 4 tends to reconstruct double-layer surfaces on regions illuminated by indirect lights as shown in Fig.~\ref{fig:ab_eik}.
    \item Our full model successfully reconstructs all surfaces of the objects. On the object ``Angel'', all models reconstruct its surface successfully as discussed in Sec.~\ref{sec:results_syn}.
\end{enumerate}
 
\begin{table}[]
    \centering
    \resizebox{\linewidth}{!}{
    \begin{tabular}{clcccc|c}
    \toprule
    Index & Description & Potion & Bell & Cat & Teapot  & Avg.\\
    \midrule
    0 & Stage I                            & 22.44  & 28.83  & 21.54  & 28.53  & 25.34  \\ 
    1 & No specular lobe sampling          & 25.94  & 22.84  & 24.69  & 24.25  & 24.43 \\ 
    2 & Only direct                        & 29.74  & 30.03  & 29.21  & 29.84  & 29.71 \\ 
    3 & Only Indirect                      & 29.95  & 30.94  & 28.46  & 30.78  & 30.03 \\ 
    4 & Without $\ell_{\rm light}$    & 30.70  & 31.07  & 29.76  & 30.80  & 30.58 \\
    5 & Without $\ell_{\rm smooth}$   & 30.90  & 30.45 &  29.41  & 31.88  & 30.66 \\
    6 & Full model                         & 31.17  & 31.05  & 30.03  & 30.95  & 30.80 \\ 
    \bottomrule
    \end{tabular}
    }
    \caption{\textbf{Ablation studies on the BRDF estimation} using objects from the Glossy-Synthetic dataset. PSNR$\uparrow$ on of the relighting images is reported.}
    \label{tab:ab_relight}
\end{table}


\textbf{Ablation studies on the BRDF estimation}. We conduct ablation studies on the importance sampling strategy and the proposed light representation. The relighting results are shown in Table~\ref{tab:ab_relight} and Fig.~\ref{fig:ab_brdf}. First, the BRDF estimated in stage I of our method usually has very small roughness because our method in stage I tends to use a smooth material to improve the ability to fit specular colors. Second, about the importance sampling, if we do not apply importance sampling on the specular lobe, the estimated surface roughness will be large because it is unable to capture the high-frequency specular colors, which is similar to the results of NeILF. Third, we compare different light representations with the same setting as the ablation studies on geometry reconstruction. Only using direct lights prevents the model from estimating a correct BRDF on surfaces illuminated by indirect lights. Only using indirect lights results in over-smooth surfaces because, on rough materials showing low-frequency color changes, the model with only indirect lights will predict a smooth material with low-frequency lights rather than predicting a rough material. We further conduct ablation studies on the neutral light loss $\ell_{\rm light}$ and the smoothness loss $\ell_{\rm smooth}$~\cite{hasselgren2022shape,munkberg2022extracting}. Neutral light loss makes the estimated albedo more accurate and reasonable while smoothness loss avoids noisy material prediction, both of which improve the relighting quality as shown in Models 4, 5, and 6 in Table~\ref{tab:ab_relight}.


\begin{figure*}
    \centering
    \setlength\tabcolsep{1pt}
    \renewcommand{\arraystretch}{0.5} 
    \begin{tabular}{cccccc}
    \includegraphics[width=0.16\textwidth]{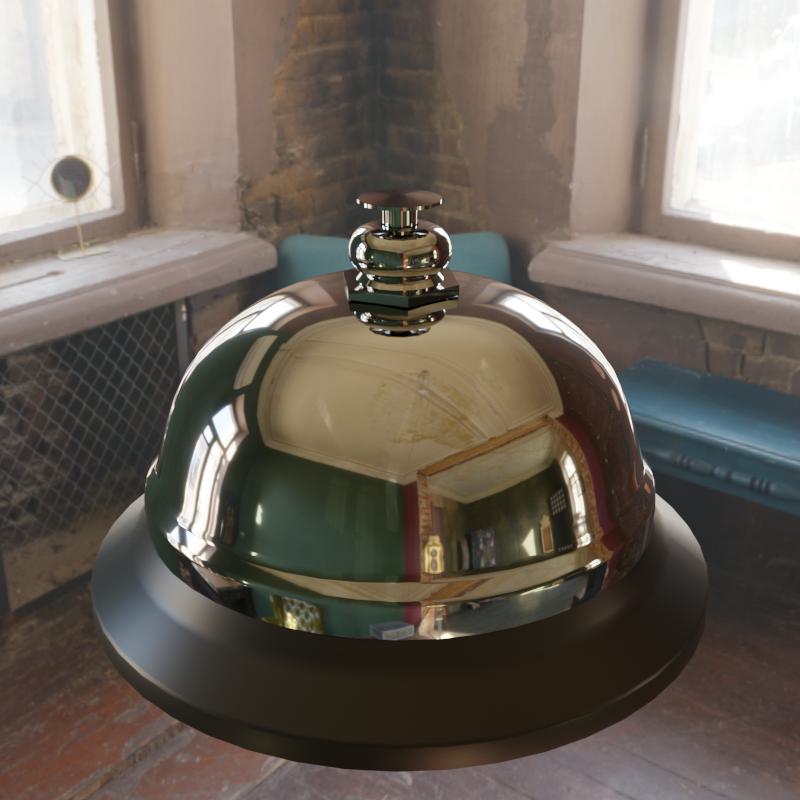} &
    \includegraphics[width=0.16\textwidth]{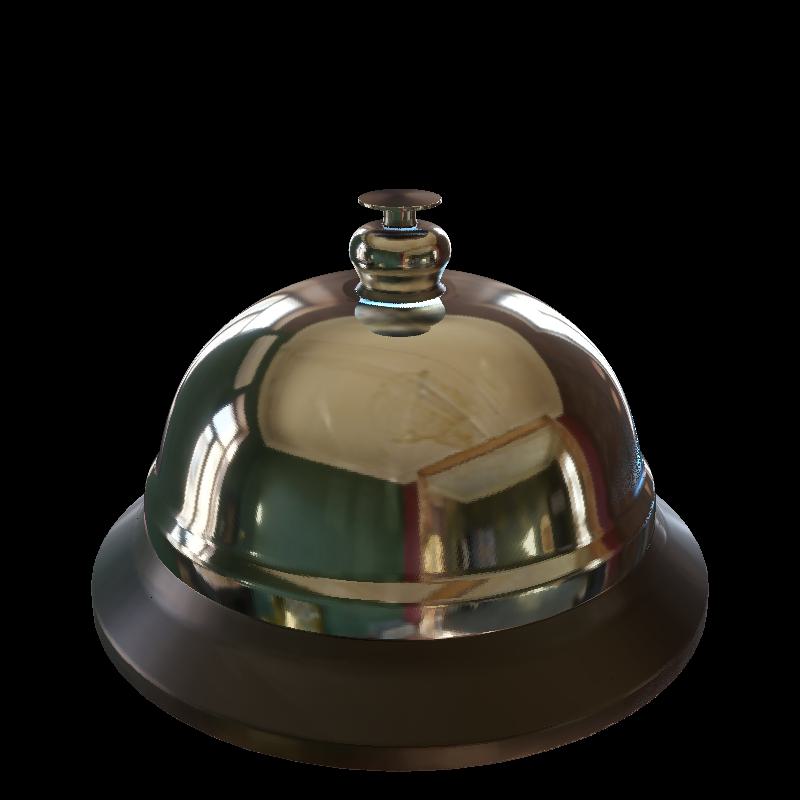} &
    \includegraphics[width=0.16\textwidth]{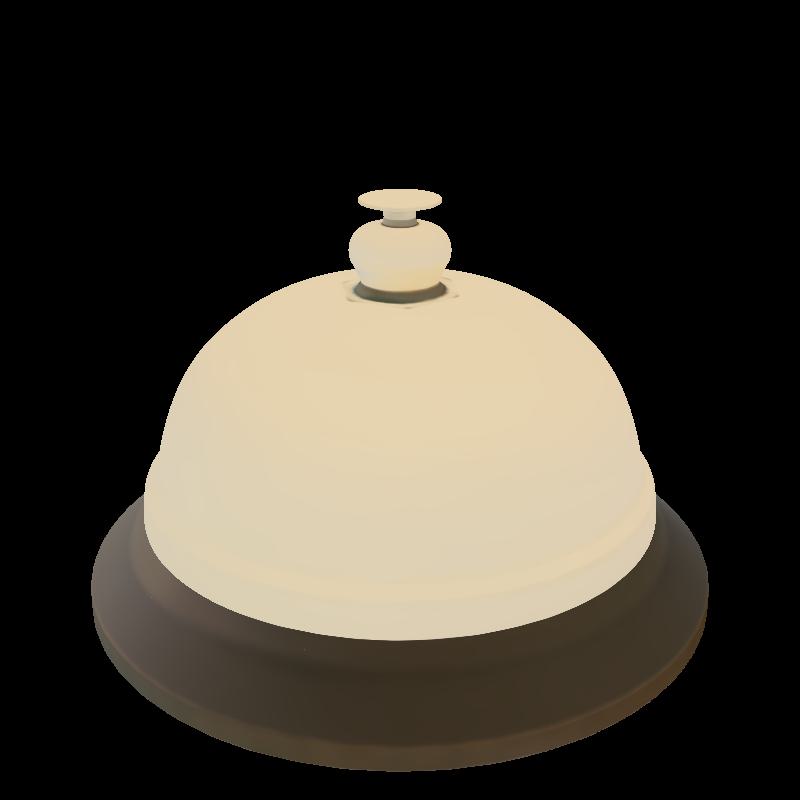} &
    \includegraphics[width=0.16\textwidth]{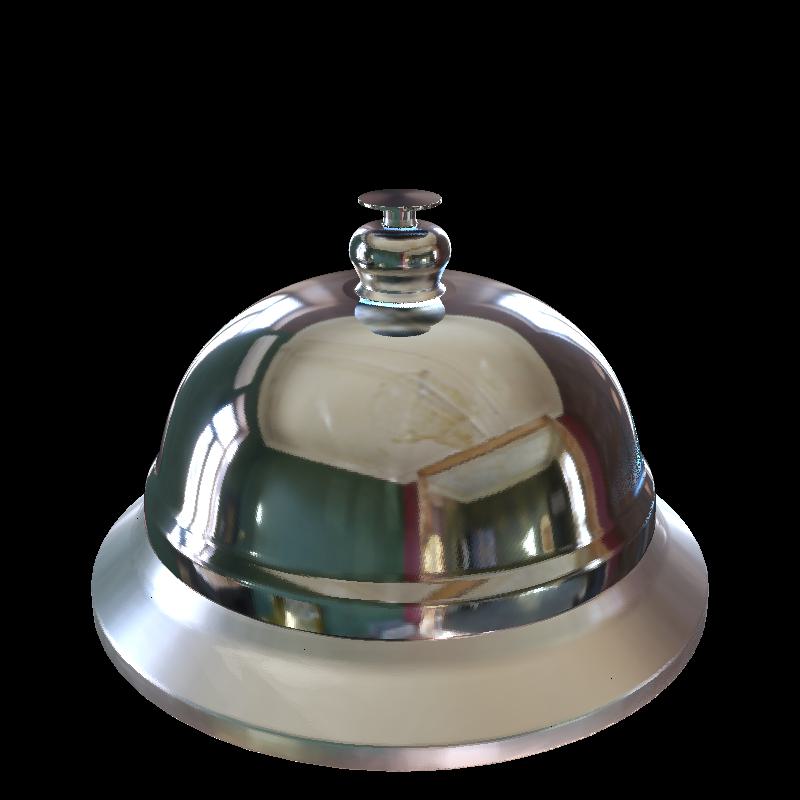} &
    \includegraphics[width=0.16\textwidth]{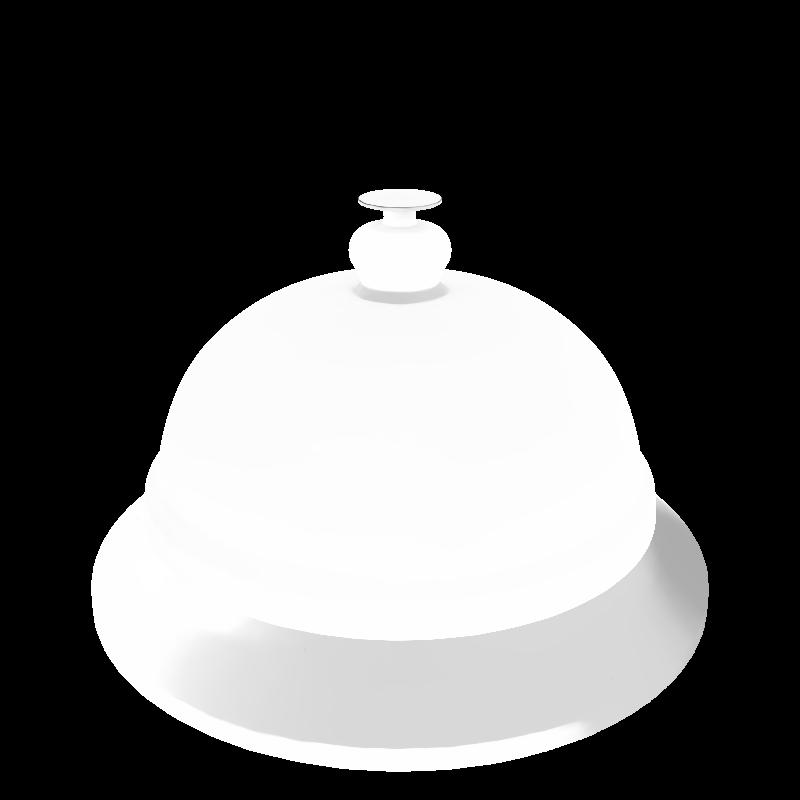} &
    \includegraphics[width=0.16\textwidth]{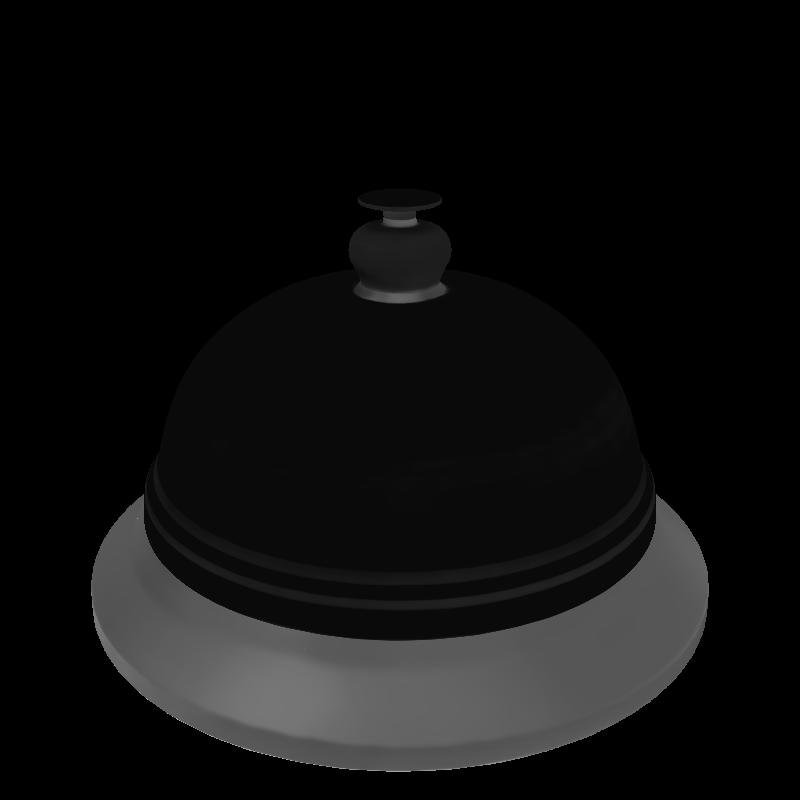} \\
    \includegraphics[width=0.16\textwidth]{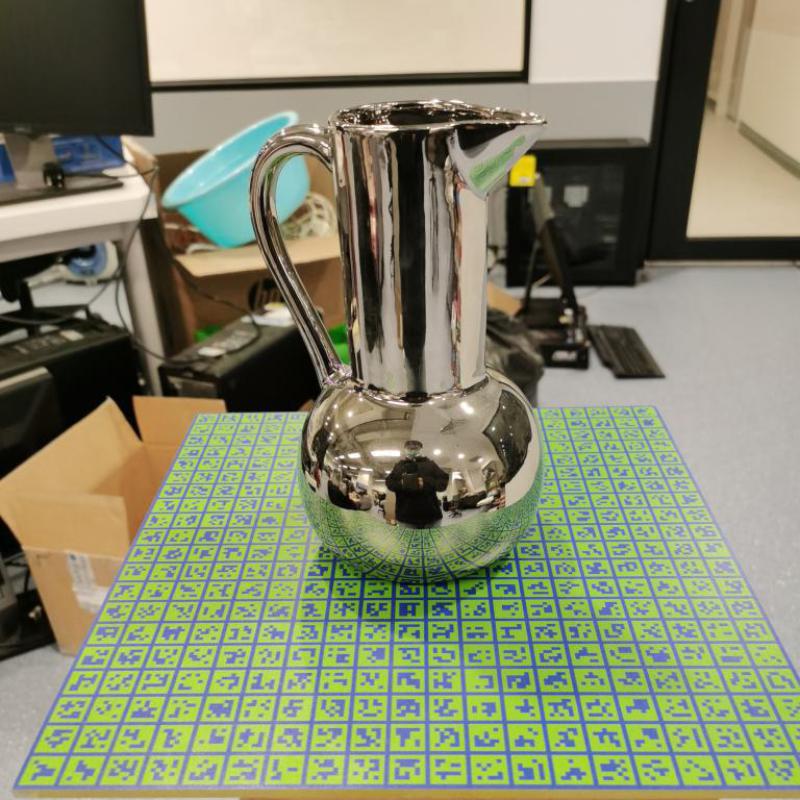} &
    \includegraphics[width=0.16\textwidth]{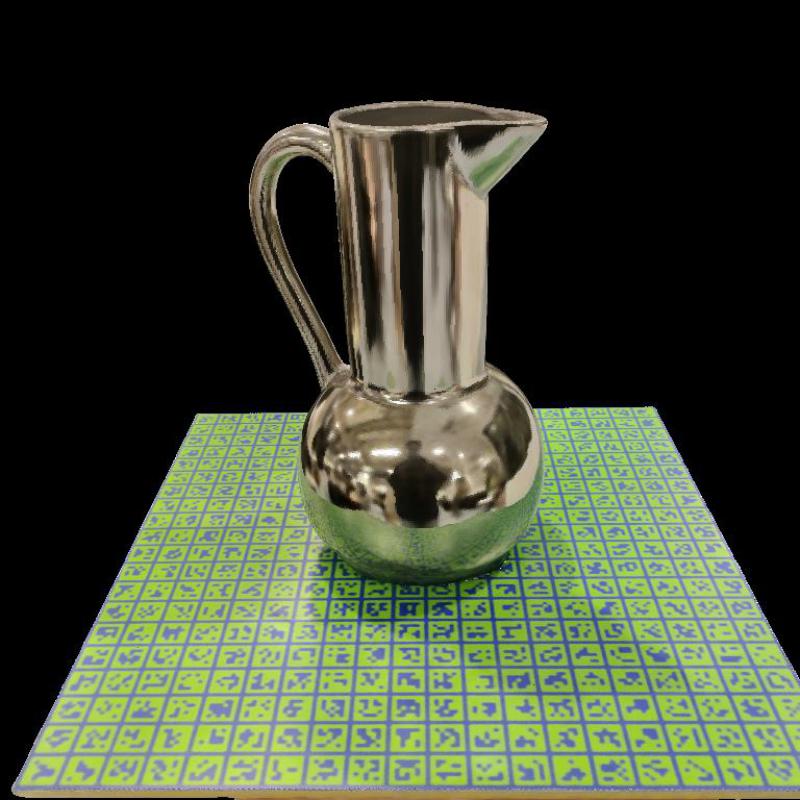} &
    \includegraphics[width=0.16\textwidth]{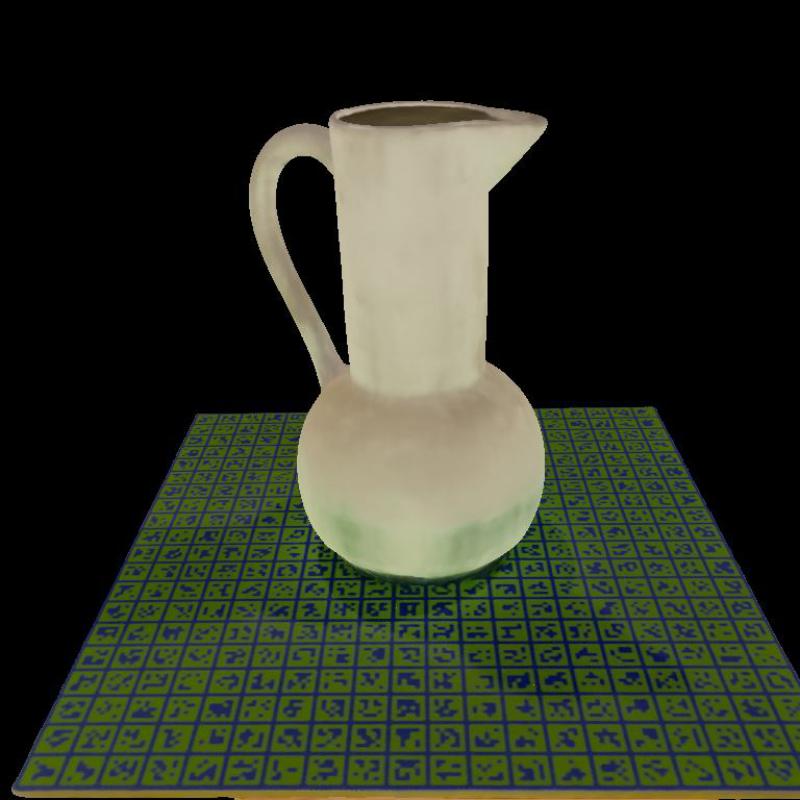} &
    \includegraphics[width=0.16\textwidth]{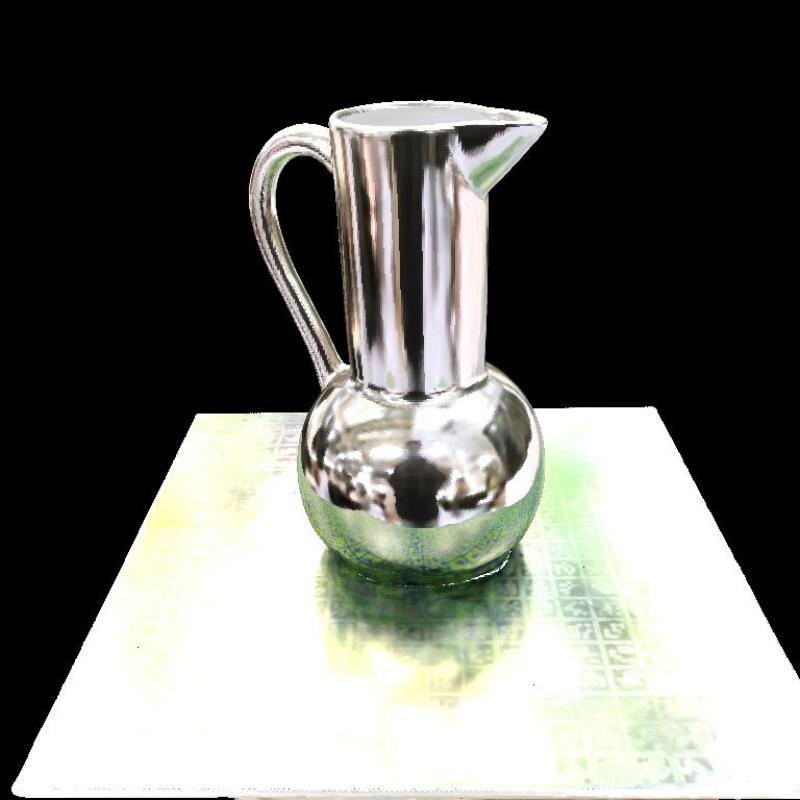} &
    \includegraphics[width=0.16\textwidth]{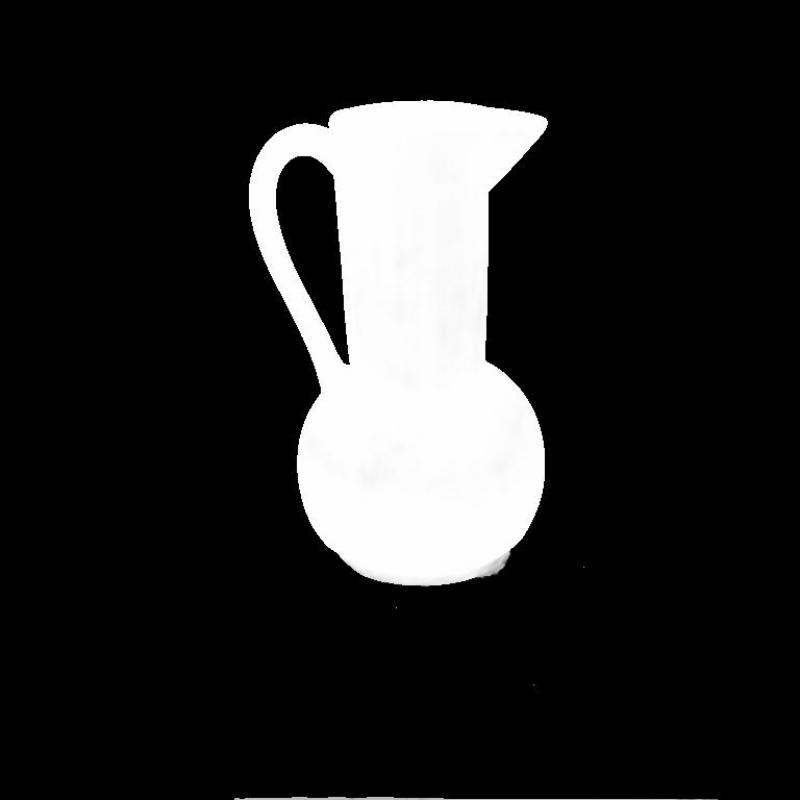} &
    \includegraphics[width=0.16\textwidth]{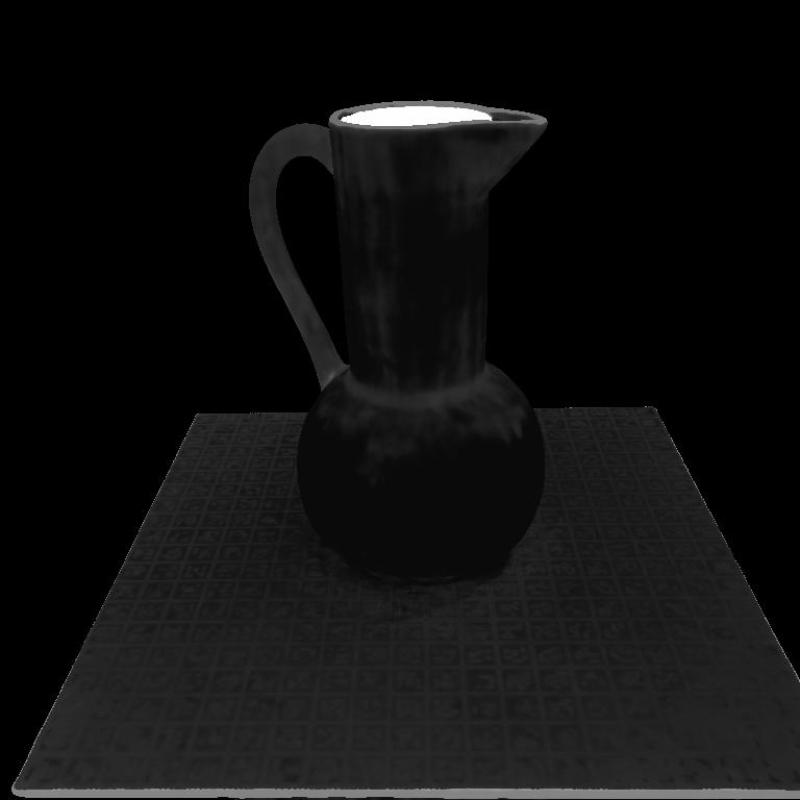} \\
    Image & Rendering & Albedo & Light & Metalness & Roughness \\
    \end{tabular}
    \caption{\textbf{Decomposition of the input image}. After the estimation of surface BRDF, our method is able to automatically decompose the input image into albedo, lights, metalness, and roughness.}
    \label{fig:decom}
\end{figure*}

\textbf{Visualization of image decomposition}. To show the quality of the estimated BRDF and lights, we provide two qualitative examples of image decomposition by our method in Fig.~\ref{fig:decom}. Though the appearances of reflective objects contain strong specular effects, our method successfully separates the high-frequency lights from the low-frequency albedo. The predicted metalness and roughness are also very reasonable. Our method accurately predicts a rougher material for the base of the table bell and a smooth material for the table bell lid in Fig.~\ref{fig:decom} (Row 1). Meanwhile, our method also distinguishes the metallic vase from the non-metallic calibration board in Fig.~\ref{fig:decom} (Row 2). 

\begin{figure}
    \centering
    \setlength\tabcolsep{1pt}
    \renewcommand{\arraystretch}{0.5} 
    \begin{tabular}{ccc}
    \includegraphics[width=0.3\linewidth]{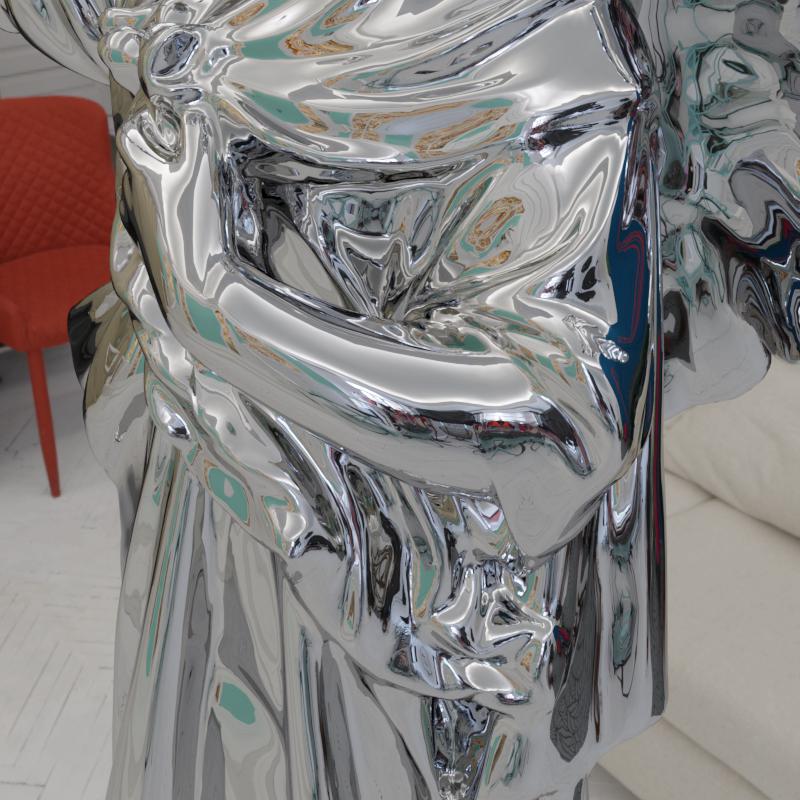} &
    \includegraphics[width=0.3\linewidth]{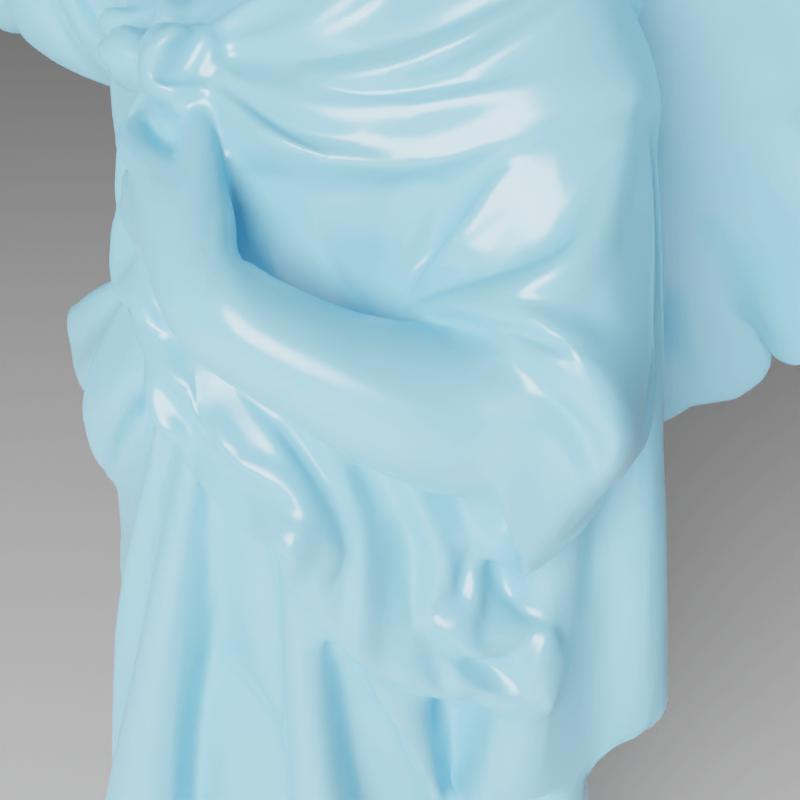} &
    \includegraphics[width=0.3\linewidth]{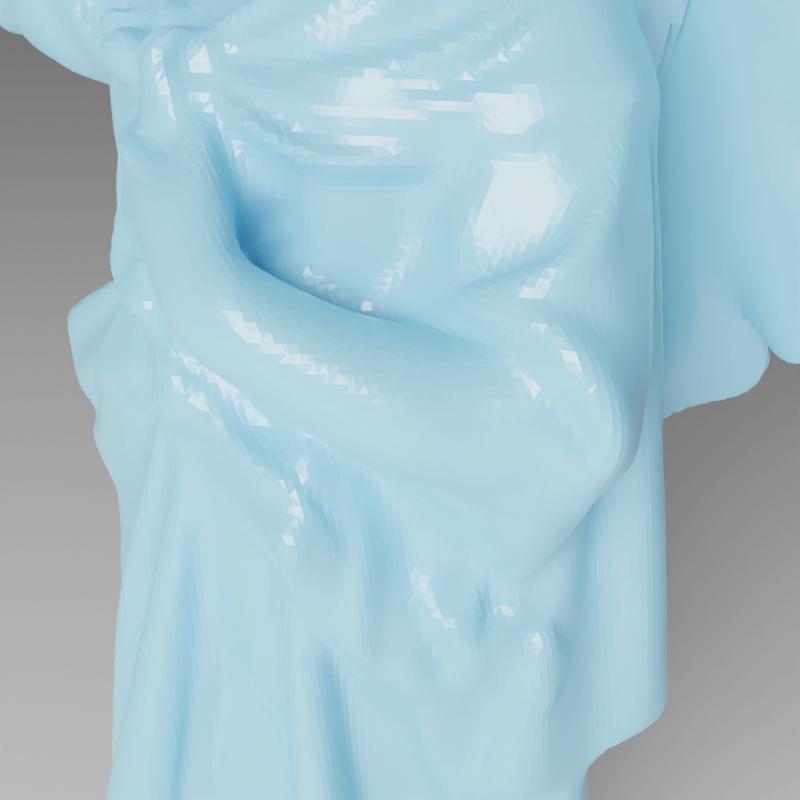} \\
    \includegraphics[width=0.3\linewidth]{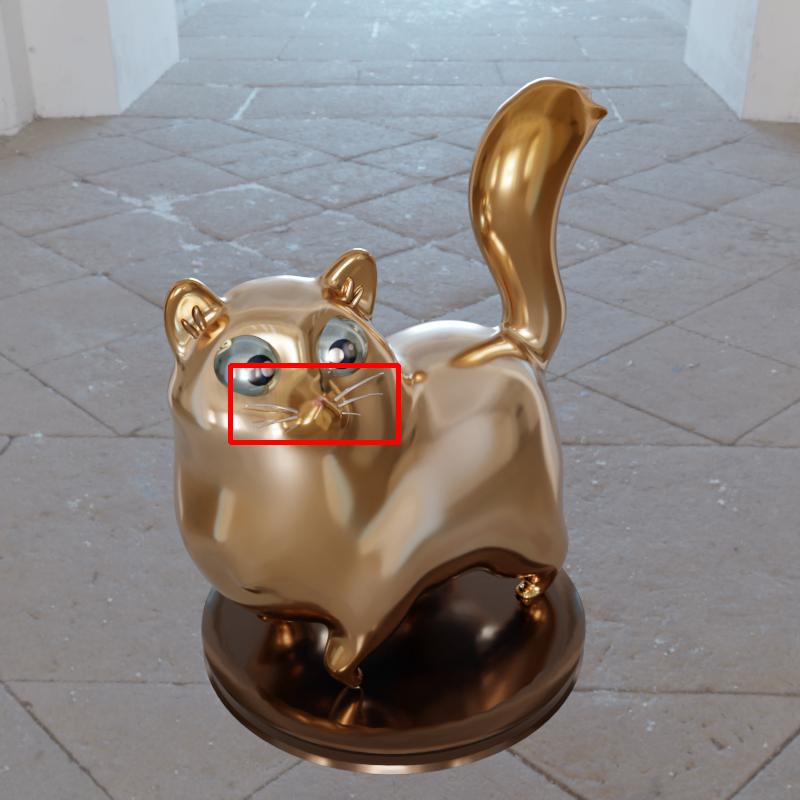} &
    \includegraphics[width=0.3\linewidth]{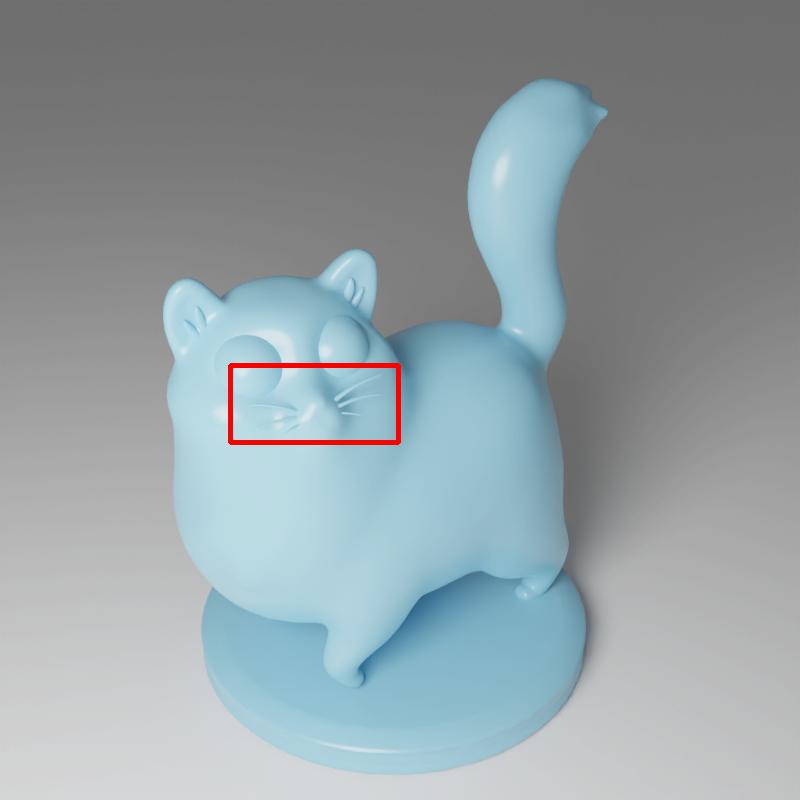} &
    \includegraphics[width=0.3\linewidth]{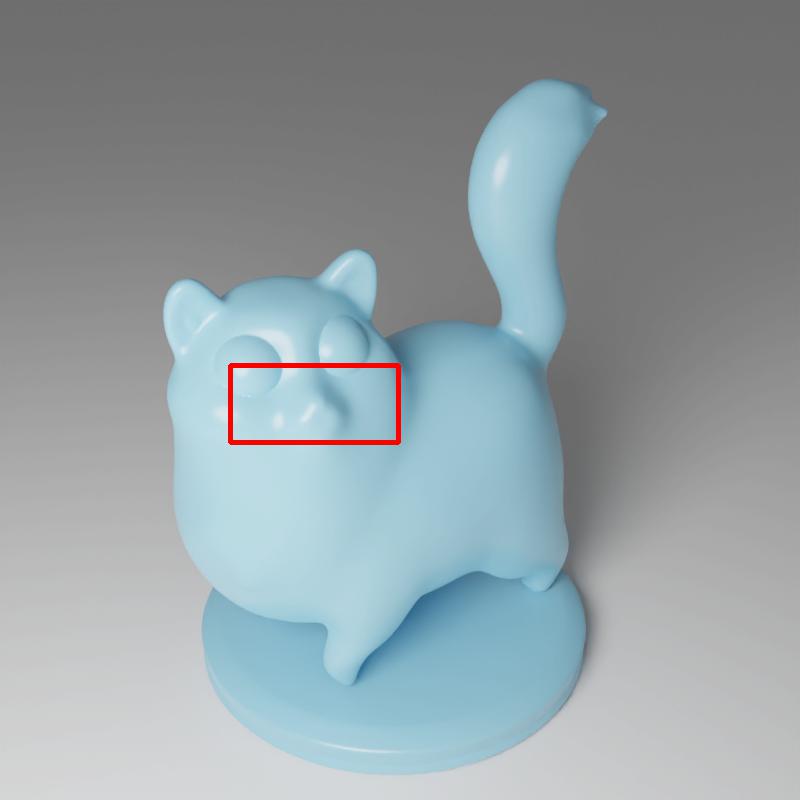} \\
    \includegraphics[width=0.3\linewidth]{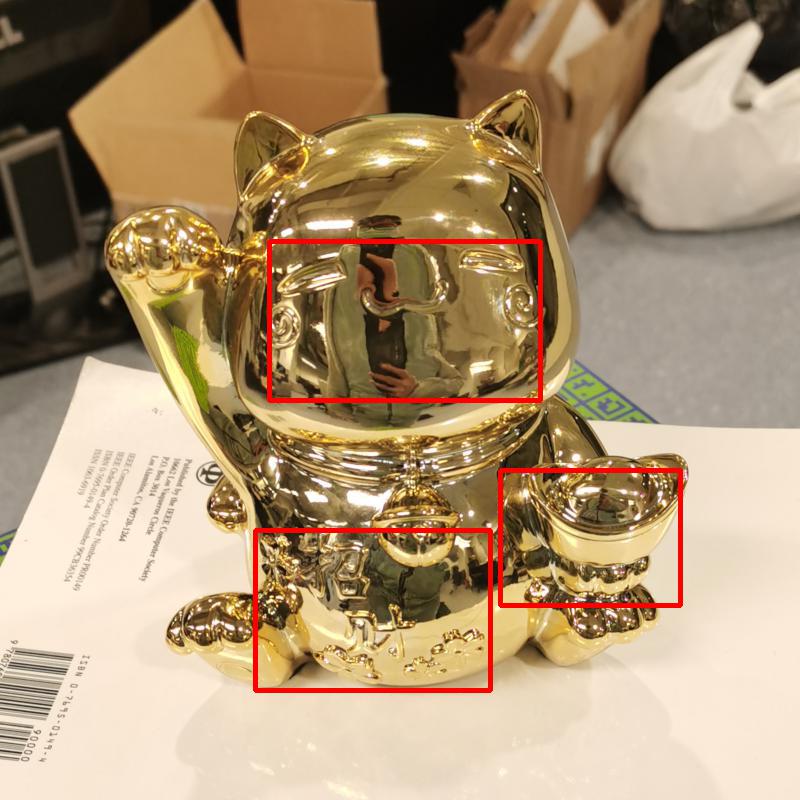} &
    \includegraphics[width=0.3\linewidth]{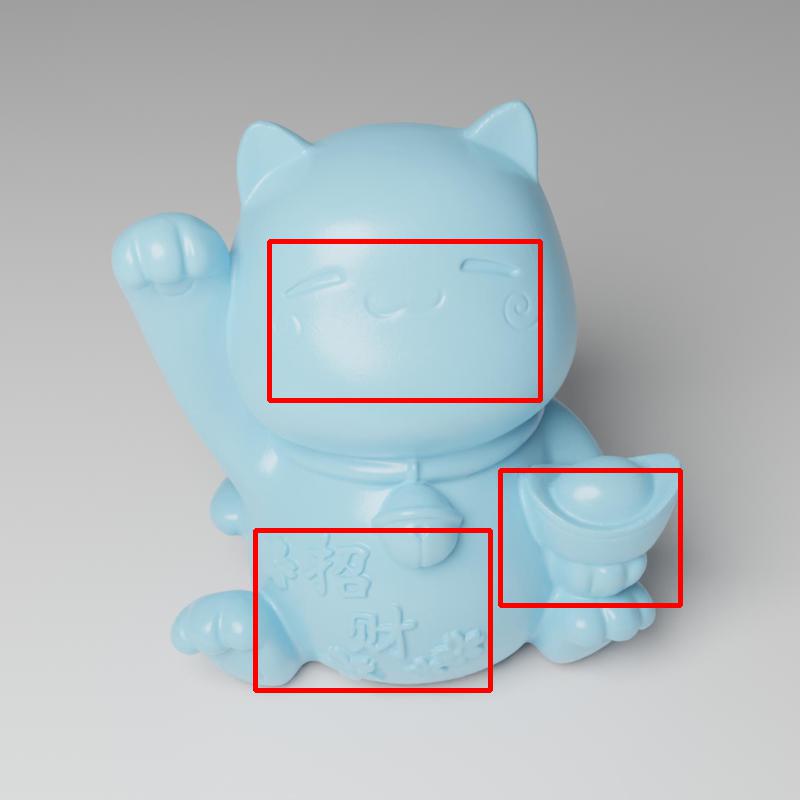} &
    \includegraphics[width=0.3\linewidth]{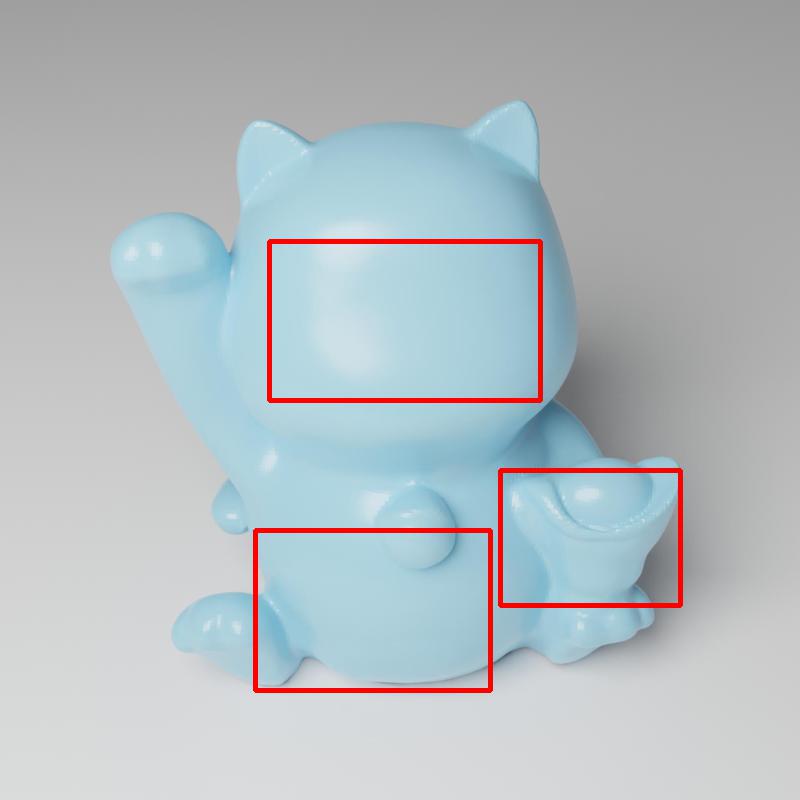} \\
    Images & Ground-truth & Ours \\
    \end{tabular}
    \caption{\textbf{Failure cases for the geometry reconstruction}. Though our method is able to correctly reconstruct the overall shapes of glossy objects, some subtle details are missing or incorrect because the neural SDF field has difficulty in producing abrupt normal changes on these subtle regions.}
    \label{fig:limit}
\end{figure}


\subsection{Limitations}
\textbf{Geometry}. Though we successfully reconstruct the shape of reflective objects, our method still fails to capture some subtle details, as shown in Fig.~\ref{fig:limit}. The main reason is that the rendering function strongly relies on the surface normals estimated by the neural SDF but a neural SDF tends to produce smooth surface normals. Thus, it is hard for the neural SDF to produce abrupt normal changes to reconstruct subtle details like the cloth textures of ``Angel'', the beards of ``Cat'', and the textures of ``Maneki''. 

\begin{figure}
    \centering
    \setlength\tabcolsep{1pt}
    \renewcommand{\arraystretch}{0.5} 
    \begin{tabular}{cccc}
    \includegraphics[width=0.24\linewidth]{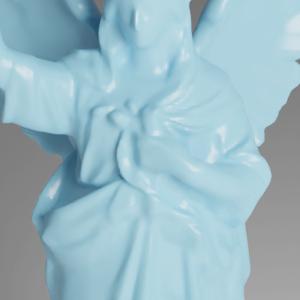} &
    \includegraphics[width=0.24\linewidth]{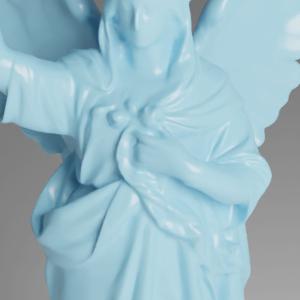} &
    \includegraphics[width=0.24\linewidth]{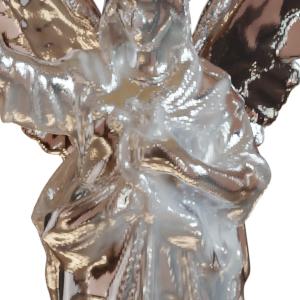} &
    \includegraphics[width=0.24\linewidth]{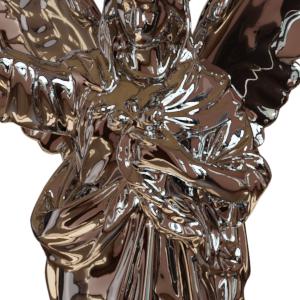} \\
    Geometry & Geometry-GT & Relighting & Relighting-GT \\
    \end{tabular}
    \caption{\textbf{Failure cases for the BRDF estimation}. Reflective appearances are very sensitive to the reflective direction. With inaccurate surface geometry, our method fails to find a correct reflective direction and thus predicts inaccurate BRDFs.}
    \label{fig:brdf_limit}
\end{figure}

\noindent{\textbf{BRDF}}. In the experiments, we observe that our BRDF estimation mainly suffers from incorrect geometry, especially on ``Angel'' as shown in Fig.~\ref{fig:brdf_limit}. Since the appearance of reflective objects strongly relies on the surface normals to compute the reflective directions, the incorrectness of surface normals will make our method struggle to fit correct colors, which leads to inaccurate BRDF estimation. Meanwhile, the BRDF in NeRO does not support advanced reflections such as anisotropic reflections.

\noindent{\textbf{Pose estimation}}. Another limitation is that our method relies on accurate input camera poses and estimating camera poses on reflective objects usually requires stable textures like calibration boards for image matching. Without calibration boards, we may recover poses from other co-visible non-reflective objects or with the help of devices like IMU.

\section{Conclusion}
We have presented NeRO, a neural reconstruction method for accurately reconstructing the geometry and the BRDF of reflective objects, without knowing the environment light conditions and the object masks. 
The key idea of NeRO is to explicitly incorporate the rendering equation in a neural reconstruction framework.
NeRO achieves this challenging goal by proposing a novel light representation and adopting a two-stage approach. In the first stage, by applying tractable approximations, we model both the direct and indirect lights with shading MLPs and learn the surface geometry faithfully. In the second stage, we fix the geometry and reconstruct a more accurate surface BRDF as well as the environment light by Monte Carlo sampling. Experiments have demonstrated that NeRO achieves better surface reconstruction quality and BRDF estimation of reflective objects compared to the state-of-the-art.

\textbf{Acknowledgement}. Lingjie Liu has been supported by the ERC Consolidator Grant 4DReply (77078). We sincerely thank the Tam Wing Fan Innovation Wing of HKU for providing the EinScan scanner to make the Glossy-Real dataset.

\bibliographystyle{ACM-Reference-Format}
\bibliography{sample-base}


\appendix

\section{Appendix}

\subsection{BRDF model}
\label{sec:app_brdf}
We adopt the Cook-Torrance BRDF~\cite{cook1982reflectance}. The basic reflection ratio $F_0=(m*\mathbf{a} + (1-m) * 0.04)$ where $\mathbf{a}$ is the albedo and $m$ is the metalness. Then, the Fresnel term is
\begin{equation}
    F=F_0 + (1-F_0)(1-(\mathbf{h}\cdot \omega_o))^5,
\end{equation}
where $\mathbf{h}$ is the half-way vector and $\omega_o$ is the viewing direction.
The Geometry function is based on the Schlick-GGX Geometry function:
\begin{equation}
    G(\mathbf{n},\omega_o,\omega_i,k)=G_{\rm sub}(\mathbf{n},\omega_o,k) G_{\rm sub}(\mathbf{n},\omega_i,k),
\end{equation}
\begin{equation}
    G_{\rm sub}(\mathbf{n},\mathbf{v},k)=\frac{\mathbf{n}\cdot \mathbf{v}}{(\mathbf{n}\cdot \mathbf{v})(1-k)+k},
\end{equation}
where $k=\rho^4/2$ and $\rho$ is the roughness.
The normal distribution for Stage II is the Trowbridge-Reitz GGX distribution
\begin{equation}
    N(\mathbf{n},\mathbf{h},\alpha) = \frac{\alpha^2}{\pi((\mathbf{n}\cdot\mathbf{h})(\alpha^2-1)+1)^2},
\end{equation}
where $\alpha=\rho^2$.

\begin{figure}
    \centering
    \includegraphics[width=0.4\linewidth]{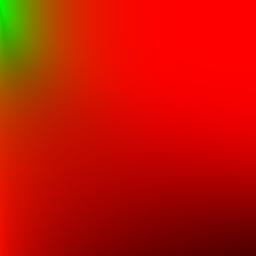}
    \caption{The prefiltered $F_1$ and $F_2$ for the integral of the BRDF.}
    \label{fig:f1f2}
\end{figure}

The prefiltered BRDF $F_1$ and $F_2$ used in the split-sum are stored in an image as shown in Fig.~\ref{fig:f1f2}, where $F_1$ is the red color, $F_2$ is the green color, x-y axis represents the roughness $\rho$ and the $\mathbf{n}\cdot \omega_o$, respectively. Given $\rho$ and $\mathbf{n} \cdot\omega_o$, we interpolate on the image of Fig.~\ref{fig:f1f2} to get $F_1$ and $F_2$.

\subsection{Discussion on the direct light representation}
\label{sec:app_light_rep}
\begin{figure}
    \centering
    \includegraphics[width=\linewidth]{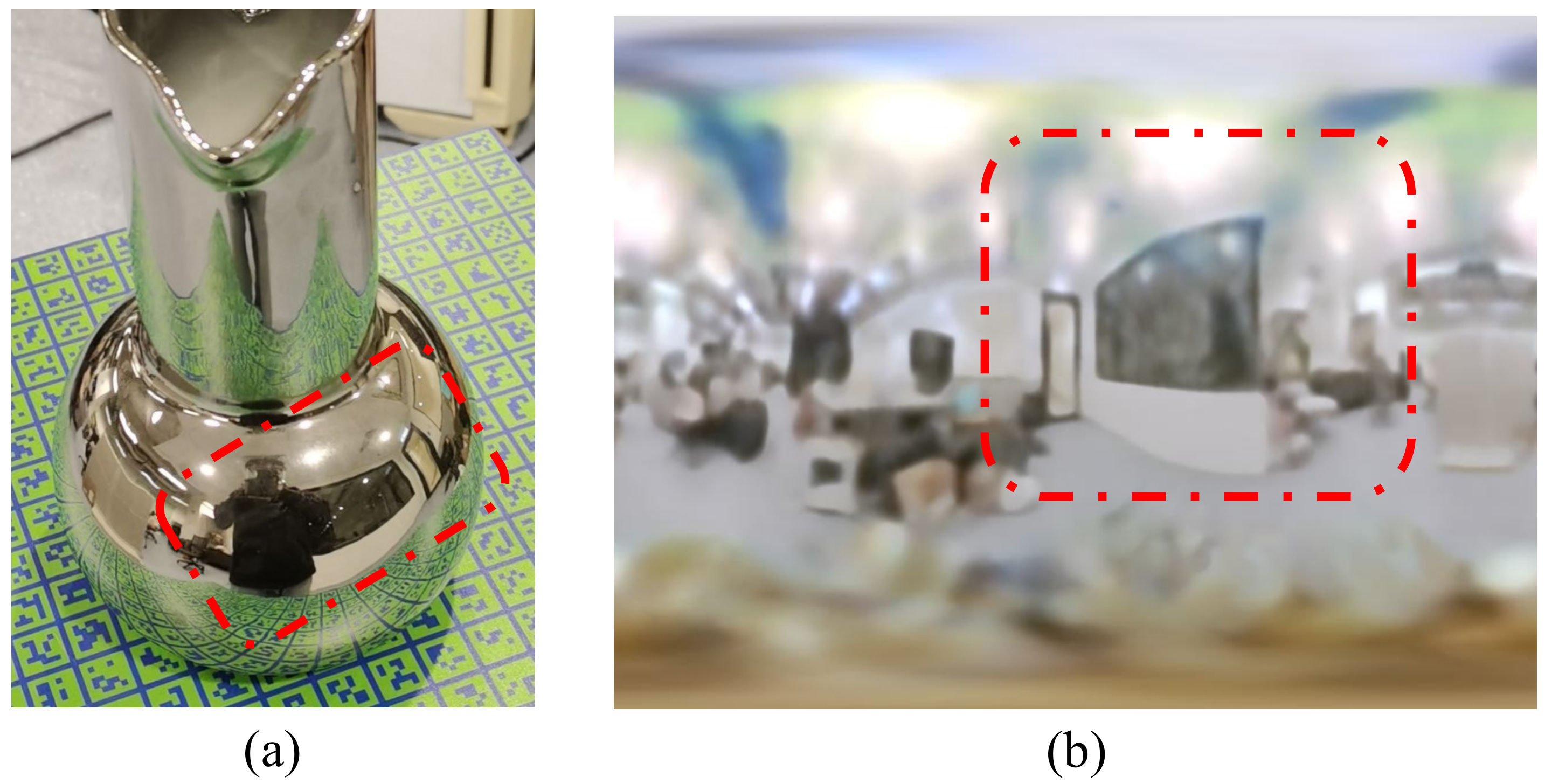}
    \caption{\textbf{Visualization of direct environment lights}. (a) An Image captured in an indoor environment. (b) Estimated direct lights. Even in an indoor environment, modeling the direct environment lights with only directions produces a satisfactory approximation for surface reconstruction.}
    \label{fig:env}
\end{figure}

In Eq.~\ref{eq:light}, the direct light is represented by $g_{\rm direct}(\omega_i)$, which only takes a direction $\omega_i$ as input. This direct light implicitly contains an assumption that direct lights all come from the light sources located at infinity. Such an assumption is accurate enough for surface reconstruction even in a challenging indoor environment. We provide an example in Fig.~\ref{fig:env} to show the environment lights estimated by our method in Stage I. 

\begin{figure}
    \centering
    \includegraphics[width=\linewidth]{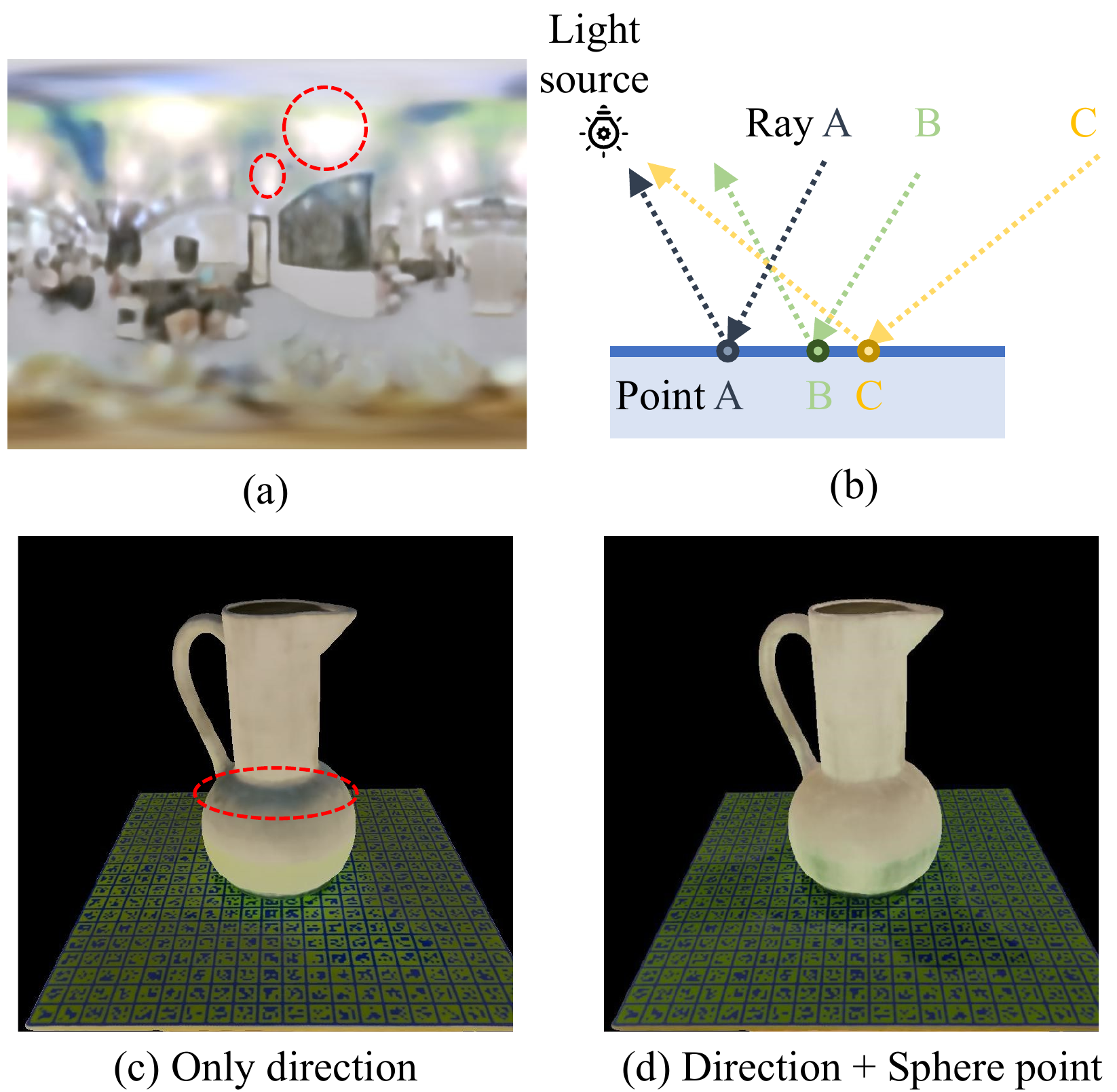}
    \caption{(a) The reconstructed direct environment lights using only directions as input. Note that the strong light sources in the red circles are enlarged. (b) A diagram containing three rays and a light source to show the sensitiveness to strong light sources. (c) Estimated albedo using only directions as the direct light representation. Red circles show the inaccurate albedo estimation. (d) Estimated albedo using both directions and sphere intersection points as the direct light representation. }
    \label{fig:direct_light}
\end{figure}
However, when using such direct light representation in the BRDF estimation of Stage II, we find the approximation is not accurate enough. The main reason for this is that the BRDF estimation is very sensitive to the locations of strong light sources at a finite distance. As shown in Fig.~\ref{fig:direct_light} (a), the estimated strong light sources in red circles are all enlarged. The reason is illustrated by Fig.~\ref{fig:direct_light} (b). There are three rays (A, B, C) corresponding to three surface points (A, B, C). Since both Ray A and Ray C are pointing to the light source, the network will increase the light intensity on the reflective directions of Ray A and C, which causes the strong light source to enlarge. Meanwhile, Ray B is not pointing to the light source but has the same reflective direction as Ray A. Increasing the light intensity on the reflective direction of Ray A will also make Ray B brighter. Therefore, the model will tend to decrease the albedo of Point B. This causes the phenomenon shown in Fig.~\ref{fig:direct_light} (c) that the region in the red circle has a darker albedo than other regions. To resolve this problem, we use the direct light representation $g_{\rm direct}(\mathbf{q}(\mathbf{p},\omega_i),\omega_i)$ in Stage II, where $\mathbf{q}(\mathbf{p},\omega_i)$ is the intersection point on the bounding sphere of the ray emitting from $\mathbf{p}$ along the direction $\omega_i$. We call this sphere intersection encoding. With sphere intersection encoding, we are able to more accurately estimate the albedo as shown in Fig.~\ref{fig:direct_light} (d).

\begin{figure}[]
    \centering
    \setlength\tabcolsep{1pt}
    \begin{tabular}{cccc}
         \multicolumn{2}{c}{With SIE} & \multicolumn{2}{c}{Without SIE}  \\
         \includegraphics[width=0.24\linewidth]{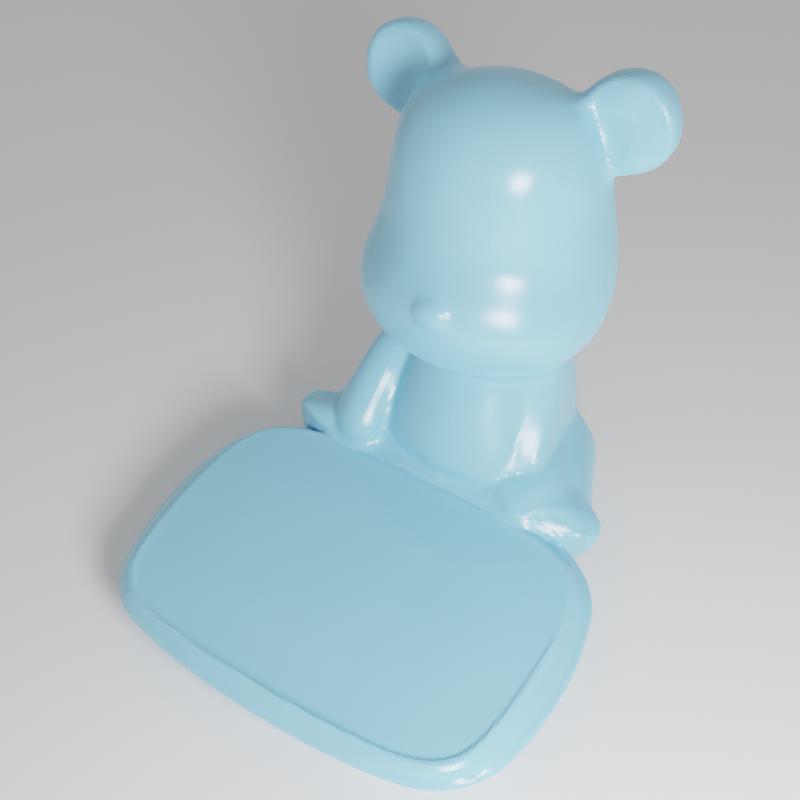} &
         \includegraphics[width=0.24\linewidth]{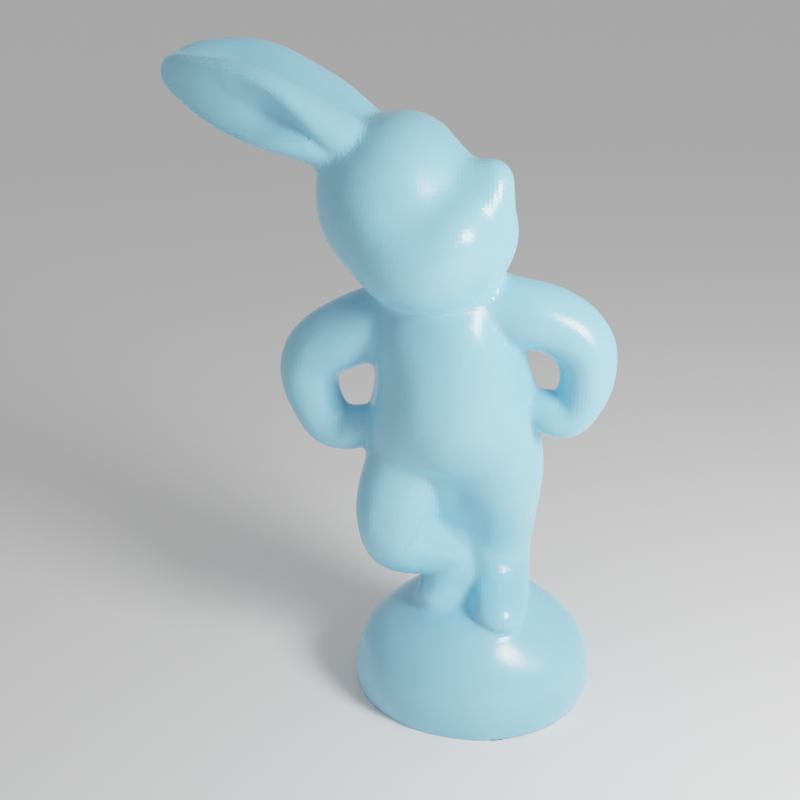} &
         \includegraphics[width=0.24\linewidth]{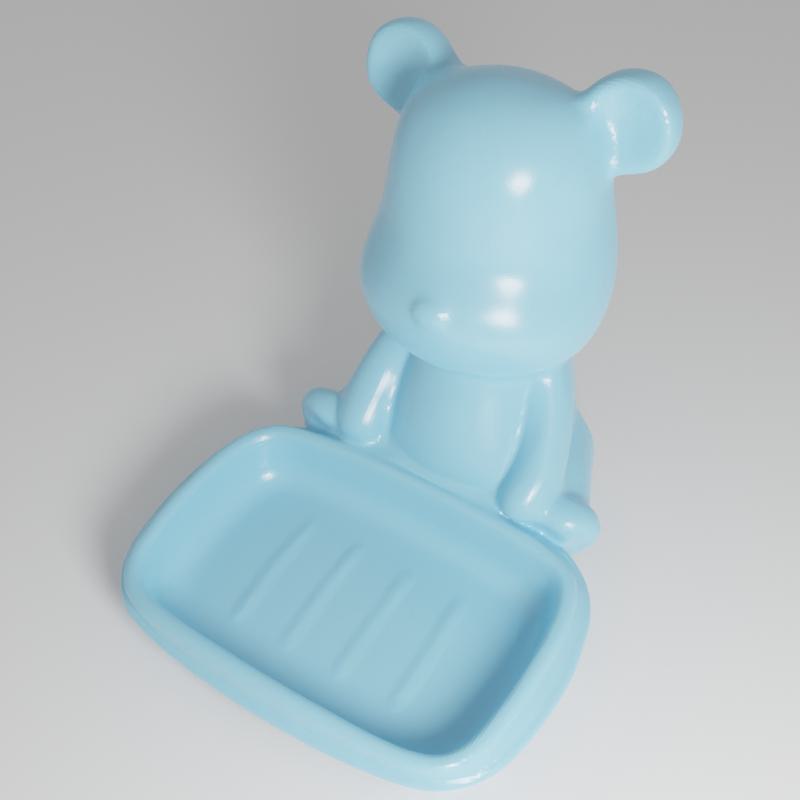} &
         \includegraphics[width=0.24\linewidth]{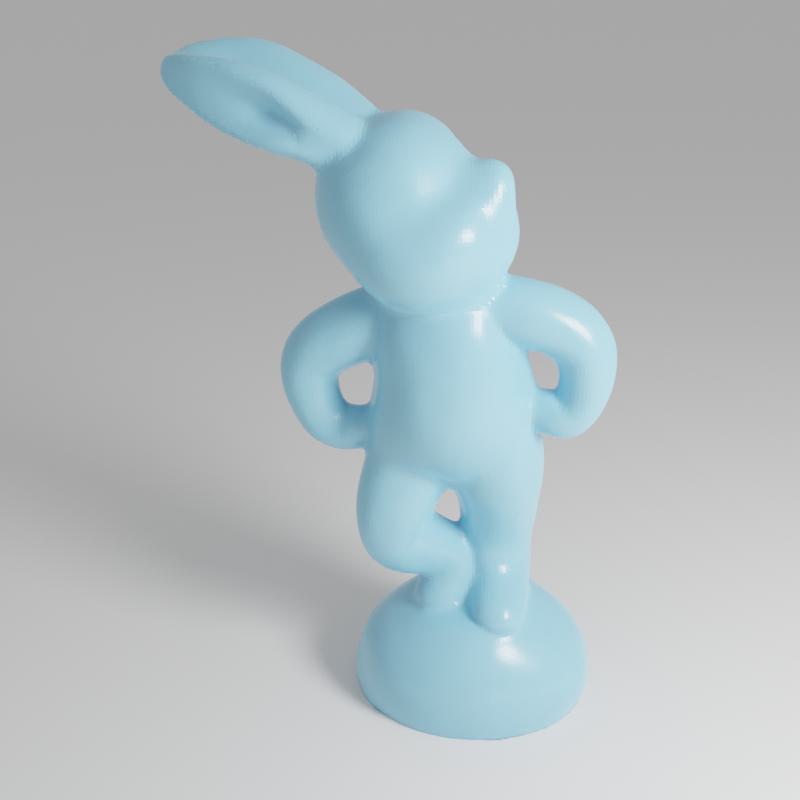} \\
    \end{tabular}
    \caption{Results with or without Sphere Intersection Encoding (SIE).}
    \label{fig:sph}
\end{figure}

An optional choice is to add the sphere intersection encoding in Stage I. We conduct experiments on the ``Bunny'' and ``Bear'' using this sphere intersection encoding. The results are shown in Fig.~\ref{fig:sph}. We find that adding such sphere intersection encoding degenerates the performance. The main reason is that adding a sphere intersection improves the fitting ability, which makes the network indiscriminately focus on color fitting even using distorted surfaces instead of recovering the faithful geometry. 
Stage II is not affected by such an overfitting problem because Stage II uses Monte Carlo sampling to sample many rays to render a single pixel, which avoids overfitting in a single direction.

\subsection{Rationale of light integral approximations}
\label{sec:app_light_int}
The rationale of the two light integral approximations is that: \begin{enumerate}
    \item On a smooth surface with a small specular lobe, the light integral is mainly determined by the light from the reflective direction. In this case, the occlusion probability of the reflective direction will be the dominant factor in the integral. Therefore, the first approximation of using the probability of the reflective direction is a good approximation for all other directions. Meanwhile, as shown in \cite{verbin2022ref}, an  integrated directional encoding (IDE) with a small roughness $\rho$ will produce a high-frequency directional encoding that is suitable to represent strong view-dependent colors on smooth surfaces. Therefore, the IDE approximation also provides a good estimation.
    \item On a rough surface with a large specular lobe, the light integral is not only affected by the reflective direction so the first approximation seems to be too assumptive. However, since the light integral on a large specular lobe will mainly be white and change slowly with the view direction, it is not essential that we use the $g_{\rm direct}$ ($s=0$) or the $g_{\rm indirect}$ ($s=1$) to predict such a slowly-changing white light. Meanwhile, the IDE with a large $\rho$ will also produce a low-frequency directional encoding for this case.
\end{enumerate}

\subsection{Stabilization loss}
\begin{figure}
    \centering
    \includegraphics[width=\linewidth]{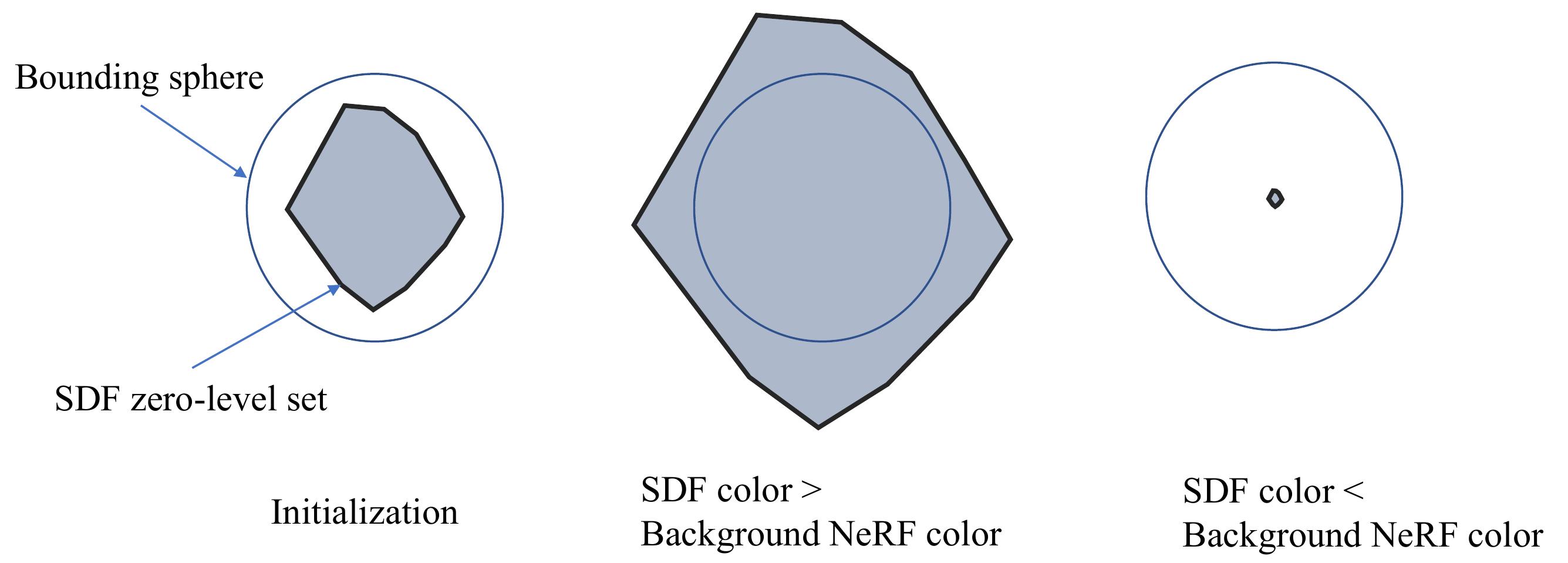}
    \caption{Imbalanced convergence speed between the outer background NeRF color and the SDF shading color will make the reconstruction fail. (left) The initialized zero-level set of neural SDF locates inside the bounding sphere. Note that only the sample points inside the bounding sphere will be used in the neural SDF to compute the opacity density~\cite{wang2021neus}. (middle) If the SDF shading color converges faster than the background NeRF color, then the surface of the neural SDF will dilate to exceed the bounding sphere and is unable to shrink back to the bounding sphere. (right) If the SDF shading color converges slower than the background NeRF color, then the surface of the neural SDF will shrink and eventually disappear. In both cases, the training of NeuS will fail.}
    \label{fig:sdf_init}
\end{figure}
\label{sec:app_stab}
As shown in Fig.~\ref{fig:sdf_init}, the initialization of the neural SDF follows~\cite{atzmon2020sal}. We follow NeuS~\cite{wang2021neus} to use a background NeRF to render the background of the image, which is actually a NeRF++~\cite{zhang2020nerf++} and should be referred to NeuS~\cite{wang2021neus} for more details.
However, the imbalanced convergence speed between the shading color of the neural SDF and the background color provided by the background NeRF will cause the training process to collapse.  If the convergence speed of the foreground shading color is faster, then the surface tends to enlarge to exceed the bounding sphere. Otherwise, it would shrink and eventually disappear. Both cases lead to the failure of training. To avoid these, we prevent the zero-level set from expanding outside the bounding sphere or shrinking to disappear by a stabilization loss $\ell_{\rm stable}$. Note that the bounding sphere is normalized to a unit sphere at the origin. We first sample some points $S_1=\{\mathbf{p}\in \mathbb{R}^3|\|\mathbf{p}\|<0.02\}$ and $S_1=\{\mathbf{p}\in \mathbb{R}^3|\|\mathbf{p}\|>0.98\}$ in the bounding sphere. Then, we penalize the SDF values of these sample points to get close to 0 in the first 1000 steps. We find that such a stabilization loss is essential for a correct convergence of NeuS~\cite{wang2021neus} and our method on these challenging reflective objects. 

\subsection{Network architectures and implementation details}
\label{sec:app_imp}
\textbf{Architectures}. We illustrate the architectures of all MLPs and implementation details in the following and more detailed structures can be referred to the codes at \href{https://github.com/liuyuan-pal/NeRO}{https://github.com/liuyuan-pal/NeRO}. Same as NeuS~\cite{wang2021neus}, $g_{\rm sdf}$ uses a positional encoding with a frequency of 6 as inputs and contains 8 linear layers with the channel number of 256 and a skip connection on the 4th layer. The output of $g_{\rm sdf}$ is an SDF value and a 256-dim feature vector. The 256-dim feature vector is fed into $g_{\rm material}$ with 4 linear layers of 256 channels to output roughness, metalness, and albedo. Both the direct light MLP $g_{\rm direct}$ and indirect light MLP $g_{\rm indirect}$ use the spherical harmonics directional encoding up to degree 5. The occlusion prob MLP $g_{\rm occ}$ and the indirect light MLP $g_{\rm indirect}$ both use the positional encoding of frequency 8. $g_{\rm occ}$, $g_{\rm direct}$, and $g_{\rm indirect}$ all contain 4 layers with a width of 256. The 2D NeRF built on the XoY plane of the camera system also contains 4 layers with a width of 256. In Stage II, we use a base feature extraction network with the same structure as the $g_{\rm sdf}$ but use a higher positional encoding of 8 to extract 256-dim feature vectors. All other MLPs ($g_{\rm direct}$, $g_{\rm indirect}$ and $g_{\rm material}$) in Stage II have the same structures as Stage I. We summarize the trainable components for both stages in Table~\ref{tab:trainable}.

\noindent{\textbf{Training details}}. We train all MLPs in Stage II from scratch. All activation functions in MLPs are ReLUs and we follow NeuS~\cite{wang2021neus} to add weight normalization on all linear weights except the SDF MLP. The final activation for the $g_{\rm material}$ is Sigmoid to get values in $[0,1]$ while the final activation function for $g_{\rm direct}$ and $g_{\rm indirect}$ is the exponential function to get light radiance in $[0,\infty)$. We apply the standard gamma correction to get colors in the sRGB space before computing the rendering loss. On all objects except the bunny object, we freeze the variance used in the computation of opacity density for the first 15k steps for better convergence. On the bunny object, we find such a freezing operation will make the SDF unable to reconstruct the hole between the legs of the bunny. The weights used in the loss computation are $\lambda_{\rm eikonal}=0.1$, $\lambda_{\rm occ}=1.0$, $\lambda_{\rm smooth}=0.05$ and $\lambda_{\rm light}=0.1$ for all experiments.

\subsection{Dataset statistics}
\label{sec:stats}
The Glossy-Blender dataset contains 128 training images for each object, which are uniformly distributed on the upper hemisphere. In order to evaluate the NVS quality, we additionally render 8 evenly-distributed test images to compute the metrics of PSNR, SSIM, and LPIPS. Both training and test images have a resolution of 800$\times$800. The number of images for each object in the Glossy-Real dataset is shown in Table~\ref{tab:stats}. All images have a resolution of 1024$\times$768 and are used for training because our target is shape and material reconstruction.

\begin{table}[]
    \centering
    \begin{tabular}{c|ccccc}
        \toprule
         Object     & Bear & Bunny & Coral & Maneki & Vase \\
         \midrule
         Image Num. & 97   & 129   & 126   & 128    & 128  \\
         \bottomrule
    \end{tabular}
    \caption{Image numbers of each object in the Glossy-Real dataset.}
    \label{tab:stats}
\end{table}

\subsection{Model size}
In this section, we compare the model sizes of NeRO with Ref-NeRF~\cite{verbin2022ref} and NeuS~\cite{wang2021neus} in Table~\ref{tab:size}.
NeRO uses the same networks as NeuS to model SDF and background. The network of NeRO is slightly larger due to the decomposition of the color function into materials and lighting. We further try to increase the width and the depth of the color network of NeuS to get an $\sim$8M model called ``NeuS-Large''. However, even with large model size, NeuS-Large is still unable to correctly reconstruct the reflective surfaces and shows similar results to the original NeuS model. 

\begin{table}[]
    \centering
    \begin{tabular}{c|cccccc}
        \toprule
                  & $g_{\rm sdf}$ & $g_{\rm material}$ & $g_{\rm direct}$ & $g_{\rm indirect}$ & $g_{\rm occ}$ & $g_{\rm camera}$ \\
        \midrule
         Stage I  & \checkmark & \checkmark & \checkmark & \checkmark & \checkmark & \checkmark \\
         Stage II &            & \checkmark & \checkmark & \checkmark &            & \checkmark \\
         \bottomrule
    \end{tabular}
    \caption{Trainable components in Stage I and Stage II.}
    \label{tab:trainable}
\end{table}

\begin{table}[]
    \centering
    \begin{tabular}{c|cccc}
        \toprule
             & Ref-NeRF & NeuS & NeuS-Large & Ours \\
        \midrule
        Model size &   5.2M   & 5.3M & 8.7M       & 7.7M \\
        Chamfer distance$\downarrow$ &
                 0.0169 &  0.0212  & 0.0200     &  \textbf{0.0038} \\
        \bottomrule
    \end{tabular}
    \caption{Model sizes and average CDs on ``Bell'' and ``Cat'' of Ref-NeRF~\cite{verbin2022ref}, NeuS~\cite{wang2021neus} and our method. ``NeuS-Large'' means that we use a deeper and wider color network for NeuS to make the model larger.}
    \label{tab:size}
\end{table}

\subsection{Relighting results in SSIM/LPIPS}
\label{sec:ssim_lpips}
We provide the evaluation of relighting quality on the Glossy-Blender dataset in terms of SSIM~\cite{wang2004image} and LPIPS~\cite{zhang2018unreasonable} in Table~\ref{tab:ssim} and Table~\ref{tab:lpips} respectively. On both metrics, we observe similar results to PSNR where our method outperforms the baselines by a significant margin on these reflective objects.

\begin{table}[]
    \centering
    \begin{tabular}{cccccc}
    \toprule
         &   NDR & NDRMC & MII   & NeILF & Ours  \\
     \midrule
    Bell    & 0.913 & 0.903 & 0.882 & 0.916 & \textbf{0.965} \\
    Cat     & 0.901 & 0.907 & 0.889 & 0.921 & \textbf{0.962} \\
    Teapot  & 0.844 & 0.899 & 0.884 & 0.918 & \textbf{0.977} \\
    Potion  & 0.824 & 0.858 & 0.885 & 0.903 & \textbf{0.950} \\
    TBell   & 0.739 & 0.883 & 0.870 & 0.897 & \textbf{0.968} \\
    Angel   & 0.864 & 0.865 & 0.867 & 0.889 & \textbf{0.911} \\
    Horse   & 0.930 & 0.935 & 0.893 & 0.945 & \textbf{0.954} \\
    Luyu    & 0.854 & 0.859 & 0.850 & 0.877 & \textbf{0.914} \\
    \midrule
    Avg.    & 0.859 & 0.889 & 0.878 & 0.908 & \textbf{0.950} \\
    \bottomrule
    \end{tabular}
    \caption{SSIM$\uparrow$ of NDR~\cite{munkberg2022extracting}, NDRMC~\cite{hasselgren2022shape}, MII~\cite{zhang2022modeling}, NeILF~\cite{yao2022neilf} and our method on the Glossy-Blender dataset.}
    \label{tab:lpips}
\end{table}

\begin{table}[]
    \centering
    \begin{tabular}{cccccc}
    \toprule
         &   NDR & NDRMC & MII   & NeILF & Ours  \\
     \midrule
    Bell    & 0.0989 & 0.1180 & 0.1477 & 0.1056 & \textbf{0.0557} \\
    Cat     & 0.1104 & 0.1150 & 0.1437 & 0.0737 & \textbf{0.0523} \\
    Teapot  & 0.1385 & 0.1207 & 0.1373 & 0.0980 & \textbf{0.0283} \\
    Potion  & 0.2062 & 0.2027 & 0.1723 & 0.1443 & \textbf{0.0843} \\
    TBell   & 0.2632 & 0.1703 & 0.2117 & 0.1407 & \textbf{0.0460} \\
    Angel   & 0.1066 & 0.1213 & 0.1183 & 0.0940 & \textbf{0.0790} \\
    Horse   & 0.0492 & 0.0530 & 0.0723 & 0.0457 & \textbf{0.0403} \\
    Luyu    & 0.1085 & 0.1080 & 0.1300 & 0.0960 & \textbf{0.0723} \\
    \midrule
    Avg.    & 0.1352 & 0.1261 & 0.1417 & 0.0996 & \textbf{0.0573} \\
    \bottomrule
    \end{tabular}
    \caption{LPIPS$\downarrow$ of NDR~\cite{munkberg2022extracting}, NDRMC~\cite{hasselgren2022shape}, MII~\cite{zhang2022modeling}, NeILF~\cite{yao2022neilf} and our method on the Glossy-Blender dataset.}
    \label{tab:ssim}
\end{table}

\subsection{Novel-view synthesis quality}
\label{sec:nvs}

\begin{table}[]
    \centering
    \begin{tabular}{cccc}
    \toprule
                        &   NeuS  & Ref-NeRF & Ours  \\
     \midrule
    PSNR$\uparrow$      &  27.80  & 27.86    & \textbf{29.73}\\
    SSIM$\uparrow$      &  0.875  & 0.878    & \textbf{0.904}\\
    LPIPS$\downarrow$   &  0.365  & 0.375    & \textbf{0.324}\\
    \bottomrule
    \end{tabular}
    \caption{NVS quality of NeuS~\cite{wang2021neus} and Ref-NeRF~\cite{verbin2022ref} on the Glossy-Blender dataset with PSNR, SSIM and LPIPS.}
    \label{tab:nvs}
\end{table}

To show the quality of novel view synthesis (NVS), we additionally render 8 novel-view images on the Glossy-Blender dataset and report the NVS quality on these images in Table~\ref{tab:nvs} in terms of PSNR, SSIM, and LPIPS.

\subsection{Results on less- or non-reflective objects}

\begin{figure*}
    \centering
    \setlength\tabcolsep{1pt}
    \begin{tabular}{cccccc}
    \includegraphics[width=0.16\textwidth]{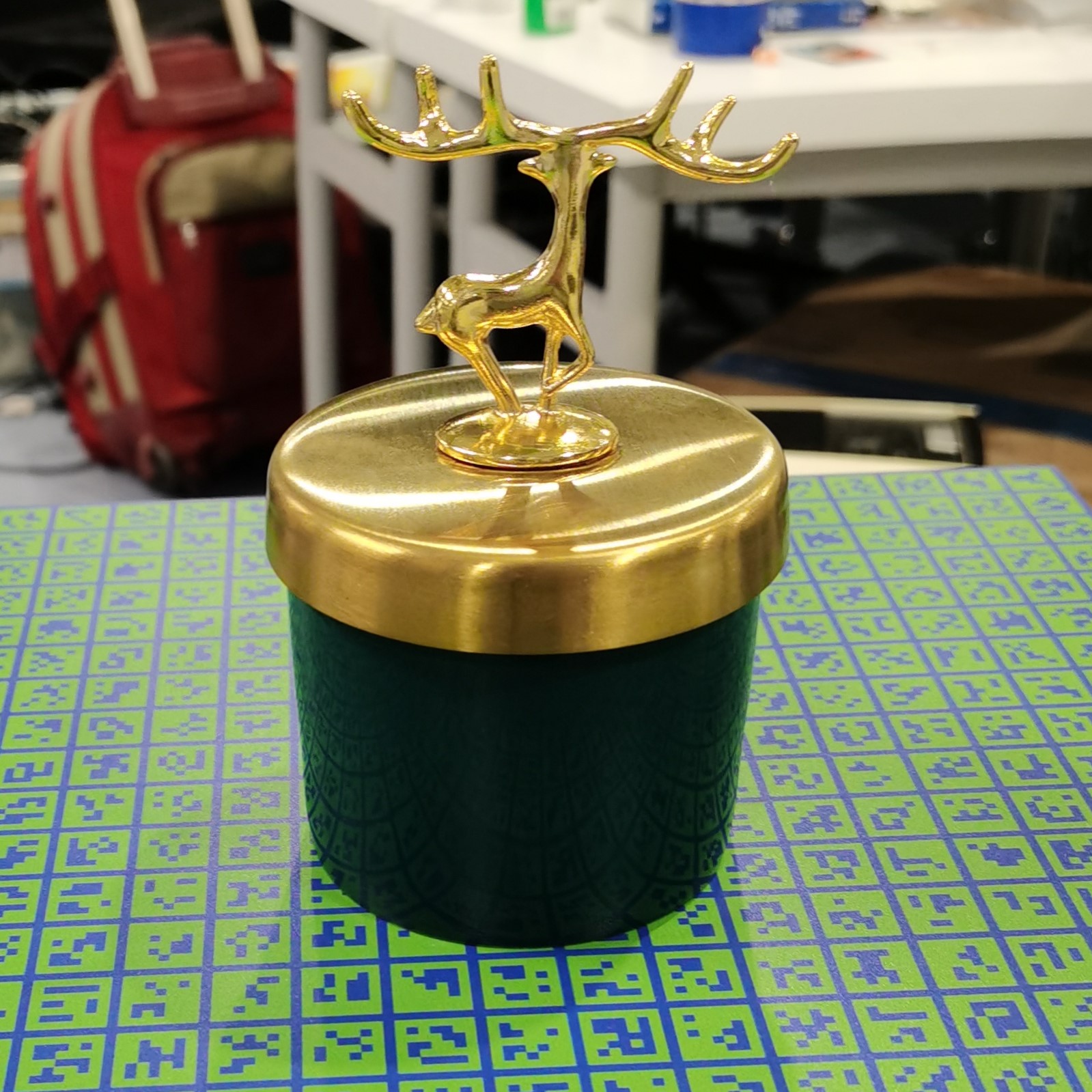} &
    \includegraphics[width=0.16\textwidth]{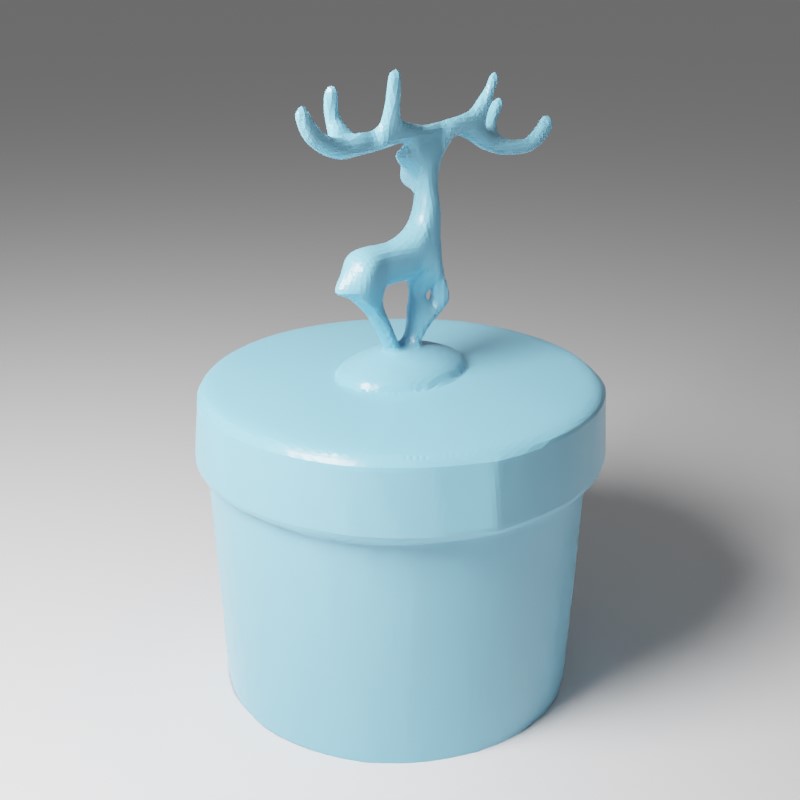} & 
    \includegraphics[width=0.16\textwidth]{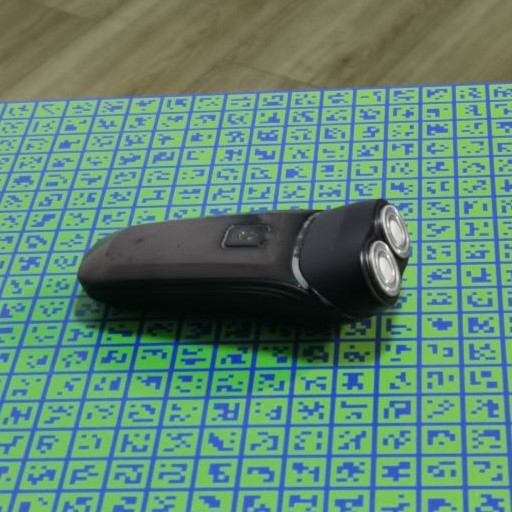} &
    \includegraphics[width=0.16\textwidth]{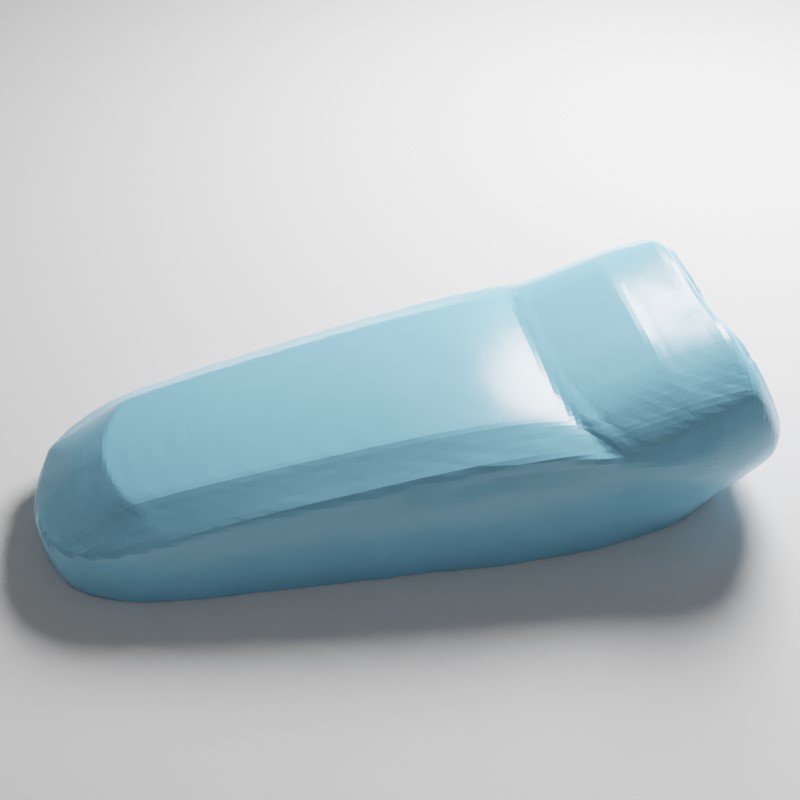} & 
    \includegraphics[width=0.16\textwidth]{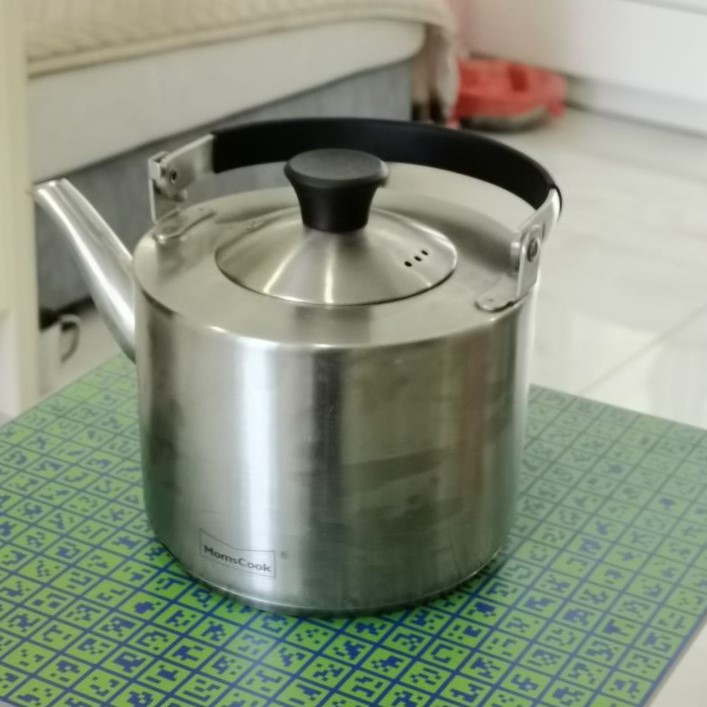} &
    \includegraphics[width=0.16\textwidth]{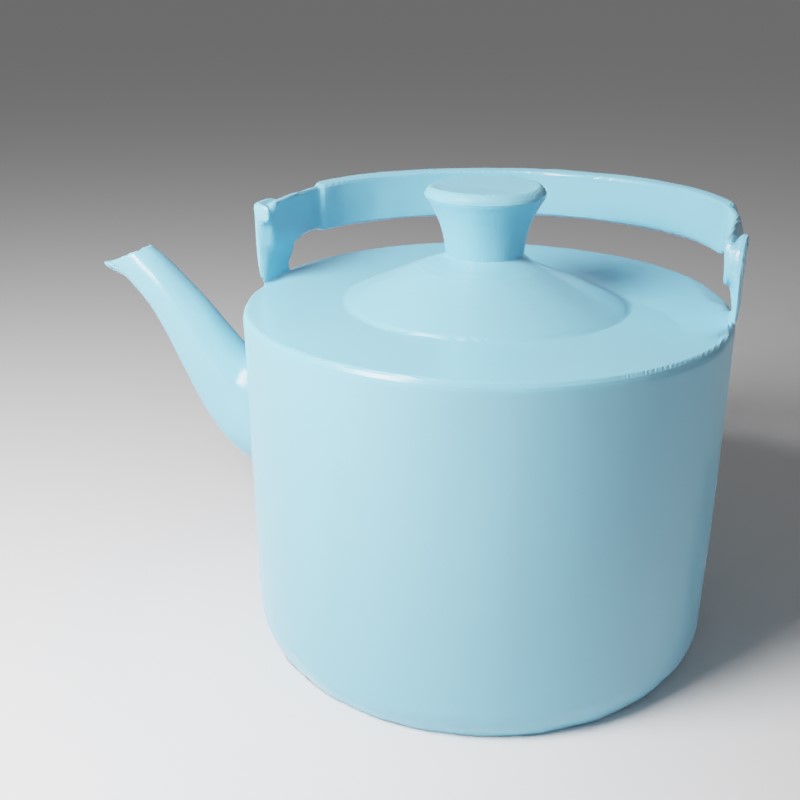} \\
    \end{tabular}
    \caption{Reconstruction results on less reflective objects.}
    \label{fig:less_ref}
\end{figure*}

\begin{figure*}
    \centering
    \setlength\tabcolsep{1pt}
    \begin{tabular}{cccc}
    \includegraphics[width=0.2\textwidth]{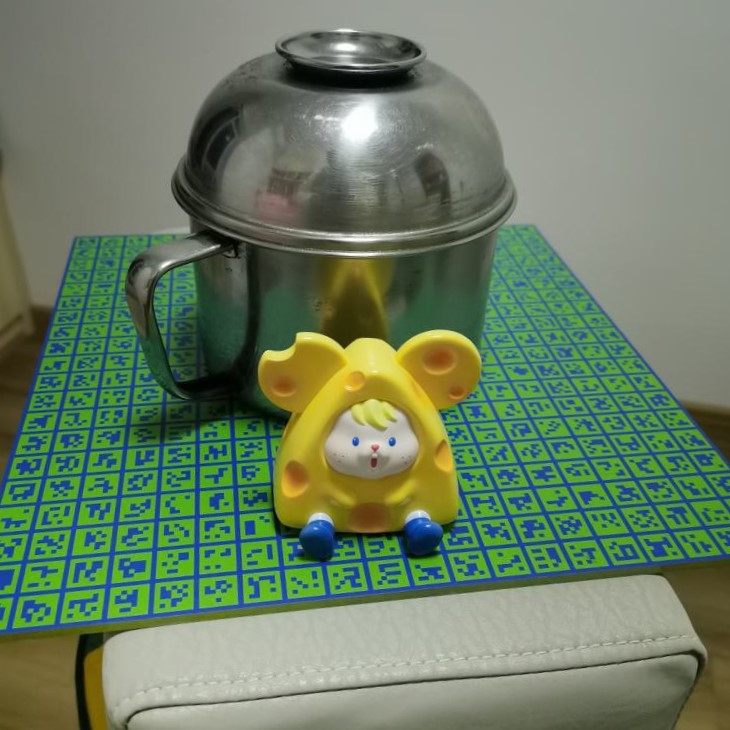} &
    \includegraphics[width=0.2\textwidth]{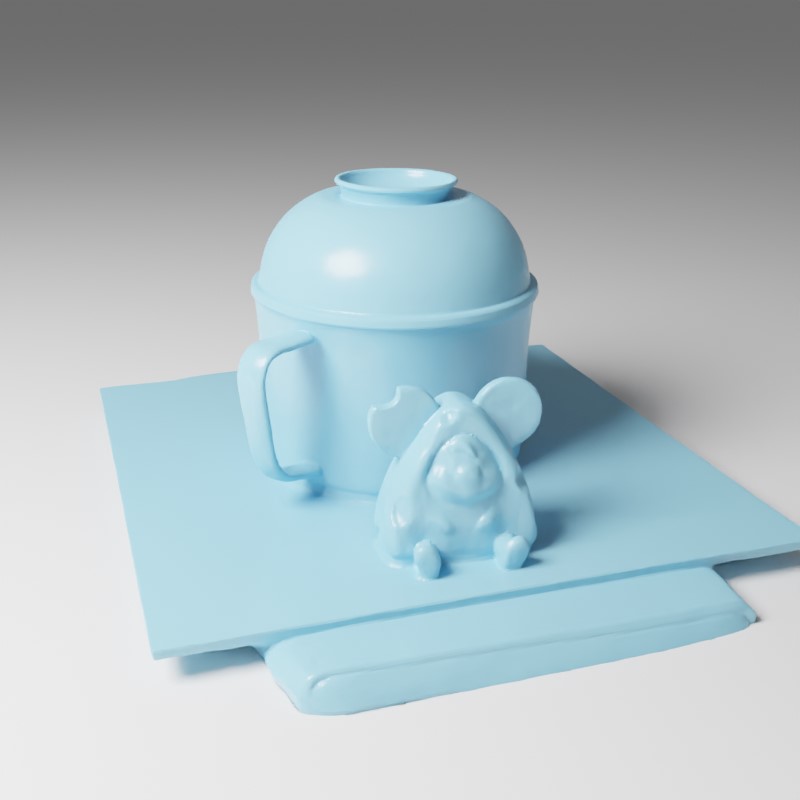} &
    \includegraphics[width=0.2\textwidth]{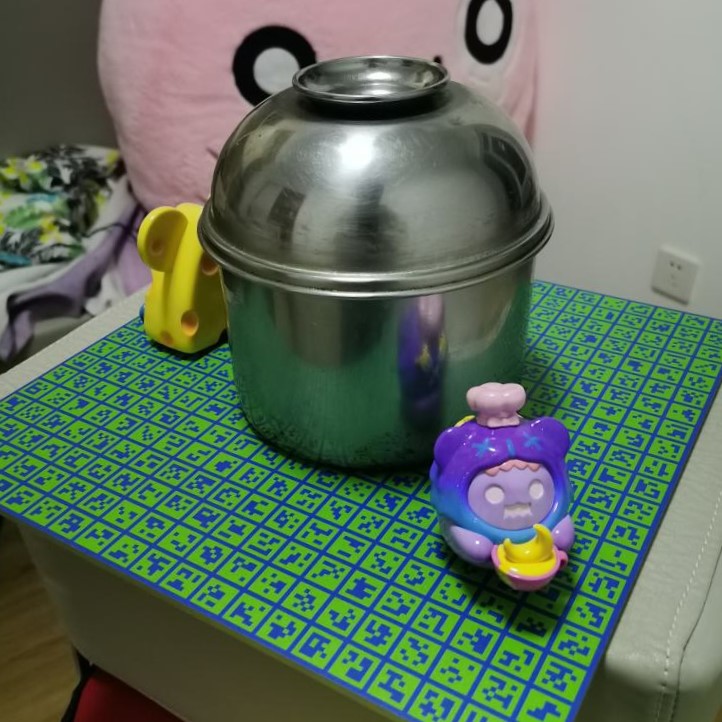} &
    \includegraphics[width=0.2\textwidth]{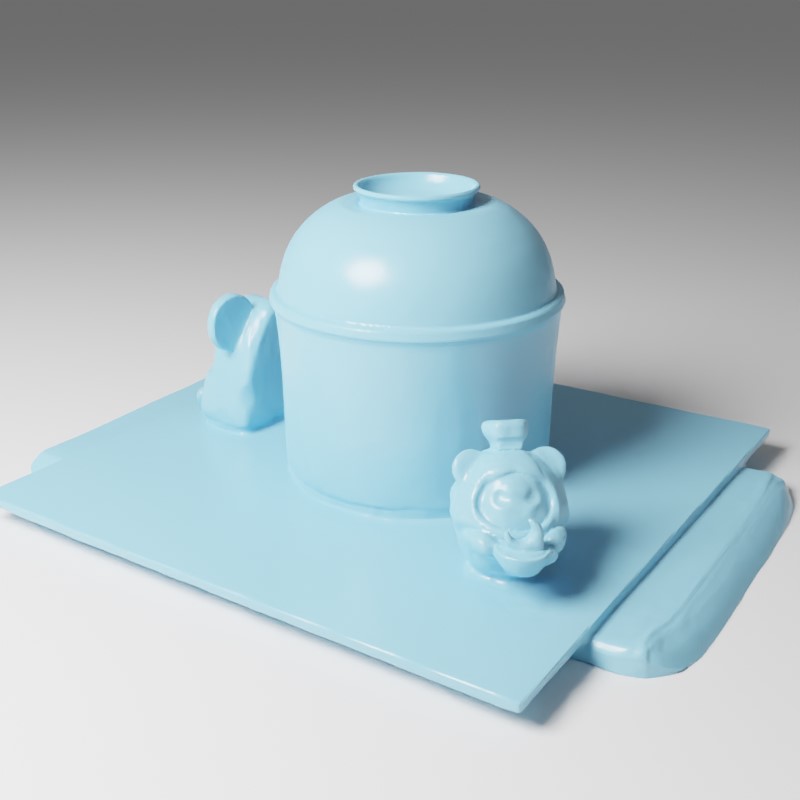} \\
    \includegraphics[width=0.2\textwidth]{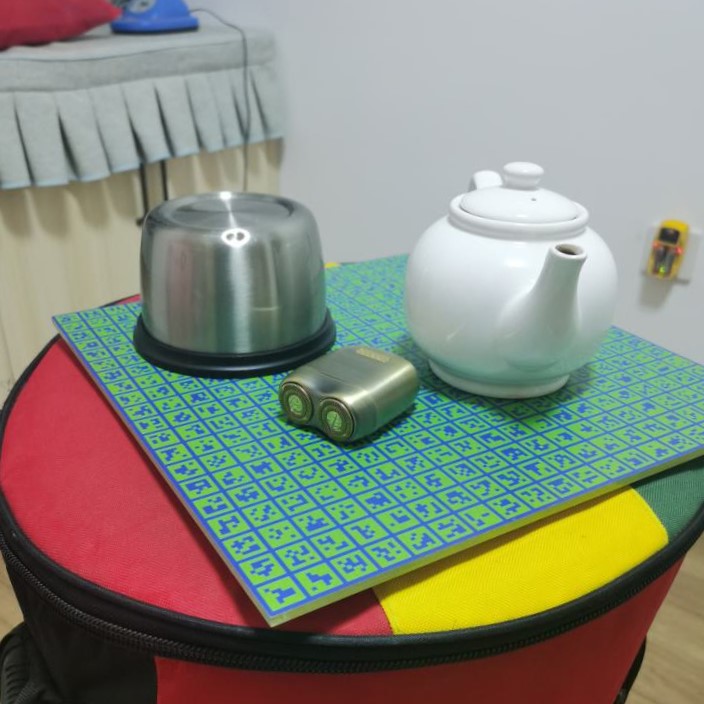} &
    \includegraphics[width=0.2\textwidth]{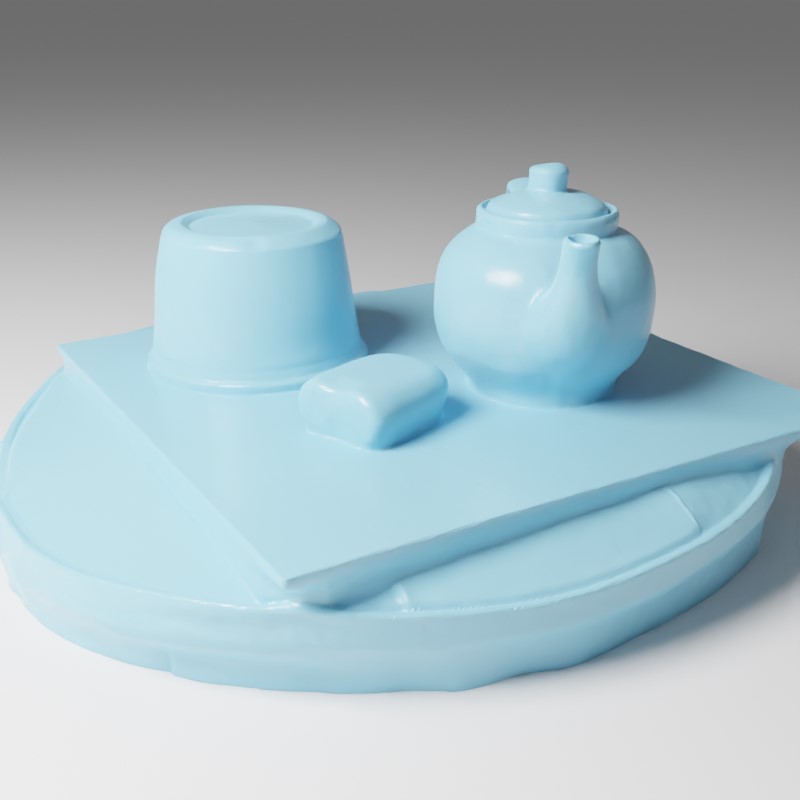} &
    \includegraphics[width=0.2\textwidth]{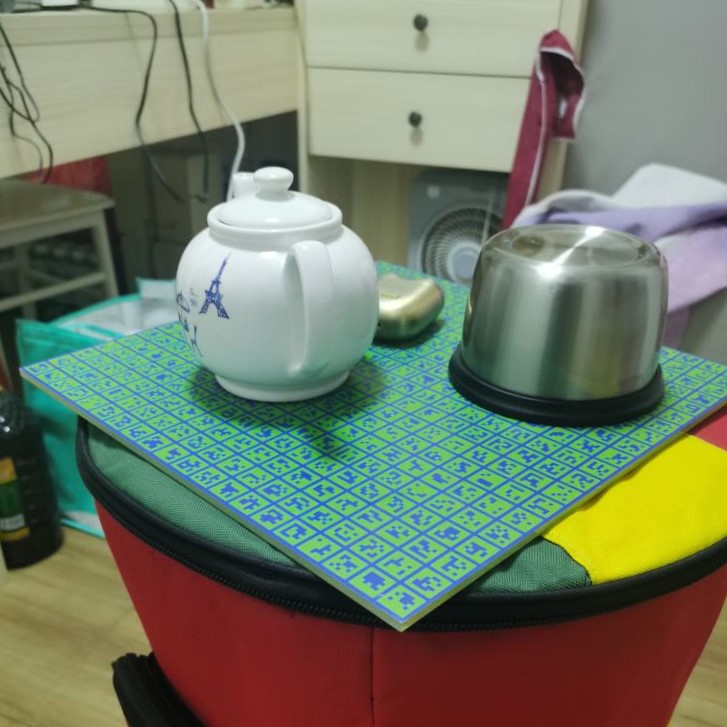} &
    \includegraphics[width=0.2\textwidth]{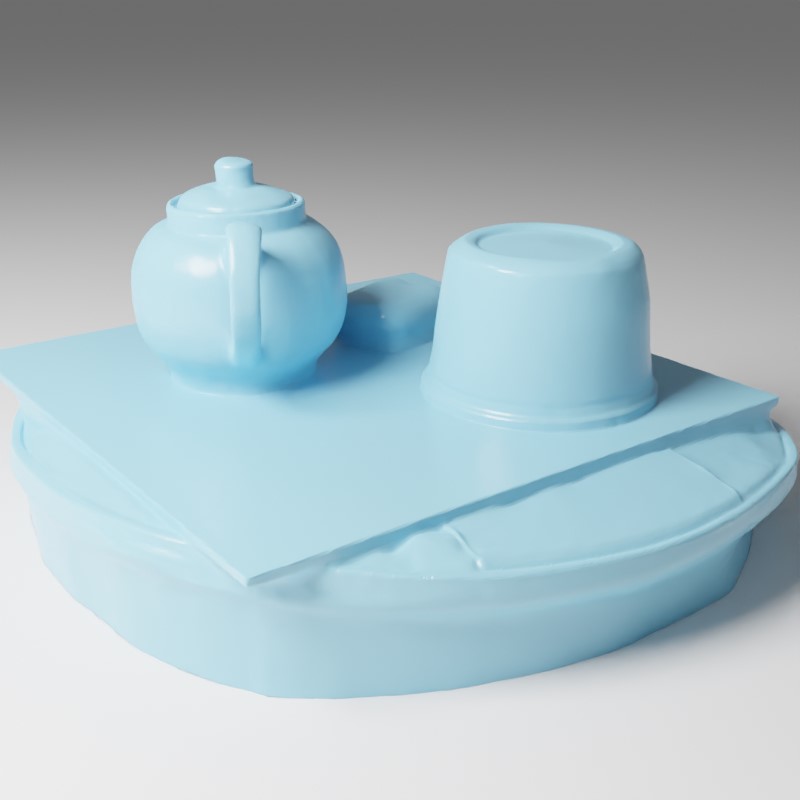} \\
    \end{tabular}
    \caption{Reconstruction results on both reflective and non-reflective objects.}
    \label{fig:set_recon}
\end{figure*}

\begin{figure*}
    \centering
    \setlength\tabcolsep{1pt}
    \begin{tabular}{cccccc}
    \includegraphics[width=0.16\textwidth]{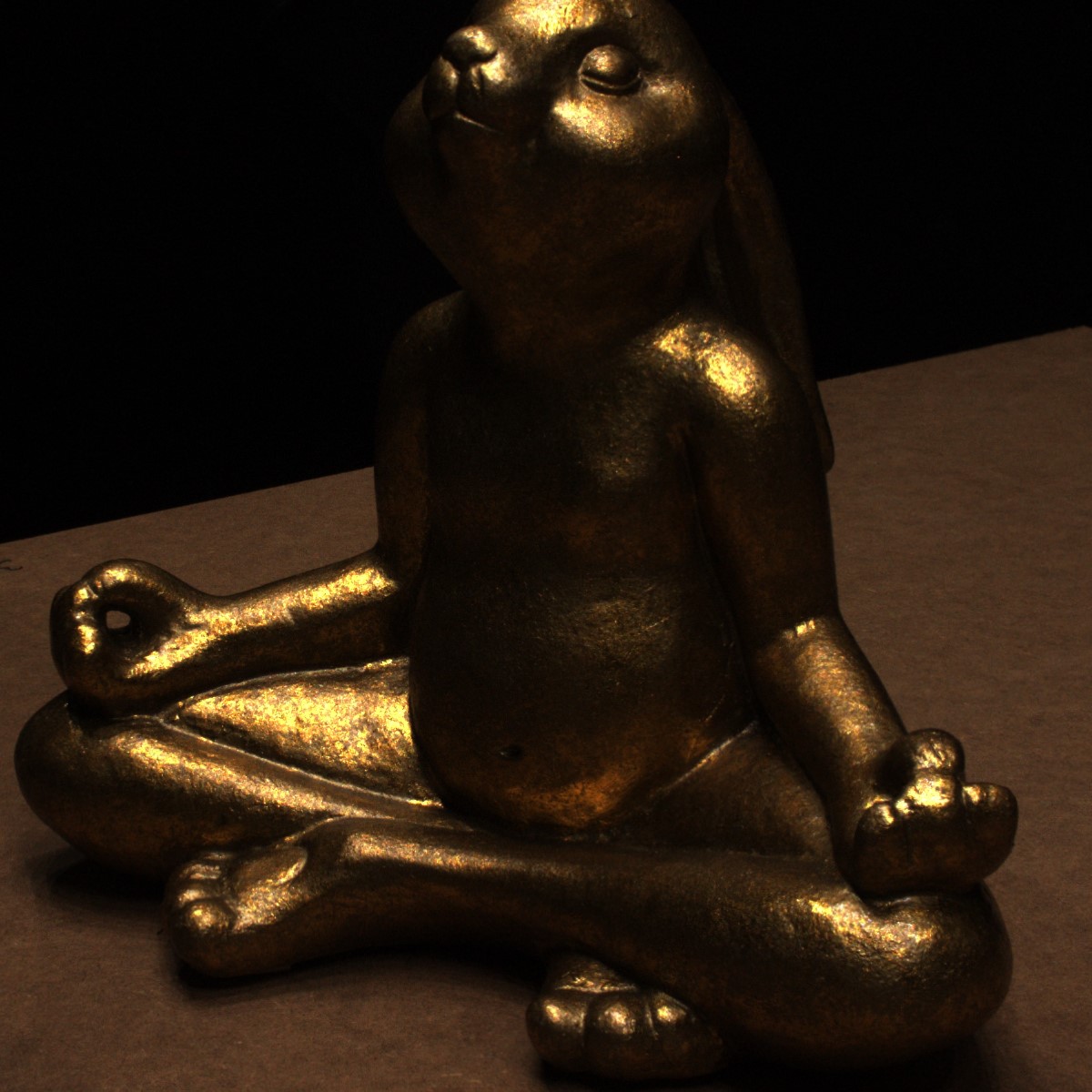} &
    \includegraphics[width=0.16\textwidth]{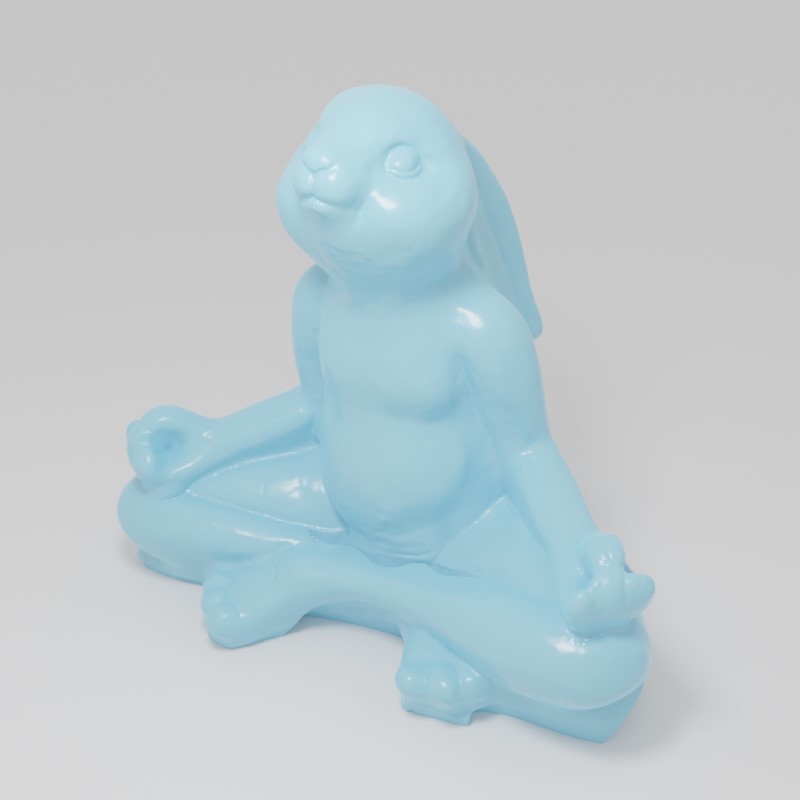} &
    \includegraphics[width=0.16\textwidth]{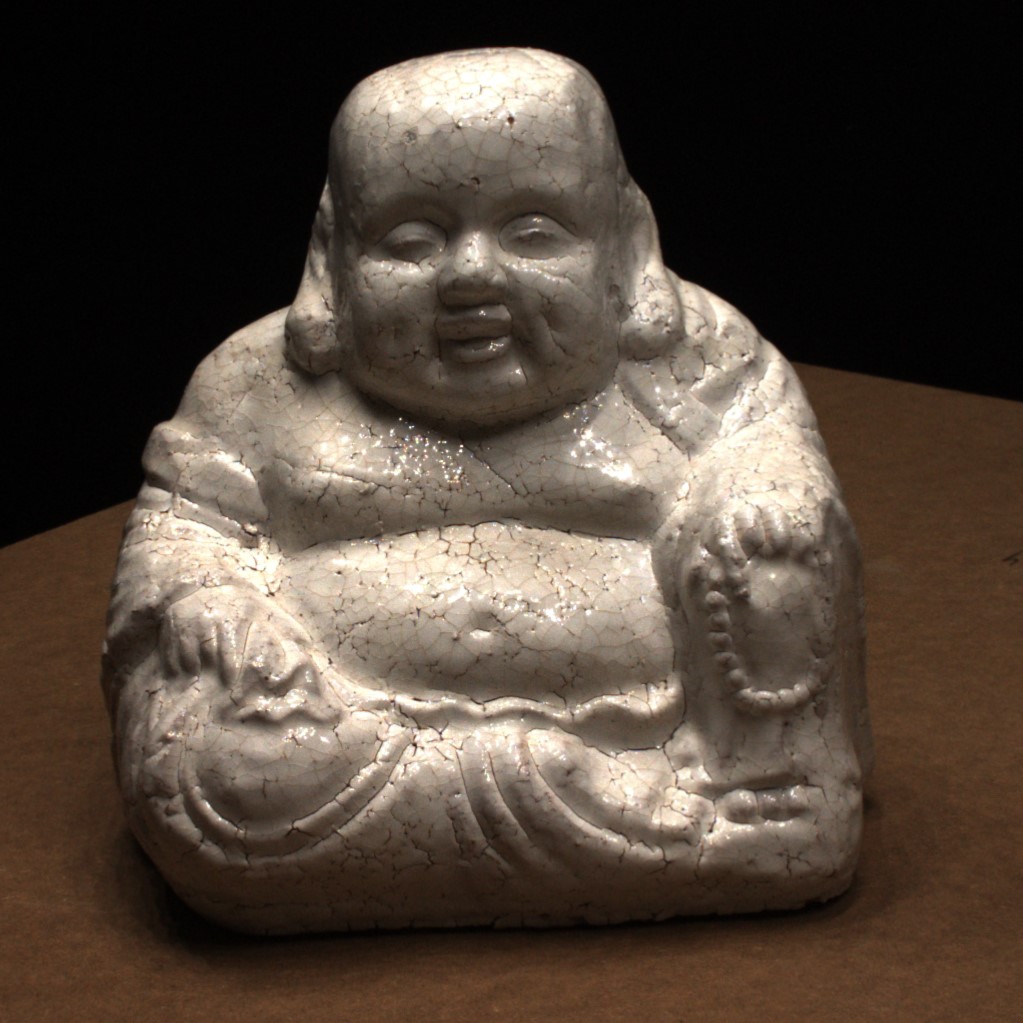} &
    \includegraphics[width=0.16\textwidth]{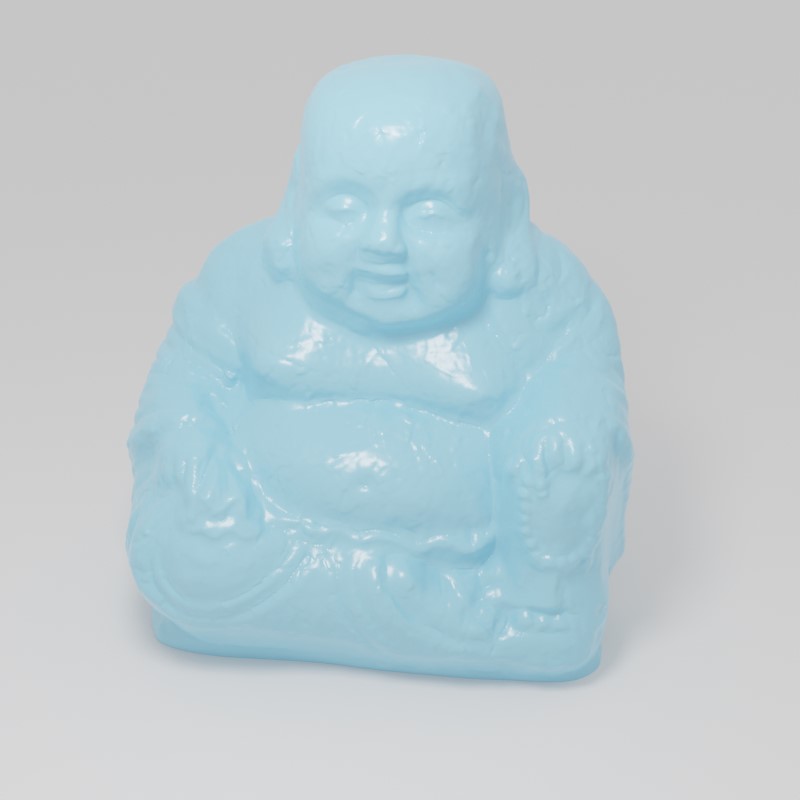} &
    \includegraphics[width=0.16\textwidth]{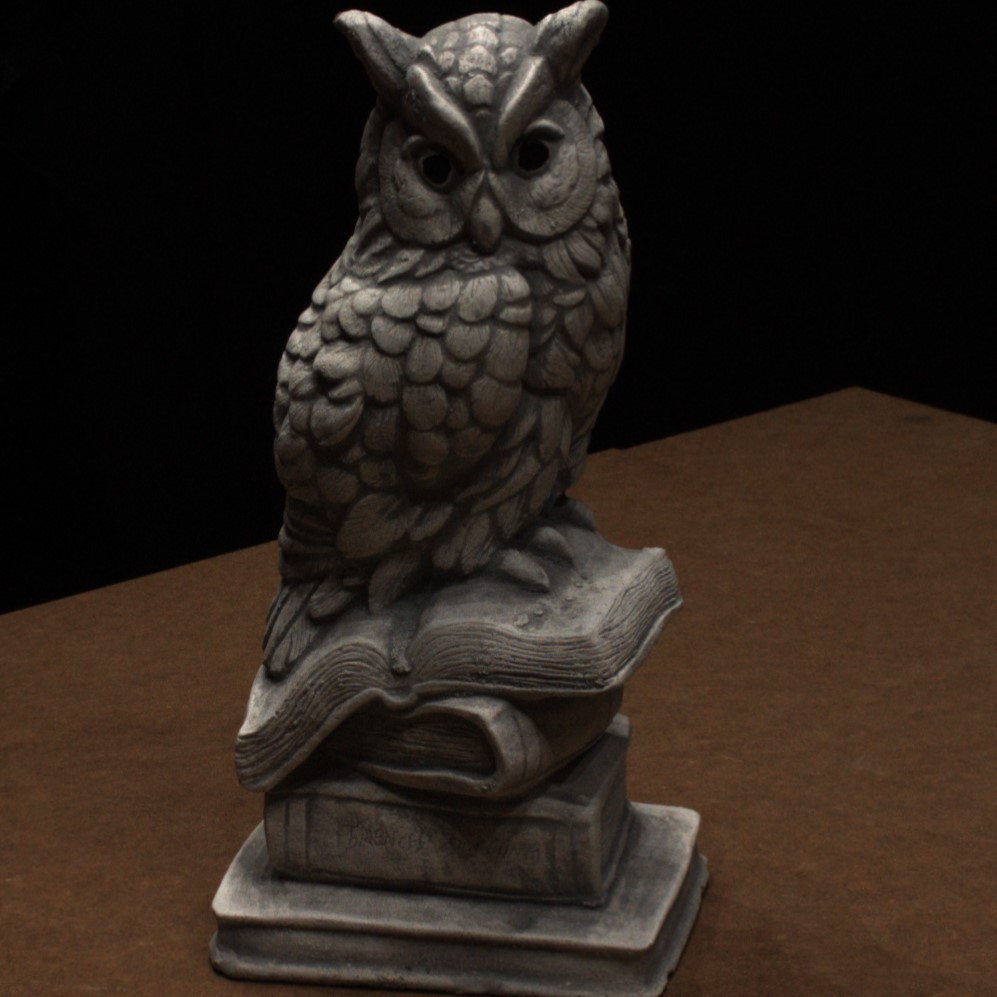} &
    \includegraphics[width=0.16\textwidth]{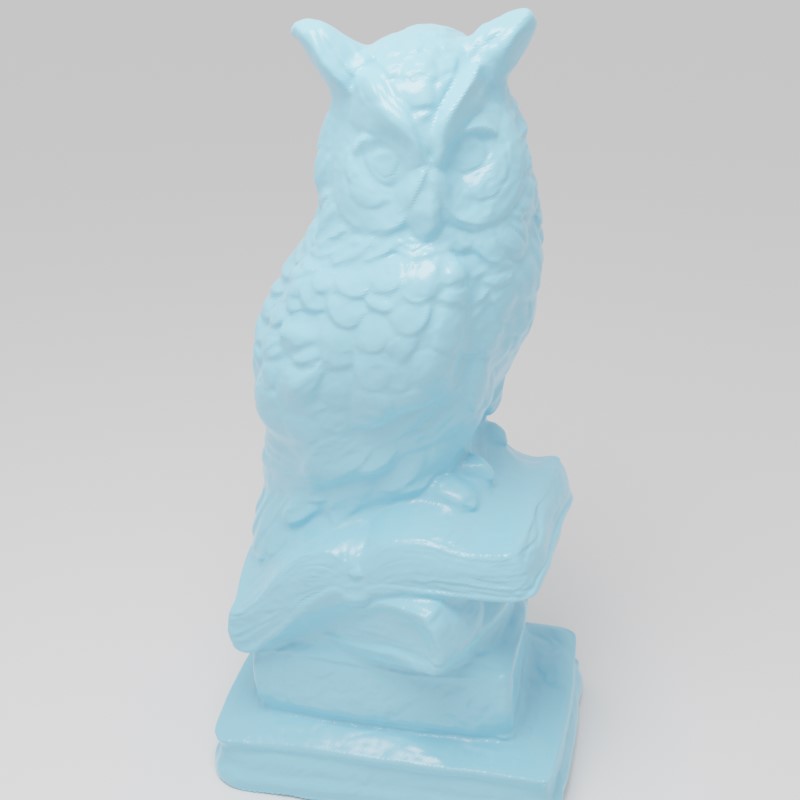} \\
    \end{tabular}
    \caption{Qualitative reconstruction results on the DTU dataset.}
    \label{fig:dtu}
\end{figure*}

NeRO is also able to reconstruct less or non-reflective objects, which we demonstrate from three aspects. First, we show the reconstruction results on three less reflective objects in Fig~\ref{fig:less_ref}, where we capture $\sim$100 images with a resolution of 1024$\times$768 on each object and recover the camera poses by COLMAP. Second, we show that NeRO can simultaneously reconstruct reflective and non-reflective objects in Fig.~\ref{fig:set_recon}, where we capture $\sim$200 images for each set of objects. 
Finally, we further evaluate NeRO on the DTU dataset. Since NeRO assumes a static light environment and the DTU dataset contains images with inconsistent light environments and shadows, we manually remove images with inconsistent light environments for training. The quantitative and qualitative results are shown in Table~\ref{tab:dtu} and Fig.~\ref{fig:dtu}.

\begin{table}[]
    \centering
    \resizebox{\linewidth}{!}{
    \begin{tabular}{ccccccc}
    \toprule
          & scan24 & scan37 & scan110 & scan114 & scan118 & scan122 \\
    \midrule
    NeuS &    1.00 &   1.37 &    1.20 &  0.35   &  0.49   & 0.54   \\
    Ours &    1.10 &   1.13 &    1.14 &  0.39   &  0.52   & 0.57   \\
    \bottomrule
    \end{tabular}
    }
    \caption{CDs$\downarrow$ on the DTU dataset.}
    \label{tab:dtu}
\end{table}

\subsection{Direct combination of Ref-NeRF with NeuS}

\label{sec:app_direct}
\begin{table}[]
    \centering
    \resizebox{\linewidth}{!}{
    \begin{tabular}{lcccc|c}
    \toprule
    Description & Angel & Bell & Cat & Teapot  & Avg.\\
    \midrule
    Only indirect lights        & 0.0037 & 0.0043 & 0.0210 & 0.0064 & 0.0089 \\
    Ref-NeRF+NeuS               & 0.0038 & 0.0062 & 0.0219 & 0.0064 & 0.0096 \\
    Ours                        & 0.0034 & 0.0032 & 0.0044 & 0.0037 & 0.0037 \\
    \bottomrule
    \end{tabular}
    }
    \caption{Comparison with the direct combination of Ref-NeRF and NeuS in terms of CD.}
    \label{tab:comb}
\end{table}

\begin{figure}
    \centering
    \setlength\tabcolsep{1pt}
    \begin{tabular}{ccc}
    \includegraphics[width=0.28\linewidth]{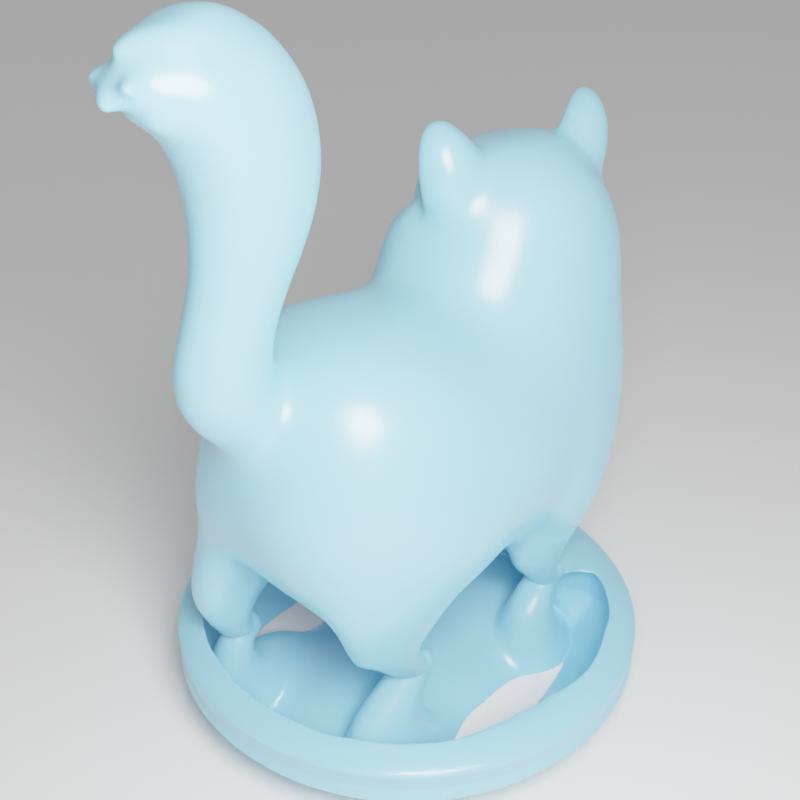} & 
    \includegraphics[width=0.28\linewidth]{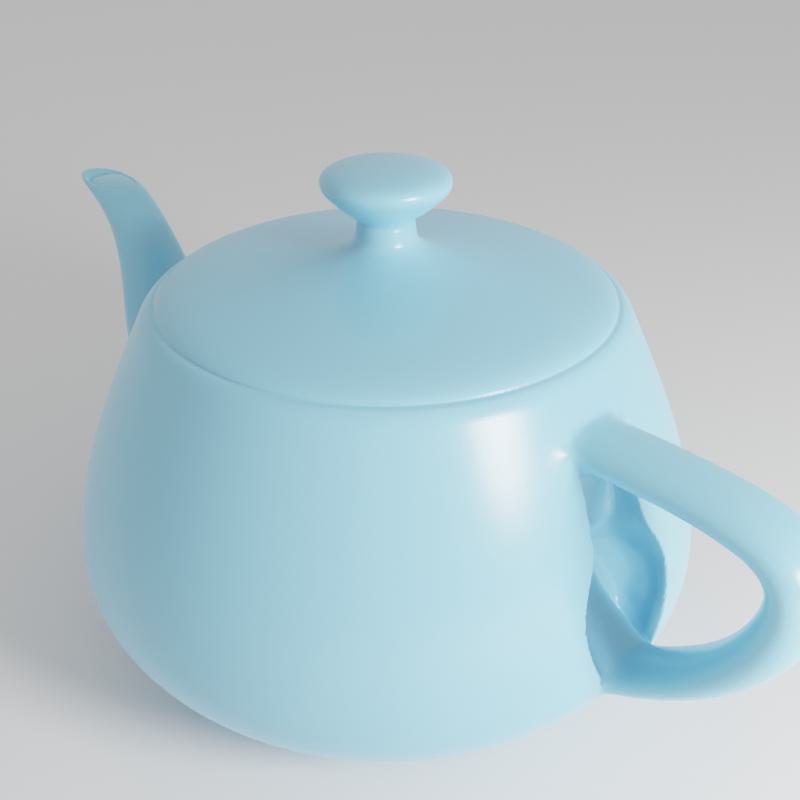} &
    \includegraphics[width=0.28\linewidth]{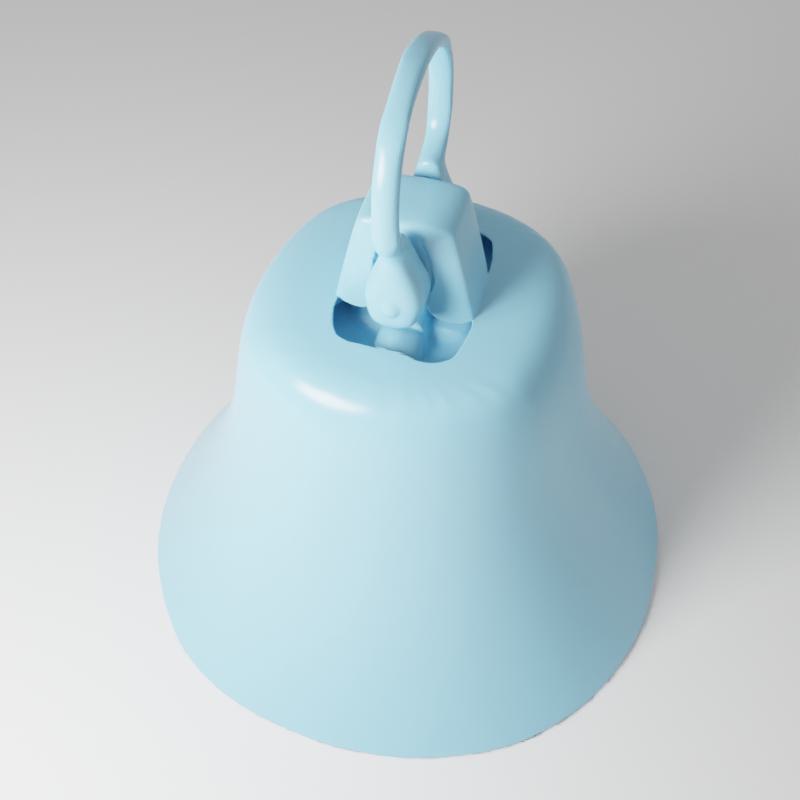} \\
    \end{tabular}
    \caption{Reconstructed surfaces of directly combining Ref-NeRF~\cite{verbin2022ref} with NeuS~\cite{wang2021neus}.}
    \label{fig:combine}
\end{figure}

To improve the reflective color fitting of NeuS~\cite{wang2021neus}, an alternative solution is to combine the color function of Ref-NeRF~\cite{verbin2022ref} with the neural SDF of NeuS. This combination leads to a model similar to Model 2 of Table~\ref{tab:ab} in the geometry ablation study, both of which predict specular lights from a PE and an IDE. The differences are that Ref-NeRF does not consider the integral of material $M_{\rm material}$ in Eq.~\ref{eq:app_spe} and that Ref-NeRF directly predicts the diffuse color from an MLP while Model 2 relies the IDE on the normal direction (Eq.~\ref{eq:app_diff} and Eq.~\ref{eq:app_diff_light}) to compute diffuse color. We report the CDs of this model in Table~\ref{tab:comb}, which are comparable to Model 2 and worse than NeRO. The qualitative results of this combination model are shown in Fig.~\ref{fig:combine}.





\subsection{Copyrights}
\label{sec:app_copy}
The Glossy-Blender dataset is created from the models listed in Table~\ref{tab:copyright}. On some objects, we modified their appearances and geometry to make the dataset. All the HDR images used in relighting or rendering are downloaded from \href{https://polyhaven.com}{https://polyhaven.com} under the CC0 license.

\begin{table}[]
    \centering
    \begin{tabular}{cccc}
    \toprule
    Model & Creator & License & Link \\
    \midrule
    Bell  & jQueary & CC BY 4.0  & \href{https://sketchfab.com/3d-models/bell-897bc8230df54a1cad474492771880d8}{here} \\
    Cat  & Suushimi & CC BY-NC 4.0  & \href{https://sketchfab.com/3d-models/cat-70a23788ef984a7a9a1c9a9fe6d5a651}{here} \\
    Teapot  & Martin Newell & N/A  & \href{https://en.wikipedia.org/wiki/Utah_teapot}{here} \\
    Luyu  & romullus & CC BY-SA 4.0  & \href{https://sketchfab.com/3d-models/lu-yu-figurine-derivative-caa5a93fa0fe4d39ad8fc391f3a4d574}{here} \\
    TBell  & gla\_bot & CC BY 4.0  & \href{https://sketchfab.com/3d-models/table-bell-77f2ea17b4c84fe1a8d2aec02caa9de3}{here} \\
    Horse  & halimi13744 & CC BY 4.0  & \href{https://sketchfab.com/3d-models/horse-2287485aa2e54f87854b0472444c5930}{here} \\
    Potion  & Blender3D & CC BY 4.0  & \href{https://sketchfab.com/3d-models/basic-bottle-b2d9a692c15e4ad980c384fe2d6a8f8c}{here} \\
    Angel  & SebastianSosnowski & CC BY 4.0  & \href{https://sketchfab.com/3d-models/angel-brass-version-1ed059cb4976440f9a595621949428f8}{here} \\
    
    \bottomrule
    \end{tabular}
    \caption{Copyrights of all models used in the Glossy-Blender dataset.}
    \label{tab:copyright}
    \vspace{-10pt}
\end{table}

\begin{figure*}
    \centering
    \setlength\tabcolsep{1pt}
    \renewcommand{\arraystretch}{0.5} 
    \begin{tabular}{cccccc}
        \includegraphics[width=0.16\textwidth]{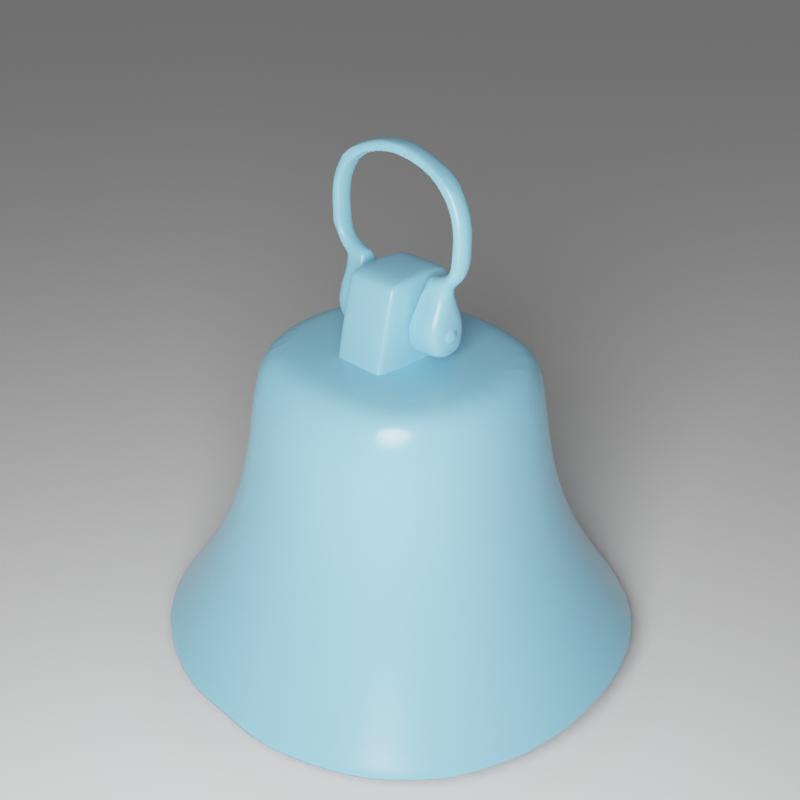} &
        \includegraphics[width=0.16\textwidth]{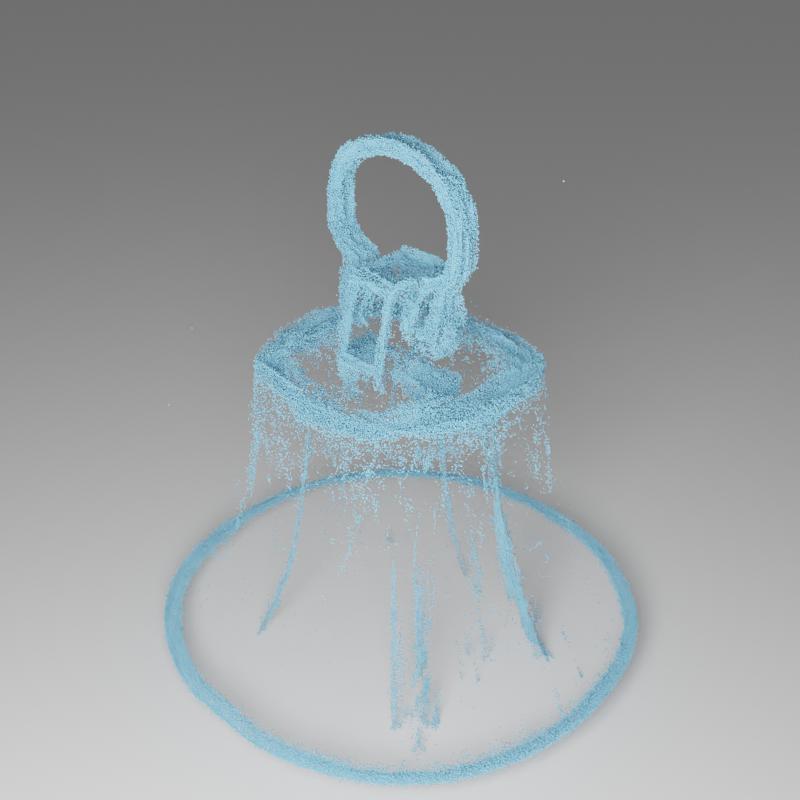} &
        \includegraphics[width=0.16\textwidth]{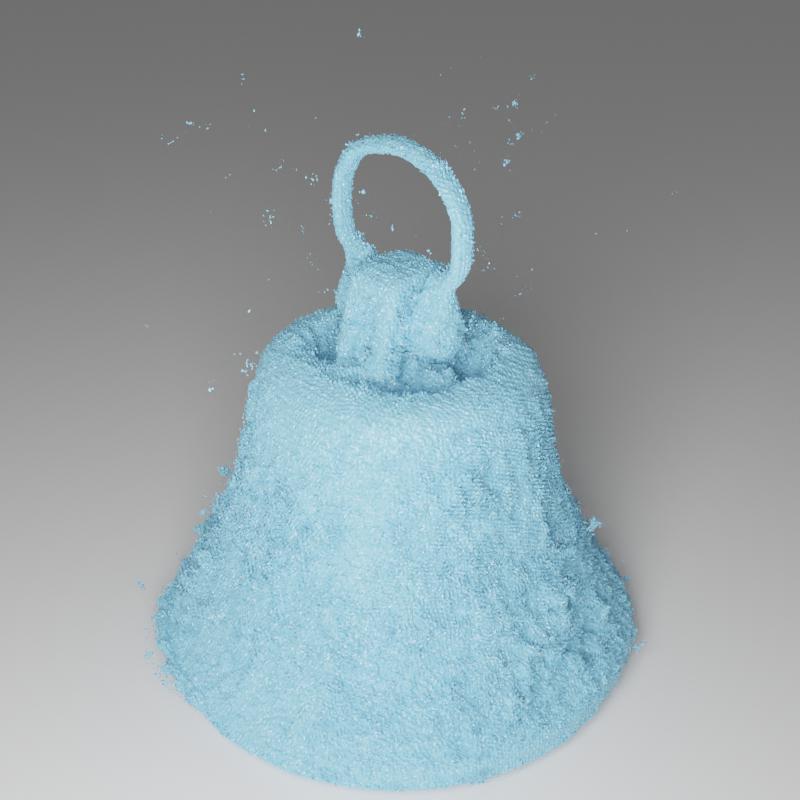} &
        \includegraphics[width=0.16\textwidth]{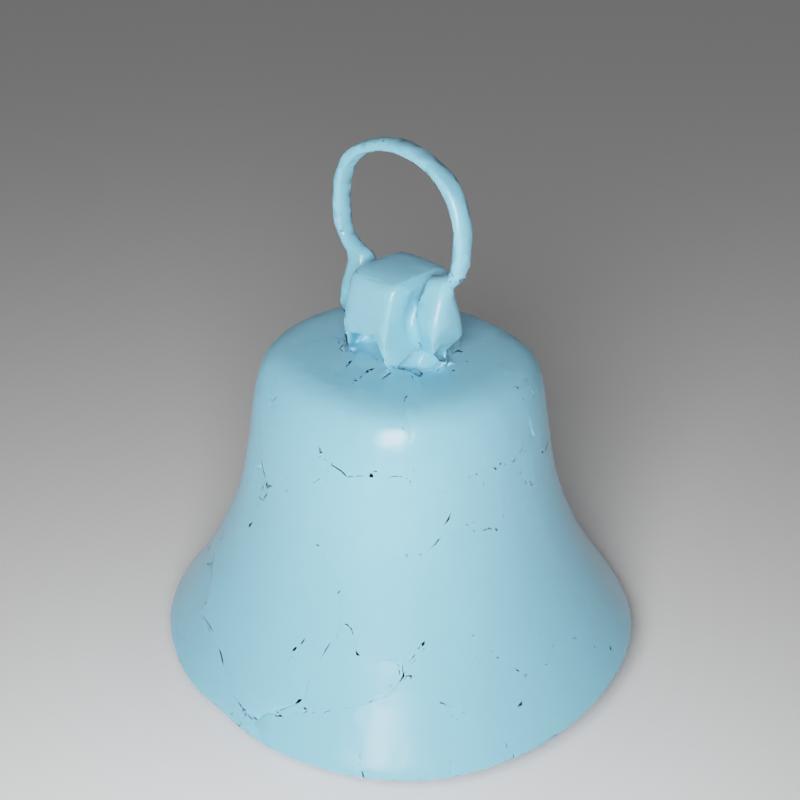} &
        \includegraphics[width=0.16\textwidth]{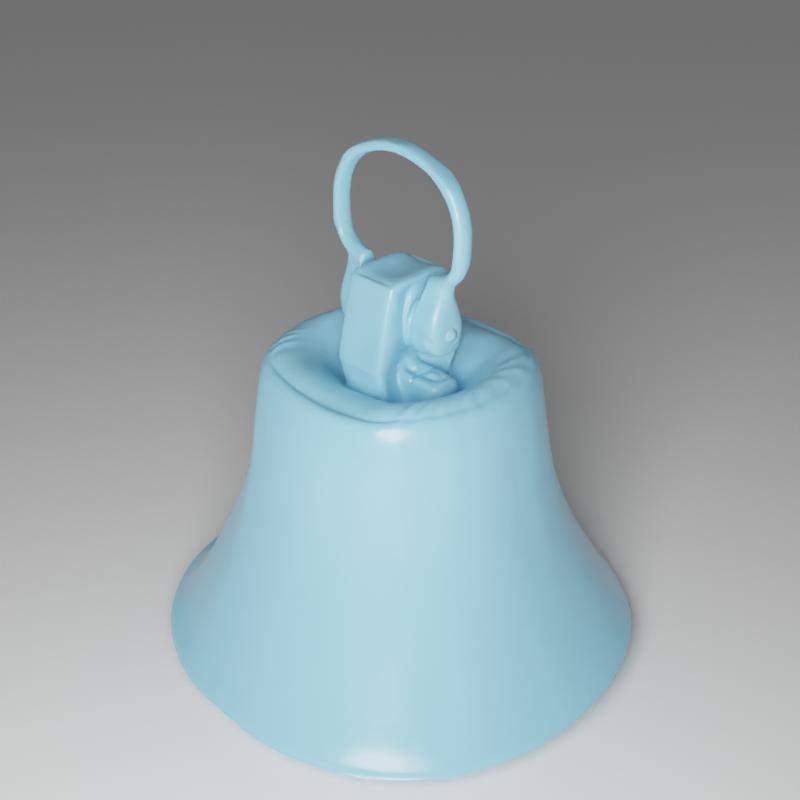} &
        \includegraphics[width=0.16\textwidth]{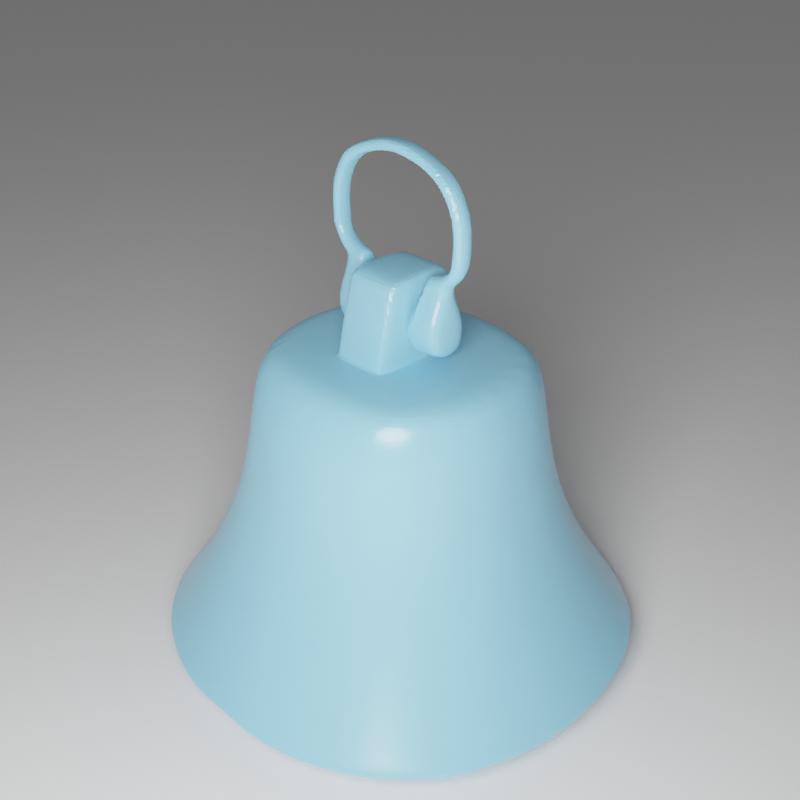} \\
        \includegraphics[width=0.16\textwidth]{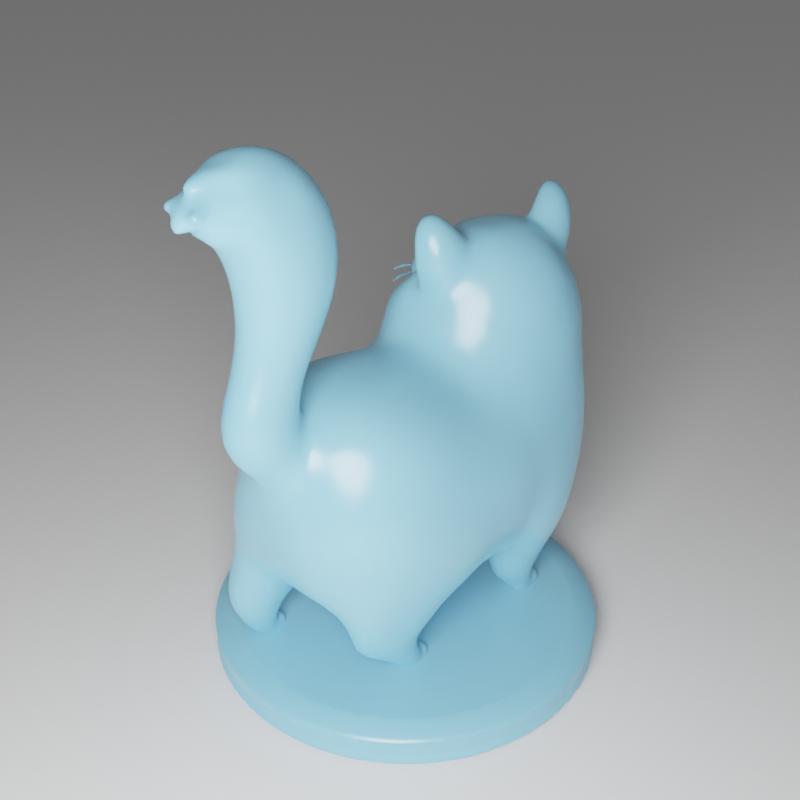} &
        \includegraphics[width=0.16\textwidth]{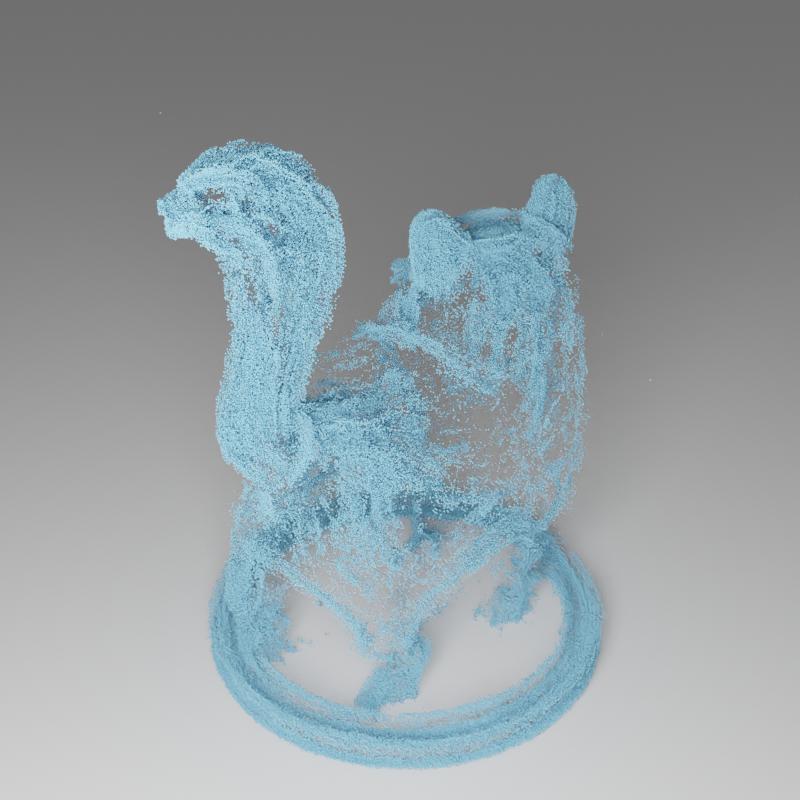} &
        \includegraphics[width=0.16\textwidth]{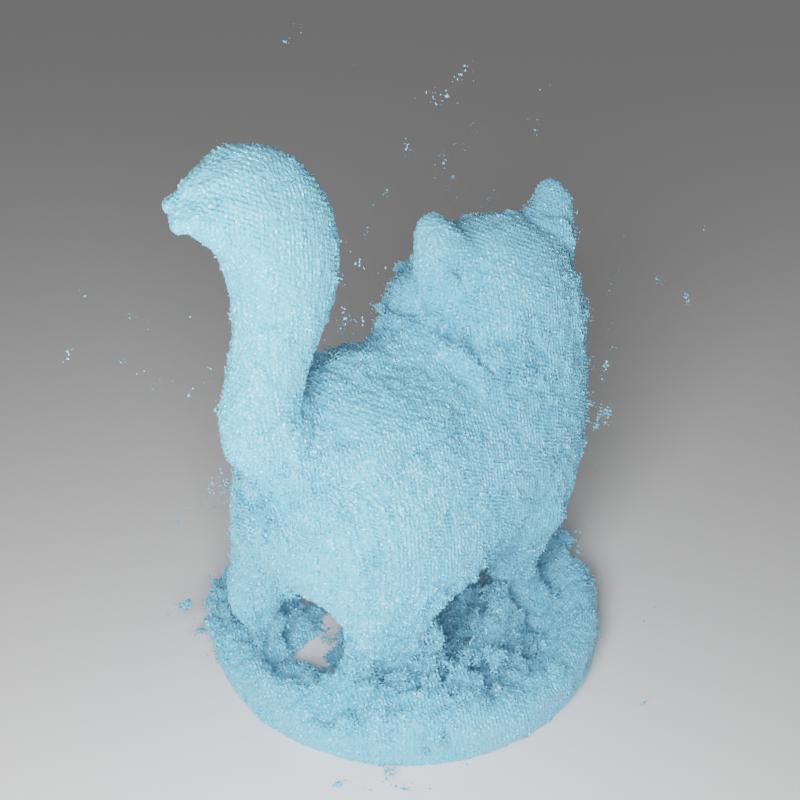} &
        \includegraphics[width=0.16\textwidth]{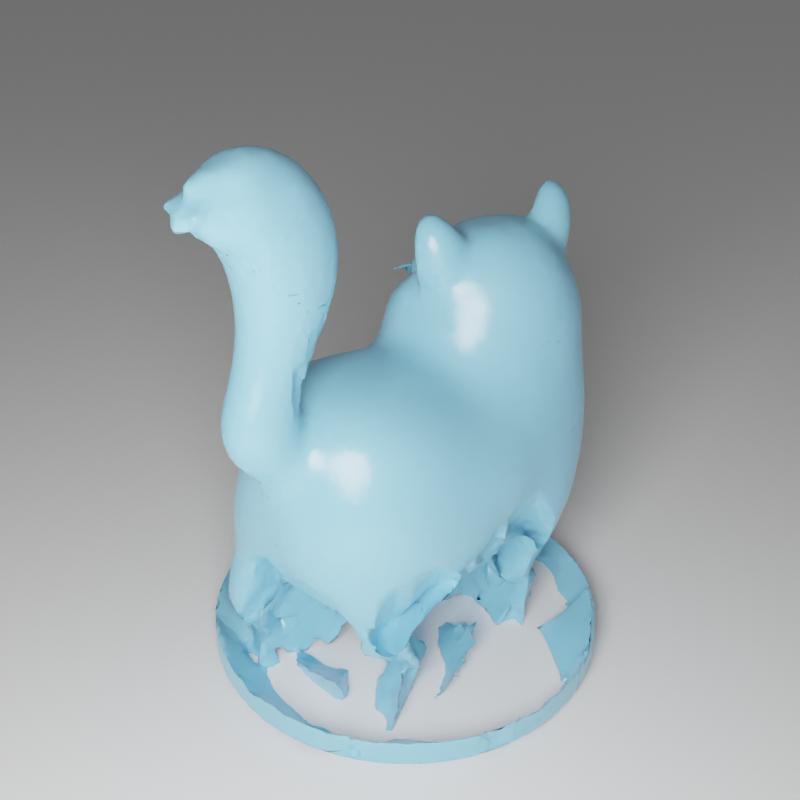} &
        \includegraphics[width=0.16\textwidth]{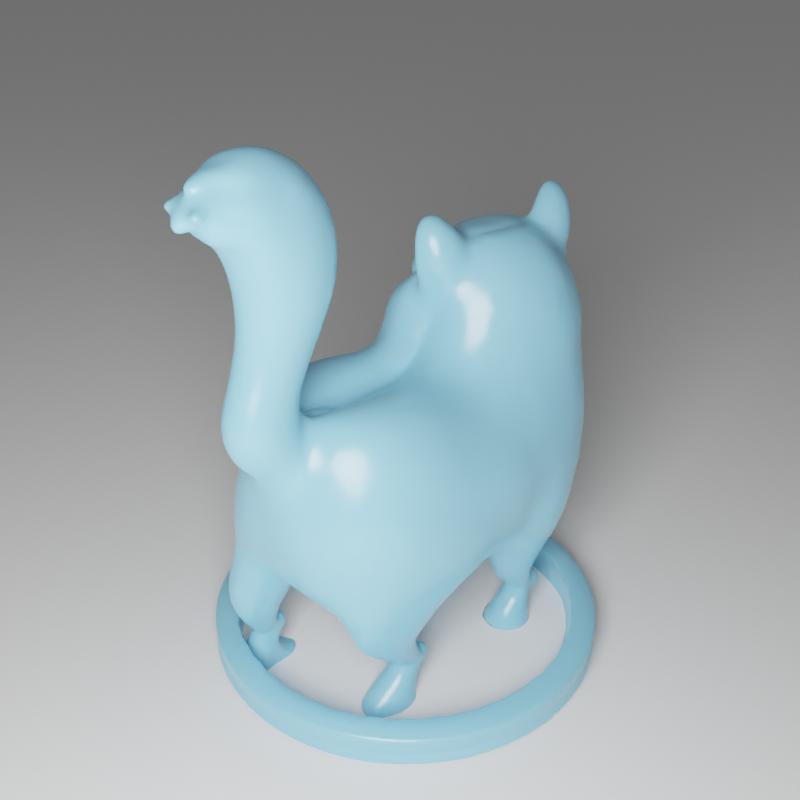} &
        \includegraphics[width=0.16\textwidth]{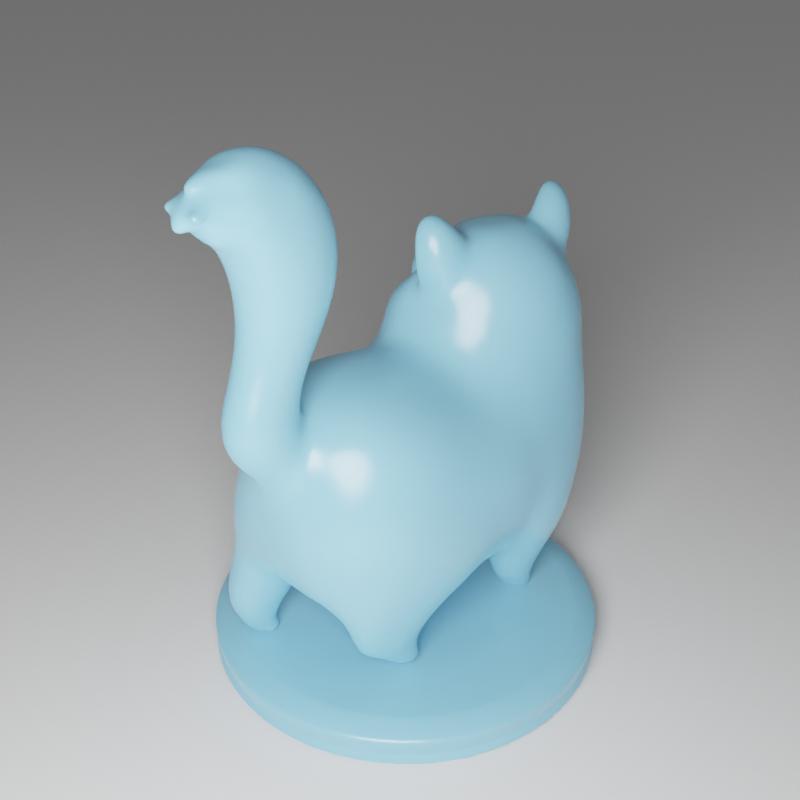} \\
        \includegraphics[width=0.16\textwidth]{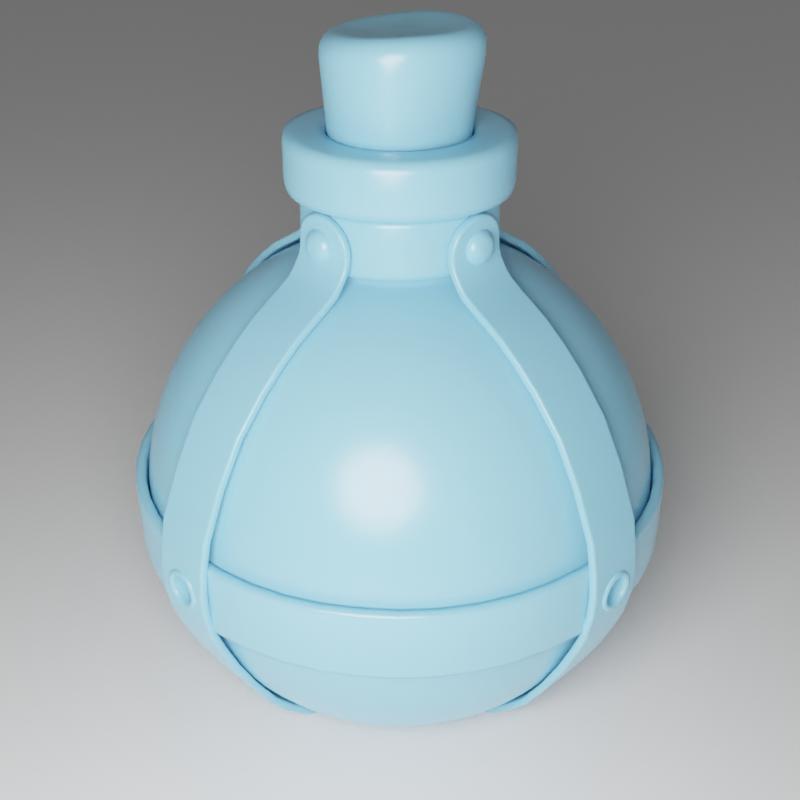} &
        \includegraphics[width=0.16\textwidth]{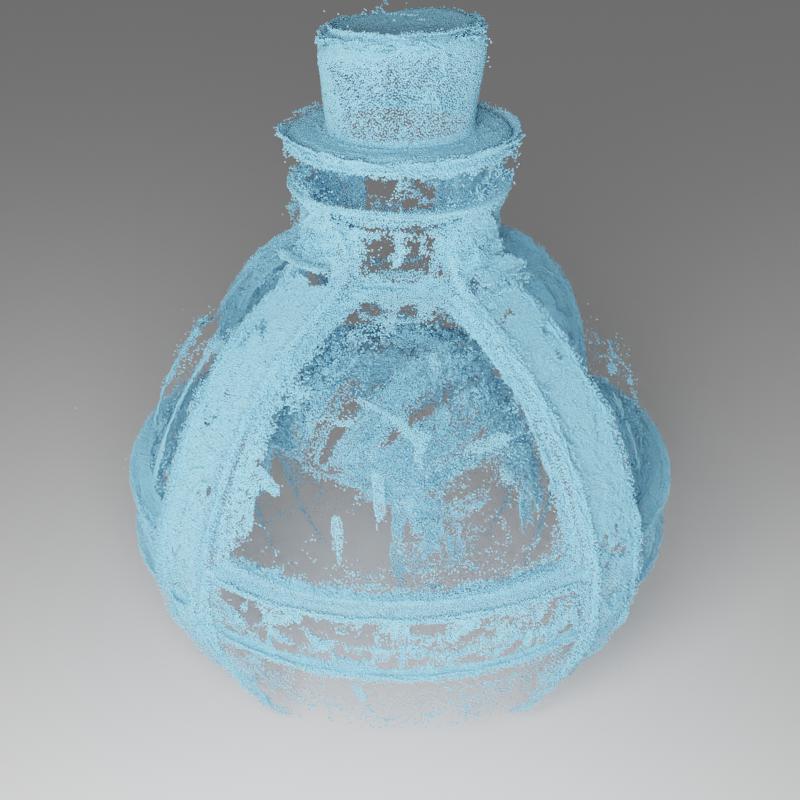} &
        \includegraphics[width=0.16\textwidth]{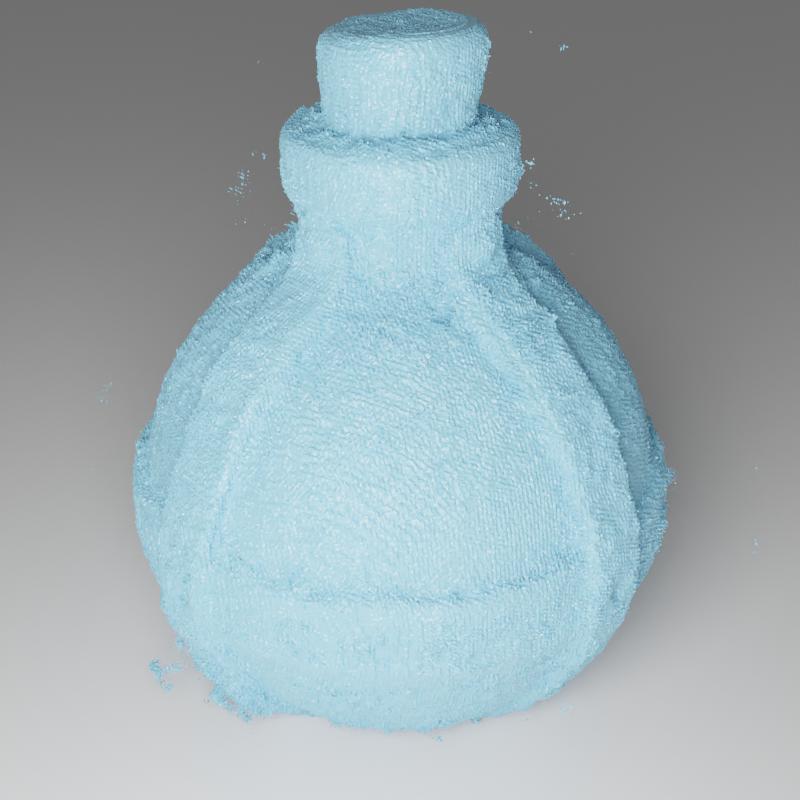} &
        \includegraphics[width=0.16\textwidth]{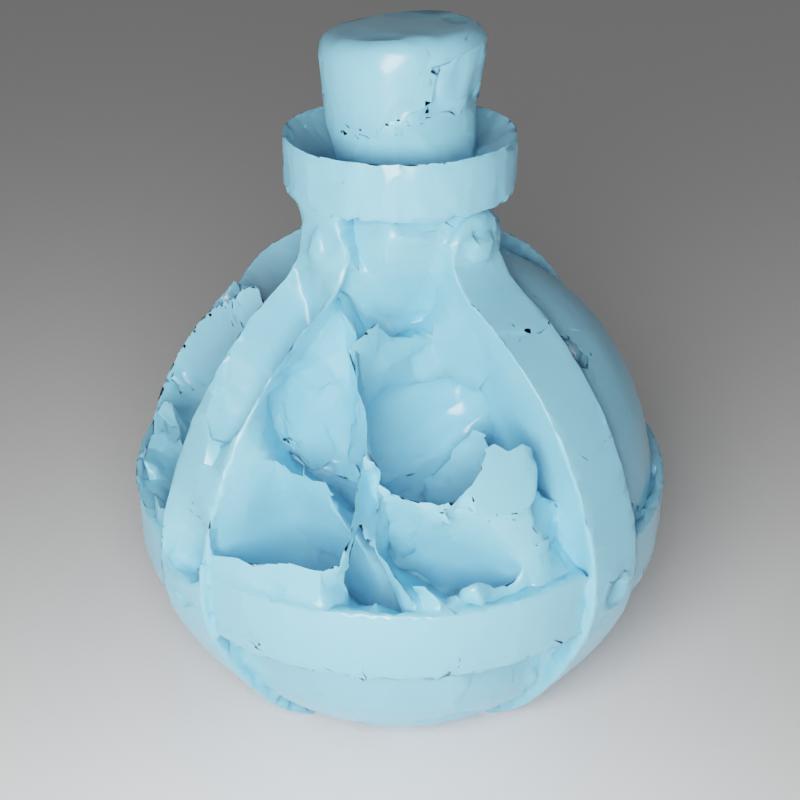} &
        \includegraphics[width=0.16\textwidth]{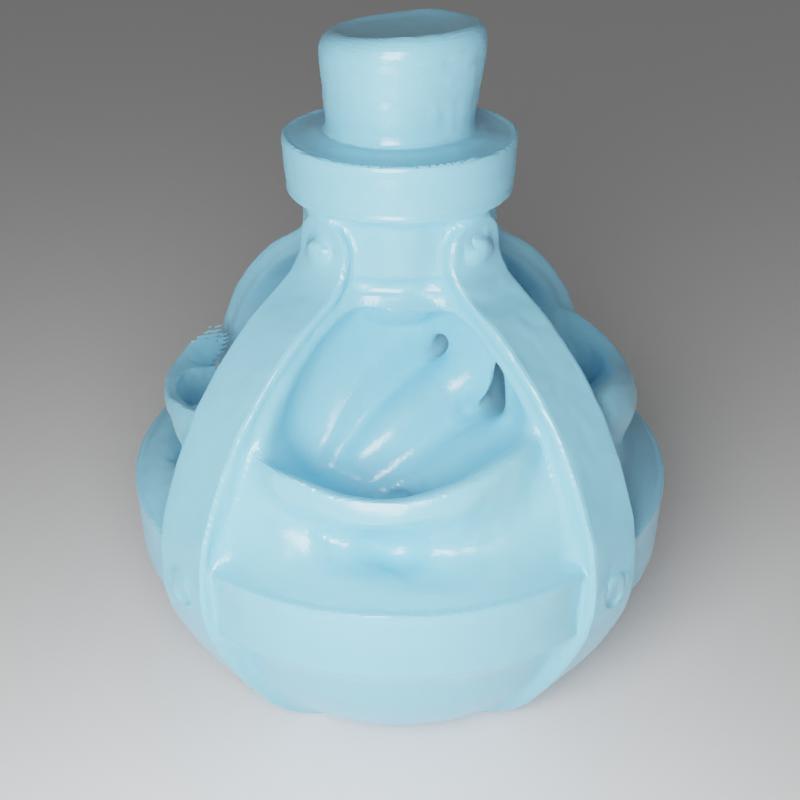} &
        \includegraphics[width=0.16\textwidth]{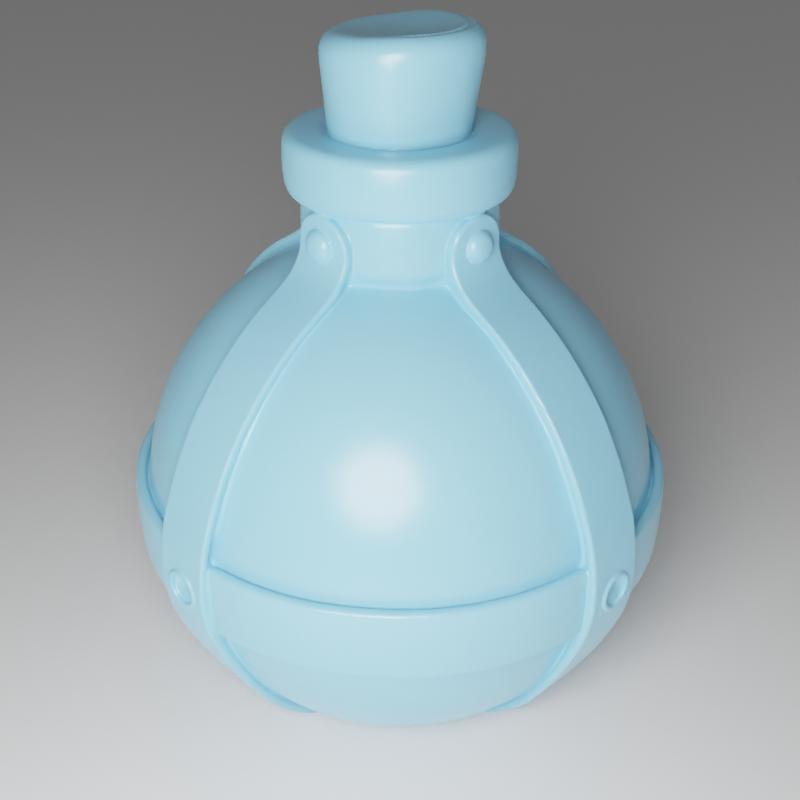} \\
        \includegraphics[width=0.16\textwidth]{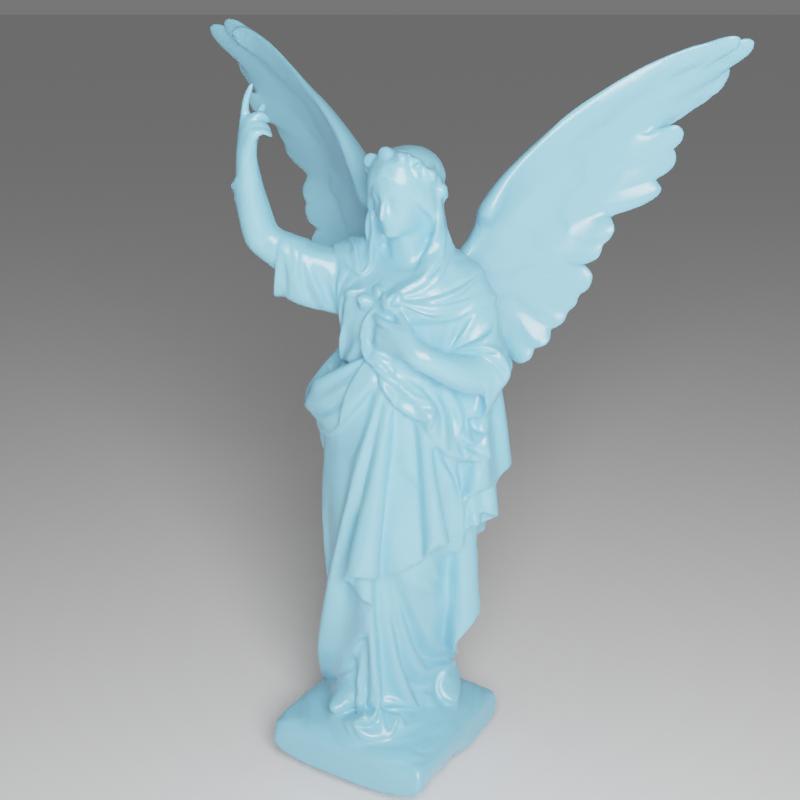} &
        \includegraphics[width=0.16\textwidth]{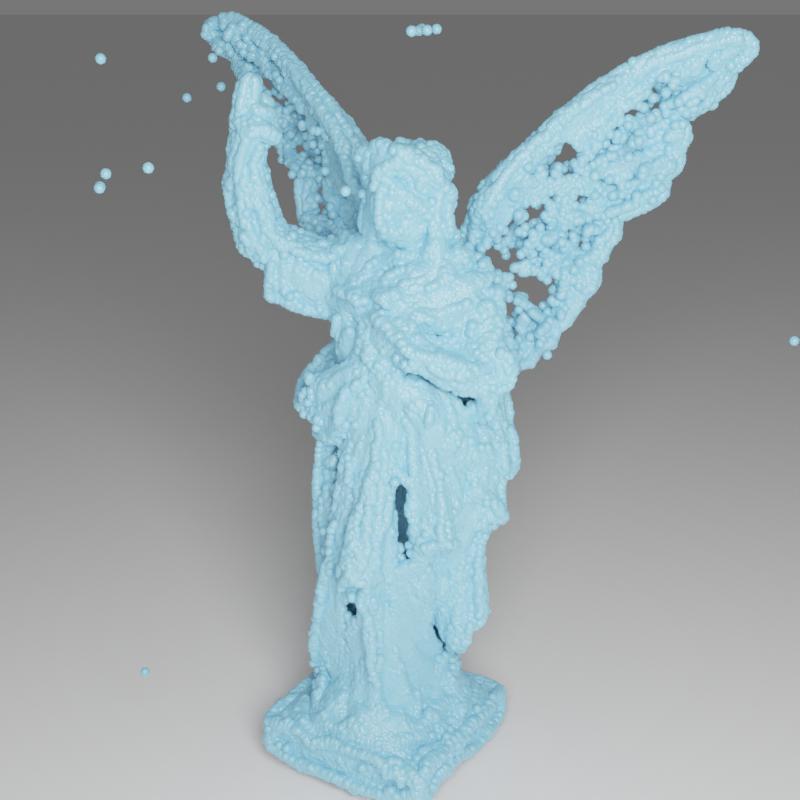} &
        \includegraphics[width=0.16\textwidth]{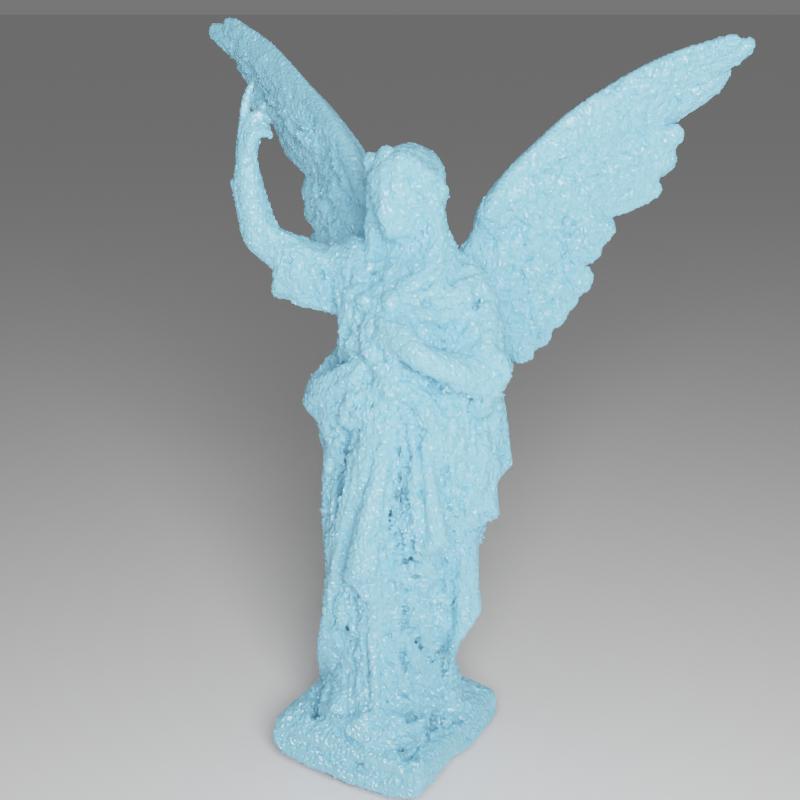} &
        \includegraphics[width=0.16\textwidth]{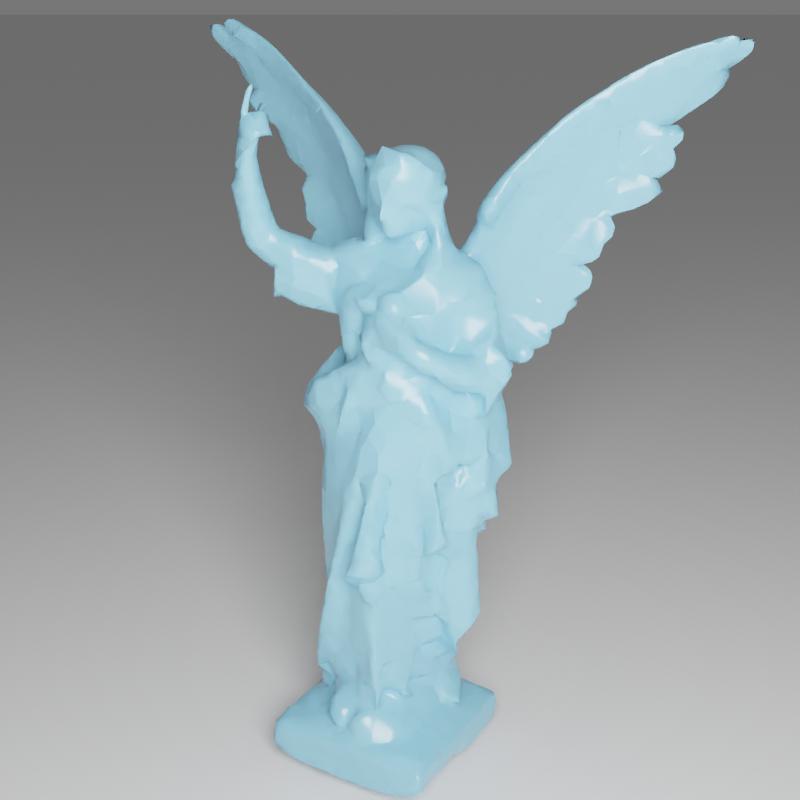} &
        \includegraphics[width=0.16\textwidth]{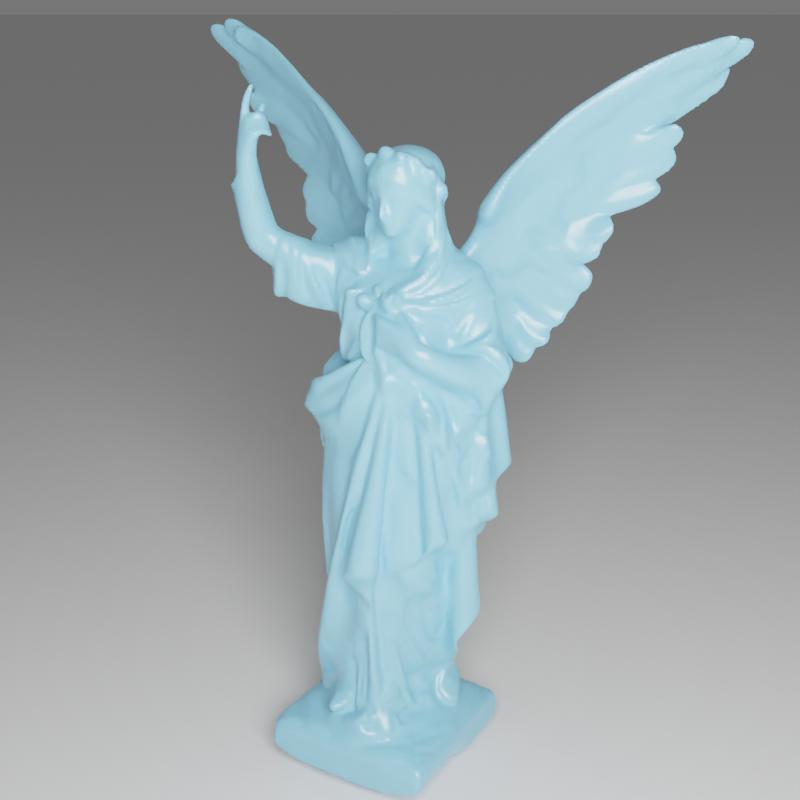} &
        \includegraphics[width=0.16\textwidth]{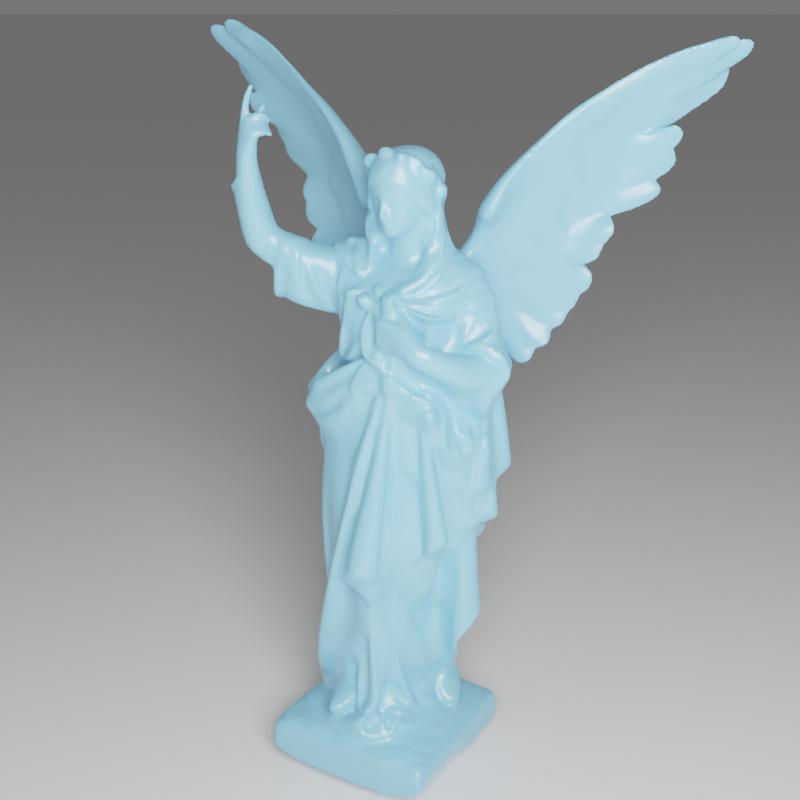} \\
        \includegraphics[width=0.16\textwidth]{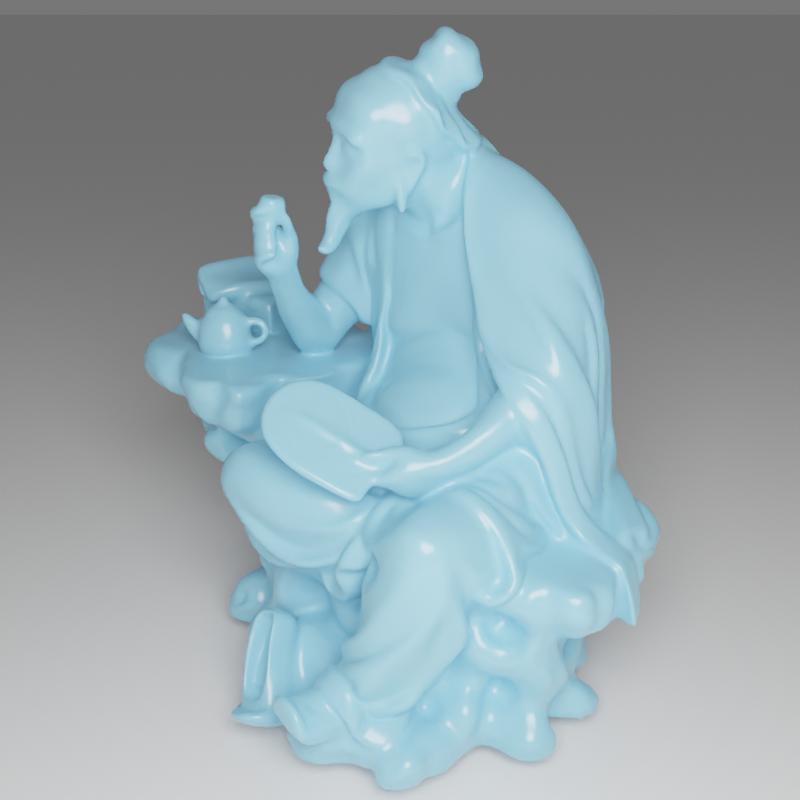} &
        \includegraphics[width=0.16\textwidth]{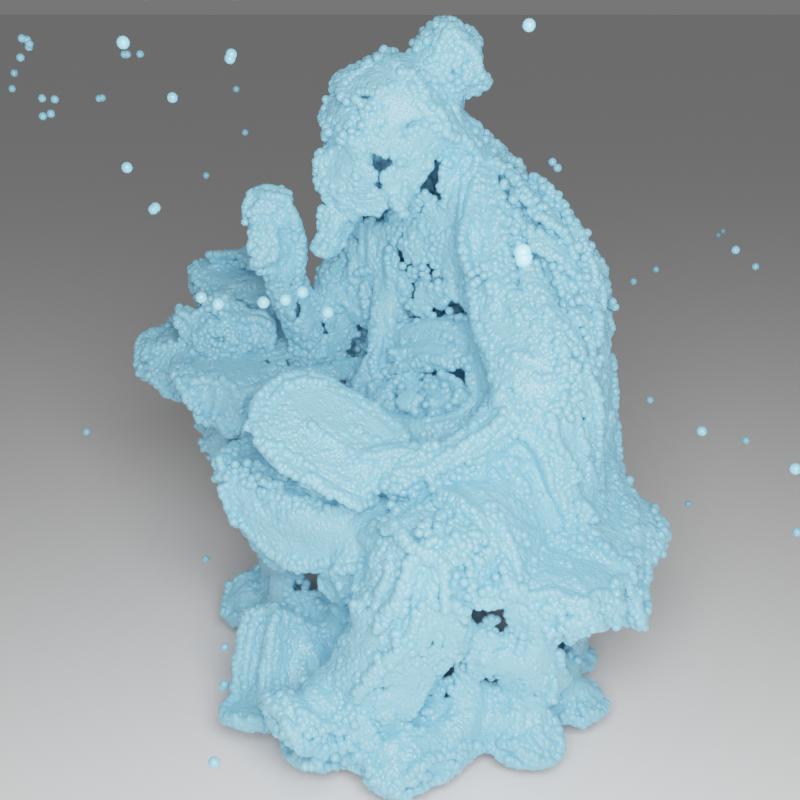} &
        \includegraphics[width=0.16\textwidth]{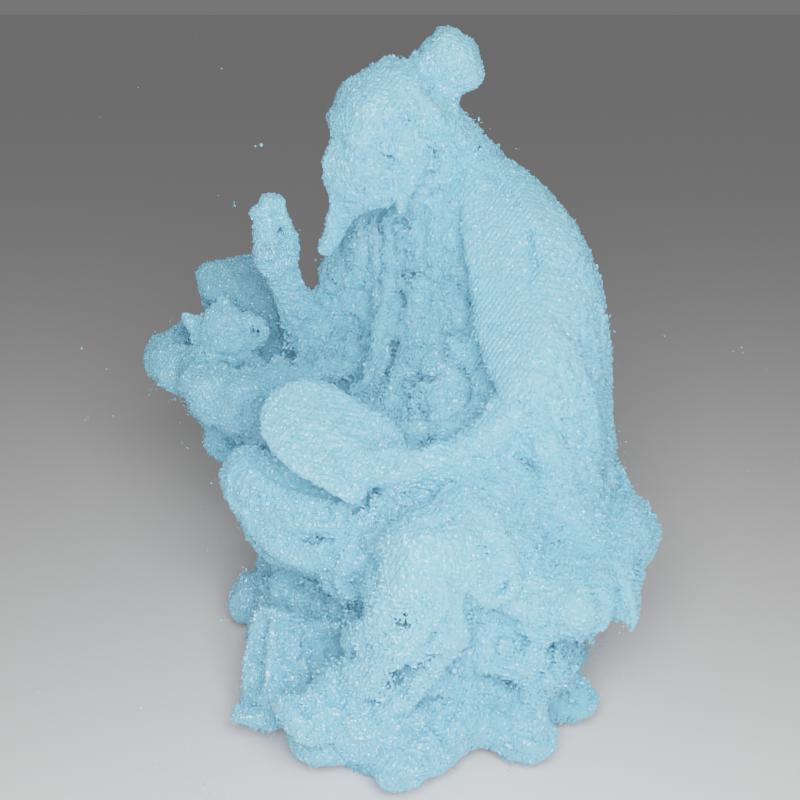} &
        \includegraphics[width=0.16\textwidth]{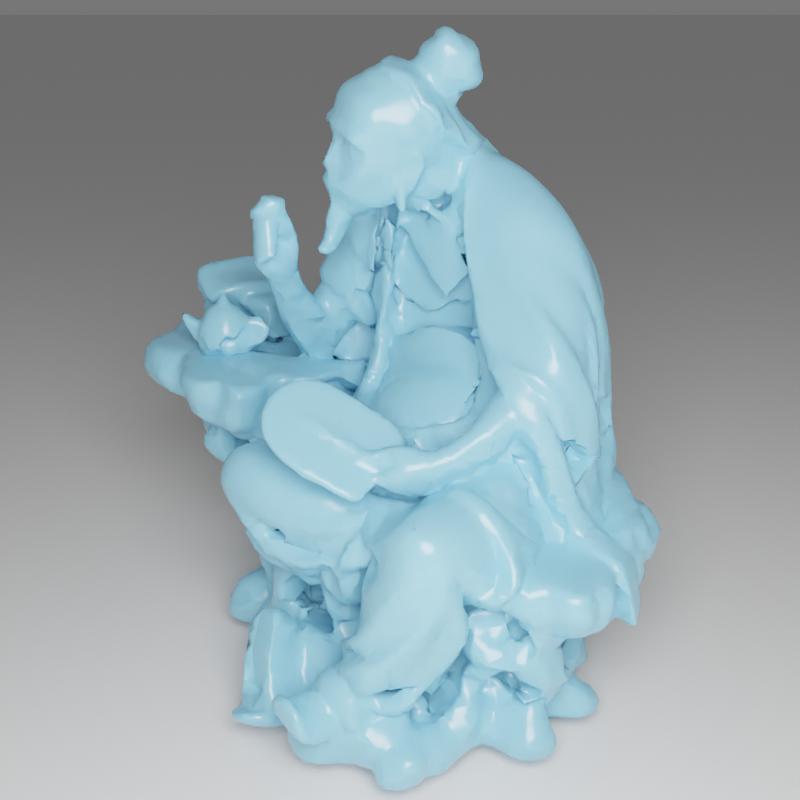} &
        \includegraphics[width=0.16\textwidth]{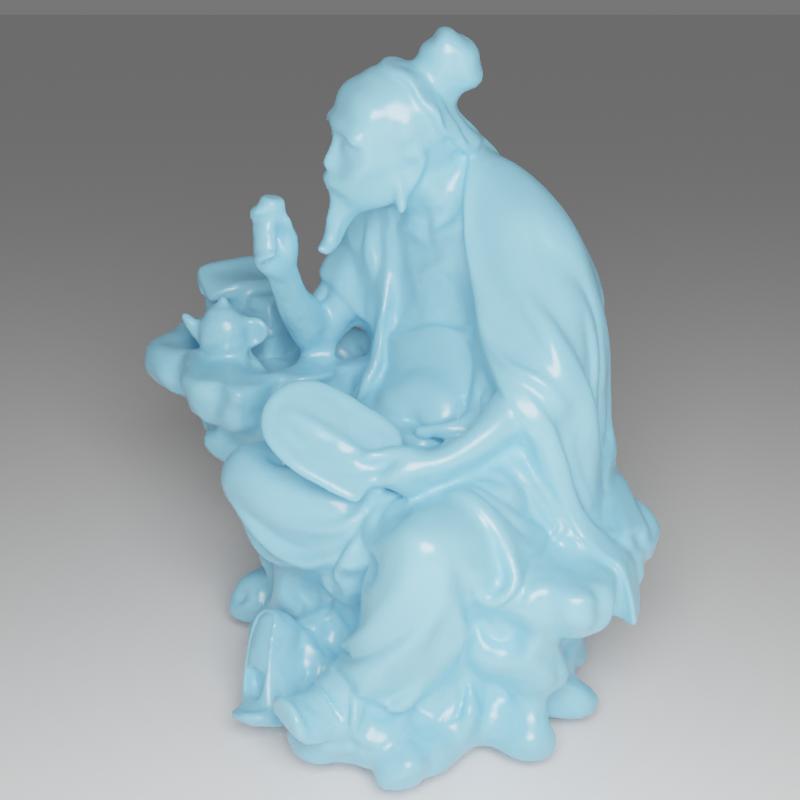} &
        \includegraphics[width=0.16\textwidth]{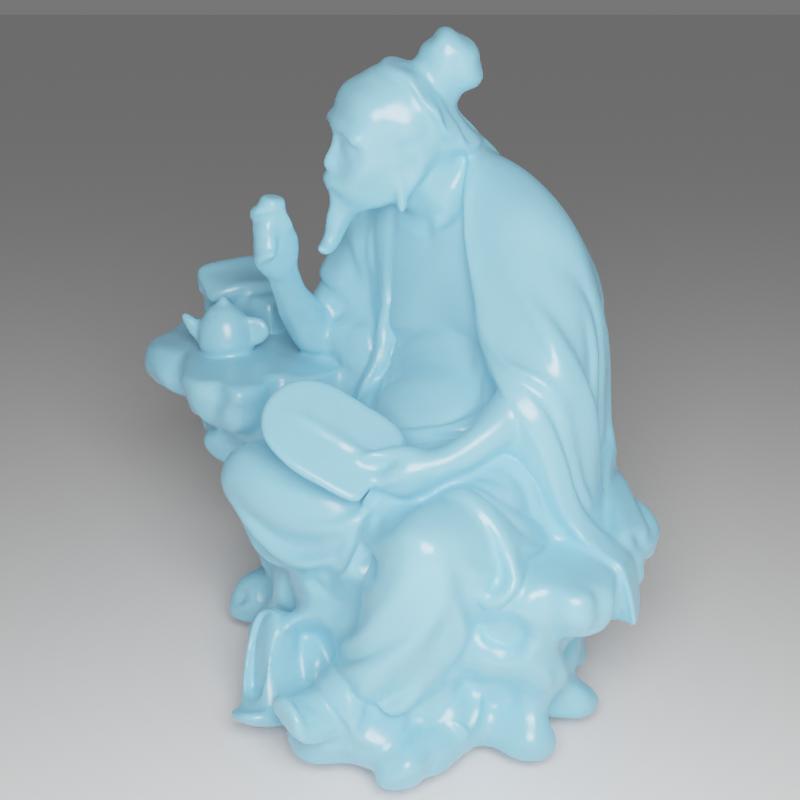} \\
        (a) Ground-truth & (b) COLMAP & (c) Ref-NeRF & (d) NDRMC$^*$ & (e) NeuS & (f) Ours
    \end{tabular}
    \caption{\textbf{Ground-truth and reconstructed surfaces of the Glossy-Blender dataset}. We compare our results with COLMAP~\cite{schoenberger2016mvs}, Ref-NeRF~\cite{verbin2022ref}, NDRMC~\cite{hasselgren2022shape}, and NeuS~\cite{wang2021neus}. $^*$NDRMC~\cite{hasselgren2022shape} is trained with ground-truth object masks while the other methods do not use object masks. The supplementary video contains more qualitative results.}
    \label{fig:syn_visual_app}
\end{figure*}

\begin{figure*}
    \centering
    \setlength\tabcolsep{1pt}
    \renewcommand{\arraystretch}{0.5} 
    \begin{tabular}{cccccc}
        \includegraphics[width=0.16\textwidth]{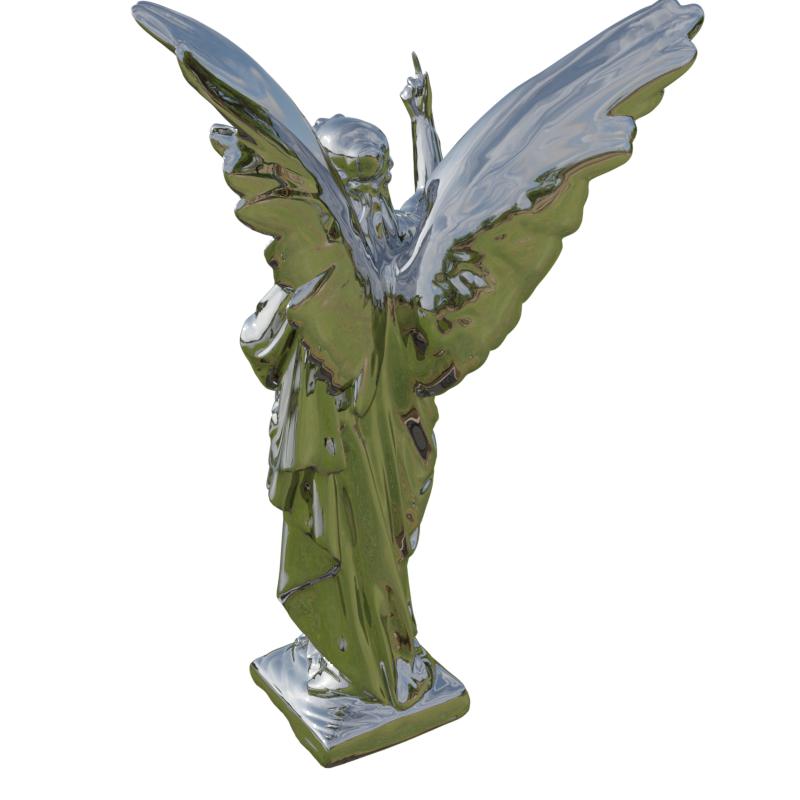} &
        \includegraphics[width=0.16\textwidth]{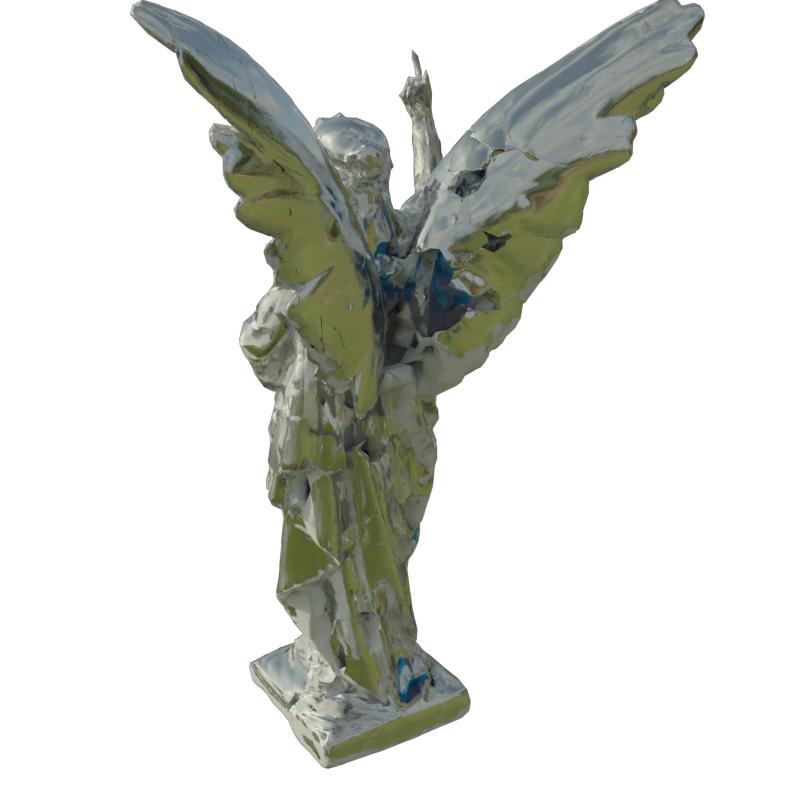} &
        \includegraphics[width=0.16\textwidth]{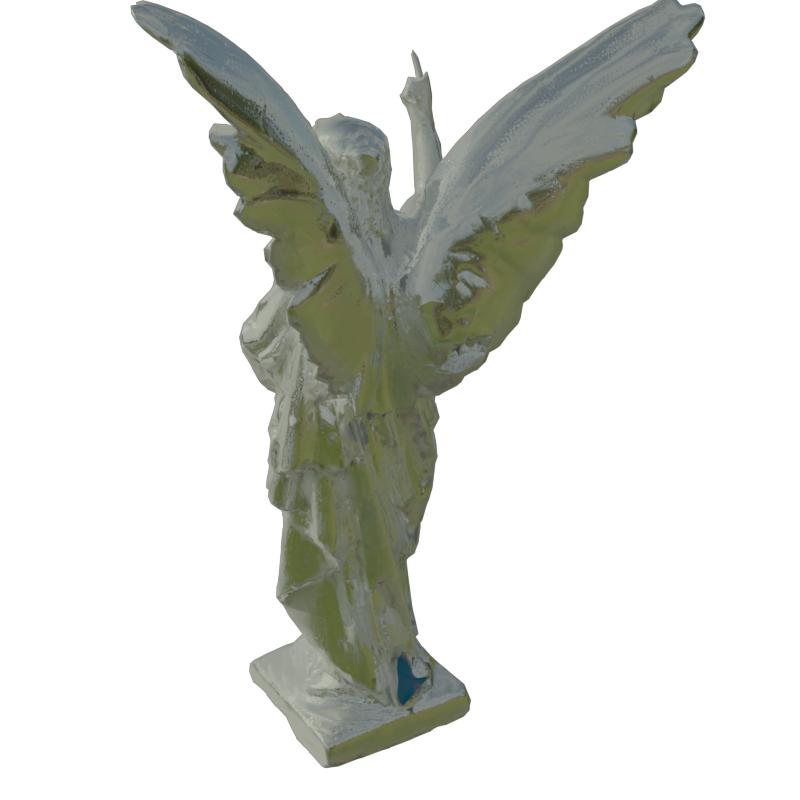} &
        \includegraphics[width=0.16\textwidth]{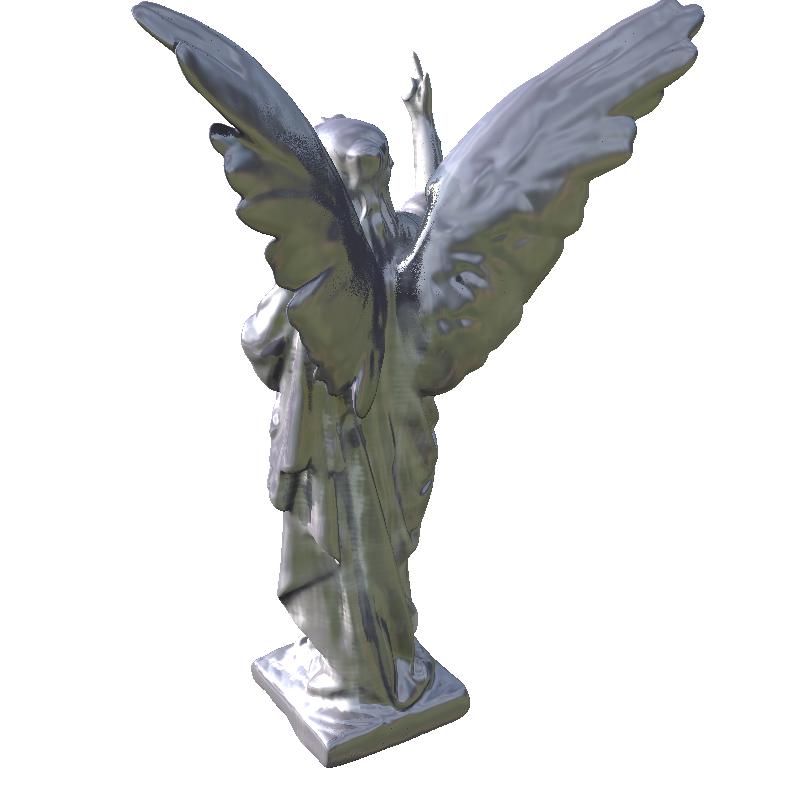} &
        \includegraphics[width=0.16\textwidth]{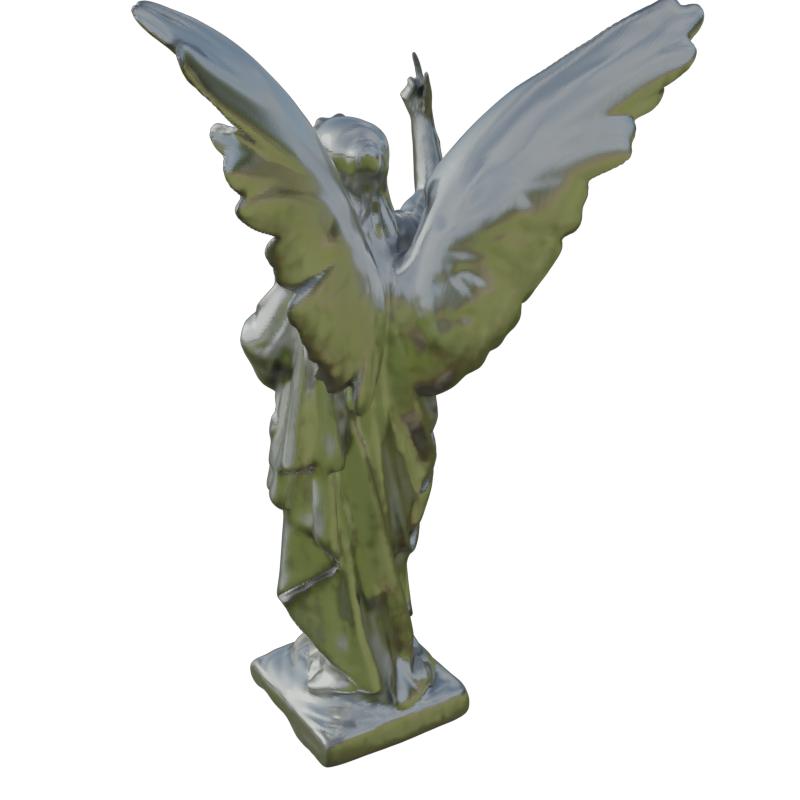} &
        \includegraphics[width=0.16\textwidth]{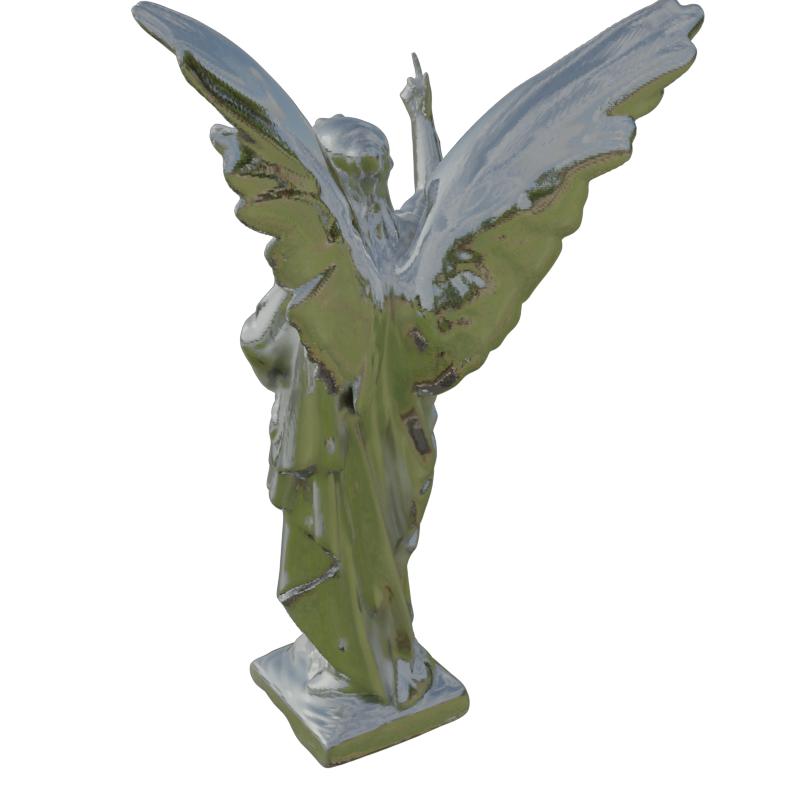} \\
        \includegraphics[width=0.16\textwidth]{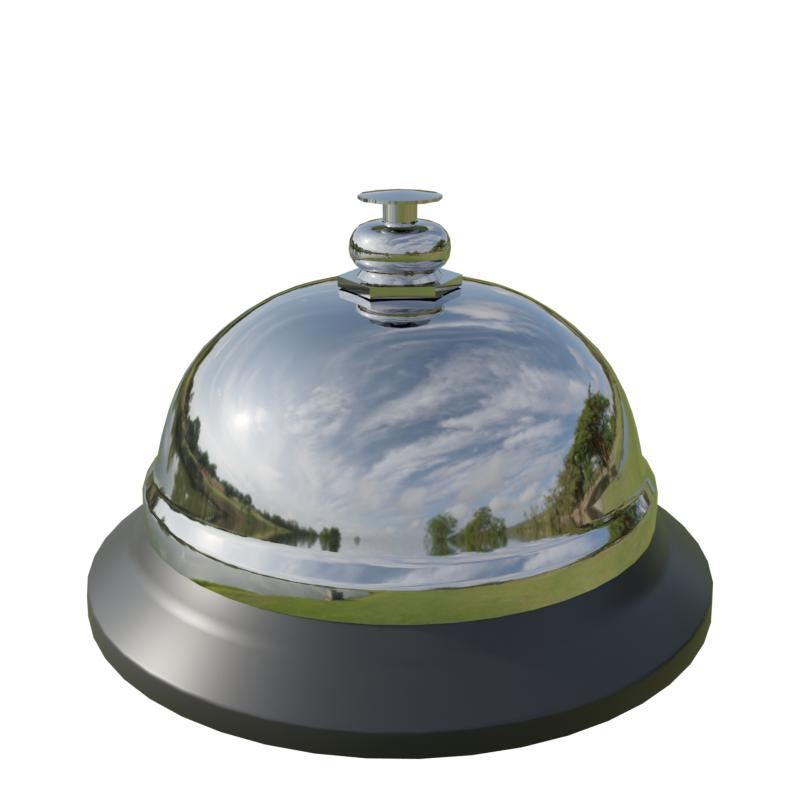} &
        \includegraphics[width=0.16\textwidth]{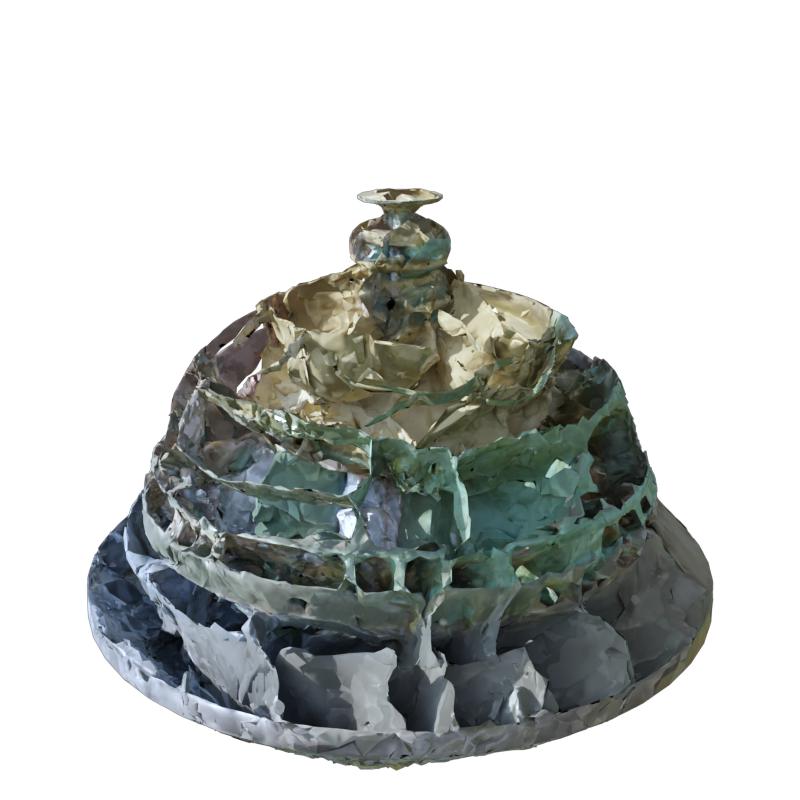} &
        \includegraphics[width=0.16\textwidth]{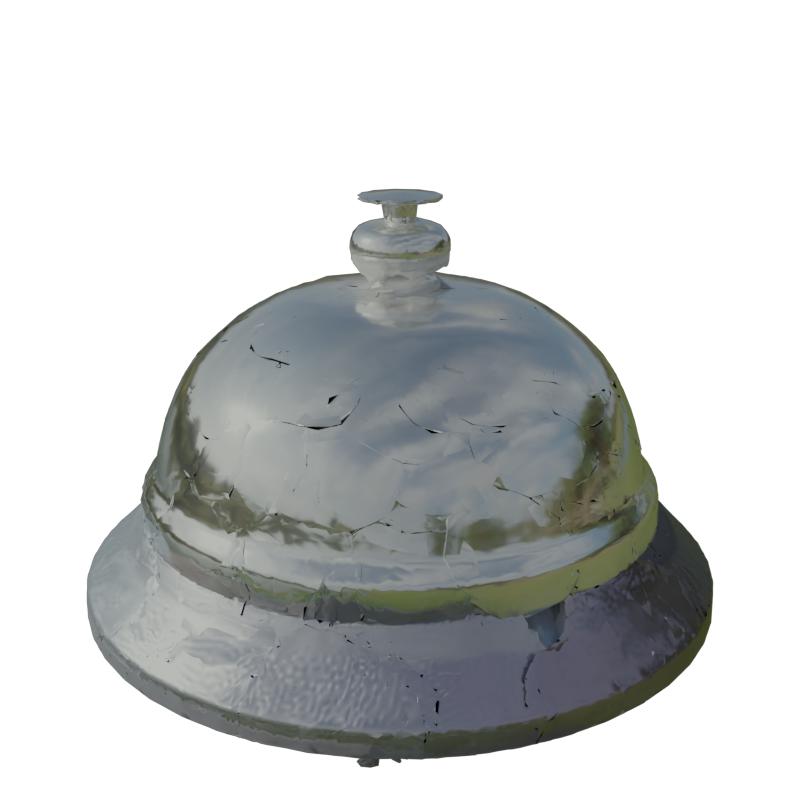} &
        \includegraphics[width=0.16\textwidth]{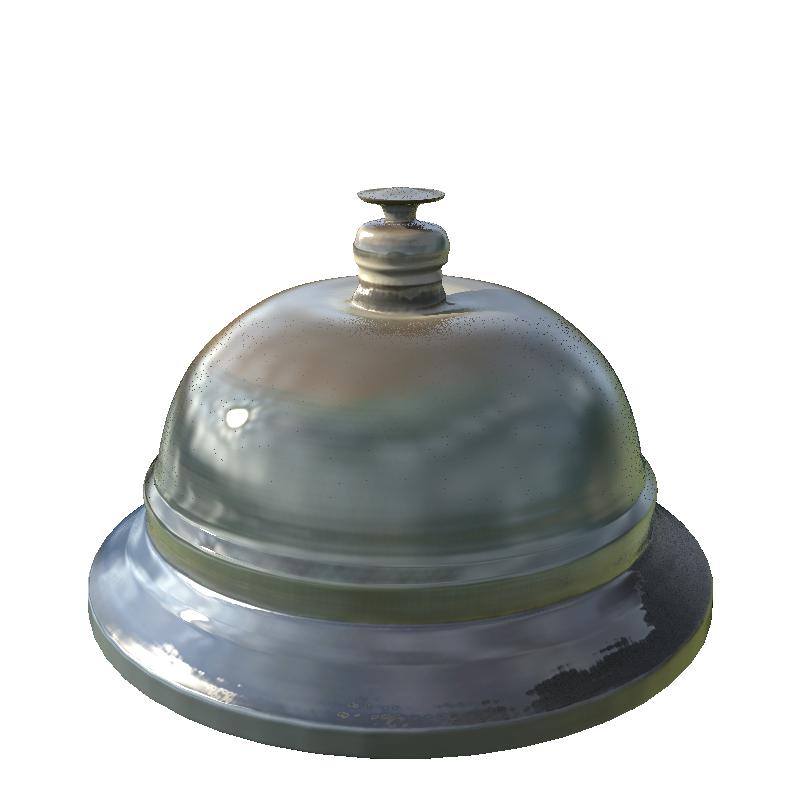} &
        \includegraphics[width=0.16\textwidth]{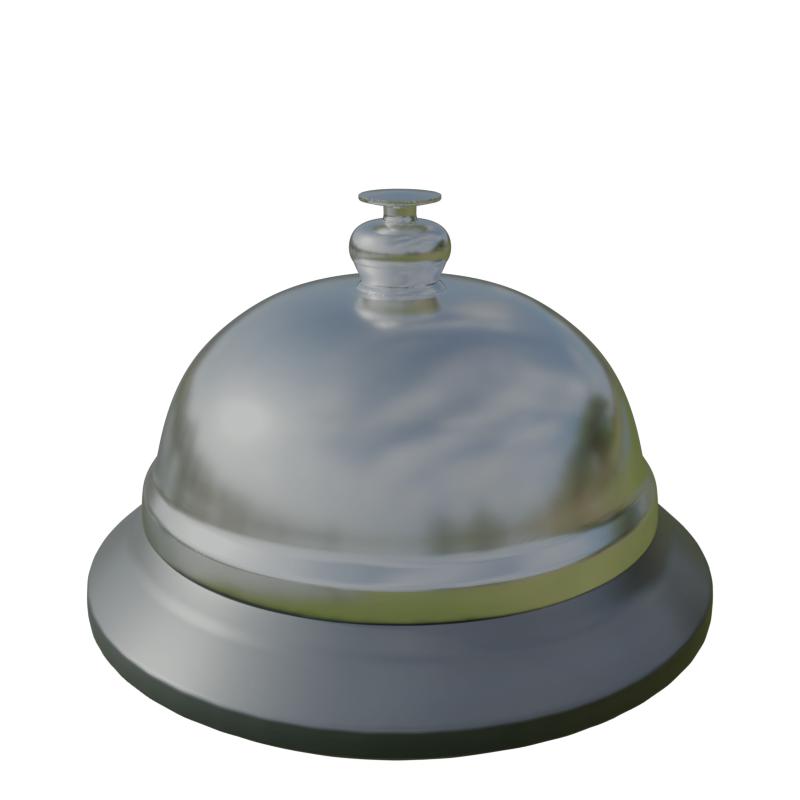} &
        \includegraphics[width=0.16\textwidth]{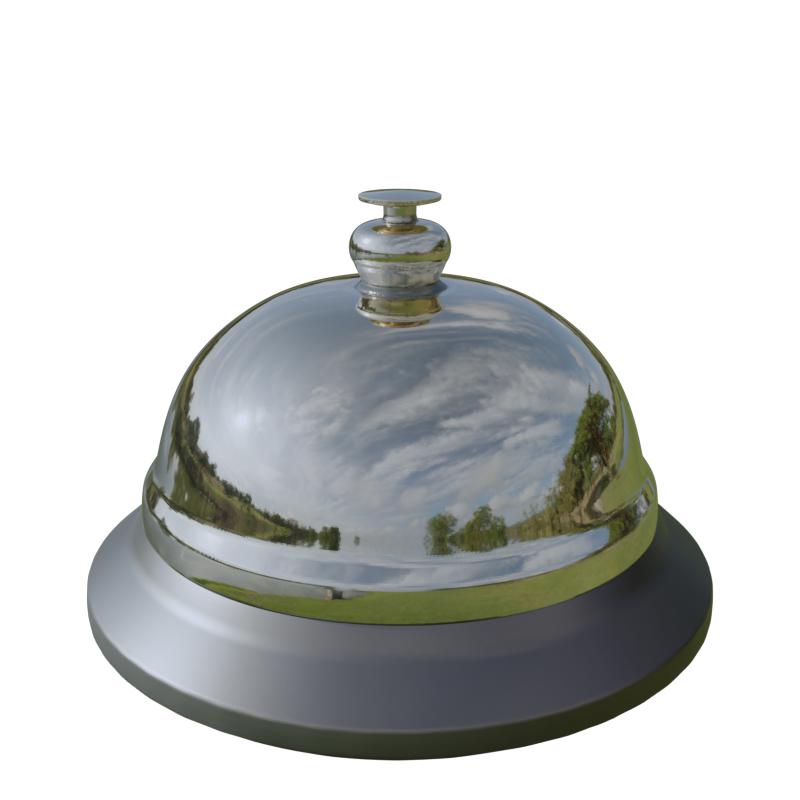} \\
        \includegraphics[width=0.16\textwidth]{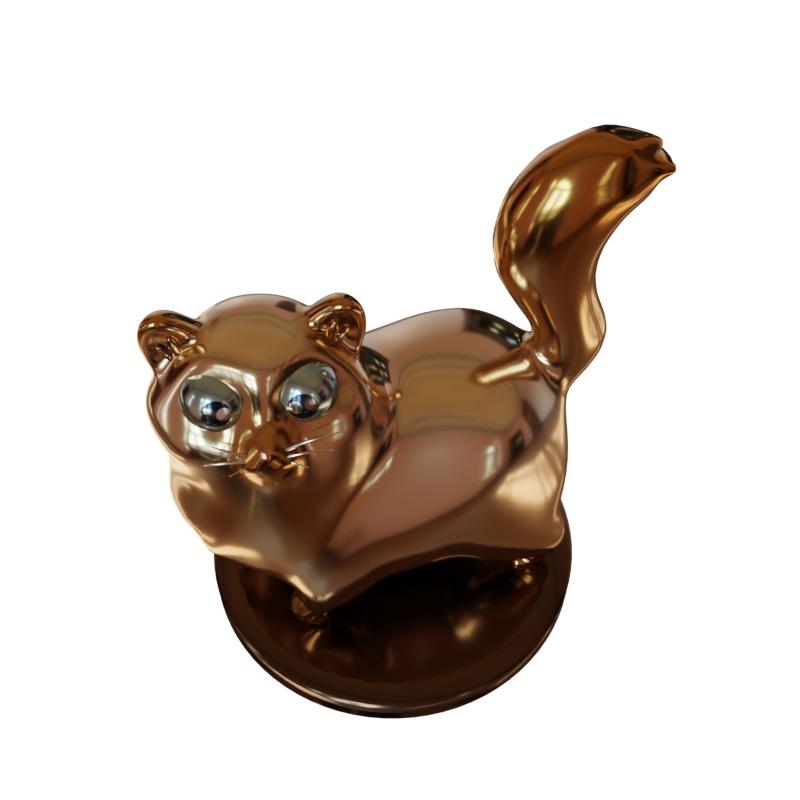} &
        \includegraphics[width=0.16\textwidth]{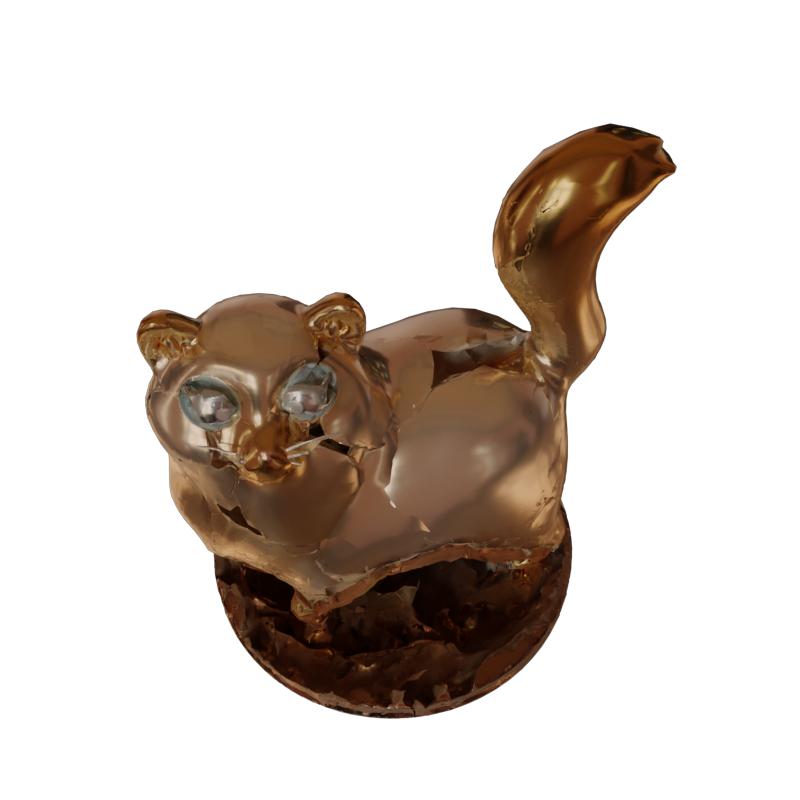} &
        \includegraphics[width=0.16\textwidth]{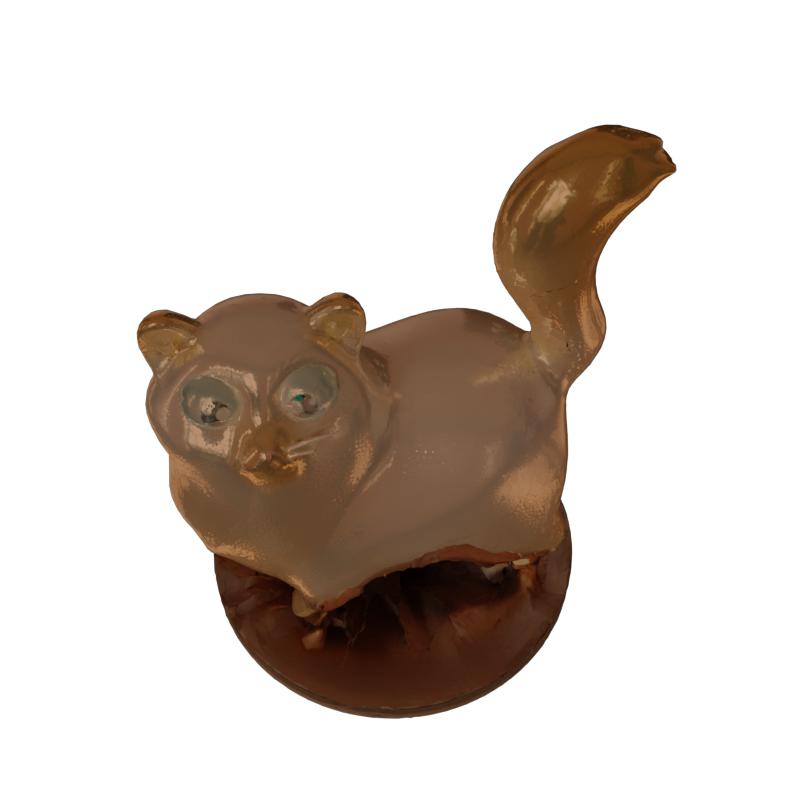} &
        \includegraphics[width=0.16\textwidth]{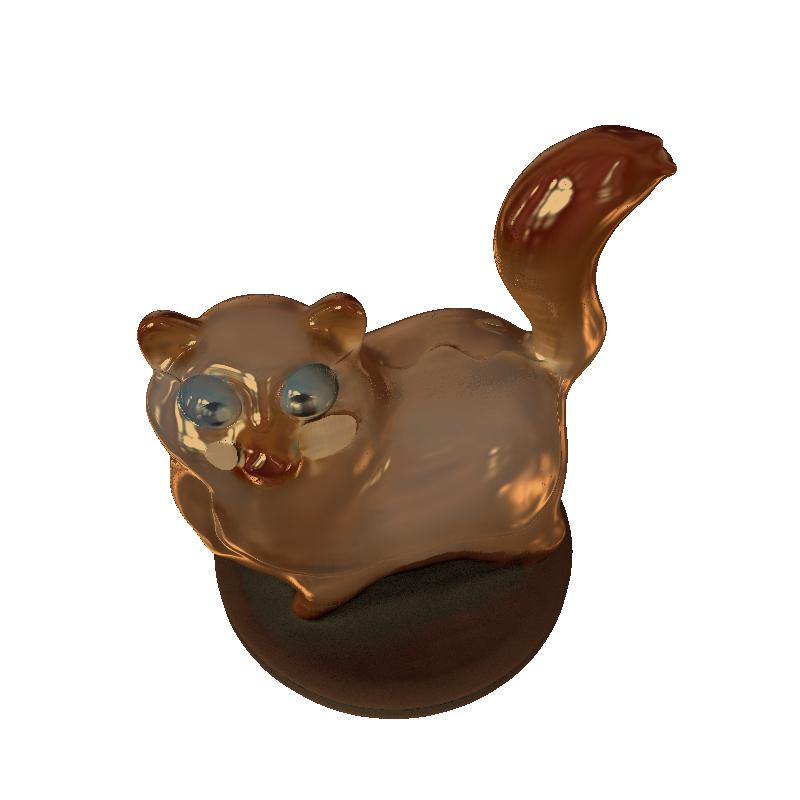} &
        \includegraphics[width=0.16\textwidth]{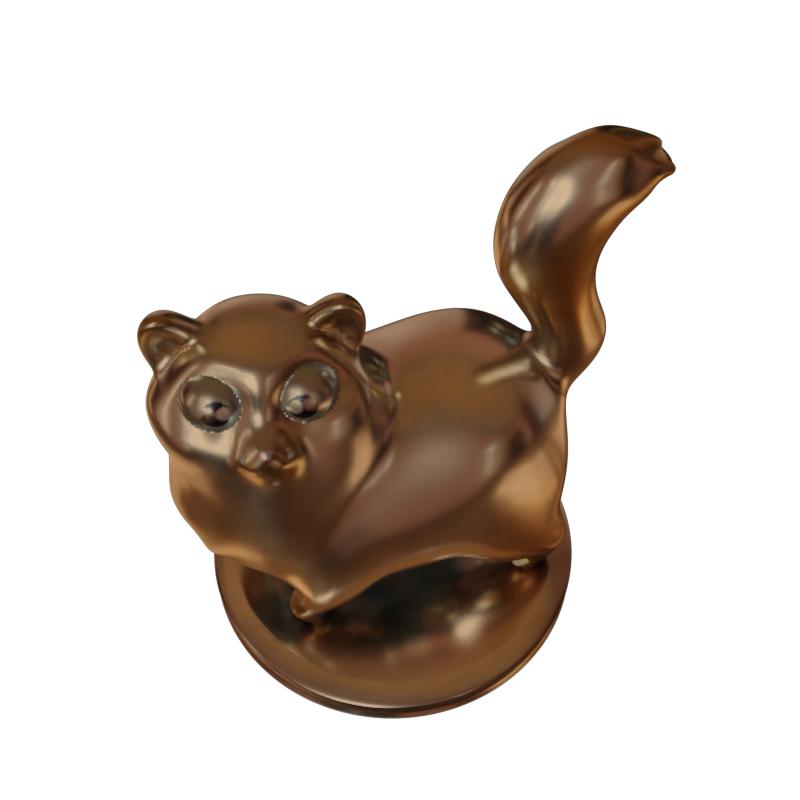} &
        \includegraphics[width=0.16\textwidth]{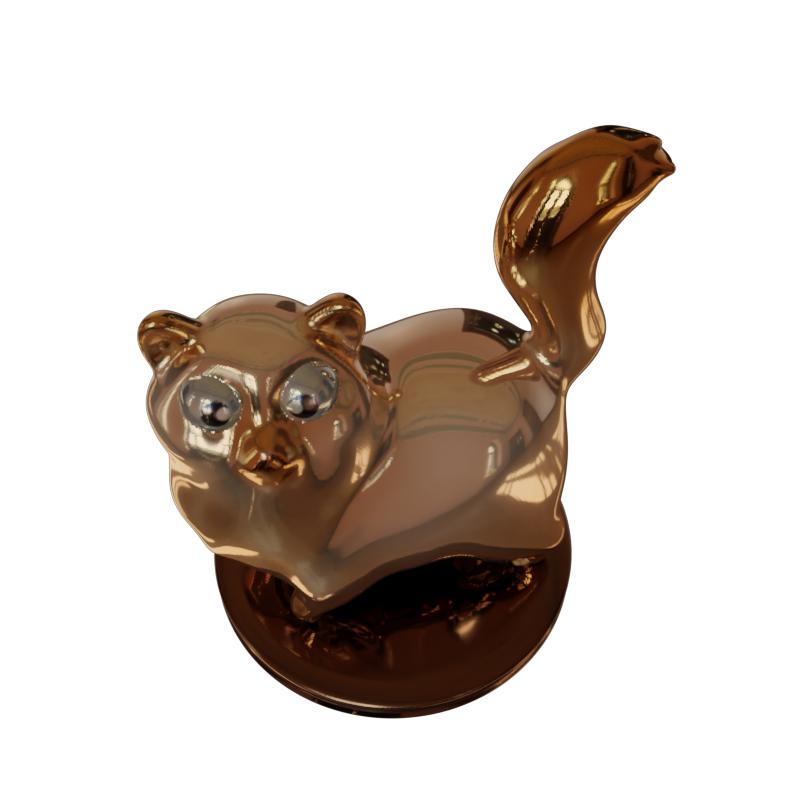} \\
        \includegraphics[width=0.16\textwidth]{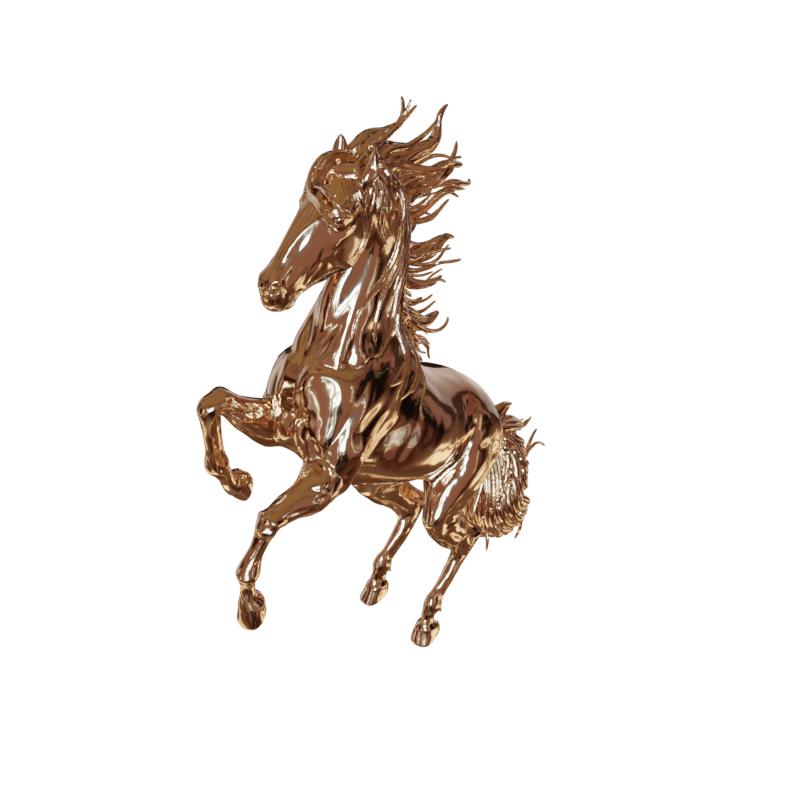} &
        \includegraphics[width=0.16\textwidth]{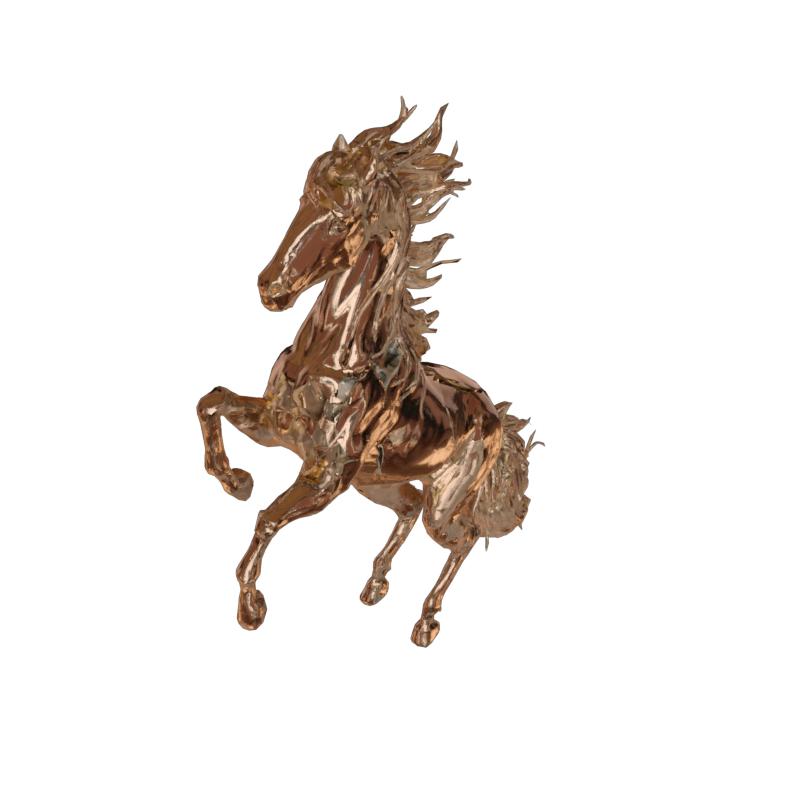} &
        \includegraphics[width=0.16\textwidth]{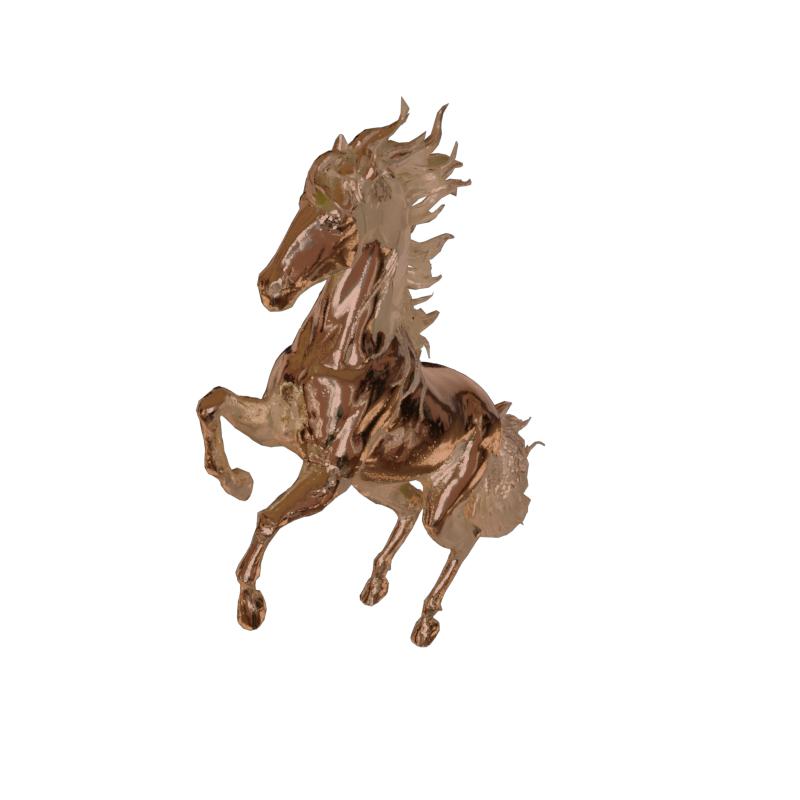} &
        \includegraphics[width=0.16\textwidth]{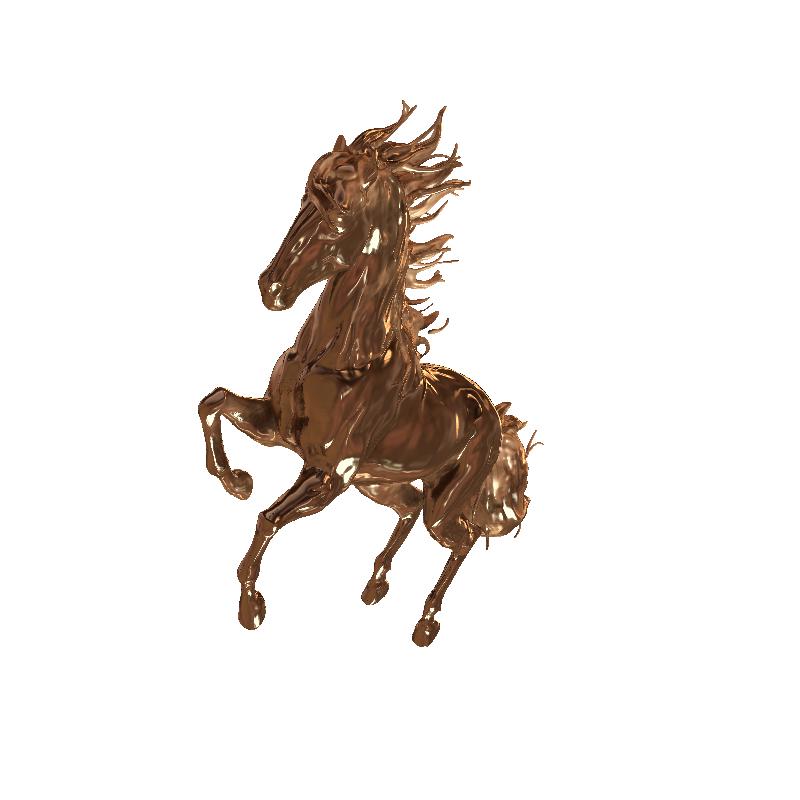} &
        \includegraphics[width=0.16\textwidth]{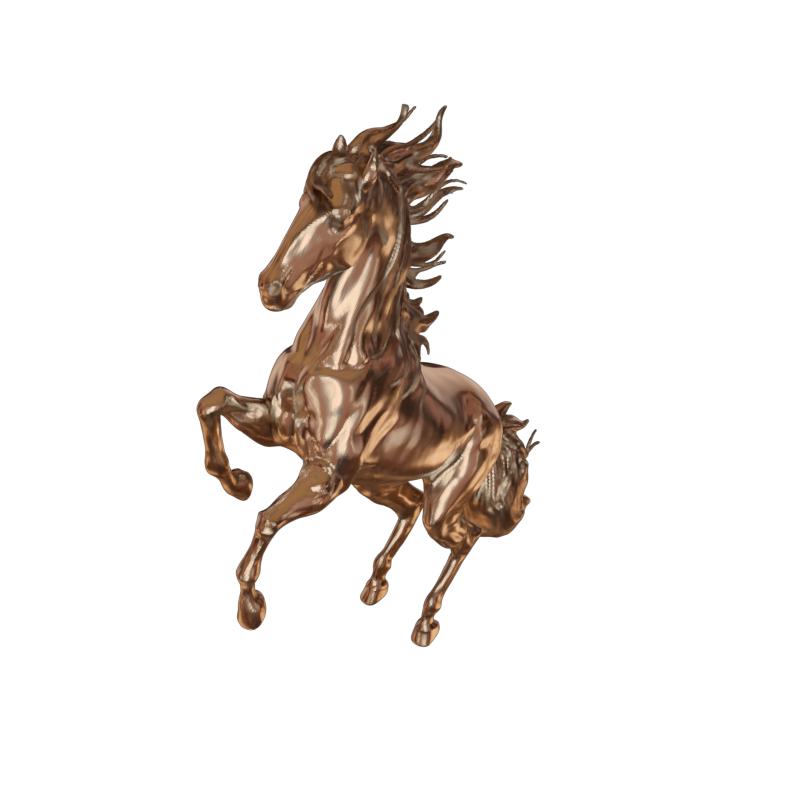} &
        \includegraphics[width=0.16\textwidth]{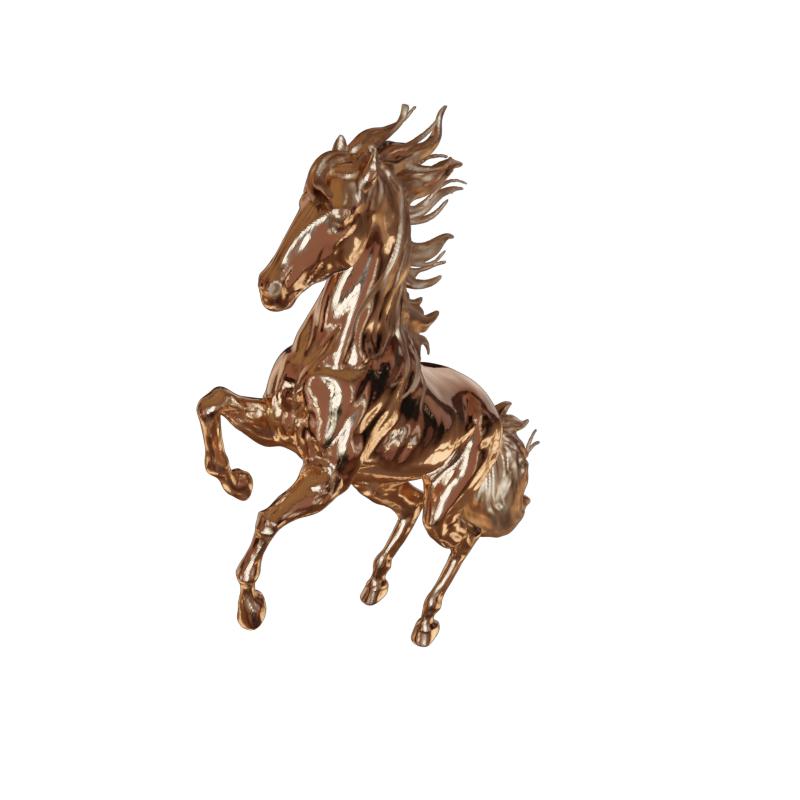} \\
        \includegraphics[width=0.16\textwidth]{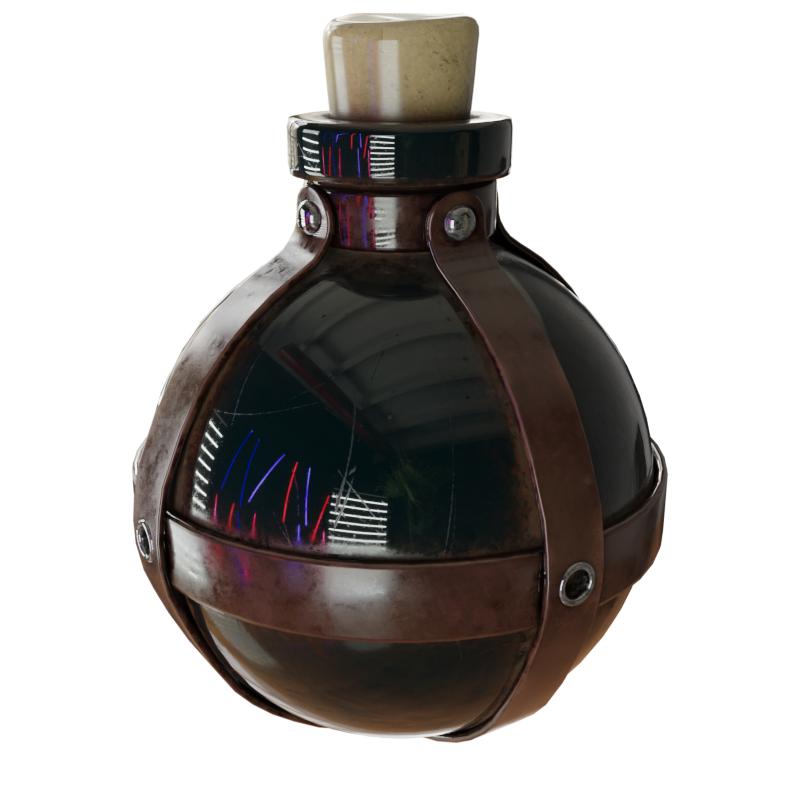} &
        \includegraphics[width=0.16\textwidth]{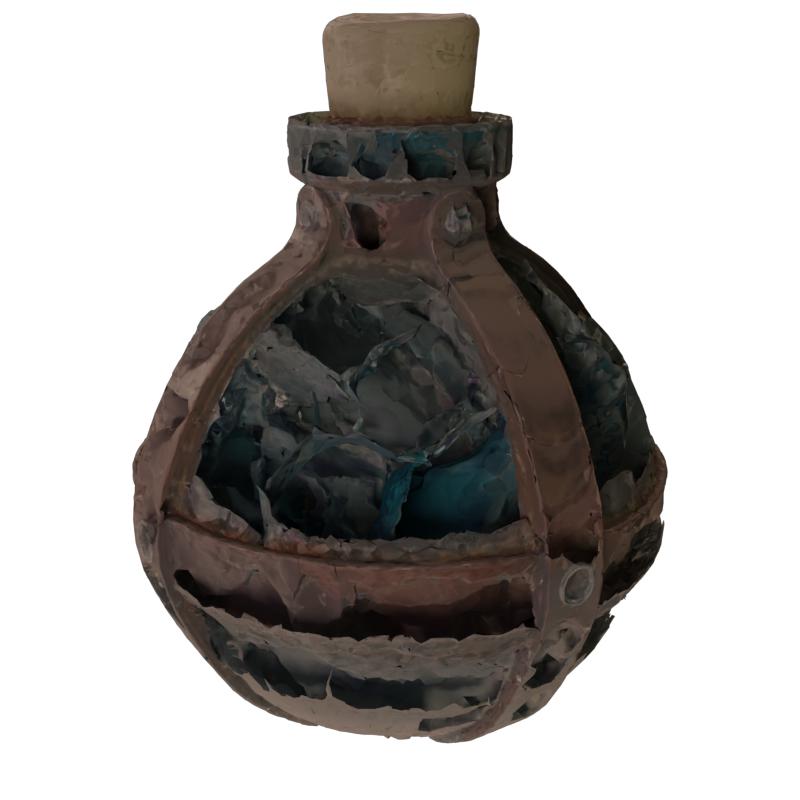} &
        \includegraphics[width=0.16\textwidth]{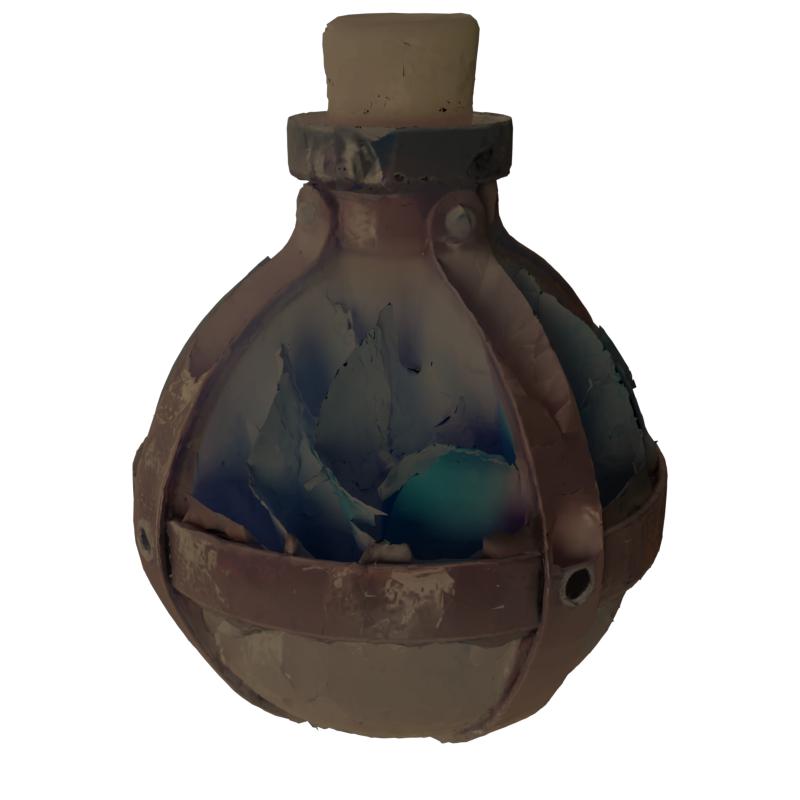} &
        \includegraphics[width=0.16\textwidth]{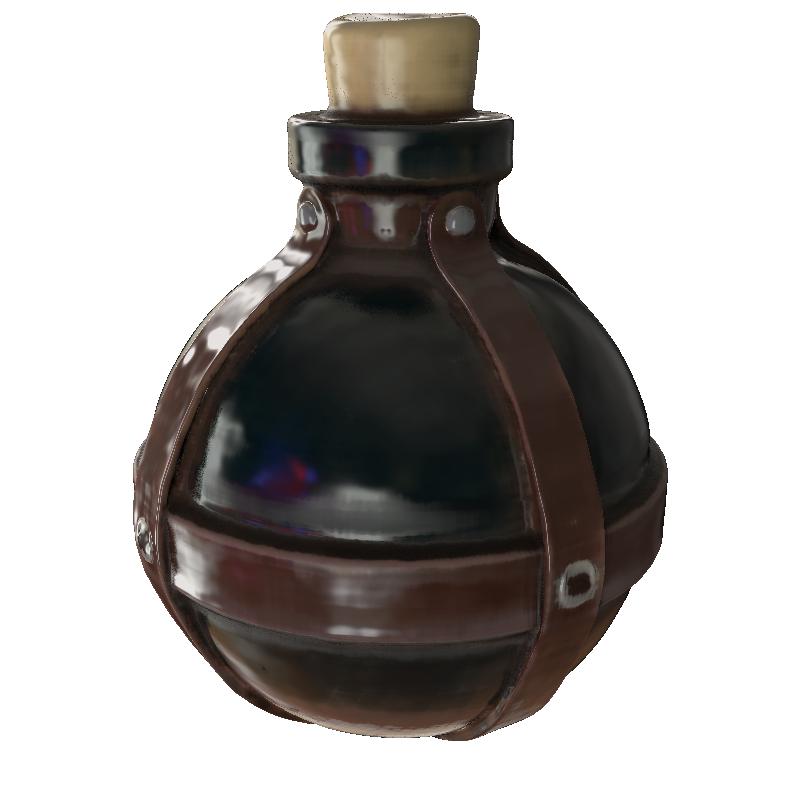} &
        \includegraphics[width=0.16\textwidth]{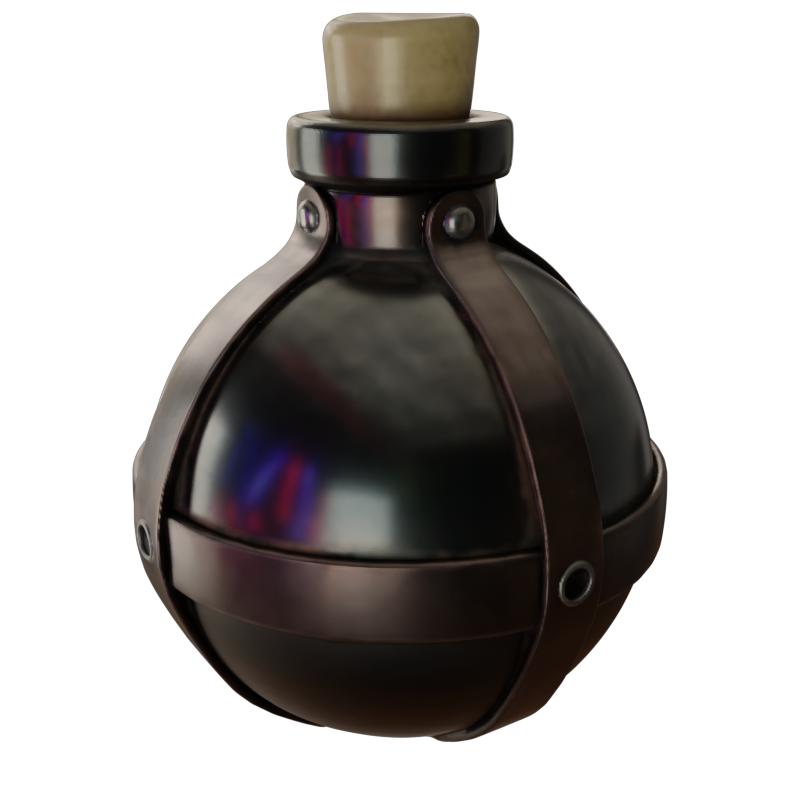} &
        \includegraphics[width=0.16\textwidth]{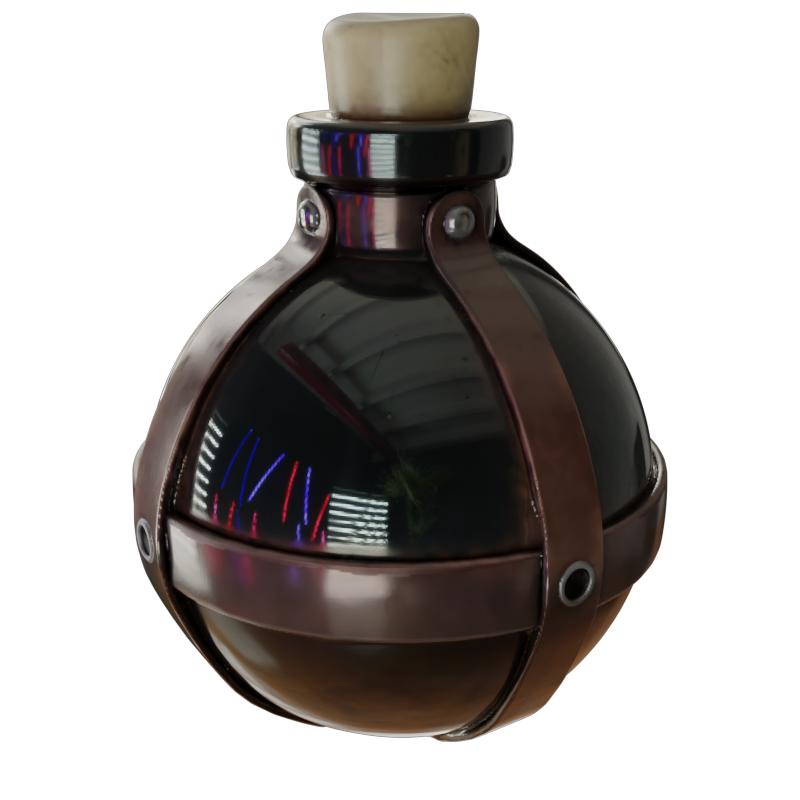} \\
        \includegraphics[width=0.16\textwidth]{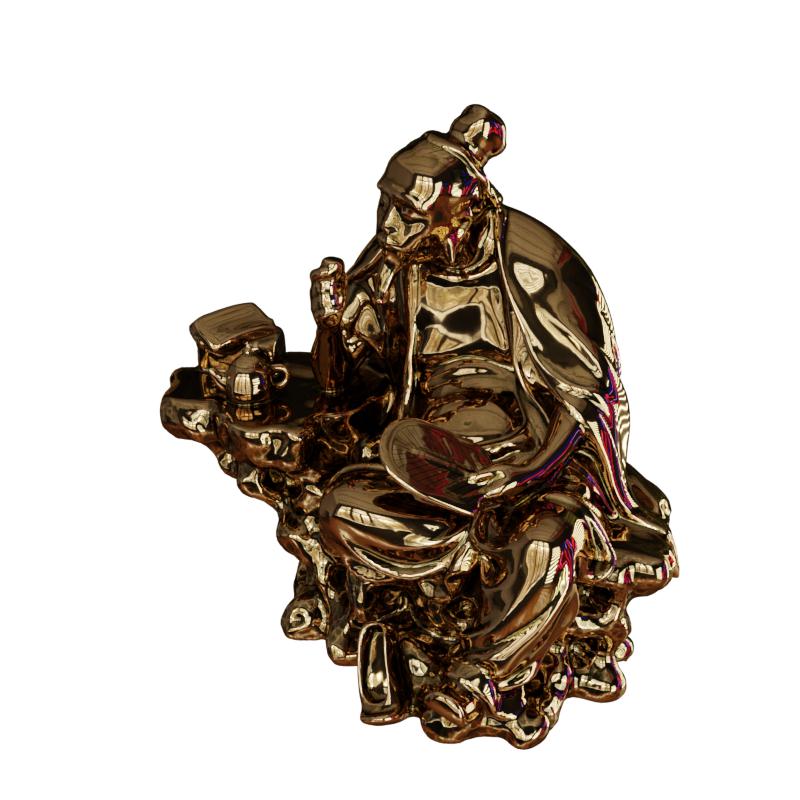} &
        \includegraphics[width=0.16\textwidth]{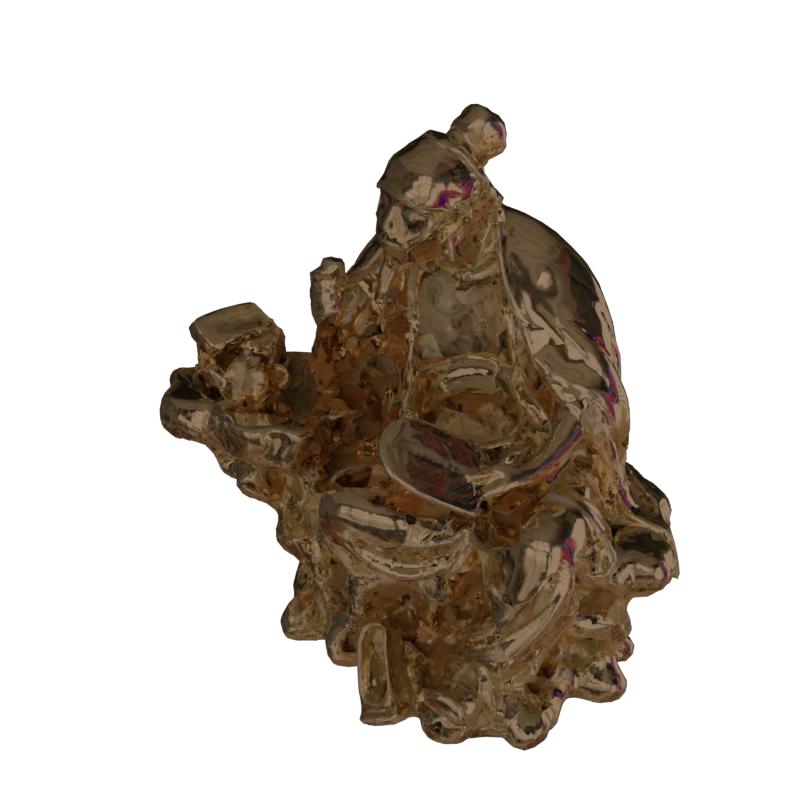} &
        \includegraphics[width=0.16\textwidth]{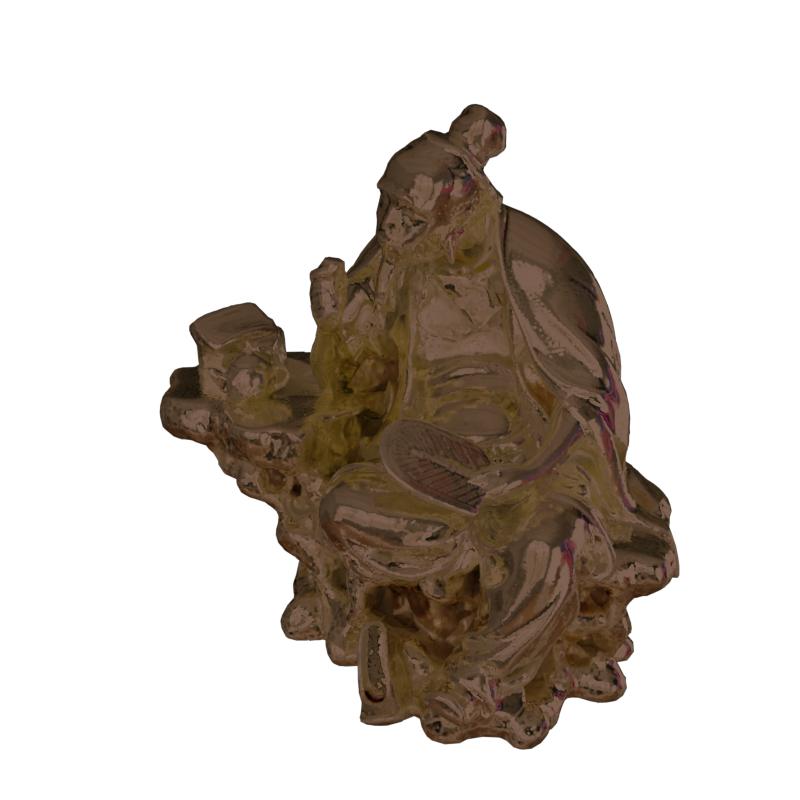} &
        \includegraphics[width=0.16\textwidth]{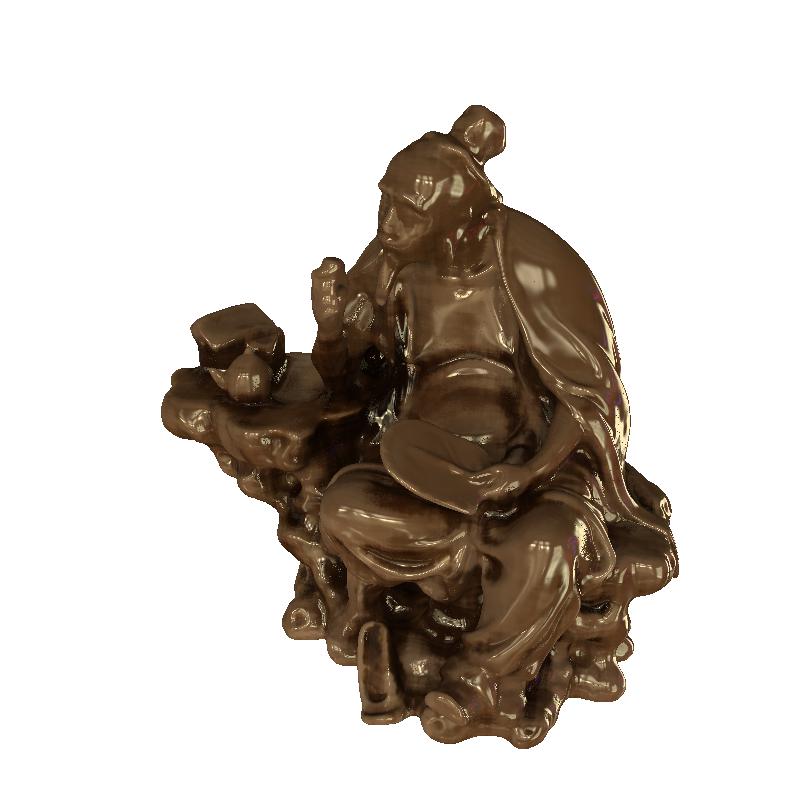} &
        \includegraphics[width=0.16\textwidth]{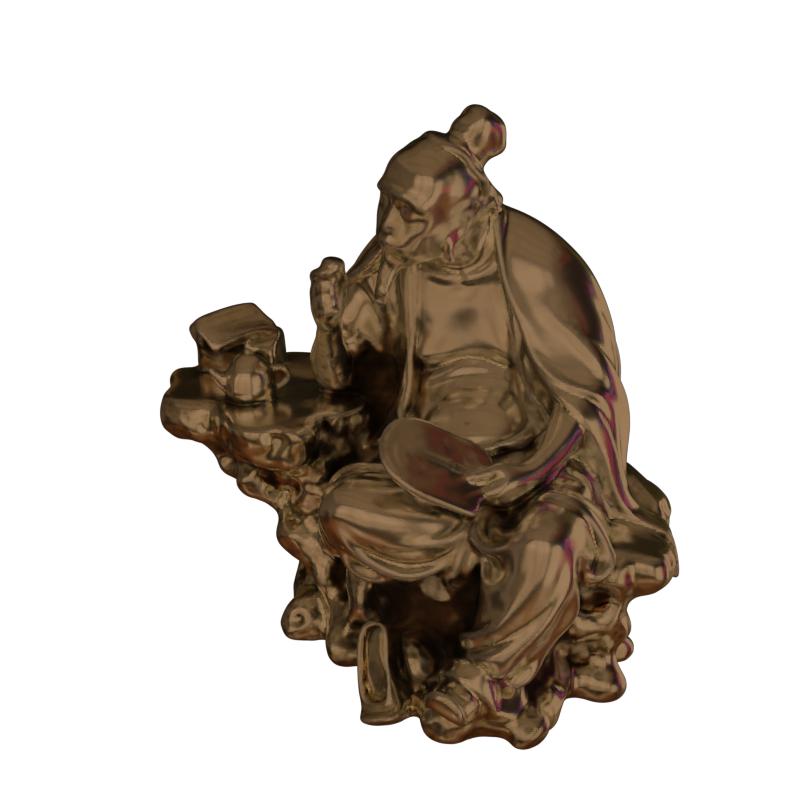} &
        \includegraphics[width=0.16\textwidth]{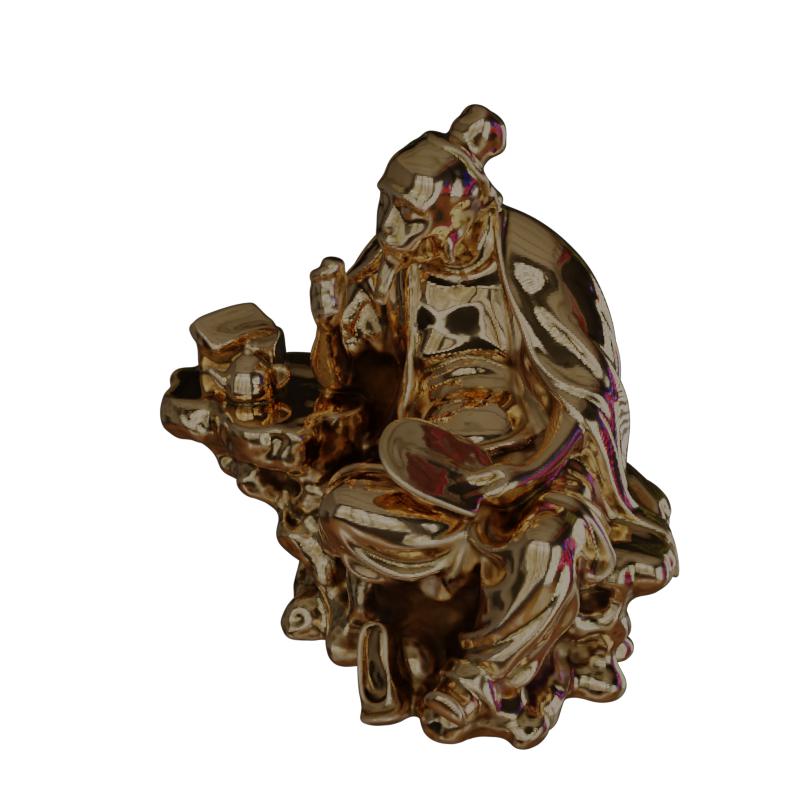} \\
        (a) Ground-truth & (b) NDR & (c) NDRMC & (d) MII & (e) NeILF & (f) Ours \\
    \end{tabular}
    \caption{\textbf{Relighting objects in the Glossy-Blender dataset}. We compare our method with NDR~\cite{munkberg2022extracting}, NDRMC~\cite{hasselgren2022shape}, MII~\cite{zhang2022modeling} and NeILF~\cite{yao2022neilf}. Note that all relighted images are normalized to match the average colors of the ground-truth images. The supplementary video contains more qualitative results.}
    \label{fig:syn_relight_app}
\end{figure*}
\begin{figure*}
    \centering
    \setlength\tabcolsep{1pt}
    \renewcommand{\arraystretch}{0.5} 
    \begin{tabular}{cccccc}
    \multicolumn{6}{c}{\includegraphics[width=0.1\linewidth]{image/real_relight/large_corridor_4k.jpg}} \\
    \includegraphics[width=0.16\textwidth]{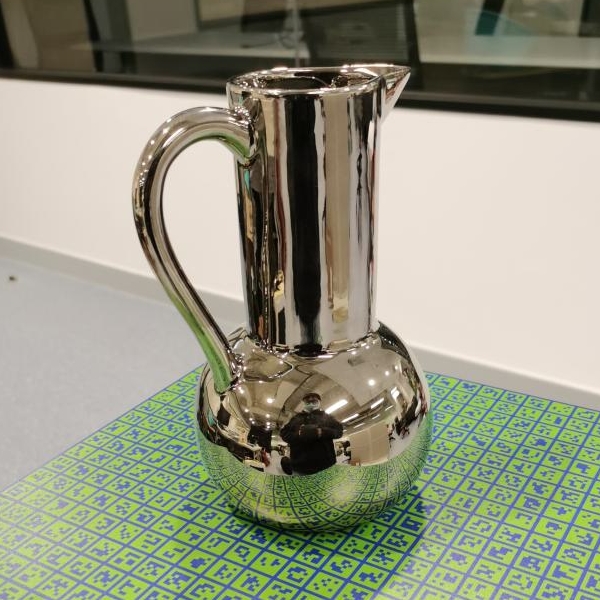} &
    \includegraphics[width=0.16\linewidth]{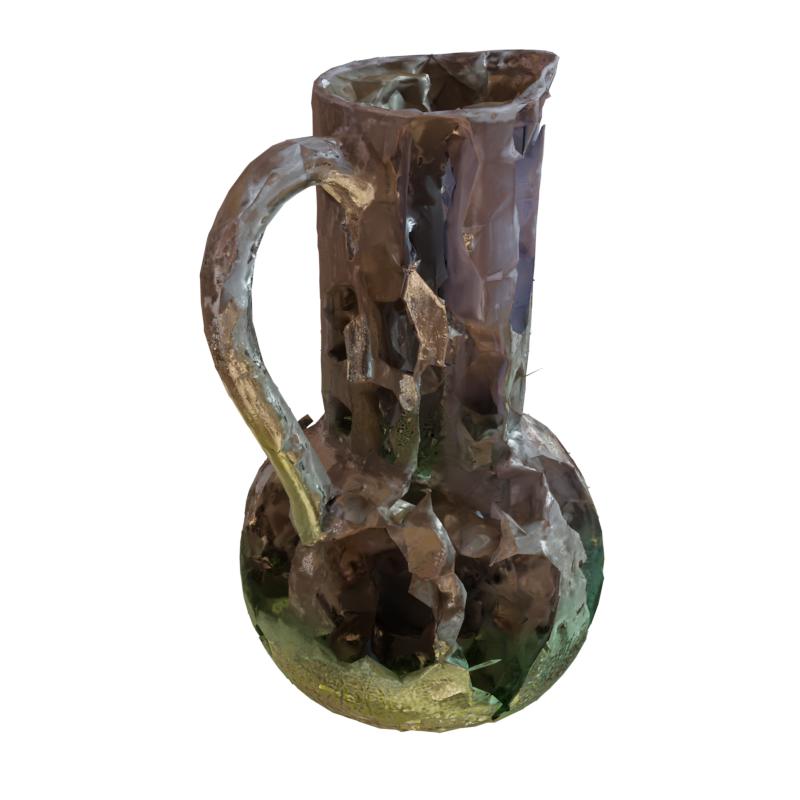} &
    \includegraphics[width=0.16\linewidth]{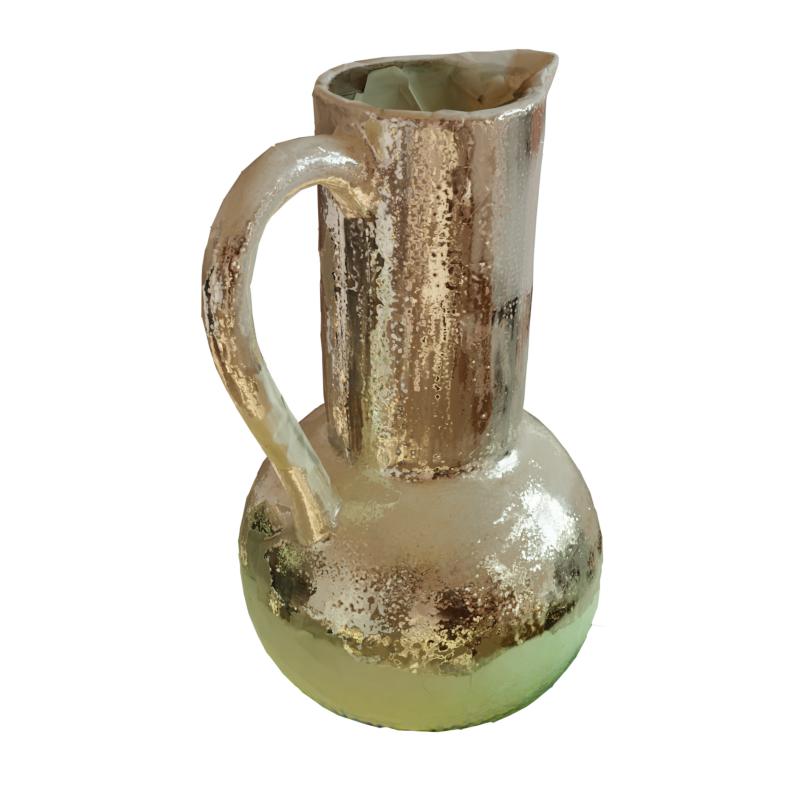} &
    \includegraphics[width=0.16\linewidth]{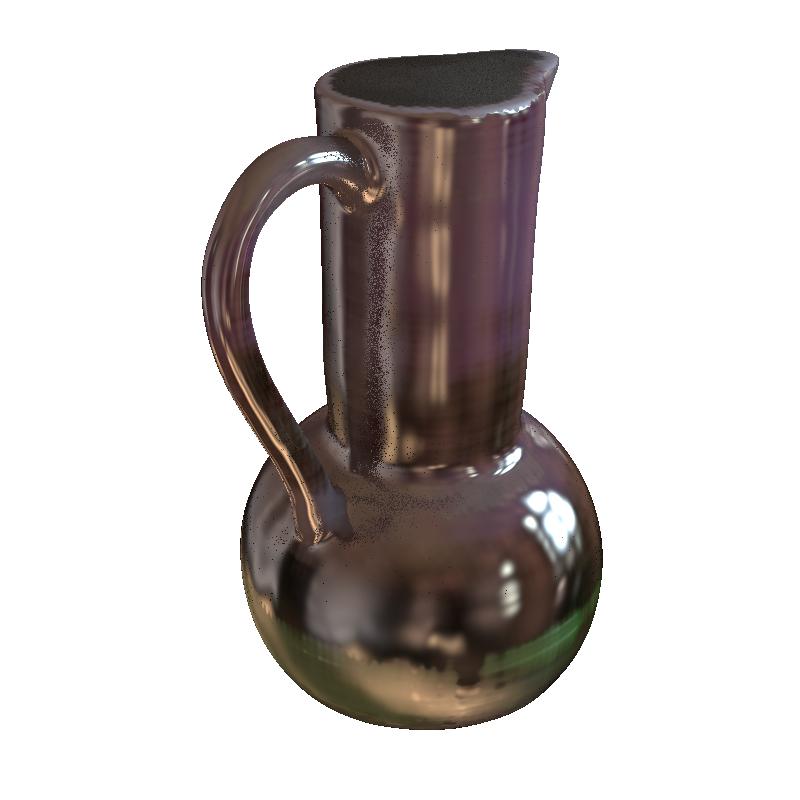} &
    \includegraphics[width=0.16\linewidth]{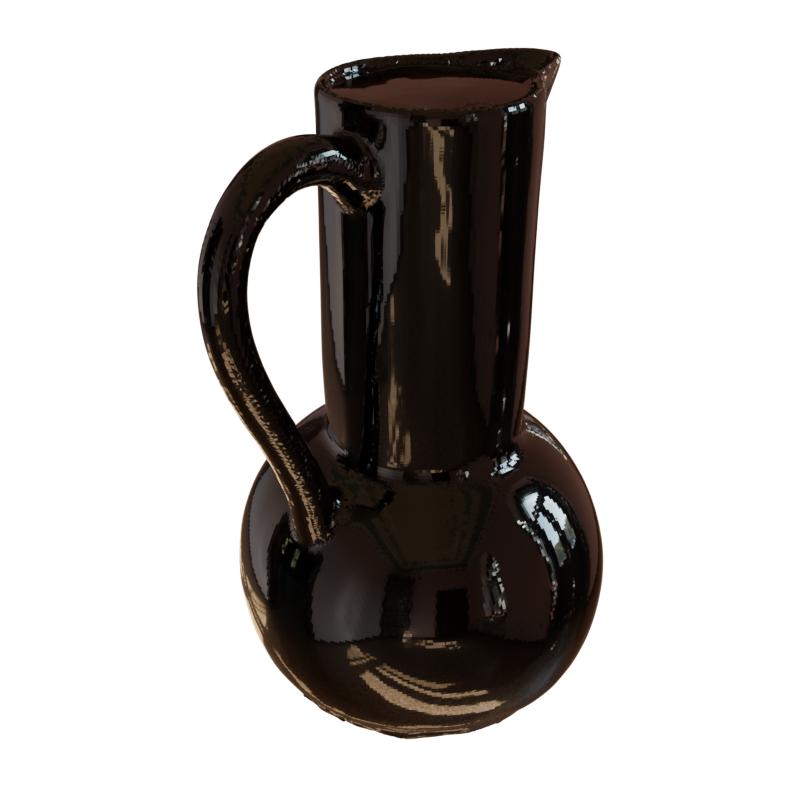} &
    \includegraphics[width=0.16\linewidth]{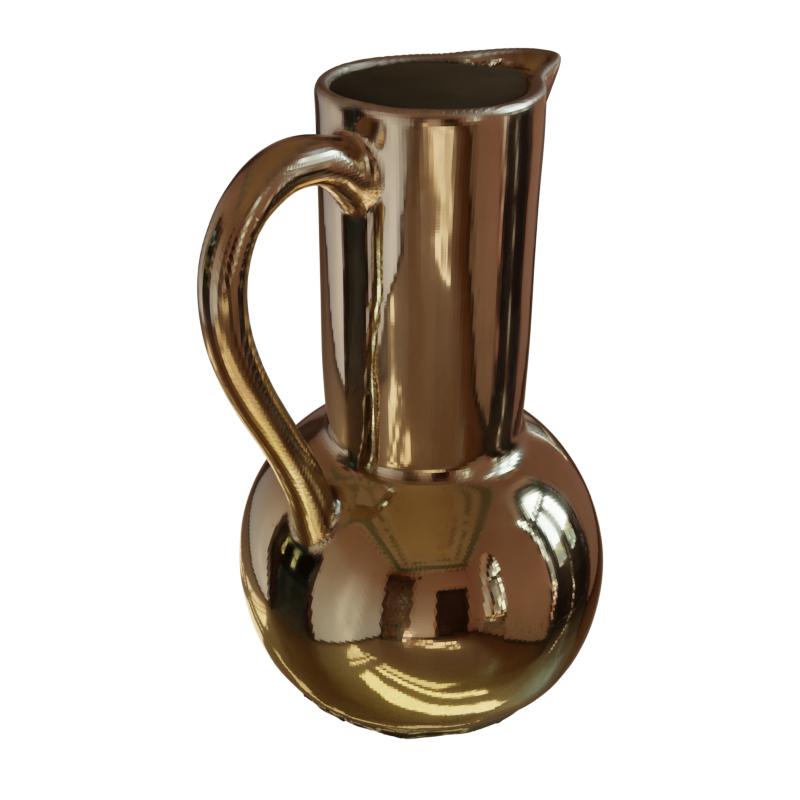} \\
    \includegraphics[width=0.16\linewidth]{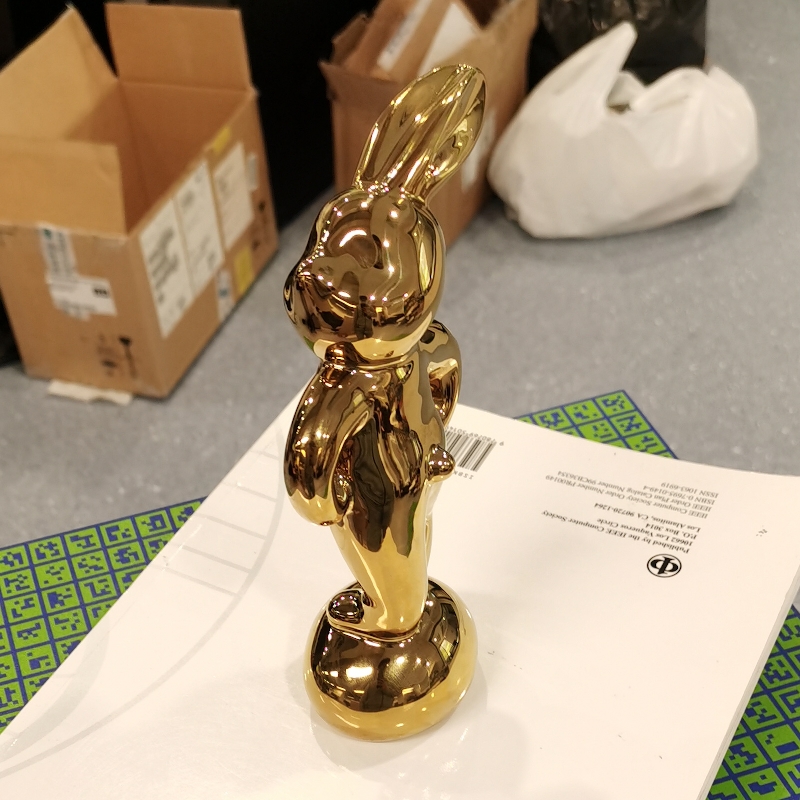} &
    \includegraphics[width=0.16\linewidth]{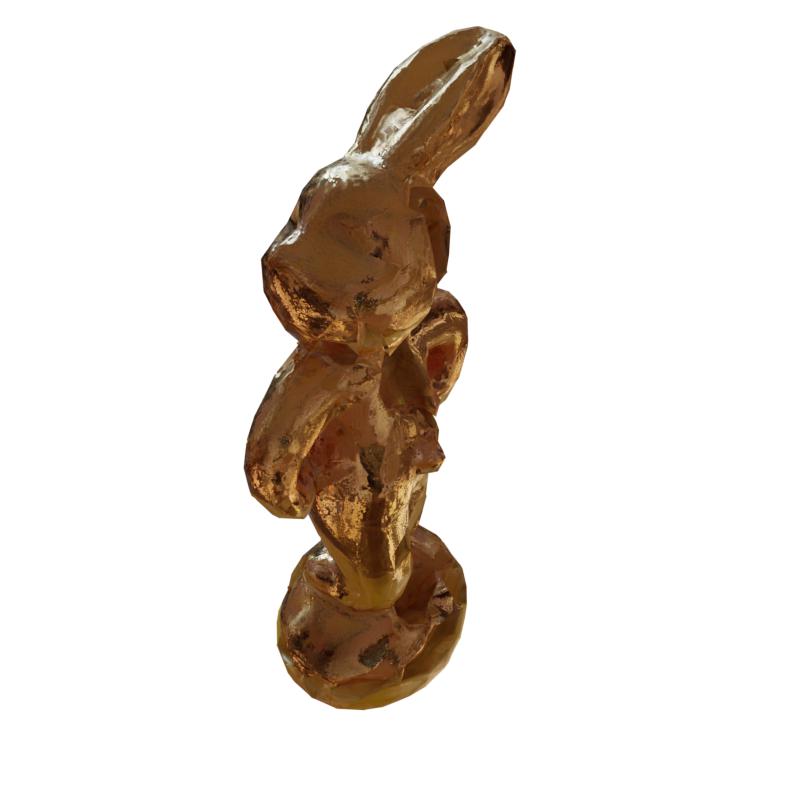} &
    \includegraphics[width=0.16\linewidth]{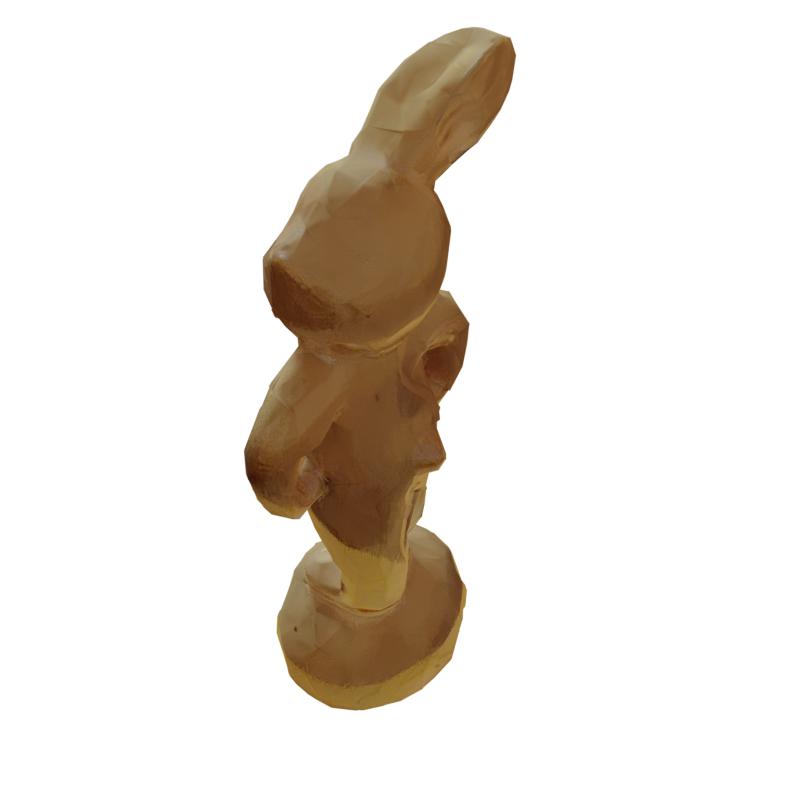} &
    \includegraphics[width=0.16\linewidth]{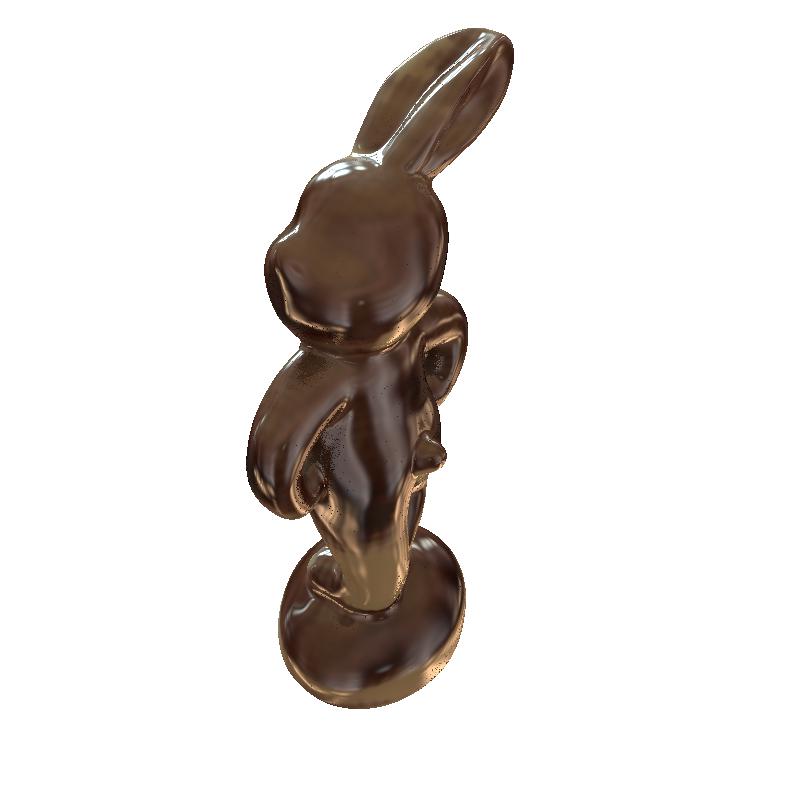} &
    \includegraphics[width=0.16\linewidth]{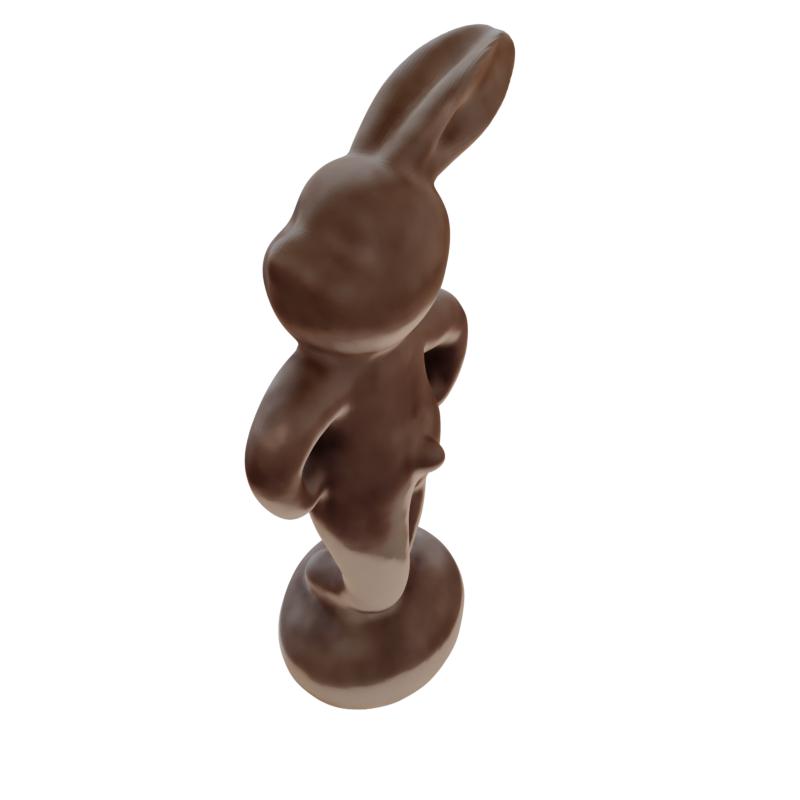} &
    \includegraphics[width=0.16\linewidth]{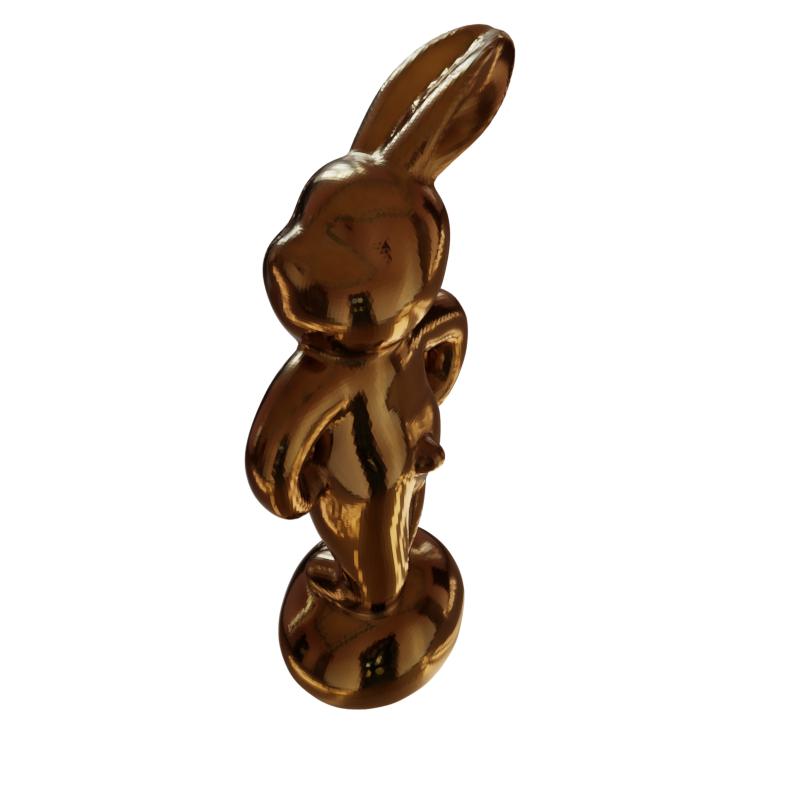} \\
    \multicolumn{6}{c}{\includegraphics[width=0.1\linewidth]{image/real_relight/neon_photostudio_4k.jpg}} \\
    \includegraphics[width=0.16\linewidth]{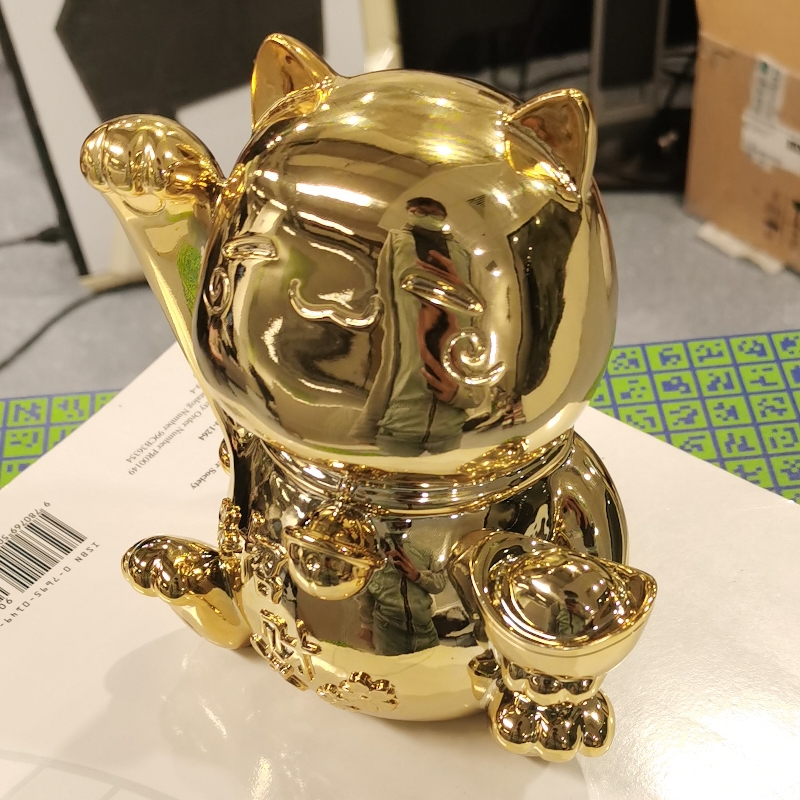} &
    \includegraphics[width=0.16\linewidth]{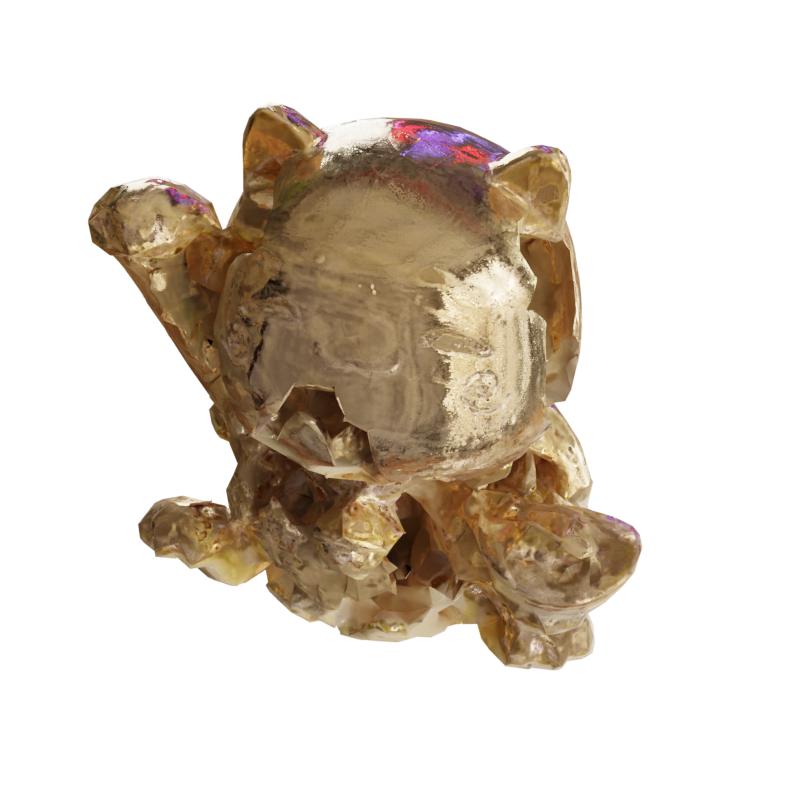} &
    \includegraphics[width=0.16\linewidth]{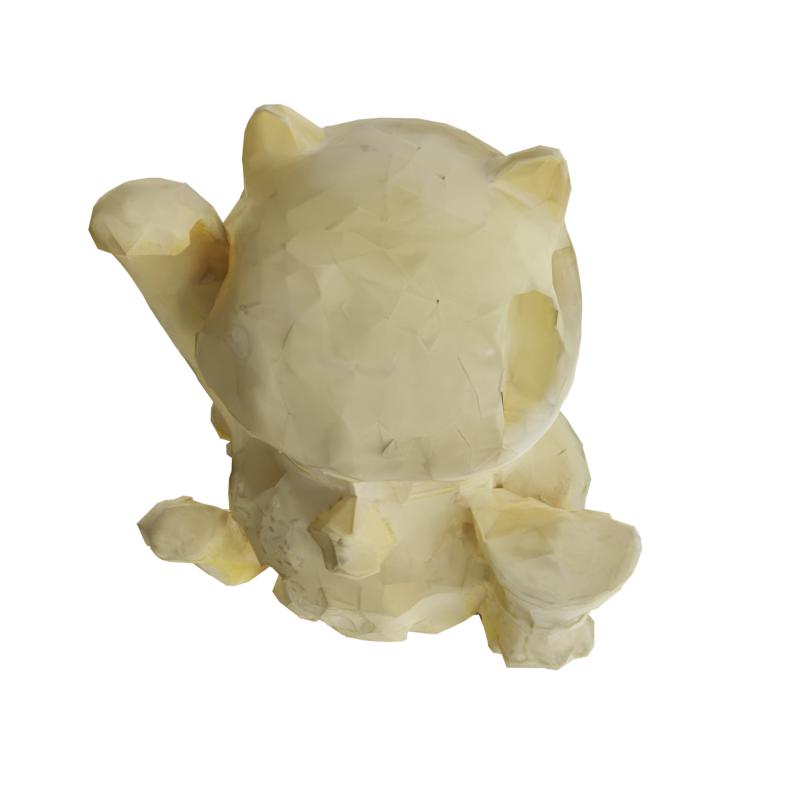} &
    \includegraphics[width=0.16\linewidth]{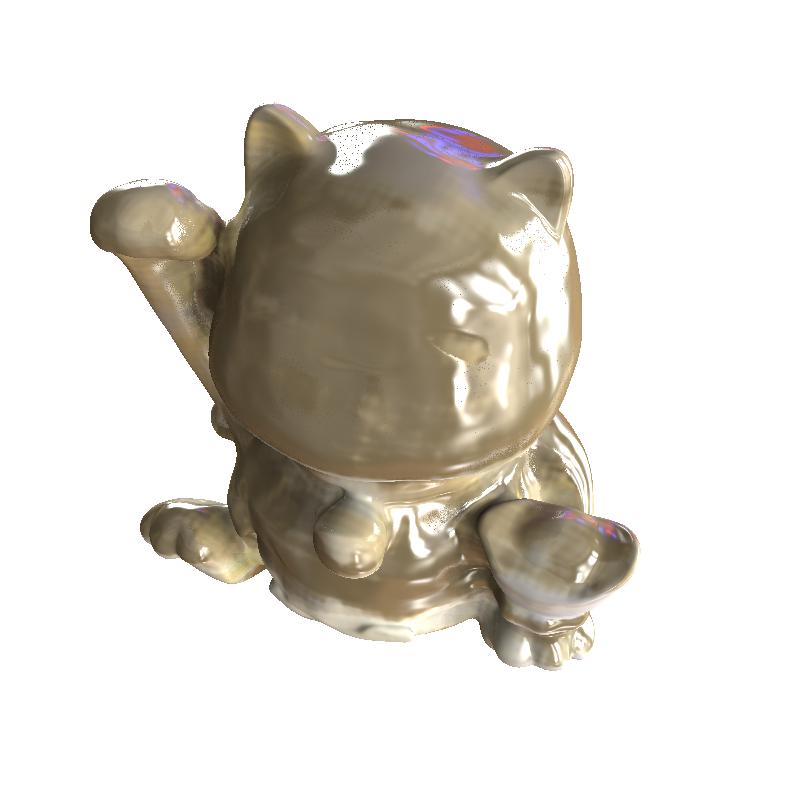} &
    \includegraphics[width=0.16\linewidth]{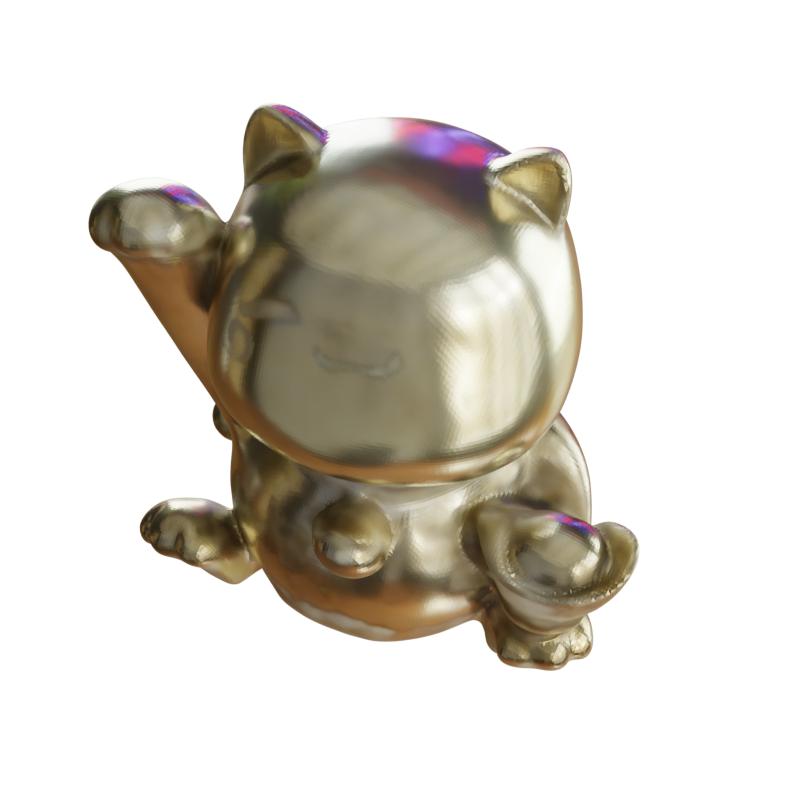} &
    \includegraphics[width=0.16\linewidth]{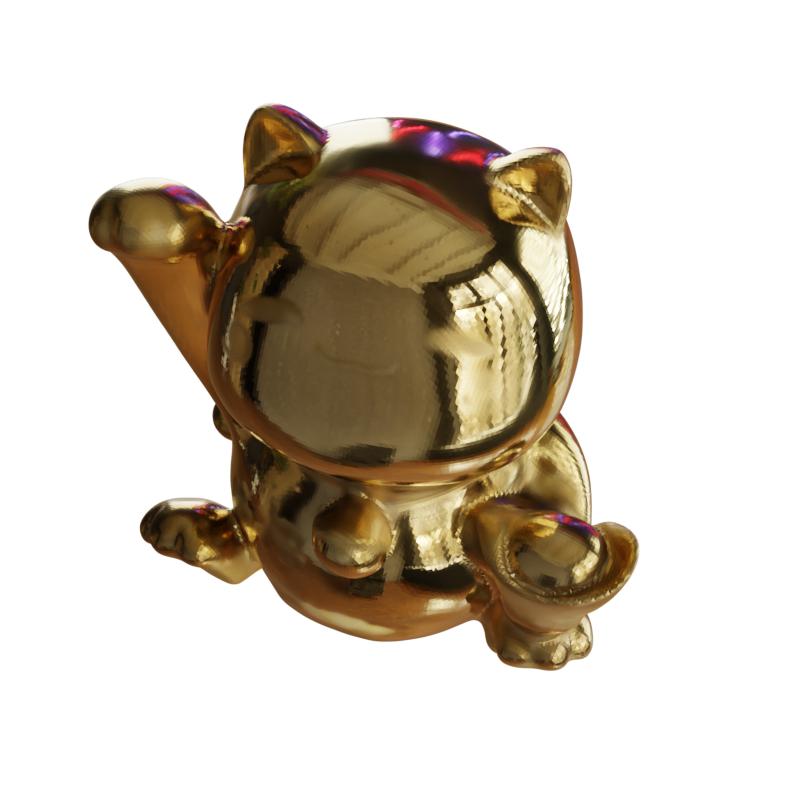} \\
    \multicolumn{6}{c}{\includegraphics[width=0.1\linewidth]{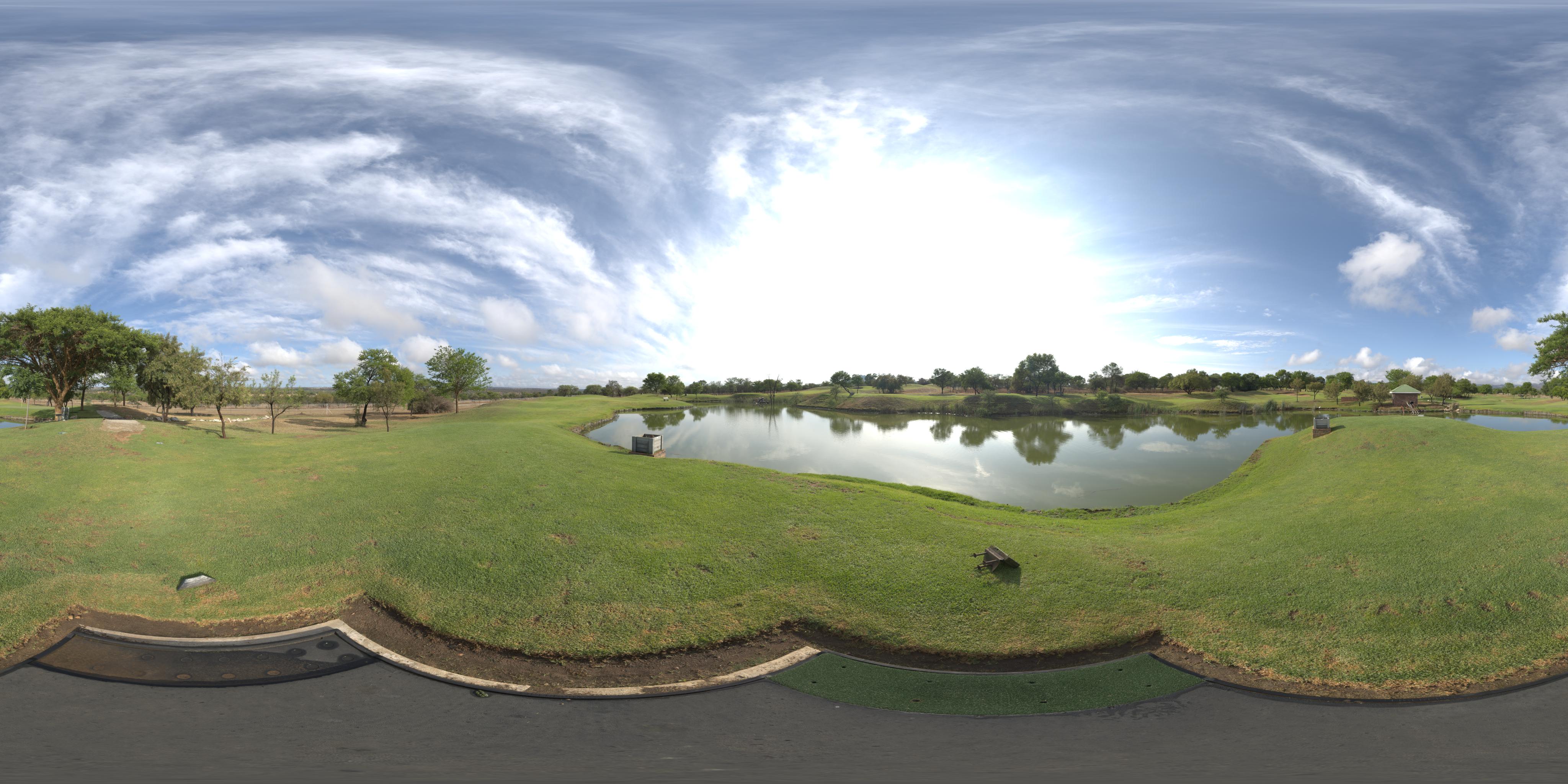}} \\
    \includegraphics[width=0.16\linewidth]{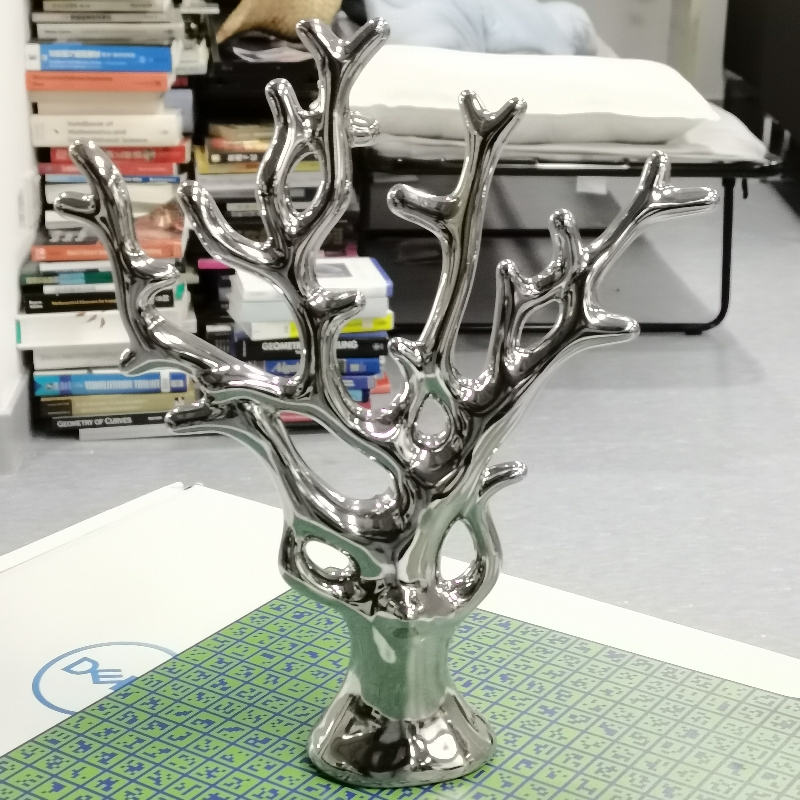} &
    \includegraphics[width=0.16\linewidth]{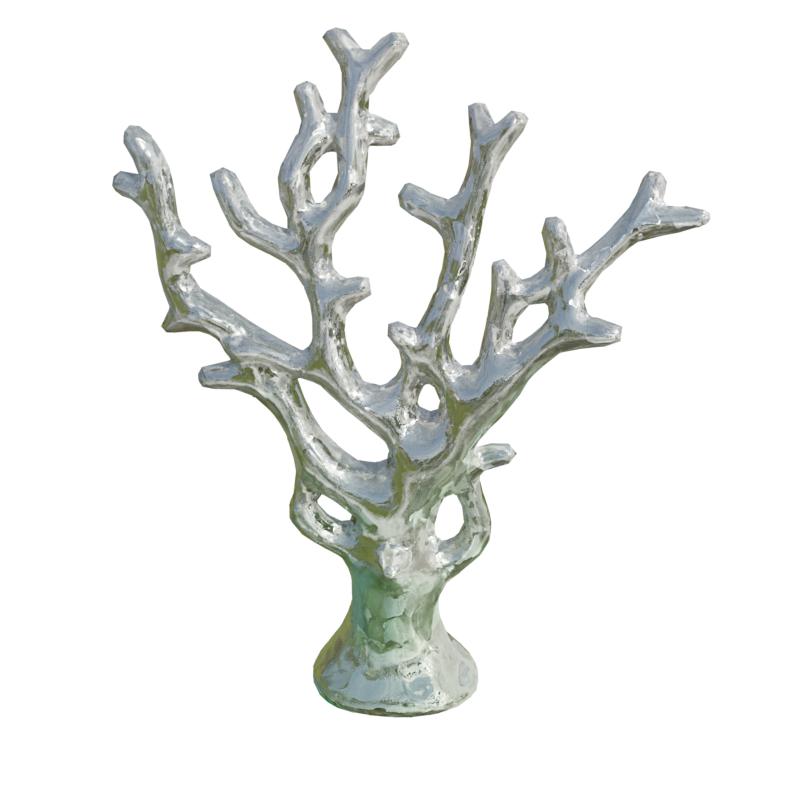} &
    \includegraphics[width=0.16\linewidth]{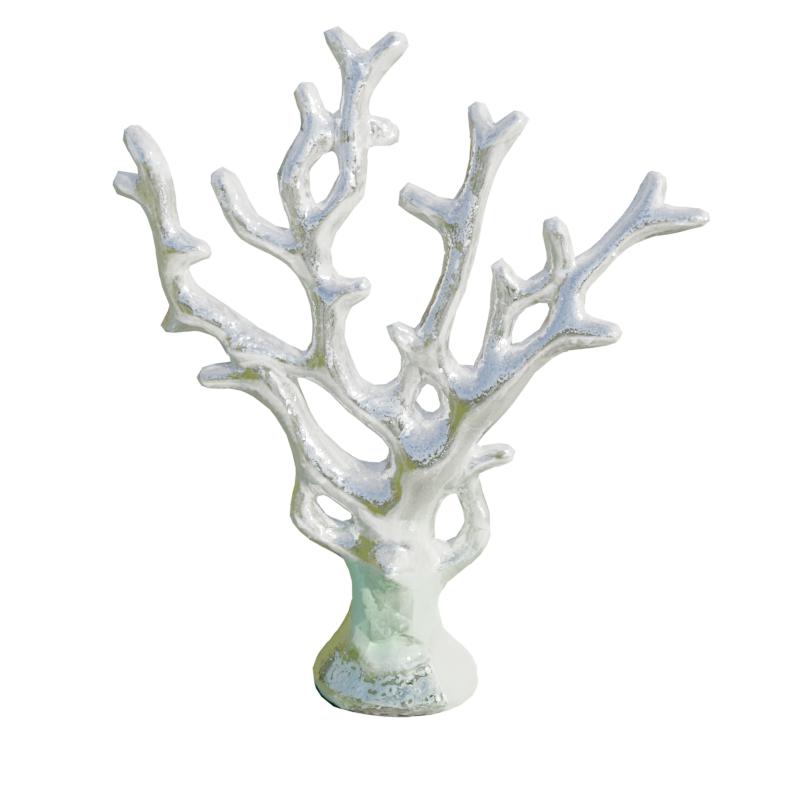} &
    \includegraphics[width=0.16\linewidth]{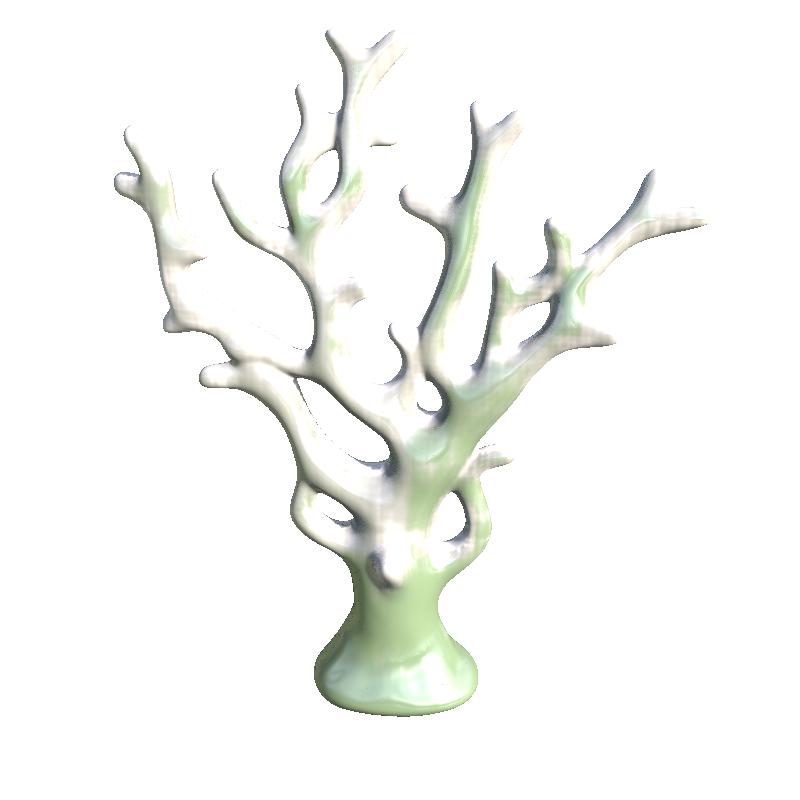} &
    \includegraphics[width=0.16\linewidth]{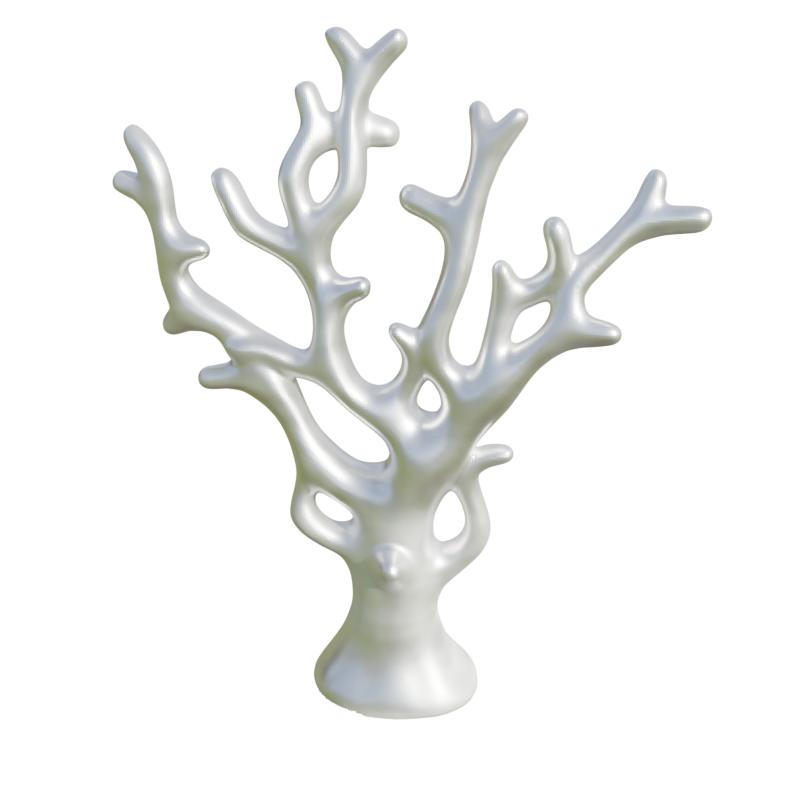} &
    \includegraphics[width=0.16\linewidth]{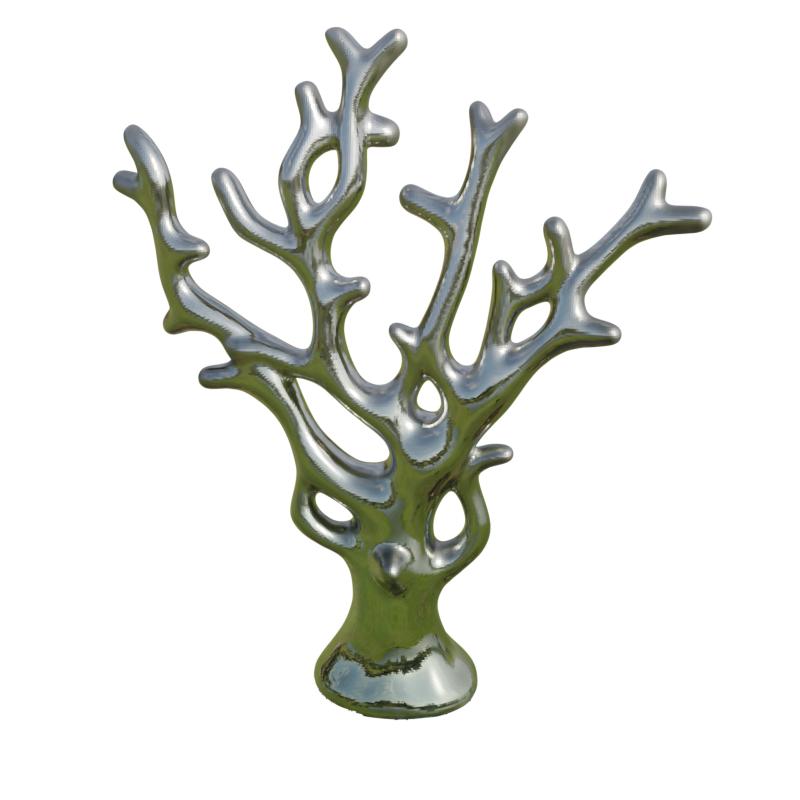} \\
    \includegraphics[width=0.16\linewidth]{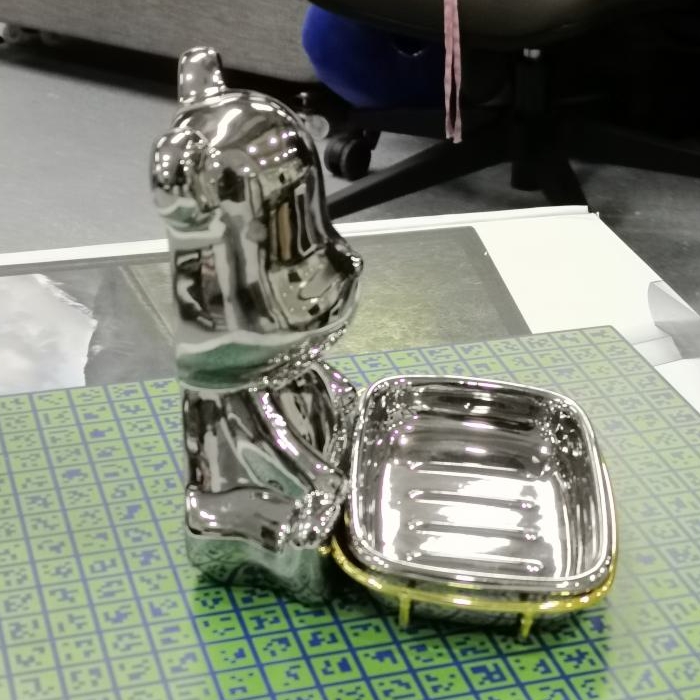} &
    \includegraphics[width=0.16\linewidth]{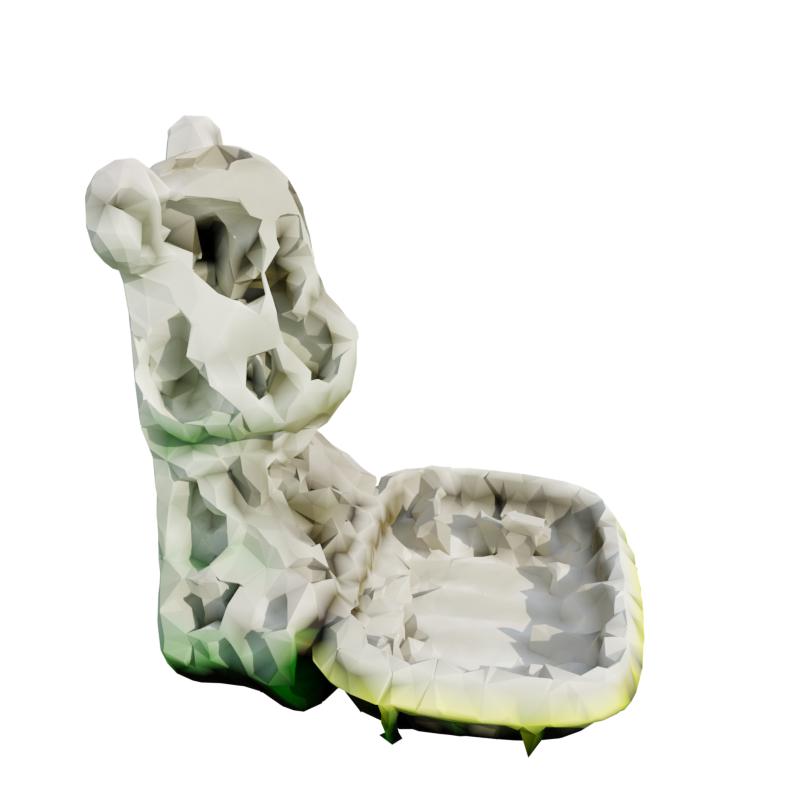} &
    \includegraphics[width=0.16\linewidth]{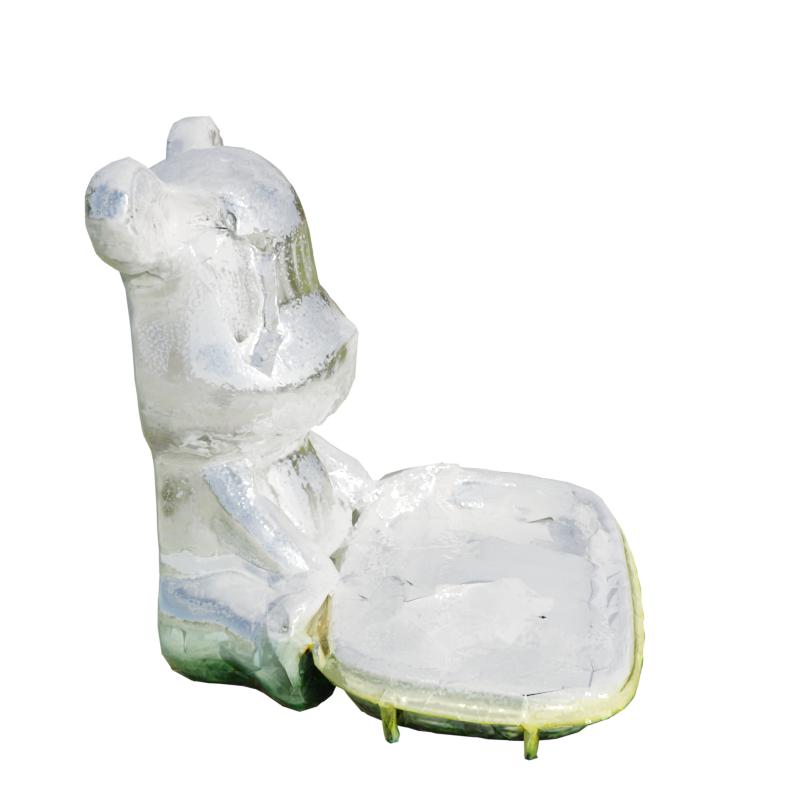} &
    \includegraphics[width=0.16\linewidth]{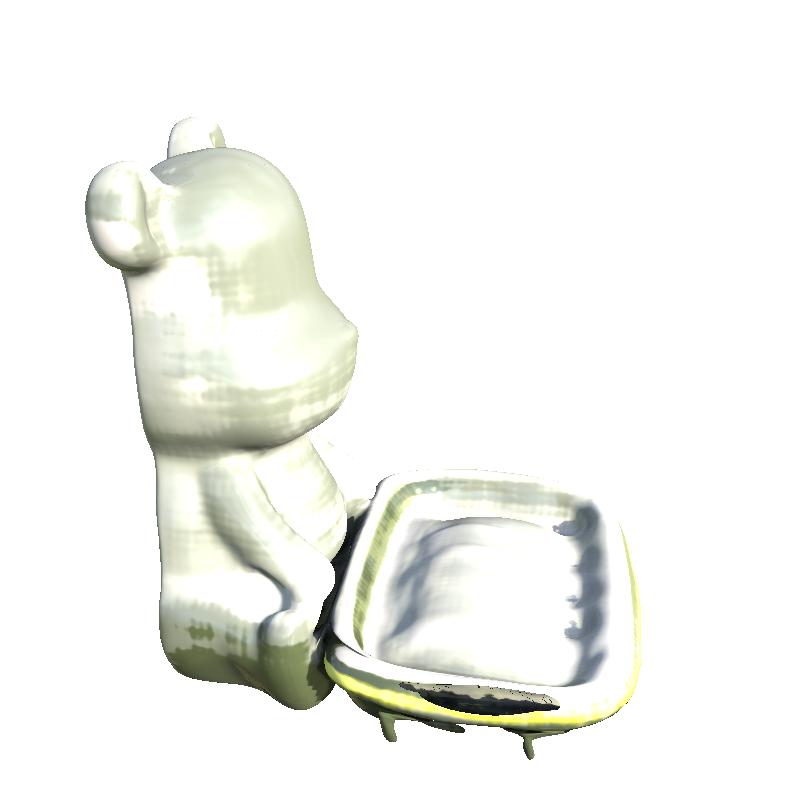} &
    \includegraphics[width=0.16\linewidth]{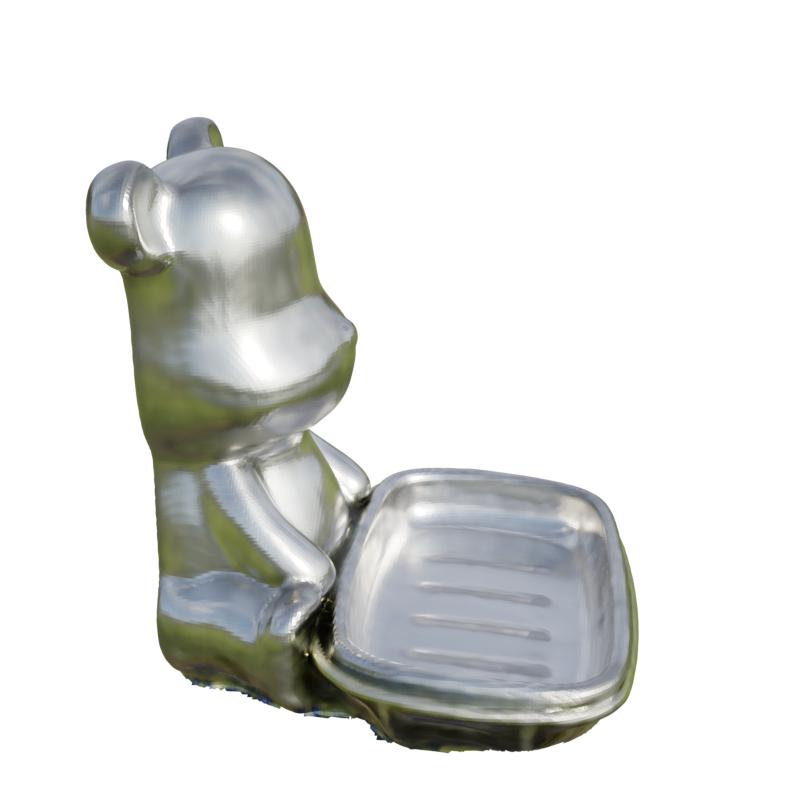} &
    \includegraphics[width=0.16\linewidth]{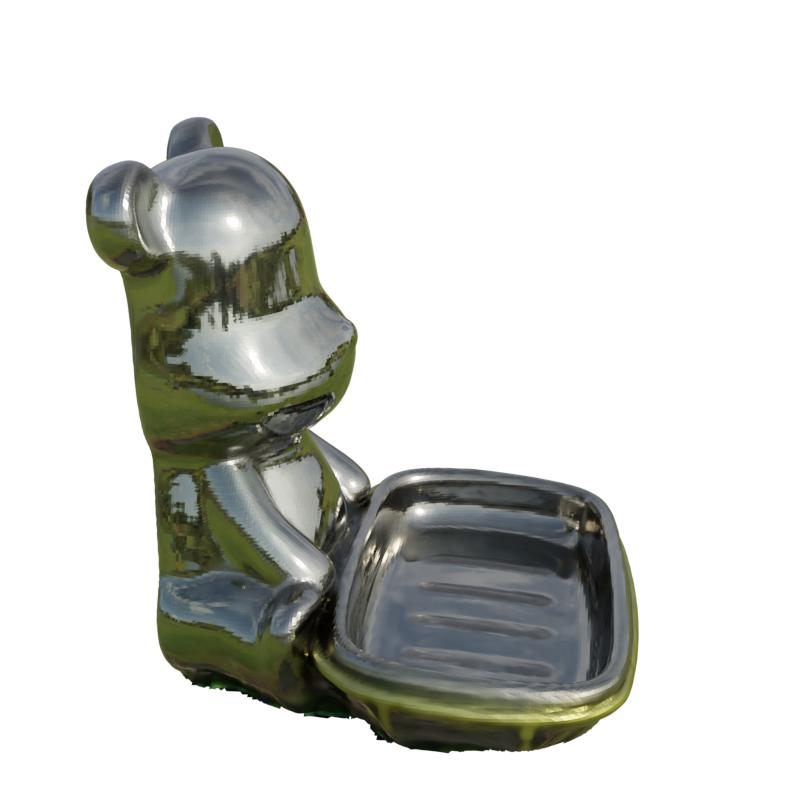} \\
    (a) Image & (b) NDR & (c) NDRMC & (d) MII & (e) NeILF & (f) Ours \\
    \end{tabular}
    \caption{\textbf{Relighting objects from the Glossy-Real dataset}. We provide a visual comparison with NDR~\cite{munkberg2022extracting}, NDRMC~\cite{hasselgren2022shape}, MII~\cite{zhang2022modeling}, and NeILF~\cite{yao2022neilf}. We provide the input image with the nearest viewpoint and the relighting HDR map as a reference. The supplementary video contains more qualitative results.
    }
    \label{fig:real_relight_app}
\end{figure*}
\begin{figure*}
    \centering
    \setlength\tabcolsep{1pt}
    \renewcommand{\arraystretch}{0.5} 
    \begin{tabular}{cccccc}

    & & & \includegraphics[width=0.16\textwidth]{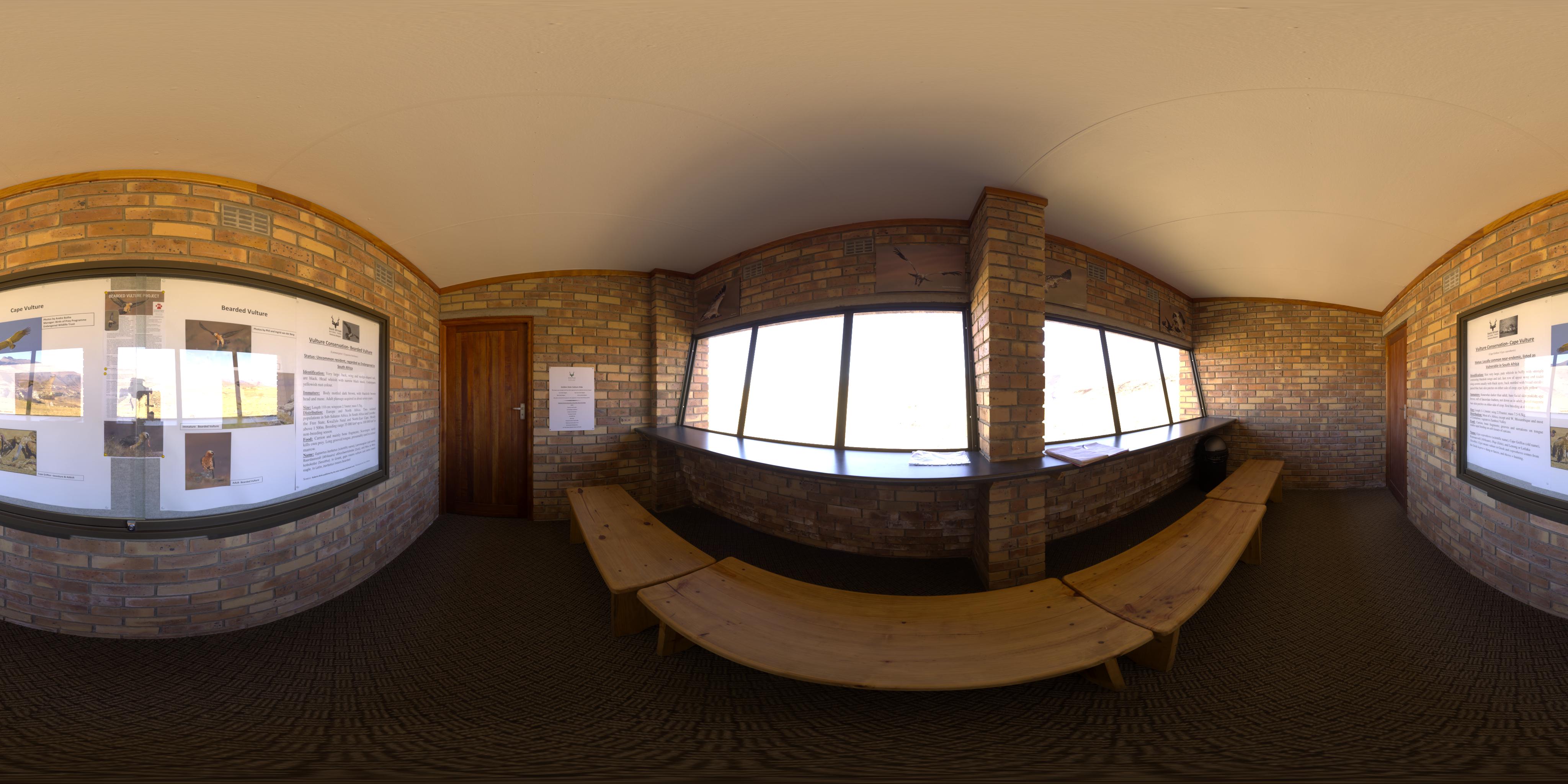} &
    \includegraphics[width=0.16\textwidth]{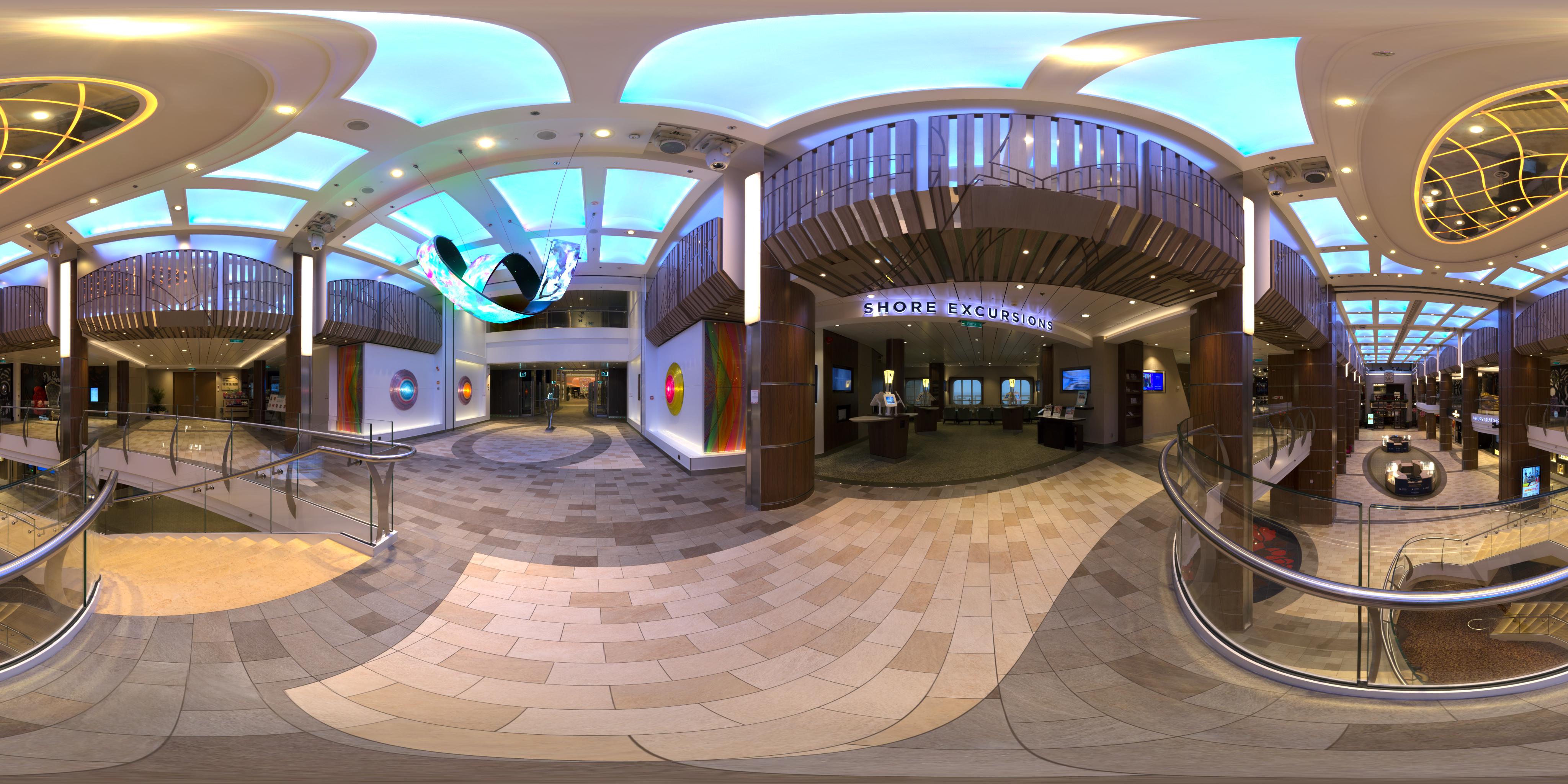} &
    \includegraphics[width=0.16\textwidth]{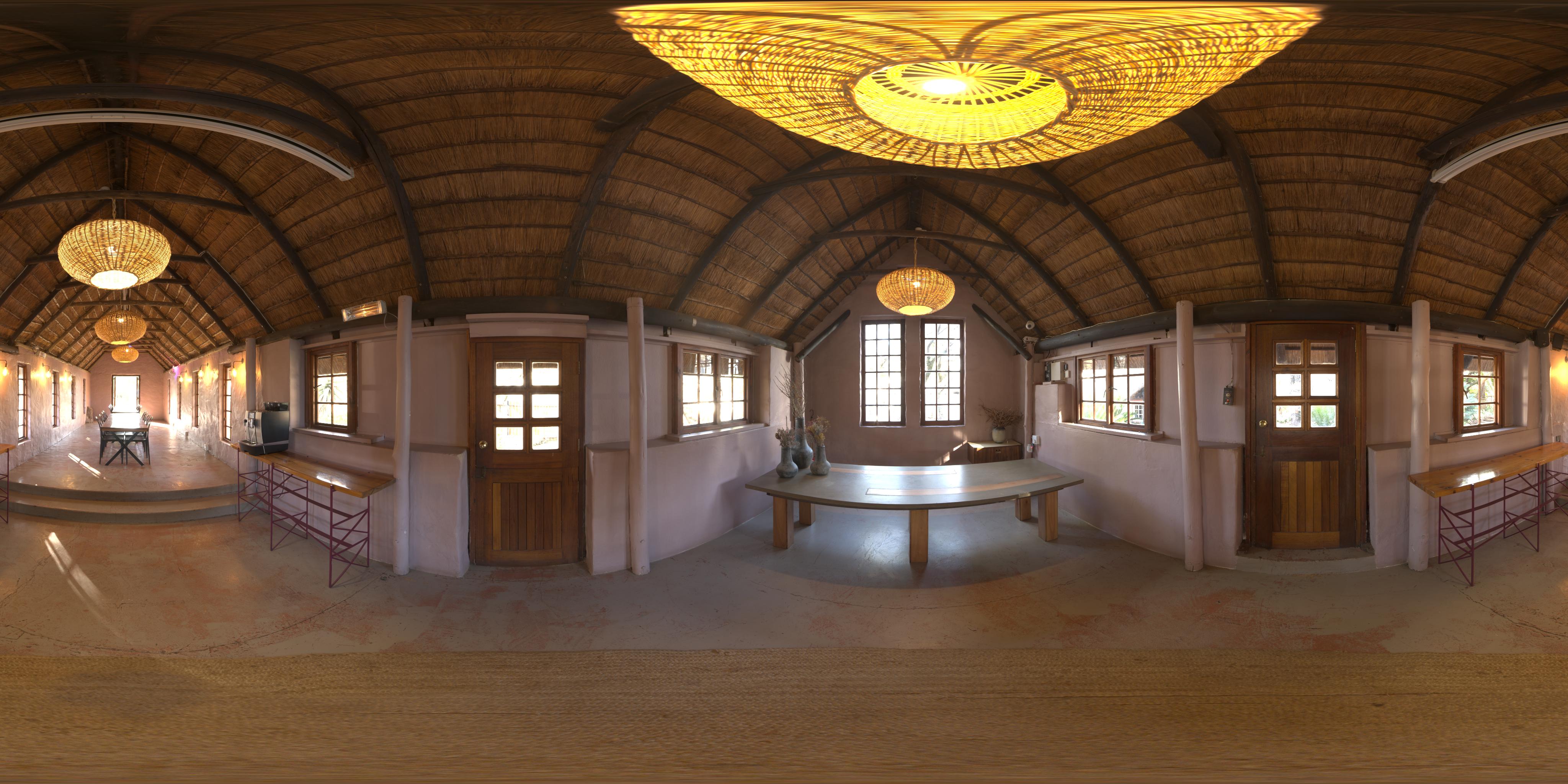} \\
    \includegraphics[width=0.16\textwidth]{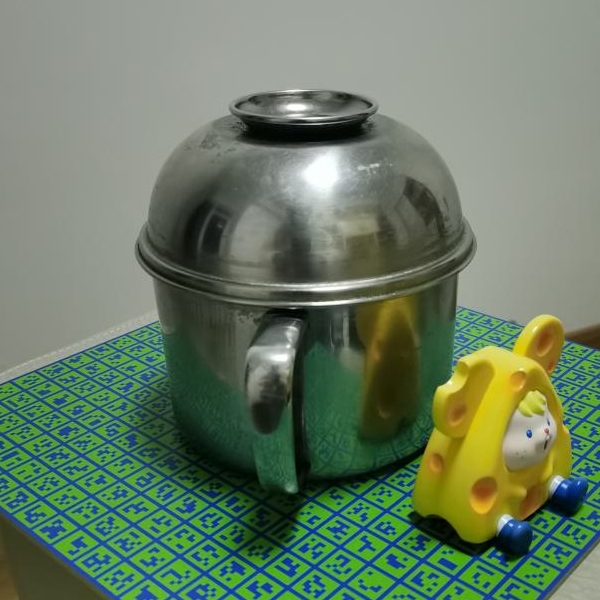} &
    \includegraphics[width=0.16\textwidth]{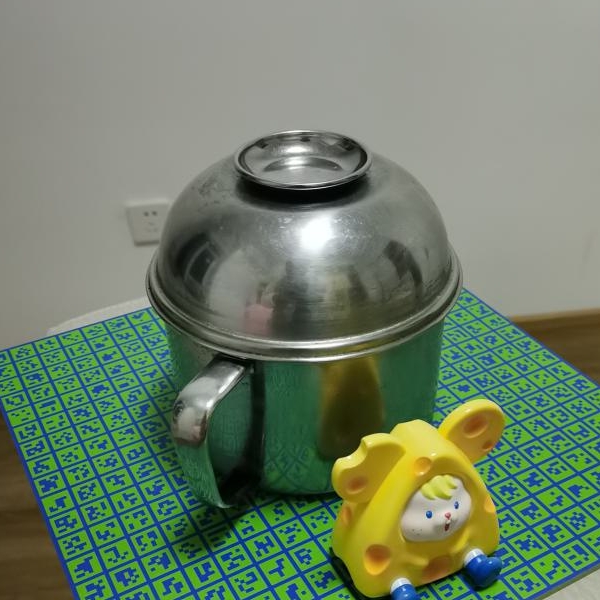} &
    \includegraphics[width=0.16\textwidth]{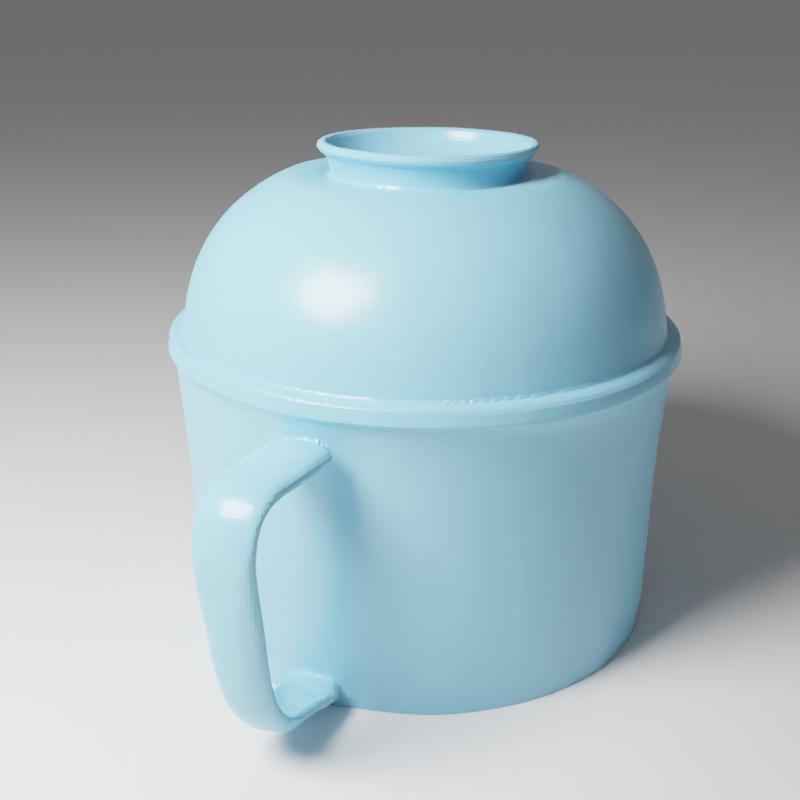} &
    \includegraphics[width=0.16\textwidth]{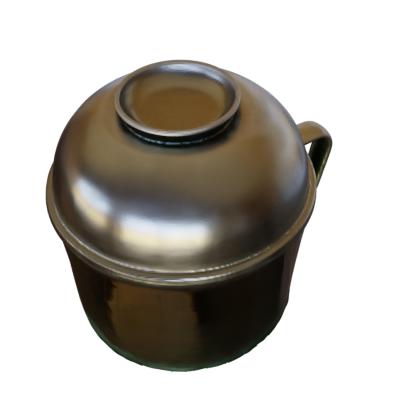} &
    \includegraphics[width=0.16\textwidth]{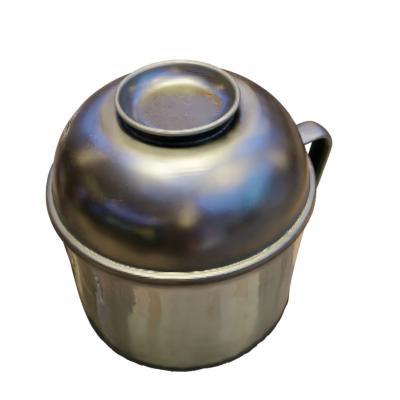} &
    \includegraphics[width=0.16\textwidth]{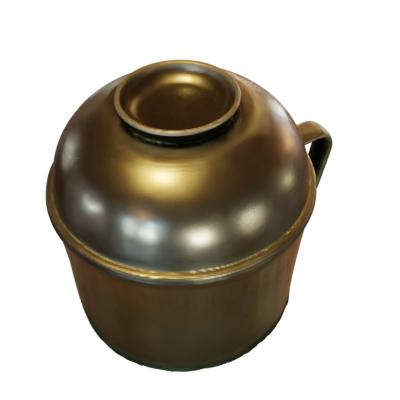} \\
    \includegraphics[width=0.16\textwidth]{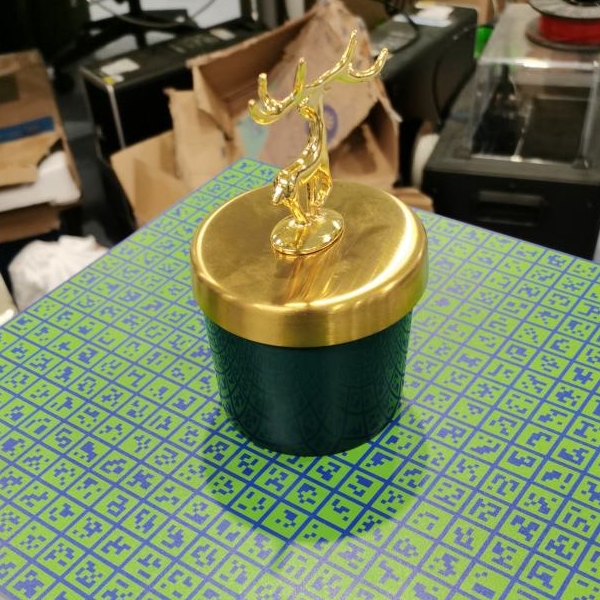} &
    \includegraphics[width=0.16\textwidth]{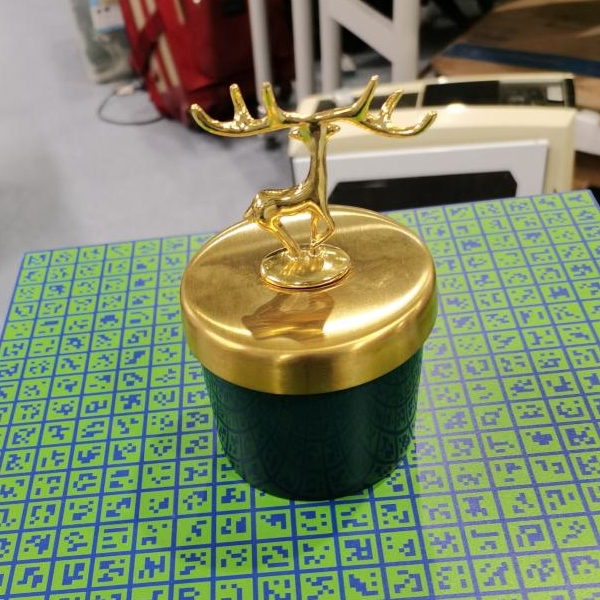} &
    \includegraphics[width=0.16\textwidth]{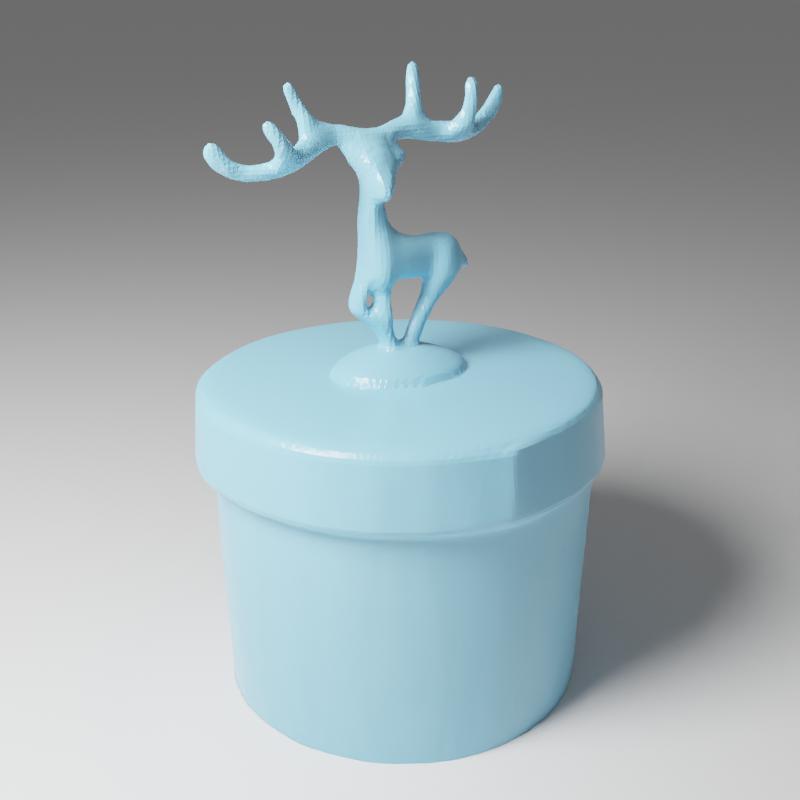} &
    \includegraphics[width=0.16\textwidth]{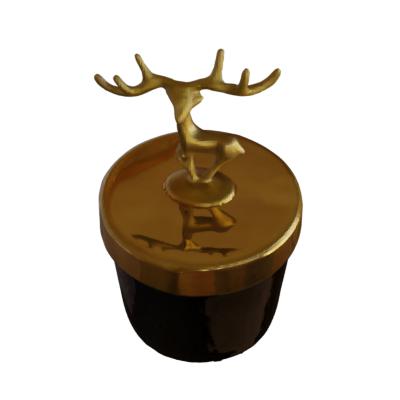} &
    \includegraphics[width=0.16\textwidth]{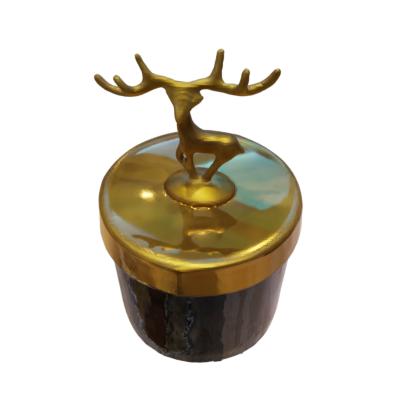} &
    \includegraphics[width=0.16\textwidth]{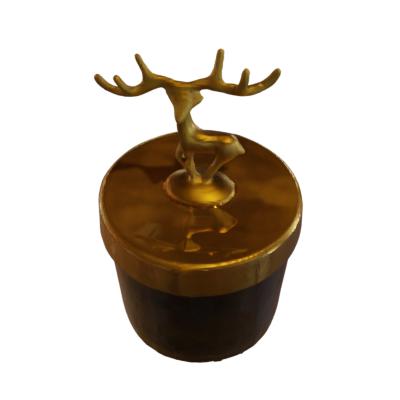} \\
    \includegraphics[width=0.16\textwidth]{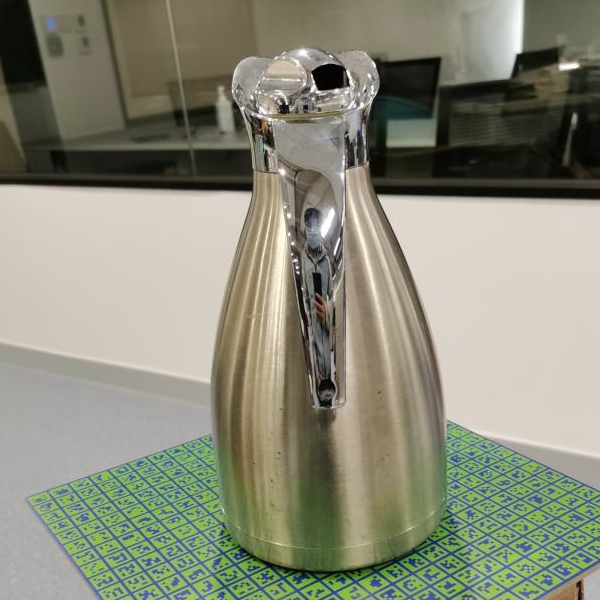} &
    \includegraphics[width=0.16\textwidth]{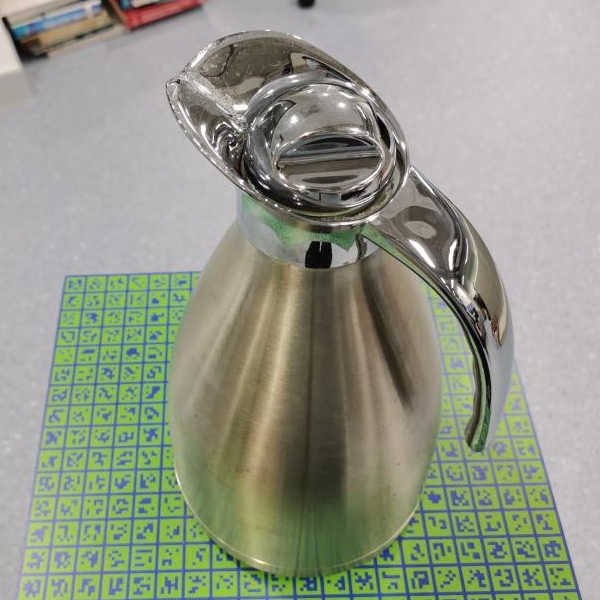} &
    \includegraphics[width=0.16\textwidth]{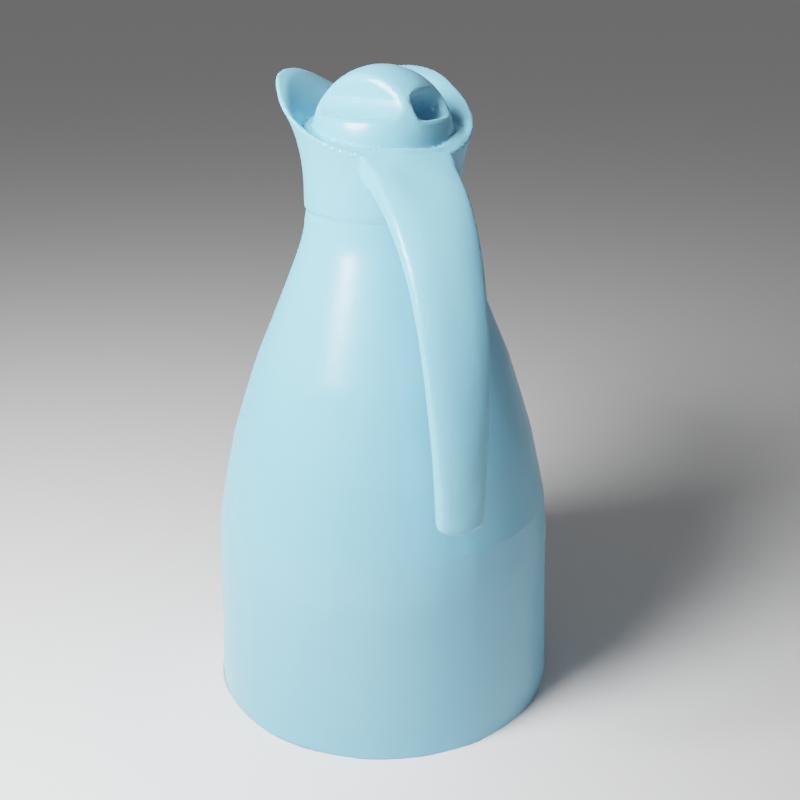} &
    \includegraphics[width=0.16\textwidth]{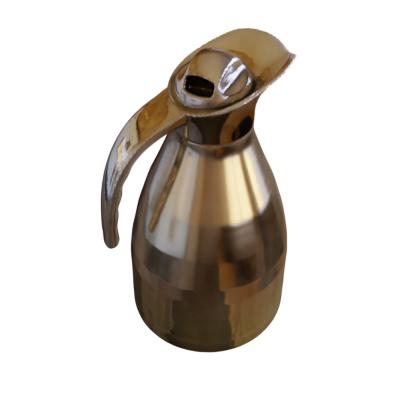} &
    \includegraphics[width=0.16\textwidth]{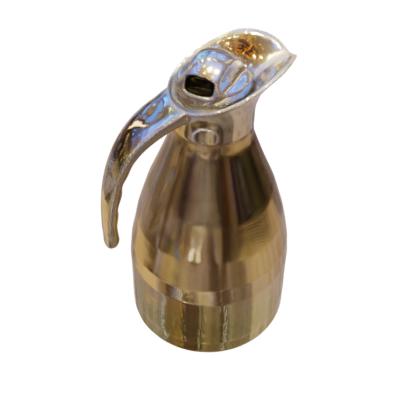} &
    \includegraphics[width=0.16\textwidth]{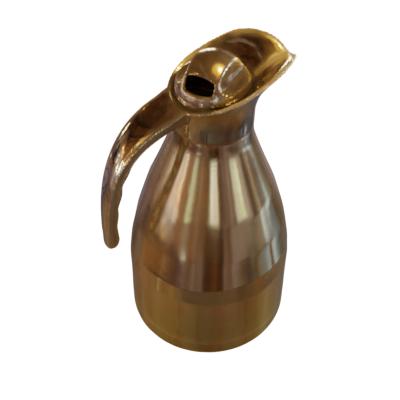} \\
    \includegraphics[width=0.16\textwidth]{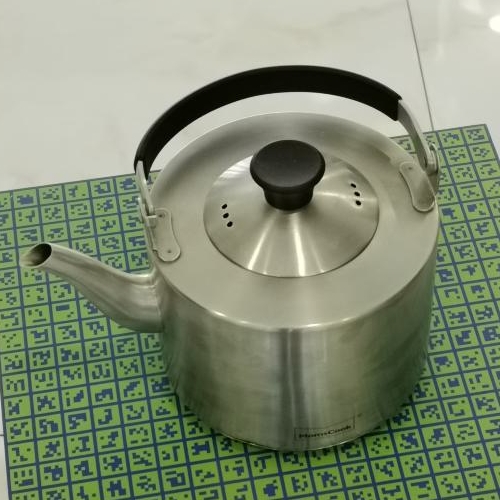} &
    \includegraphics[width=0.16\textwidth]{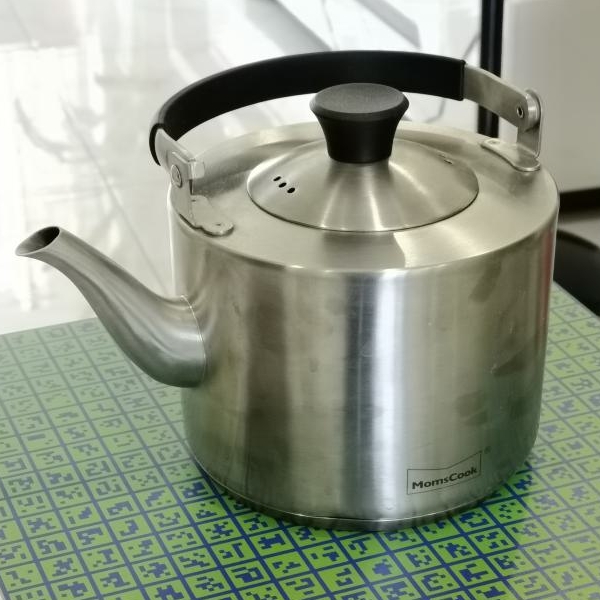} &
    \includegraphics[width=0.16\textwidth]{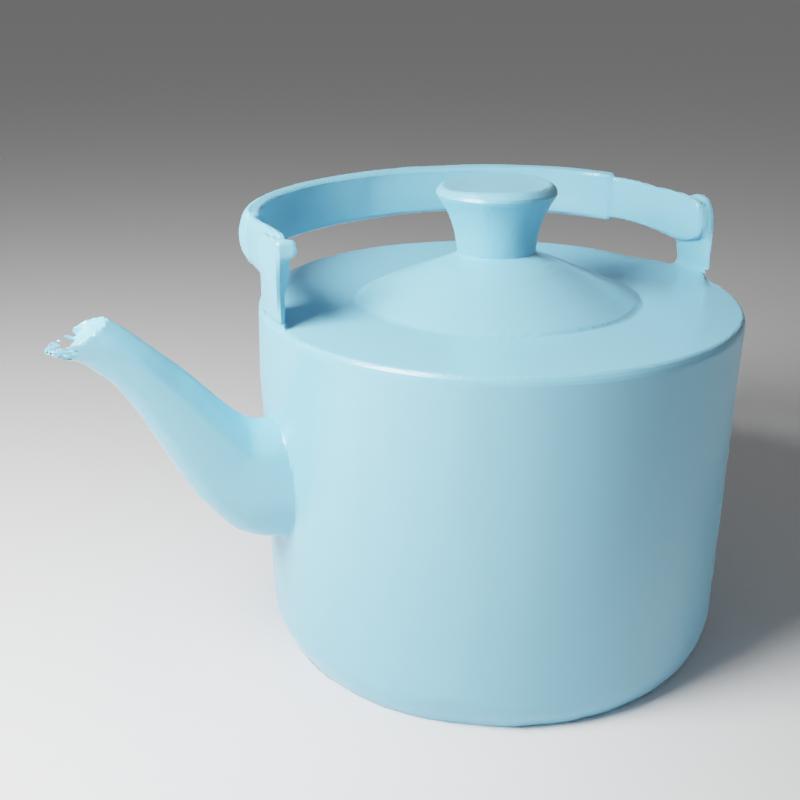} &
    \includegraphics[width=0.16\textwidth]{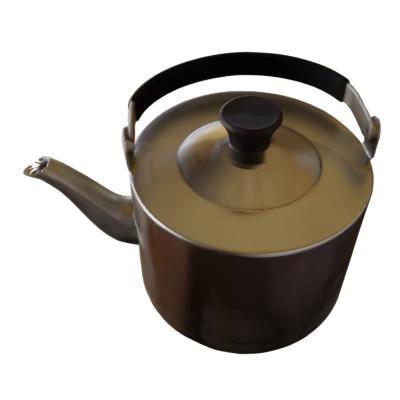} &
    \includegraphics[width=0.16\textwidth]{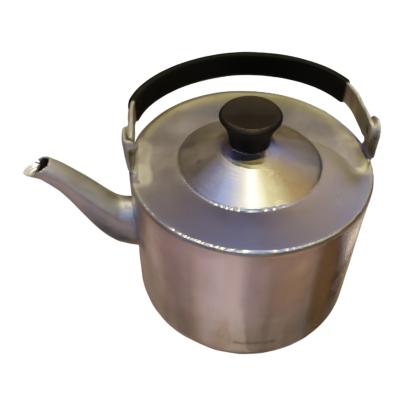} &
    \includegraphics[width=0.16\textwidth]{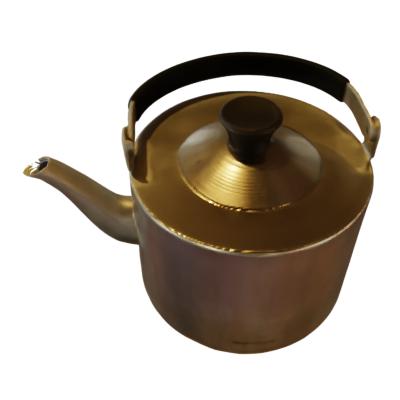} \\
    \includegraphics[width=0.16\textwidth]{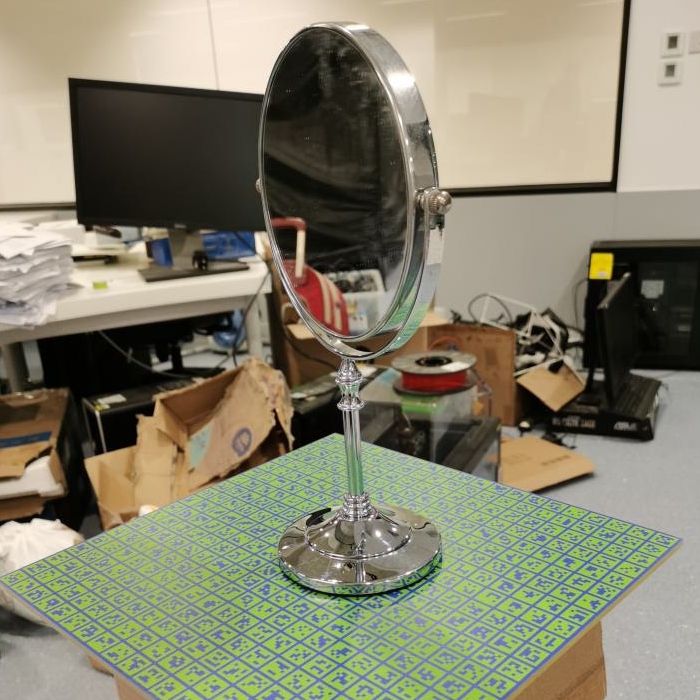} &
    \includegraphics[width=0.16\textwidth]{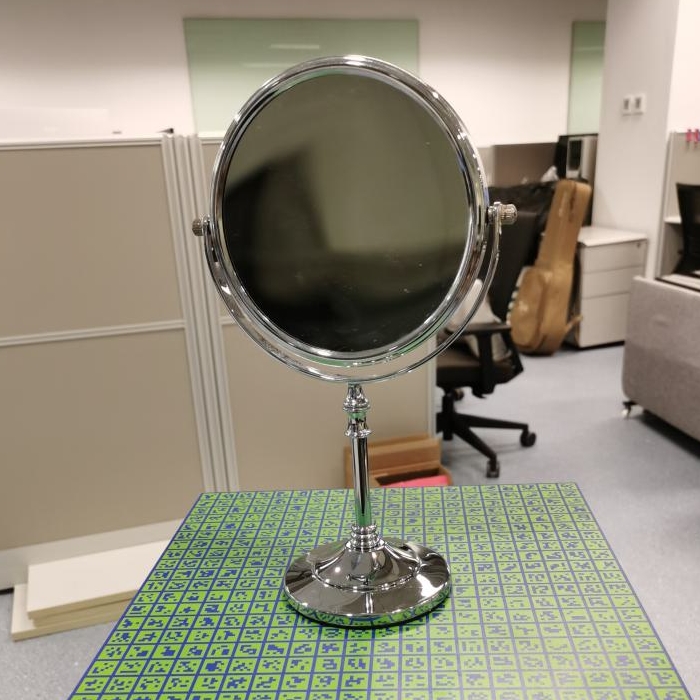} &
    \includegraphics[width=0.16\textwidth]{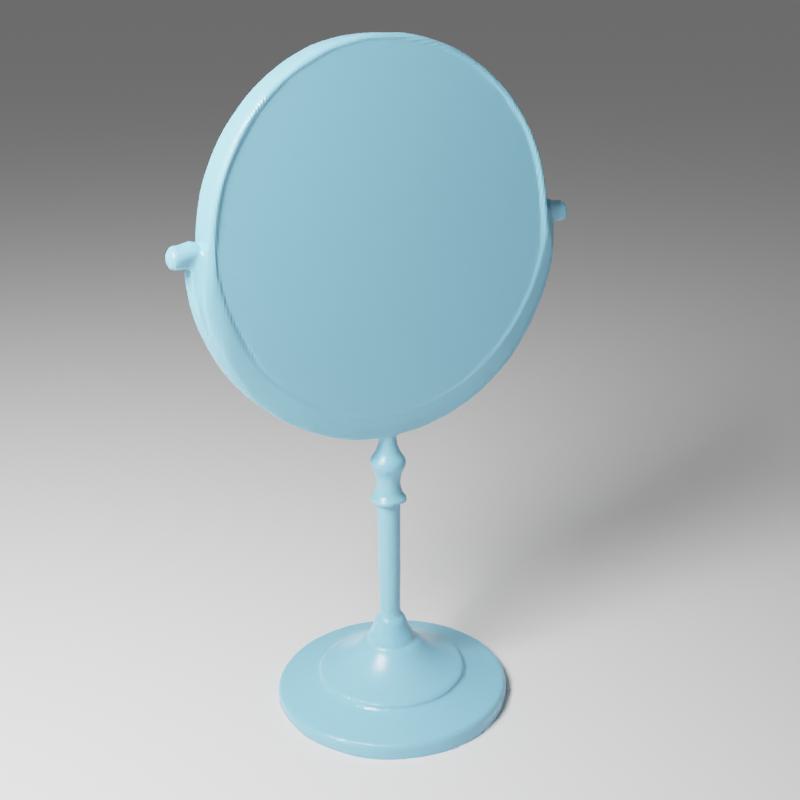} &
    \includegraphics[width=0.16\textwidth]{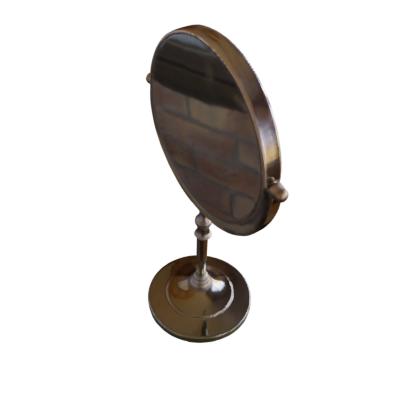} &
    \includegraphics[width=0.16\textwidth]{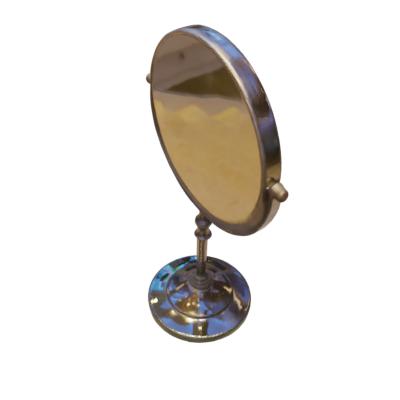} &
    \includegraphics[width=0.16\textwidth]{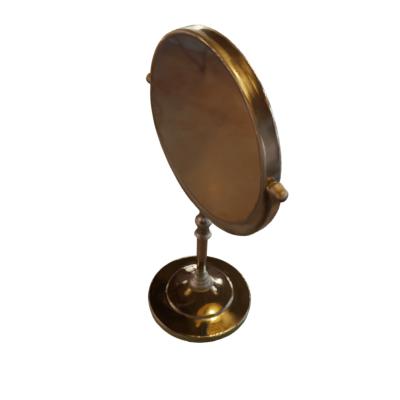} \\
    \includegraphics[width=0.16\textwidth]{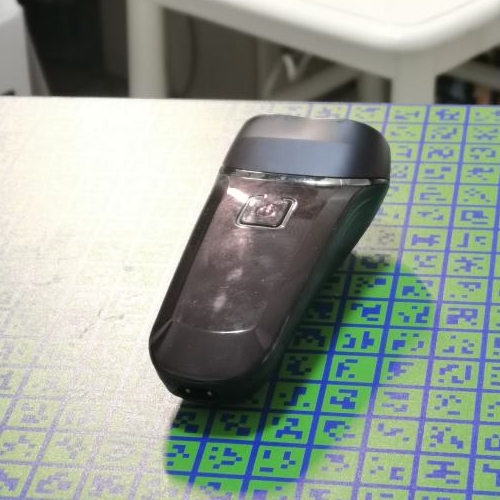} &
    \includegraphics[width=0.16\textwidth]{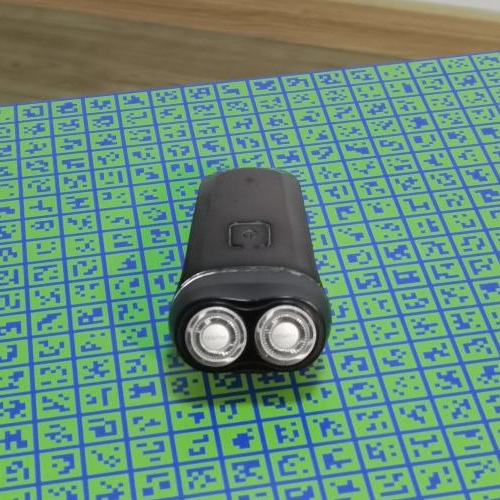} &
    \includegraphics[width=0.16\textwidth]{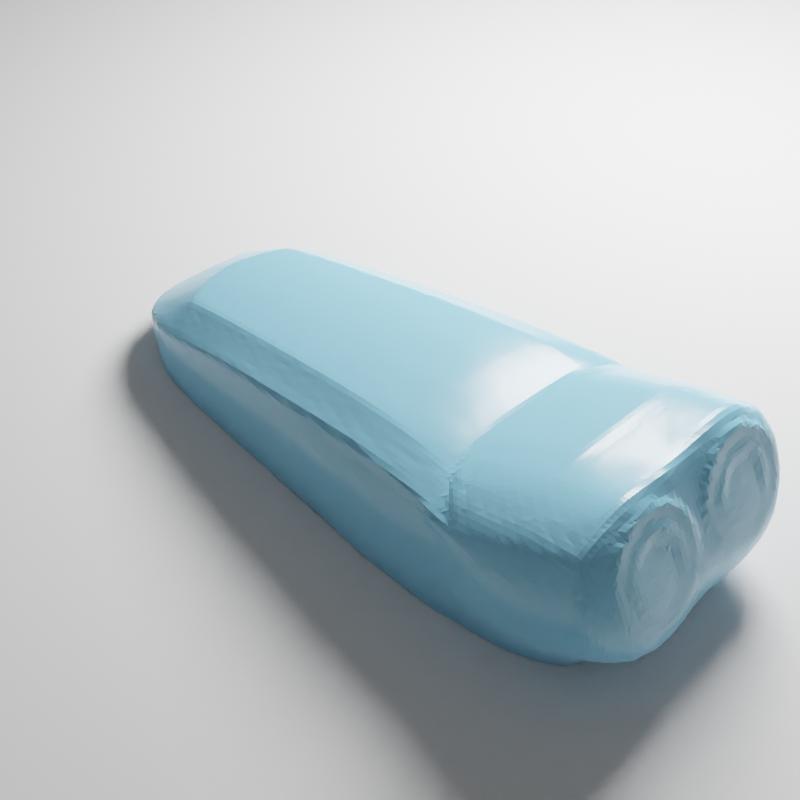} &
    \includegraphics[width=0.16\textwidth]{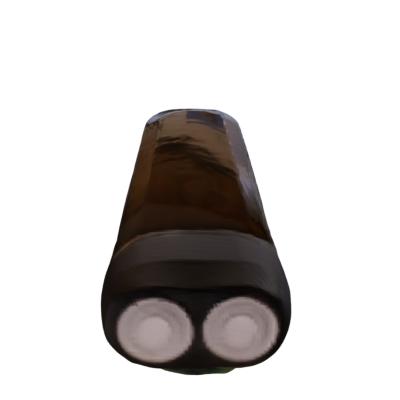} &
    \includegraphics[width=0.16\textwidth]{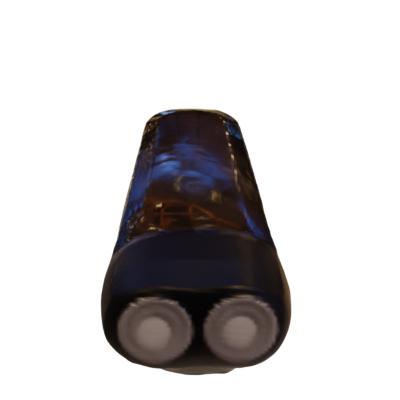} &
    \includegraphics[width=0.16\textwidth]{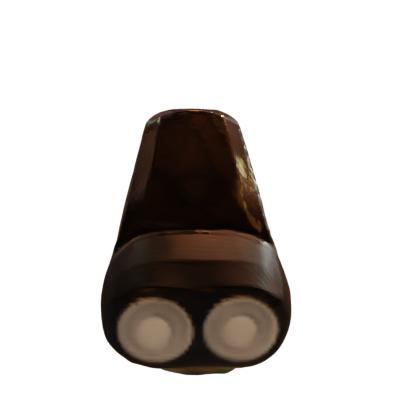} \\
    Image0 & Image1 & Geometry & Relighting0 & Relighting1 & Relighting2 \\
    \end{tabular}
    \caption{\textbf{Reconstruction results on other real objects}. The supplementary video contains more qualitative results.}
    \label{fig:add}
\end{figure*}

\end{document}